\documentclass{article}

\PassOptionsToPackage{numbers, compress}{natbib}
\PassOptionsToPackage{table}{xcolor}

\usepackage[preprint]{neurips_2026}

\usepackage{enumitem}
\usepackage{listings}    

\usepackage[utf8]{inputenc} 
\usepackage[T1]{fontenc}    
\usepackage{hyperref}       
\usepackage{url}            
\usepackage{booktabs}       
\usepackage{amsfonts}       
\usepackage{amsmath}        
\usepackage{nicefrac}       
\usepackage{microtype}      
\usepackage{xcolor}         
\usepackage{multirow}       
\usepackage{graphicx}       
\usepackage{caption}        
\usepackage{wrapfig}        
\usepackage{array}          
\usepackage[most]{tcolorbox}
\usepackage{circledsteps}
\usepackage{tabularx}               
\usepackage{arydshln} 
\usepackage{amssymb}
\usepackage{pifont}
\usepackage{algorithm}
\usepackage{algorithmic}
\usepackage{verbatim}
\usepackage{fancyhdr}
\usepackage{makecell}

\usepackage{jsonparse}
\usepackage{array}

\JSONParseFromFile{\CasesPrompts}{prompts/Cases_prompts.json}

\definecolor{LightOrange}{rgb}{1.0, 0.85, 0.7}
\definecolor{lightblue}{HTML}{42bcf5}

\newcommand{\yesmark}{\textcolor{green!60!black}{$\surd$}}
\newcommand{\nomark}{\textcolor{red}{$\times$}}

\definecolor{sae_orange}{HTML}{D67454}
\definecolor{ntublue}{HTML}{181C62}

\tcbuselibrary{most}
\tcbset{
  aibox/.style={
    width=\textwidth,
    top=10pt,
    colback=white,
    colframe=black,
    colbacktitle=black,
    enhanced,
    center,
    attach boxed title to top left={yshift=-0.1in,xshift=0.15in},
    boxed title style={boxrule=0pt,colframe=white,},
  }
}
\newtcolorbox{AIbox}[2][]{aibox,title=#2,#1}

\definecolor{qaBlue}{HTML}{eef8f9}
\definecolor{qaBlueBorder}{HTML}{d2ddf1}
\definecolor{qaBlueText}{HTML}{0070c0}
\definecolor{qualGreen}{HTML}{f0f7ec}
\definecolor{qualGreenBorder}{HTML}{d4e8c6}
\definecolor{qualGreenText}{HTML}{588e32}
\definecolor{pairPurple}{HTML}{f0eaf5}
\definecolor{pairPurpleBorder}{HTML}{e2d6ec}
\definecolor{pairPurpleText}{HTML}{7030a0}
\definecolor{pairYellow}{HTML}{fefbf1}
\definecolor{pairYellowBorder}{HTML}{ffecb2}

\newtcolorbox{QAbox}[2][]{
  width=\textwidth, top=8pt,
  colback=qaBlue, colframe=qaBlueBorder, colbacktitle=qaBlueBorder,
  coltitle=black,
  enhanced, center,
  attach boxed title to top left={yshift=-0.1in,xshift=0.15in},
  boxed title style={boxrule=0pt,colframe=white},
  title=#2, #1
}
\newtcolorbox{QualBox}[2][]{
  width=\textwidth, top=8pt,
  colback=qualGreen, colframe=qualGreenBorder, colbacktitle=qualGreenBorder,
  coltitle=black,
  enhanced, center,
  attach boxed title to top left={yshift=-0.1in,xshift=0.15in},
  boxed title style={boxrule=0pt,colframe=white},
  title=#2, #1
}
\newtcolorbox{PairBox}[2][]{
  width=\textwidth, top=8pt,
  colback=pairPurple, colframe=pairPurpleBorder, colbacktitle=pairPurpleBorder,
  coltitle=black,
  enhanced, center,
  attach boxed title to top left={yshift=-0.1in,xshift=0.15in},
  boxed title style={boxrule=0pt,colframe=white},
  title=#2, #1
}
\newtcolorbox{CASEBAX}[2][]{
  width=\textwidth, top=8pt,
  colback=pairYellow,
  colframe=pairYellowBorder,
  colbacktitle=pairYellowBorder,
  coltitle=black,
  enhanced,
  center,
  attach boxed title to top left={yshift=-0.1in,xshift=0.15in},
  boxed title style={boxrule=0pt,colframe=white},
  title=#2,
  #1
}

\newcommand{\CasePromptFromJSON}[1]{%
    \begin{minipage}[t]{\linewidth}
    \raggedright
    \scriptsize
    \textbf{Prompt:} \textit{\JSONParseValue{\CasesPrompts}{#1.prompt}}
    \end{minipage}
    }
    
    \newcommand{\CaseImage}[1]{%
    \begin{minipage}[t]{\linewidth}
    \centering
    \includegraphics[width=\linewidth,height=1.45cm,keepaspectratio]{#1}
    \end{minipage}
}

\usepackage{CJKutf8}
\usepackage[misc]{ifsym}

\definecolor{lowyellow}{RGB}{241, 196, 15}

\newcommand{\WorldReasonBench}{WorldReasonBench}
\newcommand{\WorldRewardBench}{WorldRewardBench}

\title{\WorldReasonBench{}: Human-Aligned Stress Testing of Video Generators as Future World-State Predictors}

\author{%
  \textbf{Keming Wu}$^{1*}$\quad
  \textbf{Yijing Cui}$^{1*}$\quad
  \textbf{Wenhan Xue}$^{1}$\quad
  \textbf{Qijie Wang}$^{1}$\quad
  \textbf{Xuan Luo}$^{1}$\quad
  \textbf{Zhiyuan Feng}$^{1}$\\
  \textbf{Zuhao Yang}$^{2}$\quad
  \textbf{Sudong Wang}$^{4}$\quad
  \textbf{Sicong Jiang}$^{5}$\quad
  \textbf{Haowei Zhu}$^{1}$\quad
  \textbf{Zihan Wang}$^{5}$\quad
  \textbf{Ping Nie}$^{3}$\\
  \textbf{Wenhu Chen}$^{3}$\quad
  \textbf{Bin Wang}$^{1 \textsuperscript{\Letter} }$\\[4pt]
  $^{1}$Tsinghua University\quad
  $^{2}$Nanyang Technological University\quad
  $^{3}$University of Waterloo\\
  $^{4}$Hong Kong University of Science and Technology (Guangzhou)\quad
  $^{5}$2077 AI\\[2pt]
  \footnotesize
    \vspace{0.15cm}
    $^\ast$Equal contribution ~~~~~~\textsuperscript{\Letter} Corresponding author \\
    \vspace{0.15cm}
    \tt\small{Project page: \url{https://unix-ai-lab.github.io/WorldReasonBench/}}\\
  \texttt{\{wukm25, cuiyj25\}@mails.tsinghua.edu.cn}
}

\begin{document}

\maketitle


\begin{abstract}
  Commercial video generation systems such as Seedance2.0 and Veo3.1 have rapidly improved, strengthening the view that video generators may be evolving into ``world simulators.'' Yet the community still lacks a benchmark that directly tests whether a model can reason about how an observed world should evolve over time. We introduce \textbf{\WorldReasonBench{}}, which reframes video generation evaluation as \emph{world-state prediction}: given an initial state and an action, can a model generate a future video whose state evolution remains physically, socially, logically, and informationally consistent? \WorldReasonBench{} contains 436 curated test cases with structured ground-truth QA annotations spanning four reasoning dimensions and 22 subcategories. We evaluate generated videos with a human-aligned two-part methodology: \emph{Process-aware Reasoning Verification} uses structured QA and reasoning-phase diagnostics to detect temporal and causal failures, while \emph{Multi-dimensional Quality Assessment} scores reasoning quality, temporal consistency, and visual aesthetics for ranking and reward modeling. We further introduce \textbf{\WorldRewardBench{}}, a preference benchmark with approximately 6K expert-annotated pairs over 1.4K videos, supporting pair-wise and point-wise reward-model evaluation. Across modern video generators, our results expose a persistent gap between visual plausibility and world reasoning: videos can look convincing while failing dynamics, causality, or information preservation. We will release our benchmarks and evaluation toolkit to support community research on genuinely world-aware video generation at \url{https://github.com/UniX-AI-Lab/WorldReasonBench/}.
\end{abstract}


\section{Introduction}
\label{sec:intro}

The rapid advance of large-scale video generation models~\citep{polyak2024movie, kong2024hunyuanvideo, wan2025wan, yang2024cogvideox,wu2026visual} has shifted the central question in video generation. Frontier systems in the Seedance, Veo, and Sora families~\citep{gao2025seedance, wiedemer2025video, brooks2024video} now produce longer, cleaner, and more controllable videos, while recent studies suggest that video models may already exhibit zero-shot learning and reasoning-like behavior in selected settings~\citep{wiedemer2025video}. These advances make it increasingly plausible to ask whether modern video generators are beginning to act as \emph{world models} rather than only powerful pixel synthesizers.

\begin{figure}[t]
\centering
\includegraphics[width=0.9\textwidth]{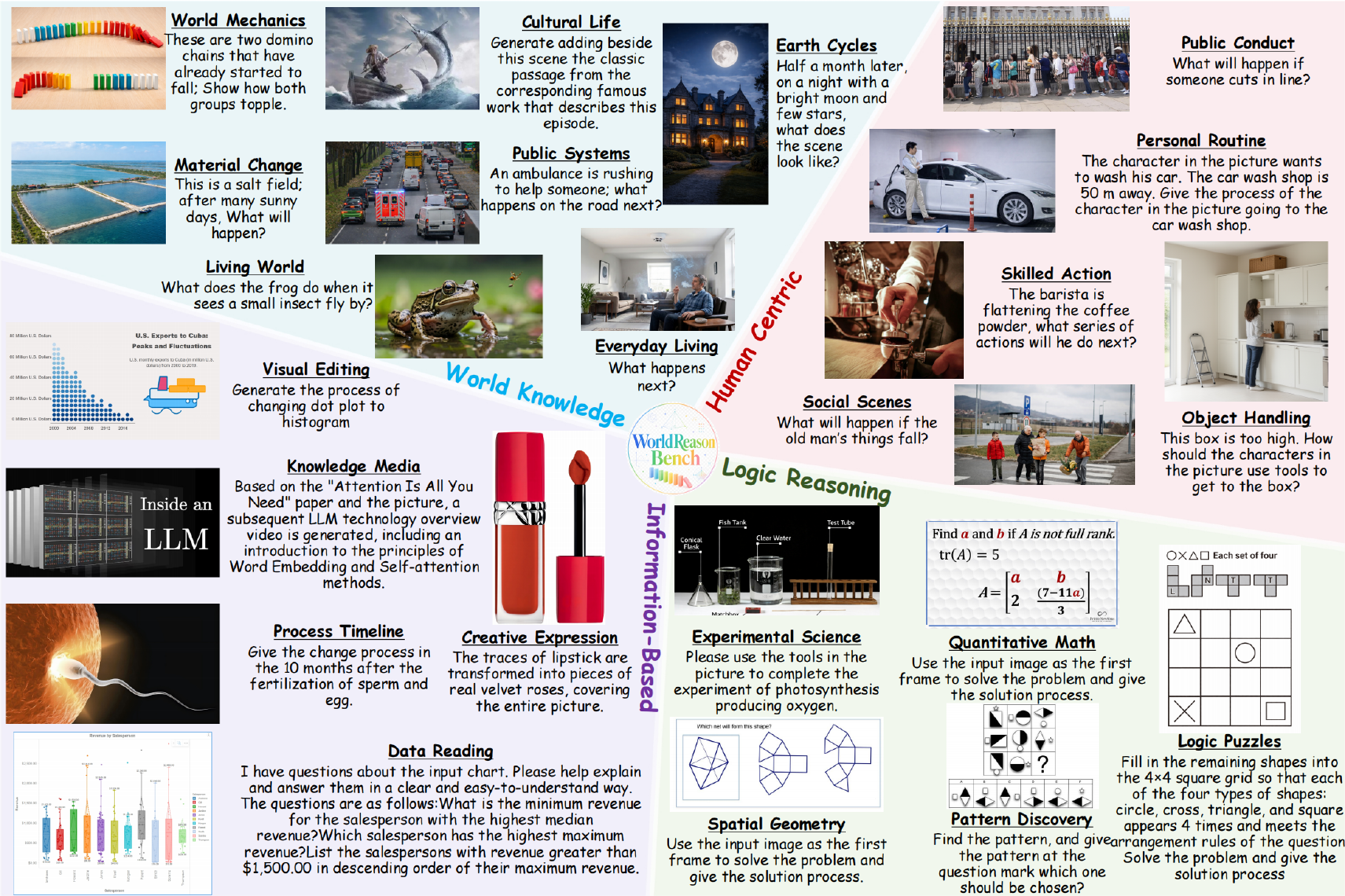}
\caption{\textbf{Overview of \WorldReasonBench{}.} We evaluate video generators as world-state predictors: given an initial visual state and an action or instruction, the model must generate a future video whose state evolution remains physically, socially, logically, and informationally consistent. \WorldReasonBench{} spans four reasoning dimensions organized into 22 concise, dimension-specific subcategories, and is paired with complementary automated and human-aligned evaluation pipelines.}
\label{fig:overview}
\end{figure}

Evaluation, however, has not kept pace with this shift. Most existing benchmarks still emphasize perceptual quality, motion smoothness, or prompt alignment. Recent reasoning-oriented efforts each cover a useful slice of the problem but stop short of open-domain world-state prediction: V-ReasonBench~\citep{luo2025v} and Gen-ViRe~\citep{liu2025can} target answer-verifiable cognitive tasks, VIPER~\citep{li2025viper} formalizes process-aware diagnostics on procedural settings, WorldSimBench~\citep{qin2024worldsimbench} focuses on embodied control, and VideoVerse~\citep{wang2025videoverse} evaluates single-event causality with binary QA. None of them asks, end-to-end and on open-domain content, whether a generator that observes an initial visual state can correctly infer and simulate the \emph{future evolution of the world}, and none releases calibrated expert preference data for reward-model evaluation. This gap is especially consequential for the open-source community: as frontier commercial systems improve rapidly, the field needs a common benchmark that can tell whether open-source progress reflects genuine reasoning gains or simply better visual polish.

Consider a simple example: a generator given an image of an apple on a branch and instructed to drop it may produce a visually impressive clip---smooth motion, realistic textures, attractive lighting---yet fail as a world model if the apple accelerates upward, splits in mid-air, or traces a linear rather than parabolic trajectory. Standard quality metrics reward such a video for realism while missing its failure to obey basic dynamics. The core question is therefore not only \emph{how good the video looks}, but whether the model has generated the \emph{right future state transition}. We accordingly recast video generation evaluation as \emph{world-state prediction}: given an initial visual state and an action or instruction, can the model roll the world forward into temporally consistent future states? We further separate transitions that are inferable from visual evidence alone from those that benefit from explicit textual guidance, probing reasoning under different levels of external help. We introduce \textbf{\WorldReasonBench{}}, a reasoning-aware benchmark with 436 curated test cases and structured ground-truth QA annotations, guided by the principle that \emph{a true world model should be interrogable}---one should be able to ask reasoning-oriented questions about the video and obtain answers consistent with real-world knowledge. Since binary QA alone may hide process failures, we evaluate each model through two complementary components, Process-aware Reasoning Verification and Multi-dimensional Quality Assessment. Our contributions are: \textbf{(1)} \WorldReasonBench{}, a reasoning-aware benchmark covering four dimensions and 22 subcategories that tests whether 11 closed- and open-source generators roll an observed initial state into a coherent future sequence (Figure~\ref{fig:overview}); \textbf{(2)} a human-aligned evaluation methodology combining Process-aware Reasoning Verification with Multi-dimensional Quality Assessment, validated against expert human preferences; and \textbf{(3)} \WorldRewardBench{}, a preference-based calibration benchmark with approximately 6K expert-annotated pairs over 1.4K videos supporting pair-wise and point-wise reward-model evaluation.


\section{Related Work}
\label{sec:related}

\paragraph{Video generation models as world simulators.}
Popularized by Sora~\citep{brooks2024video}, the view of video generators as world simulators has become more compelling as commercial systems such as Seedance and Veo improve in long-horizon coherence, controllability, and realism~\citep{gao2025seedance, wiedemer2025video}, with recent studies even suggesting zero-shot learning and reasoning-like behavior in selected settings~\citep{wiedemer2025video}. Capability demos alone do not establish robust world understanding, however: physical-law analyses show that even strong models fail on gravity, object permanence, and causal consistency~\citep{kang2024far}. We therefore aim to test these claims systematically rather than infer them from isolated examples.

\paragraph{Benchmarks and automatic evaluation for video generation.}
Existing video benchmarks mostly target perceptual quality or prompt alignment via reference metrics (FID~\citep{heusel2017gans}, FVD~\citep{unterthiner2019fvd}, LPIPS~\citep{zhang2018unreasonable}) and aesthetics/compositionality suites~\citep{huang2024vbench,zheng2025vbench,liu2024evalcrafter,liu2023fetv,sun2025t2v}, none of which provide structured reasoning verification. Reasoning-oriented benchmarks each cover one slice---embodied task-success~\citep{qin2024worldsimbench}, small-scale answer-verifiable puzzles~\citep{luo2025v,liu2025can}, procedural process-aware tasks~\citep{li2025viper}, single-event causality with Likert ratings~\citep{wang2025videoverse}, physical-law or rule-governed transitions~\citep{meng2024towards,he2025ruler}, and video understanding rather than generation~\citep{wang2026very}. VLM-as-Judge pipelines~\citep{zheng2023judging,ma2025videoeval,he2025videoscore2} scale evaluation but single-pass judges over-reward visual plausibility and miss process-level errors. \WorldReasonBench{} instead pairs an initial image with a text instruction to probe open-domain future-state evolution, annotates each case with 5--7 QA pairs across four reasoning phases (state, process, fidelity, mechanism), and releases \WorldRewardBench{} with ${\sim}6$K expert preference pairs over 1{,}432 videos from 11 generators to calibrate automatic metrics.


\begin{table*}[t]
  \caption{\textbf{Comparison with existing video generation reasoning benchmarks along auditable axes.} \emph{\#Cases}: N/R if not stated. \emph{Input}: T2V, TI2V(initial image + text), or Embodied. \emph{Reward Data}: publicly released preferences. \emph{Process Phases}: $\geq 2$ phase-level scores. \emph{Human Calib.}: rank correlation or pairwise agreement vs.\ experts (Likert user study alone does not qualify).}
  \label{tab:benchmark_comparison}
  \centering
  \scriptsize
  \setlength{\tabcolsep}{3.5pt}
  \renewcommand{\arraystretch}{1.10}
  \resizebox{\textwidth}{!}{
  \begin{tabular}{@{}lcccllccc@{}}
    \toprule
    \textbf{Benchmark}
    & \begin{tabular}[c]{@{}c@{}}\textbf{Dim. /}\\\textbf{Sub-dim.}\end{tabular}
    & \textbf{\#Cases}
    & \textbf{Input}
    & \textbf{Domain}
    & \textbf{Annotation Unit}
    & \begin{tabular}[c]{@{}c@{}}\textbf{Reward}\\\textbf{Data}\end{tabular}
    & \begin{tabular}[c]{@{}c@{}}\textbf{Process}\\\textbf{Phases}\end{tabular}
    & \begin{tabular}[c]{@{}c@{}}\textbf{Human}\\\textbf{Calib.}\end{tabular} \\
    \midrule
    VBench-2.0~\citep{zheng2025vbench}
    & 5/18 & $\approx$1260 & T2V
    & curated open prompts
    & Likert + auto metrics
    & \nomark & \nomark & user study \\

    WorldSimBench~\citep{qin2024worldsimbench}
    & 3/20 & 2,831 & T2V/TI2V
    & embodied scenarios
    & perceptual + manipulative
    & \yesmark & \nomark & task success \\

    V-ReasonBench~\citep{luo2025v}
    & 4/13 & 326 & TI2V
    & answer-verifiable tasks
    & rule-based verifier
    & \nomark & \nomark & manual val. \\

    Gen-ViRe~\citep{liu2025can}
    & 6/24 & 72 & T2V
    & cognitive subtasks
    & per-task verifier
    & \nomark & \nomark & human val. \\

    VIPER~\citep{li2025viper}
    & 6/16 & 309 & T2V
    & procedural / rule-following
    & per-step process score
    & \nomark & \yesmark & -- \\

    VideoVerse~\citep{wang2025videoverse}
    & 10/-- & 300 & T2V
    & single-event causality
    & binary QA / event (793)
    & \nomark & partial & user study \\

    \midrule
    \textbf{\WorldReasonBench{}}
    & \textbf{4/22}
    & \textbf{436}
    & \textbf{TI2V}
    & \textbf{open-domain world-state prediction}
    & \textbf{5--7 QA pairs / case}
    & \textbf{${\sim}$6K pref.\ pairs}
    & \textbf{\yesmark{} (4 phases)}
    & \textbf{expert Elo + Spearman} \\
    \bottomrule
  \end{tabular}}
\end{table*}

\section{\WorldReasonBench{}}
\label{sec:method}

We frame video generation as \emph{world-state prediction}: given an observed initial state and an instruction, a generator should produce a future video that follows the intended world evolution rather than merely appearing realistic.

\paragraph{Problem formulation and instruction regimes.}
Let $x_0$ be the initial world state and $a$ the intended action or transition; a generator produces $\hat{V}=\mathcal{G}(x_0,a)$, and evaluation asks whether $\hat{V}$ faithfully realizes the state evolution implied by both inputs. To measure how much textual guidance helps, we evaluate each case under two regimes: $a_{\mathrm{implicit}}$ provides only a high-level intent, while $a_{\mathrm{hinted}}$ adds explicit transition guidance, and the resulting gap $\Delta_{\mathrm{hint}}=\mathrm{Score}(\hat{V}^{(1)})-\mathrm{Score}(\hat{V}^{(0)})$ measures the \emph{reasoning assistance benefit}.

\begin{figure}[t]
  \centering
  \includegraphics[width=\linewidth]{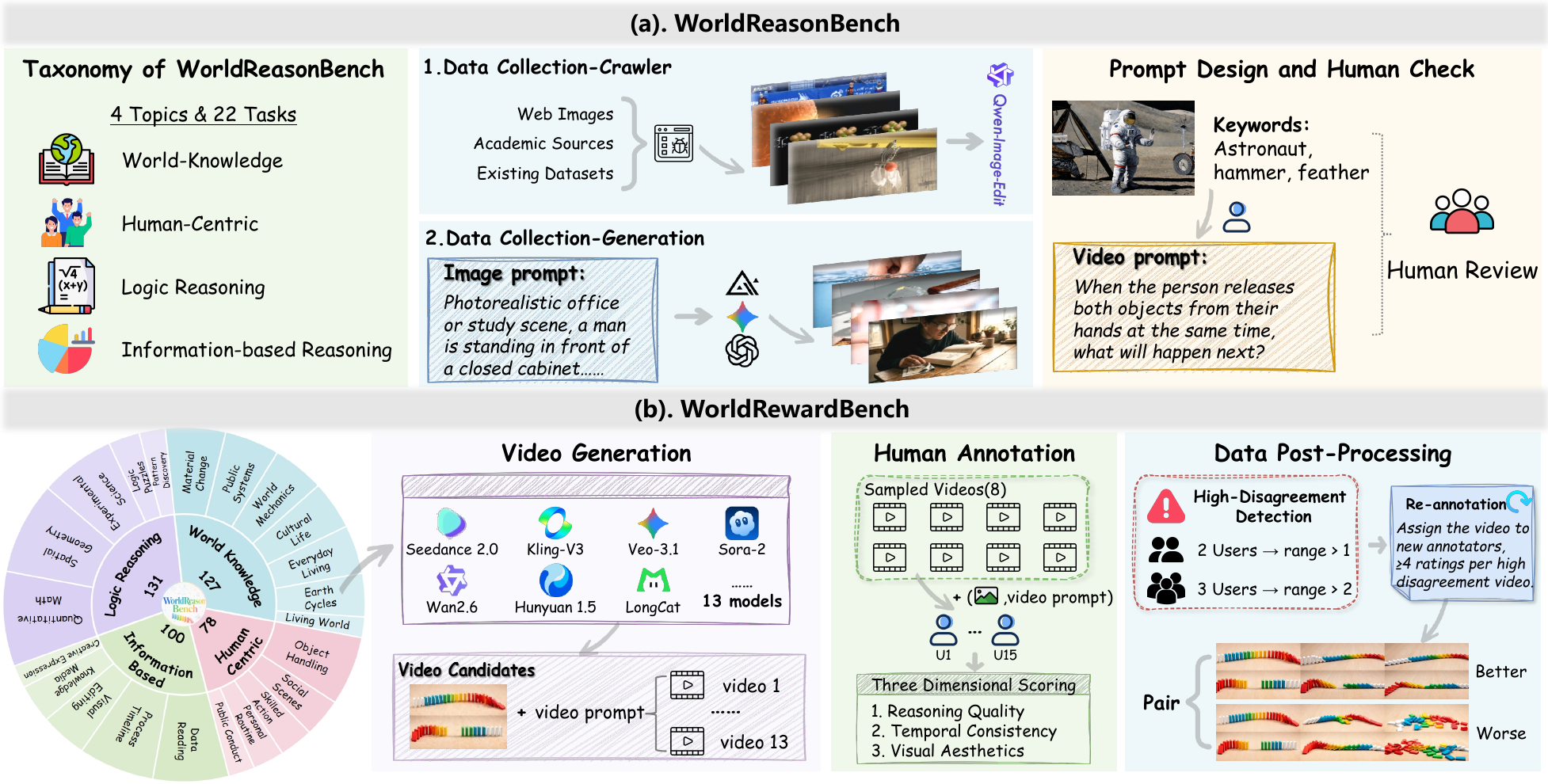}
  \caption{\textbf{Benchmark construction pipeline.} \emph{A}: \WorldReasonBench{} construction, including taxonomy-aware captioning, prompt generation, and QA generation. \emph{B}: \WorldRewardBench{} construction, including video sampling, expert scoring, preference-pair construction, and human-alignment evaluation.}
  \label{fig:pipeline}
\end{figure}

\subsection{\WorldReasonBench{} Construction}
\label{subsec:wrbench_construction}
\label{subsec:taxonomy}
\label{subsec:data}

\WorldReasonBench{} is constructed to evaluate whether a video generator can predict future world states from an observed initial state. As shown in Figure~\ref{fig:pipeline}(A), construction consists of a compact reasoning taxonomy and a three-stage VLM-assisted data pipeline.

\paragraph{Reasoning taxonomy.}
We organize world reasoning into four high-level dimensions and 22 short, interpretable subcategories. The complete taxonomy is visualized in Figure~\ref{fig:overview}, with detailed definitions, examples, and inclusion criteria provided in Appendix~\ref{app:taxonomy-details}.

\paragraph{Question design.}
Each test case is associated with a compact set of structured QA pairs spanning four question types: \emph{factual} (28.4\%, direct visual verification), \emph{reasoning} (27.1\%, causal mechanism understanding), \emph{detail} (24.7\%, fine-grained element verification), and \emph{temporal} (19.7\%, sequence and timing verification). Questions are further stratified into \emph{easy}, \emph{medium}, and \emph{hard} difficulty levels, enabling fine-grained analysis across both reasoning type and difficulty.

\paragraph{Data curation pipeline.}
We construct each benchmark case through three VLM-assisted stages. First, Qwen3.5~\citep{qwen35blog} produces a structured caption covering subjects, spatial relations, visual attributes, text/numeric elements, scene context, and potential dynamics. Second, Qwen3.5-27B generates reasoning-aware prompts conditioned on the target dimension, subcategory, and instruction regime. Third, Gemini3.1-Pro generates ground-truth QA pairs with expected answers, question-type labels, difficulty labels, and evaluation criteria. We use iterative JSON validation and repair to ensure reliable structured annotations. To control for VLM bias in the generated QA, two trained auditors further audit a stratified random subset on answerability, ground-truth correctness, and answer uniqueness, and rejected cases are rewritten or removed; the audit protocol and statistics are reported in Appendix~\ref{app:qa-audit}.

\subsection{\WorldRewardBench{} Construction}
\label{subsec:reward_bench}

\WorldRewardBench{} provides a human-aligned preference benchmark for evaluating whether automatic video judges recover expert preferences over world-reasoning failures. As summarized in Figure~\ref{fig:pipeline}(B), we build it from a high-quality subset of \WorldReasonBench{}: for each selected case, we collect generations from \textbf{11 video generation models} and sample \textbf{8 videos per case} to form a diverse annotation pool.

\paragraph{Human annotation and preference construction.}
Fifteen trained annotators rate each video on \emph{reasoning quality}, \emph{temporal consistency}, and \emph{visual aesthetics} using a 1--5 scale. We aggregate these ratings as $S(v)=0.4\,s_r(v)+0.3\,s_c(v)+0.3\,s_a(v)$, then rank videos within each benchmark case to derive candidate pairwise preferences. We apply confidence-aware filtering over score margins, relabel near-equal pairs ($\Delta_{ij}<0.1$) as ties, and randomize left/right order to reduce presentation bias. The resulting benchmark contains approximately 6K balanced preference pairs over 1.4K unique videos; implementation details and exact statistics are in Appendix~\ref{app:reward-postprocess}.

\WorldRewardBench{} supports pair-wise and point-wise reward-model evaluation through preference agreement, rank correlation, and tie/divergence diagnostics, providing the human-aligned calibration layer for the automatic evaluation methodology described next.

\subsection{Evaluation Framework}
\label{subsec:eval}

\begin{figure}[t]
  \centering
  \includegraphics[width=\linewidth]{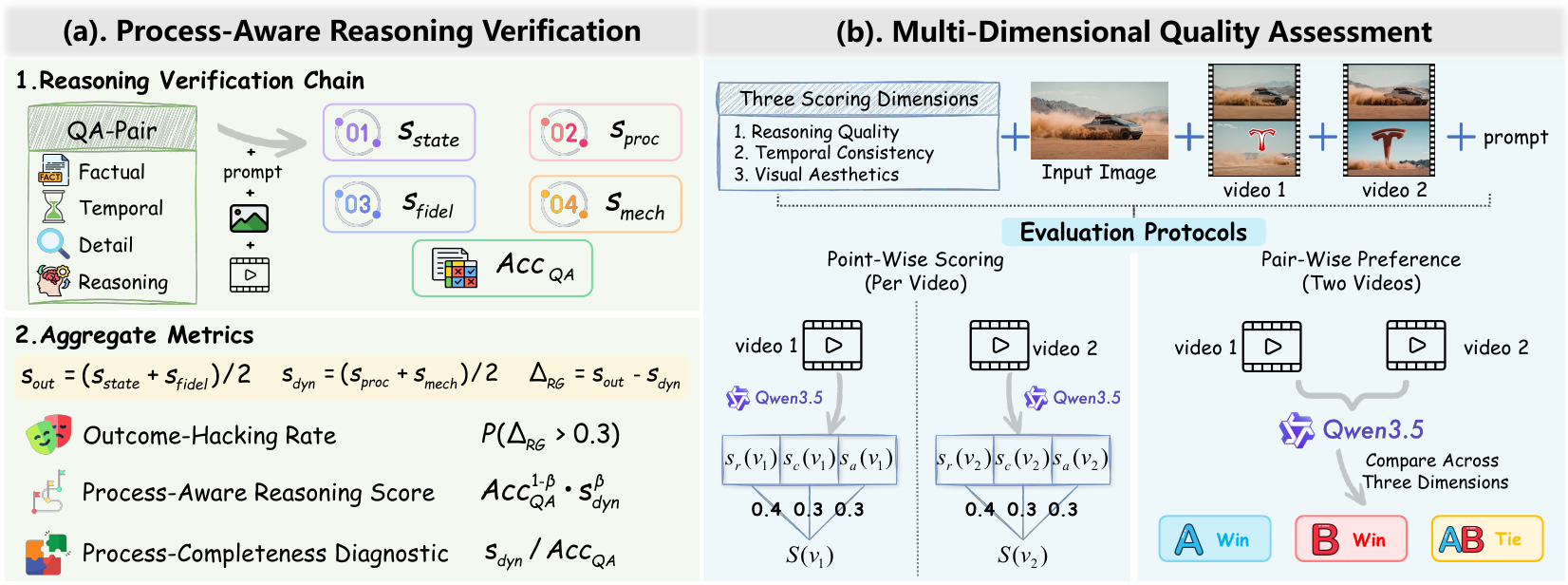}
  \caption{\textbf{Evaluation pipeline.} \emph{A}: Process-aware Reasoning Verification, which answers structured QA pairs from generated videos and converts them into reasoning-phase diagnostics. \emph{B}: Multi-dimensional Quality Assessment, which scores each video along reasoning quality, temporal consistency, and visual aesthetics for ranking and reward-model evaluation.}
  \label{fig:eval_pipeline}
\end{figure}

As shown in Figure~\ref{fig:eval_pipeline}, \WorldReasonBench{} evaluates reasoning with two complementary components. \emph{Process-aware Reasoning Verification} uses structured QA to check both outcome correctness and process faithfulness, while \emph{Multi-dimensional Quality Assessment} scores each video on reasoning quality, temporal consistency, and visual aesthetics. Together, they provide binary-verifiable diagnostic signals and continuous quality scores for ranking, reward-model training, and human-alignment analysis.

\subsubsection{Process-aware Reasoning Verification}
\label{subsubsec:reasoning_verification}

This component checks whether a generated video reaches the correct final state along a plausible world-state transition, using a two-stage structured QA protocol: a VLM answers each video-grounded question from visible evidence, then a separate LLM judge assigns a binary score against the ground truth.

\paragraph{Reasoning verification chain.}
Each test case has multiple QA pairs across four question types, which we map to complementary reasoning phases: \emph{factual} (initial or final state content), \emph{temporal} (event order), \emph{detail} (fine-grained visual fidelity), and \emph{reasoning} (causal or physical mechanisms). The corresponding phase scores $s_{\mathrm{state}}, s_{\mathrm{proc}}, s_{\mathrm{fidel}}, s_{\mathrm{mech}}$ are mean binary accuracies within each type, and overall accuracy is $\mathrm{Acc}_{\mathrm{QA}} = \frac{1}{N}\sum_{i=1}^{N}\mathbb{1}[\hat{y}_i = y_i^{\mathrm{gt}}]$.

\paragraph{Reasoning gap and process-aware score.}
To expose \emph{outcome hacking}---videos that look correct in static frames but fail dynamically---we contrast static outcome performance $s_{\mathrm{out}}=(s_{\mathrm{state}}+s_{\mathrm{fidel}})/2$ with dynamic performance $s_{\mathrm{dyn}}=(s_{\mathrm{proc}}+s_{\mathrm{mech}})/2$ and define the reasoning gap $\Delta_{\mathrm{RG}}=s_{\mathrm{out}}-s_{\mathrm{dyn}}$; a large positive $\Delta_{\mathrm{RG}}$ signals strong static appearance but weak process reasoning. For the headline metric we use $\mathrm{Score}_{\mathrm{PR}}=\mathrm{Acc}_{\mathrm{QA}}^{0.8}\cdot s_{\mathrm{dyn}}^{0.2}$, which keeps QA accuracy interpretable while discounting models that succeed mainly on static questions, and we use $s_{\mathrm{dyn}}/\mathrm{Acc}_{\mathrm{QA}}$ as a process-completeness diagnostic. Auxiliary metrics are in Appendix~\ref{app:aux-process-metrics}.

\subsubsection{Multi-dimensional Quality Assessment}
\label{subsubsec:quality}

Reward-model training, model ranking, and human-alignment analysis all need continuous calibrated per-video scores. Multi-dimensional Quality Assessment asks a VLM judge to rate each video on a 1--5 scale along three interpretable dimensions: \textbf{Reasoning Quality} ($s_r$, whether the intended world-state transition is realized), \textbf{Temporal Consistency} ($s_c$, coherence and stability across time), and \textbf{Visual Aesthetics} ($s_a$, frame stability, motion naturalness, composition, and overall appeal). The three are aggregated into $S(v)=0.4\,s_r(v)+0.3\,s_c(v)+0.3\,s_a(v)$, with the largest weight on reasoning quality to match both the benchmark's focus and the \WorldRewardBench{} annotation protocol (Section~\ref{subsec:reward_bench}) for direct human-vs-automatic comparability.

\paragraph{Evaluation protocols.}
We report two complementary protocols. In the \emph{point-wise} protocol, the judge scores each video independently and pairwise preferences are induced from $S(v_i)$ vs.\ $S(v_j)$ with a tie threshold of $0.1$, supporting reward-model training and score-based ranking. In the \emph{pair-wise} protocol, the judge compares two videos in a single call and emits A wins / B wins / tie, giving a stronger ordinal signal for preference recovery and judge calibration at the cost of per-video continuous scores.


\section{Experiments}
\label{sec:exp}

\begin{table*}[t]
  \caption{\textbf{Main evaluation results across \WorldReasonBench{} dimensions.} Per-dimension $\mathrm{Score}_{\mathrm{PR}}$ and $S(v)$ ($0$--$100$) computed for every generator on a shared evaluation set for fully controlled cross-model comparison. \textbf{Bold}/\underline{underline}: best/second-best across all $11$ models. Full subcategory results, $95\%$ bootstrap CIs, and additional open-source coverage are in Appendix Table~\ref{tab:main_results_full} and Appendix~\ref{app:bootstrap-ci}.}
  \label{tab:main_results}
  \centering
  \small
  \setlength{\tabcolsep}{3.5pt}
  \renewcommand{\arraystretch}{1.05}
  \resizebox{\textwidth}{!}{
  \begin{tabular}{@{}llccccccccccc@{}}
    \toprule
    \multicolumn{2}{c}{} & \multicolumn{5}{c}{\textbf{Closed-Source Models}} & \multicolumn{6}{c}{\textbf{Open-Source Models}} \\
    \cmidrule(lr){3-7} \cmidrule(lr){8-13}
    \textbf{Dimension} & \textbf{Metric} & \textbf{Sora2} & \textbf{Kling} & \textbf{Wan2.6} & \textbf{Seedance2.0} & \textbf{Veo3.1-Fast} & \textbf{LTX2.3} & \textbf{Wan2.2-14B} & \textbf{UniVideo} & \textbf{HunyuanVideo-1.5} & \textbf{Cosmos-Predict2.5} & \textbf{LongCat-Video} \\
    \midrule
    \multirow{2}{*}{World Knowledge}
      & $\mathrm{Score}_{\mathrm{PR}}$ & 36.9 & 42.2 & 35.2 & \underline{43.2} & \textbf{55.0} & 15.6 & 22.9 & 13.8 & 21.6 & 15.2 & 13.3 \\
      & $S(v)$ & 62.6 & \underline{72.0} & 61.8 & 70.4 & \textbf{80.1} & 35.1 & 39.4 & 29.4 & 37.7 & 40.8 & 35.1 \\
    \midrule
    \multirow{2}{*}{Human-Centric}
      & $\mathrm{Score}_{\mathrm{PR}}$ & \textbf{44.7} & 32.5 & 34.5 & \underline{35.9} & 35.1 & 19.3 & 14.5 & 15.8 & 8.1 & 22.2 & 22.8 \\
      & $S(v)$ & 76.7 & \textbf{87.2} & 64.2 & \underline{83.9} & 77.2 & 27.8 & 38.1 & 37.2 & 35.3 & 30.8 & 42.8 \\
    \midrule
    \multirow{2}{*}{Logic Reasoning}
      & $\mathrm{Score}_{\mathrm{PR}}$ & 25.9 & 22.4 & \underline{26.2} & \textbf{31.7} & 25.7 & 11.9 & 16.4 & 11.2 & 12.7 & 7.1 & 12.6 \\
      & $S(v)$ & \underline{43.0} & 37.3 & 42.3 & \textbf{56.7} & 31.5 & 24.7 & 19.5 & 14.4 & 19.8 & 26.7 & 16.3 \\
    \midrule
    \multirow{2}{*}{Information-Based}
      & $\mathrm{Score}_{\mathrm{PR}}$ & \underline{37.3} & 35.7 & 35.5 & \textbf{47.6} & 28.6 & 22.7 & 15.0 & 17.3 & 24.2 & 24.7 & 22.8 \\
      & $S(v)$ & \textbf{58.0} & \underline{48.8} & 42.6 & 42.5 & 47.2 & 25.8 & 30.5 & 16.0 & 22.5 & 26.1 & 20.2 \\
    \midrule
    \multirow{2}{*}{\textbf{Overall}}
      & $\mathrm{Score}_{\mathrm{PR}}$ & 34.3 & 32.7 & 32.4 & \textbf{39.8} & \underline{35.3} & 16.8 & 17.5 & 14.4 & 17.9 & 16.9 & 17.4 \\
      & $S(v)$ & \underline{56.9} & 55.4 & 50.3 & \textbf{59.4} & 54.8 & 28.1 & 30.0 & 21.3 & 27.0 & 30.5 & 25.3 \\
    \bottomrule
  \end{tabular}}
\end{table*}

\subsection{Experimental Setup}
\label{subsec:setup}

\paragraph{Evaluation settings.} We evaluate eleven video generators: five closed-source systems (Sora2, Kling, Wan2.6, Seedance2.0, Veo3.1-Fast) and six open-source models (LTX2.3, Wan2.2-14B, UniVideo, HunyuanVideo-1.5, Cosmos-Predict2.5, LongCat-Video). All automatic evaluation uses Qwen3.5-27B~\citep{team2026qwen3}; the QA pipeline enables extended thinking for video question answering, disables it for binary judging, and processes videos at 4~FPS.

\paragraph{Metrics.} We report $\mathrm{Score}_{\mathrm{PR}}$ as the headline metric for Process-aware Reasoning Verification, with $\mathrm{Acc}_{\mathrm{QA}}$, phase scores, process completeness, and $\Delta_{\mathrm{RG}}$ as diagnostics. Multi-dimensional Quality Assessment reports the weighted per-video score $S(v)$ over reasoning quality, temporal consistency, and visual aesthetics, and uses pairwise agreement and Spearman~$\rho$ for reward-model alignment. Auxiliary process-aware metrics are defined in Appendix~\ref{app:aux-process-metrics}.

\paragraph{Reward-model baselines.} On \WorldRewardBench{}, we evaluate five reward/judge models (GPT-5.4, Gemini-3.1, Qwen3.5-9B, Qwen3.5-27B, and our method) under both pair-wise and point-wise protocols to measure recovery of human video preferences.

\subsection{Generator Performance on \WorldReasonBench{}}
\label{subsec:main_results}

\paragraph{Closed-source models lead by a robust factor on both reasoning and quality.}
Under controlled cross-model comparison (Table~\ref{tab:main_results}), closed-source generators sit at $32.4$--$39.8$ overall $\mathrm{Score}_{\mathrm{PR}}$ and $50.3$--$59.4$ on $S(v)$, while open-source generators stay at $14.4$--$17.9$ and $21.3$--$30.5$, respectively---a roughly two-fold gap on both axes, with no open-source $95\%$ CI overlapping any closed-source one. Even the strongest system (Seedance2.0, $\mathrm{Score}_{\mathrm{PR}}{=}39.8$) sits well below saturation, so today's most capable generators remain incomplete world models. The gap is not driven by raw visual fidelity: the process-completeness ratio $s_{\mathrm{dyn}}/\mathrm{Acc}_{\mathrm{QA}}$ in Section~\ref{subsec:human_alignment} shows that open-source failures concentrate on dynamic-phase reasoning rather than static appearance.

\paragraph{Difficulty is dominated by Logic Reasoning and Information-Based categories.}
Performance is highly uneven across dimensions. Logic Reasoning is the hardest: the best closed-source $\mathrm{Score}_{\mathrm{PR}}$ is only $31.7$ (Seedance2.0), and five of the six open-source models score below $14$. Information-Based is second hardest, with per-subcategory residuals (Appendix Table~\ref{tab:main_results_full}) concentrating in \emph{World Mechanics}, \emph{Material Change}, and \emph{Data Reading}---categories needing physically-grounded transitions or exact text/data preservation. World Knowledge and Human-Centric exceed $35$ for every closed-source model and reach $55.0$ (Veo3.1-Fast on WK) and $44.7$ (Sora2 on HC), so the bottleneck is mechanism- and information-level reasoning rather than visual recognition.

\begin{wraptable}{r}{0.5\textwidth}
  \caption{\textbf{Reasoning assistance benefit.} QA accuracy under implicit (Diff.) vs.\ hinted (Easy) prompts; hint gain is absolute / relative.}
  \label{tab:easy_vs_difficult}
  \centering
  \scriptsize
  \renewcommand{\arraystretch}{1.08}
  \setlength{\tabcolsep}{2.5pt}
  \begin{tabular}{@{}lccc@{}}
    \toprule
    \textbf{Model} & \textbf{Diff.} & \textbf{Easy} & \textbf{Hint gain} \\
    \midrule
    Sora2-8s & 35.1 & 45.4 & +10.3 / +29.2\% \\
    \midrule
    LTX2.3 & 17.5 & 32.3 & +14.8 / +84.9\% \\
    Wan2.2-14B & 21.6 & 35.2 & +13.6 / +63.2\% \\
    UniVideo & 17.8 & 27.6 & +9.9 / +55.5\% \\
    HunyuanVideo-1.5 & 20.8 & 33.0 & +12.2 / +58.4\% \\
    Cosmos-Predict2.5 & 19.4 & 30.8 & +11.4 / +58.9\% \\
    \bottomrule
  \end{tabular}
  \vspace{-0.8em}
\end{wraptable}

\paragraph{Hint gain is larger for open-source models.}
With explicit transition hints, every open-source model gains $9.9$--$14.8$ absolute QA points ($+56$--$85$\% relative), whereas Sora2-8s---the only closed-source system run under both regimes---gains only $+10.3$ points ($+29$\%) (Table~\ref{tab:easy_vs_difficult}). This indicates open-source generators rely more on prompt-side guidance, though ceiling effects, prompt-length sensitivity, and instruction-following gaps may also contribute; the substantive outcome-vs-process attribution is carried by $\mathrm{Score}_{\mathrm{PR}}$ and $s_{\mathrm{dyn}}/\mathrm{Acc}_{\mathrm{QA}}$ in Section~\ref{subsec:human_alignment}.

\paragraph{Statistical significance and rank stability.}
We compute $95\%$ bootstrap confidence intervals ($B{=}2000$, case-level resampling with replacement) for $\mathrm{Score}_{\mathrm{PR}}$, $\mathrm{Acc}_{\mathrm{QA}}$, and $S(v)$ at overall and per-dimension level on the shared evaluation set behind Table~\ref{tab:main_results}. The closed-vs.-open separation is statistically robust: every open-source overall-$\mathrm{Score}_{\mathrm{PR}}$ CI lies strictly below every closed-source CI (open-source upper bound $\leq 23.1$ vs.\ closed-source lower bound $\geq 26.4$). Joint rank bootstrap shows that the two tiers \emph{never} swap, and Seedance2.0 has a clearly favoured rank inside the closed tier (modal rank~$1$ in $89.3\%$ of bootstraps, $95\%$ rank interval $[1,2]$); the other five closed-source models share rank slots with overlapping CIs, so we report their cluster rather than a strict ordering. Within open-source, UniVideo is the only generator with a tightly concentrated rank (modal rank~$12$ in $69.7\%$); the remaining five sit in slots $[7,11]$ as a tied cluster. Full per-model CIs, per-dimension CIs, and the rank-distribution table are reported in Appendix~\ref{app:bootstrap-ci}.

\subsection{Validating Process-aware Metrics against Human Preferences}
\label{subsec:human_alignment}

\begin{table}[!htp]
  \caption{\textbf{Human-aligned ranking and metric validation.} Models ordered by Human Elo; $|\Delta r|$ is the absolute rank displacement from the human ranking. Horizontal line separates closed- and open-source models.}
  \label{tab:elo_ranking}
  \centering
  \small
  \renewcommand{\arraystretch}{1.05}
  \setlength{\tabcolsep}{3pt}
  \begin{tabular}{clcccccccc}
    \toprule
    \begin{tabular}[c]{@{}c@{}}\textbf{Human}\\\textbf{Rank}\end{tabular}
    & \textbf{Model}
    & \begin{tabular}[c]{@{}c@{}}\textbf{Human}\\\textbf{Elo}\end{tabular}
    & \begin{tabular}[c]{@{}c@{}}\textbf{Judge}\\\textbf{Elo}\end{tabular}
    & \begin{tabular}[c]{@{}c@{}}\textbf{Judge}\\\textbf{Rank}\end{tabular}
    & \begin{tabular}[c]{@{}c@{}}$\mathrm{Acc}_{\mathrm{QA}}$\\(\%)\end{tabular}
    & $|\Delta r|$
    & \begin{tabular}[c]{@{}c@{}}$\mathrm{Score}_{\mathrm{PR}}$\\(\%)\end{tabular}
    & $|\Delta r|$
    & \begin{tabular}[c]{@{}c@{}}$s_\mathrm{dyn}/$\\$\mathrm{Acc}_\mathrm{QA}$\end{tabular} \\
    \midrule
    1 & Seedance2.0       & \textbf{1471} & 1183 & 3 & \textbf{41.2} & 0 & \textbf{39.8} & 0 & 0.84 \\
    2 & Veo3.1-Fast       & 1253 & 1151 & 4 & 36.0 & 0 & 35.3 & 0 & 0.91 \\
    3 & Kling             & 1240 & 1142 & 5 & 34.0 & 2 & 32.7 & 1 & 0.82 \\
    4 & Wan2.6            & 1211 & 1130 & 6 & 34.7 & 0 & 32.4 & 1 & 0.71 \\
    5 & Sora2-8s          & 1118 & \textbf{1222} & 1 & 35.3 & 2 & 34.3 & 2 & 0.86 \\
    6 & Sora2-12s         & 1109 & 1217 & 2 & 33.5 & 0 & 32.4 & 0 & 0.84 \\
    \midrule
    7 & Wan2.2-14B        & 953  & 913 & 7 & 19.6 & 2 & 17.5 & 1 & 0.57 \\
    8 & HunyuanVideo-1.5  & 911  & 841 & 9 & 20.2 & 1 & 17.9 & 1 & 0.56 \\
    9 & LongCat-Video     & 904  & 876 & 8 & 19.7 & 1 & 17.4 & 0 & 0.54 \\
    10 & UniVideo          & 665  & 737 & 11 & 16.2 & 1 & 14.4 & 1 & 0.56 \\
    11 & LTX2.3            & 587  & 802 & 10 & 18.5 & 1 & 16.8 & 1 & 0.63 \\
    \bottomrule
  \end{tabular}
\end{table}

\paragraph{Process-aware QA metrics outperform pairwise VLM judges on human alignment.}
Using ${\sim}6$K expert preference pairs from \WorldRewardBench{}, we fit a Bradley-Terry model with Davidson ties (Appendix~\ref{app:elo-details}) for a Human Elo ranking and compare it with three automatic rankings (Table~\ref{tab:elo_ranking}). $\mathrm{Score}_{\mathrm{PR}}$ and $\mathrm{Acc}_{\mathrm{QA}}$ reach Spearman $\rho{=}0.955$ and $0.927$, both well above the pairwise VLM-judge Elo ($\rho{=}0.804$). The process-completeness ratio $s_\mathrm{dyn}/\mathrm{Acc}_{\mathrm{QA}}$ stays at $0.71$--$0.91$ for closed-source vs.\ $0.54$--$0.63$ for open-source, attributing the open-source deficit to dynamic reasoning rather than static-frame errors.

\begin{figure}[t]
  \centering
  \includegraphics[width=0.95\linewidth]{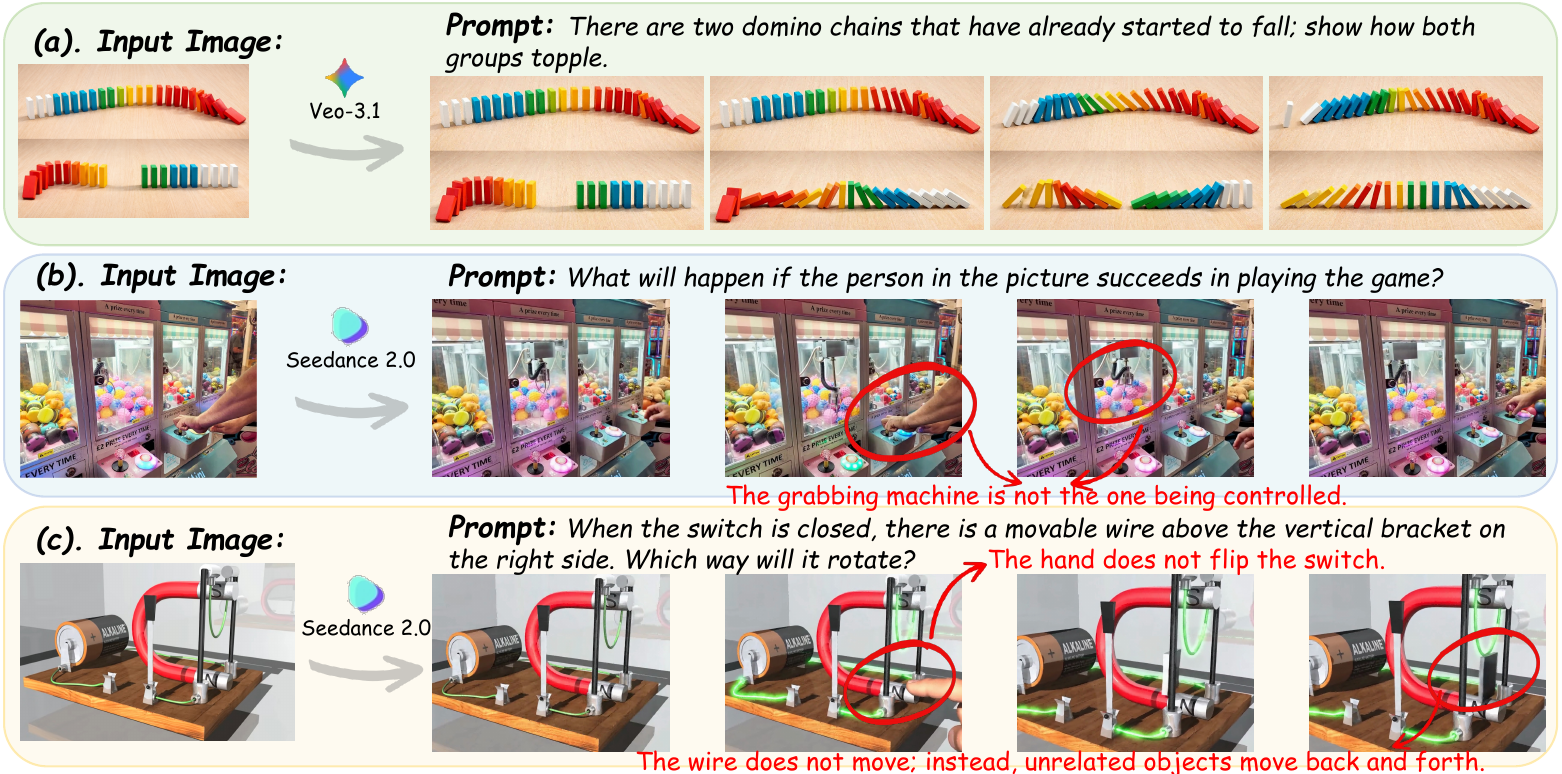}
  \caption{\textbf{Qualitative comparison on representative reasoning cases.} Visually plausible generations can still fail process-level world reasoning, while higher-scoring models better preserve the intended state transition and temporal dynamics.}
  \label{fig:qualitative_comparison}
\end{figure}

\paragraph{Diagnosing the residual judge--human disagreement.}
The largest remaining inconsistency in Table~\ref{tab:elo_ranking} is the closed-source ordering: humans place Seedance2.0 first, but the pairwise judge places Sora2-8s and Sora2-12s on top. We trace this to two pairwise-protocol effects. (i) The judge consumes a fixed budget of 8 frames per video, so 8s/12s Sora2 clips expose more events at lower temporal density and the judge often reads this as richer reasoning evidence; Figure~\ref{fig:qualitative_comparison} shows cases where Seedance2.0 instead produces smoother, more physically faithful motion that humans reward but the fixed-frame judge misses. (ii) Judge accuracy drops sharply on close pairs ($89.1\%$ when the human gap is $>1.5$, $47.5\%$ for $\leq 0.5$), and such close pairs disproportionately involve Seedance2.0 against the Sora2 family, suppressing Seedance2.0's Elo. $\mathrm{Score}_{\mathrm{PR}}$ avoids this duration mismatch and matches the human ordering up to a single one-rank swap.

\subsection{\WorldRewardBench{}: VLM Judges as Reward Models}
\label{subsec:reward_bench_results}

\begin{table*}[!htp]
  \caption{\textbf{Reward-model alignment on \WorldRewardBench{}.} Pair-wise: agreement (\%) \emph{w/ Ties} / \emph{w/o Ties}; point-wise: induced pairwise accuracy / Spearman $\rho$. GPT-5.4 pair-wise uses $5{,}947/5{,}969$ pairs after Azure OpenAI safety filtering ($0.37\%$ refused; details in Appendix~\ref{app:reward-bench-fullset}).}
  \label{tab:reward_alignment}
  \centering
  \small
  \renewcommand{\arraystretch}{1.08}
  \resizebox{\textwidth}{!}{
  \begin{tabular}{ll|cc|cccc}
    \toprule
    \multicolumn{2}{c|}{} & \multicolumn{2}{c|}{\textbf{Closed-Source Models}} & \multicolumn{4}{c}{\textbf{Open-Source Models}} \\
    \cmidrule(lr){3-4} \cmidrule(lr){5-8}
    \textbf{Dimension} & \textbf{Protocol / Metric} & \textbf{GPT-5.4} & \textbf{Gemini-3.1-Flash} & \begin{tabular}[c]{@{}c@{}}\textbf{Qwen3.5-9B}\\\textbf{Thinking}\end{tabular} & \begin{tabular}[c]{@{}c@{}}\textbf{Qwen3.5-27B}\\\textbf{Instruct}\end{tabular} & \begin{tabular}[c]{@{}c@{}}\textbf{Qwen3.5-27B}\\\textbf{Thinking}\end{tabular} & \begin{tabular}[c]{@{}c@{}}\textbf{Qwen3.5-27B}\\\textbf{Thinking (4\,FPS)}\end{tabular} \\
    \midrule
    \multicolumn{2}{l|}{\textbf{Frames Used}} & 8 & 1FPS & ${\sim}$10 & ${\sim}$10 & ${\sim}$10 & 4\,FPS \\
    \midrule
    \multirow{2}{*}{World Knowledge}
      & Pair w/ / w/o & 60.77 / 67.84 & 51.50 / 60.44 & 70.81 / 76.19 & 69.37 / 74.16 & 69.94 / 74.64 & 69.51 / 74.23 \\
      & Point Acc / $\rho$ & 54.55 / 0.592 & 59.86 / 0.582 & 60.70 / 0.720 & 54.01 / 0.658 & 60.57 / 0.687 & 62.09 / 0.711 \\
    \cmidrule(lr){1-8}
    \multirow{2}{*}{Human-Centric}
      & Pair w/ / w/o & 68.37 / 76.80 & 58.22 / 66.27 & 71.71 / 77.52 & 71.25 / 76.05 & 72.61 / 77.81 & 69.08 / 74.41 \\
      & Point Acc / $\rho$ & 59.14 / 0.626 & 60.06 / 0.675 & 59.54 / 0.702 & 55.94 / 0.682 & 62.81 / 0.713 & 60.49 / 0.703 \\
    \cmidrule(lr){1-8}
    \multirow{2}{*}{Logic Reasoning}
      & Pair w/ / w/o & 67.41 / 78.43 & 58.23 / 67.68 & 69.33 / 77.13 & 68.46 / 74.51 & 70.16 / 76.23 & 68.53 / 74.97 \\
      & Point Acc / $\rho$ & 53.42 / 0.523 & 57.65 / 0.562 & 57.50 / 0.617 & 55.71 / 0.573 & 60.17 / 0.606 & 58.40 / 0.597 \\
    \cmidrule(lr){1-8}
    \multirow{2}{*}{Information-Based}
      & Pair w/ / w/o & 56.95 / 63.68 & 50.21 / 58.10 & 52.45 / 61.76 & 60.44 / 65.22 & 60.24 / 65.32 & 61.50 / 66.39 \\
      & Point Acc / $\rho$ & 48.15 / \underline{0.484} & 47.89 / 0.432 & 53.59 / 0.471 & 47.95 / 0.408 & 50.15 / 0.445 & 52.41 / \textbf{0.526} \\
    \midrule
    \multirow{2}{*}{\textbf{Overall}}
      & Pair w/ / w/o & 63.04 / 71.36 & 54.39 / 62.99 & \underline{67.14} / \textbf{74.35} & 66.89 / 72.07 & \textbf{67.74} / \underline{73.05} & 66.90 / 72.30 \\
      & Point Acc / $\rho$ & 53.43 / 0.565 & 55.84 / 0.568 & 57.76 / \textbf{0.655} & 53.15 / 0.591 & \textbf{57.85} / 0.626 & \underline{57.83} / \underline{0.644} \\
    \bottomrule
  \end{tabular}}
\end{table*}

We evaluate whether the Multi-dimensional Quality Assessment protocol can also serve as an automatic reward model (Table~\ref{tab:reward_alignment}). Pair-wise judging directly compares two candidate videos; point-wise scoring induces preferences from the aggregate $S(v)$. Subcategory-level results, model settings, and parsing statistics are in Appendices~\ref{app:reward-bench-subcategory}--\ref{app:reward-bench-fullset}.

\paragraph{Pair-wise wins on agreement; point-wise wins on calibration.}
The strongest pair-wise judge is Qwen3.5-9B-Thinking ($74.35\%$ w/o ties), with Qwen3.5-27B-Thinking close behind ($73.05\%$) and both ahead of every point-wise variant; Qwen3.5-9B-Thinking also has the top point-wise $\rho{=}0.655$ (27B-Thinking $0.626$). The two protocols are therefore complementary: pair-wise is preferable for selecting the better video among close candidates, point-wise gives calibrated per-video signals suitable for reward-model training. Gemini-3.1 lags pair-wise by ${>}10$pp despite competitive point-wise scores, so explicit comparison at the prompt level matters as much as raw judging capacity. The Information-Based bottleneck transfers from generators to judges: pair-wise agreement drops from $74$--$78\%$ on the other dimensions to $58$--$65\%$, and point-wise $\rho$ from $0.6$--$0.7$ to $0.4$--$0.5$, making Information-Based the most discriminative dimension for future reward models.

\subsection{Ablation Studies}
\label{subsec:ablation}

\paragraph{Point-wise protocol and frame rate.}
Vanilla single-call point-wise scoring is both more efficient and at least as effective as Sequential Dimension Evaluation (SDE), reaching the best $\rho{=}0.626$ and $67.63\%$ w/o-ties accuracy with one judge call versus three for SDE. The frame-rate ablation in Appendix Tables~\ref{tab:fps_ablation}--\ref{tab:fps_ablation_full} shows $4$~FPS gives the best cost--accuracy trade-off ($37.2\%$ vs.\ $37.6\%$ at $8$~FPS, with ${\sim}9$k vs.\ ${\sim}12$k visual tokens per $5$s video; $34.9\%$ at $2$~FPS). We therefore default to vanilla point-wise scoring at $4$~FPS; full tables and halo-effect analysis are in Appendix~\ref{app:pointwise-ablation}.

\paragraph{Weight design and sensitivity.}
Since $s_{\mathrm{proc}}$ and $s_{\mathrm{mech}}$ already enter $\mathrm{Acc}_{\mathrm{QA}}$ as one quarter each, the $s_{\mathrm{dyn}}^{0.2}$ term in $\mathrm{Score}_{\mathrm{PR}}=\mathrm{Acc}_{\mathrm{QA}}^{0.8}\cdot s_{\mathrm{dyn}}^{0.2}$ acts as a second-order penalty on outcome-hacking rather than a substitute for outcome accuracy. Re-ranking the eleven \WorldRewardBench{} models in Table~\ref{tab:elo_ranking} under $\alpha\in\{0,0.2,0.5,0.7,0.8,0.9,1\}$, arithmetic / geometric / $\min$ aggregators, and a $231$-point simplex grid over $(w_r,w_c,w_a)$ keeps Spearman $\rho$ vs.\ human Elo in $[0.83,0.96]$ for $\mathrm{Score}_{\mathrm{PR}}$ and $\rho\geq 0.95$ on $67.5\%$ of the $S(v)$ simplex; the paper exponent $\alpha{=}0.8$ in fact attains the \emph{highest} $\rho{=}0.955$, so the chosen weights are an empirical optimum rather than a compromise. Full grids in Appendix~\ref{app:weight-sensitivity}.

\paragraph{Cross-family judge robustness.}
We cross-compare three judge families on the same \WorldRewardBench{} pairs (Table~\ref{tab:reward_alignment}). Within Qwen3.5, scaling 9B$\to$27B and toggling extended thinking moves overall pair-wise w/o-ties agreement by at most $2.3$pp ($72.07$--$74.35\%$) and point-wise $\rho$ by at most $0.064$ ($0.591$--$0.655$). Across families, Gemini-3.1-Flash trails Qwen on pair-wise agreement by ${\sim}10$pp while tracking it point-wise ($\rho{=}0.568$); GPT-5.4 sits between, with $71.36\%$ pair-wise agreement and $\rho{=}0.565$ matching Gemini. All three families flag \emph{Information-Based} as the hardest category and recover the same closed-vs-open ordering as Table~\ref{tab:elo_ranking}, so the reasoning gap and Information-Based bottleneck are not artefacts of a single judge family.


\section{Conclusion}
\label{sec:conclusion}

We introduced \WorldReasonBench{}, a world-state prediction benchmark with 436 cases and structured QA annotations spanning four reasoning dimensions and 22 subcategories, together with \WorldRewardBench{}, a preference benchmark with approximately 6K expert-annotated pairs over 1.4K videos from 11 generators. Building on this data, we proposed a two-part evaluation methodology---Process-aware Reasoning Verification and Multi-dimensional Quality Assessment---and validated it directly against human Elo, where $\mathrm{Score}_{\mathrm{PR}}$ reaches Spearman $\rho{=}0.955$ and clearly outperforms a pairwise VLM judge. The results expose a persistent gap between visual plausibility and world reasoning: closed- and open-source generators differ by roughly a factor of two on both reasoning and quality, and the process-completeness ratio $s_{\mathrm{dyn}}/\mathrm{Acc}_{\mathrm{QA}}$ attributes the open-source deficit to dynamic-phase failures rather than static appearance. Logic Reasoning and Information-Based content remain the most challenging dimensions for both generators and judges, suggesting that progress on world-aware video generation will be driven less by visual polish and more by mechanism-level reasoning and information preservation. We release \WorldReasonBench{}, \WorldRewardBench{}, and the evaluation toolkit so that the community can audit reward models, calibrate new judges, and extend the reasoning taxonomy as video generators continue to evolve.

\bibliographystyle{plainnat}
\bibliography{references}

\newpage

\appendix
\section{Representative Examples of WorldReason-Bench}
\label{app:representative-examples}

To provide a more intuitive understanding of the data distribution and reasoning requirements in \WorldReasonBench{}, we present representative examples from each major category in this appendix. These examples cover the four reasoning categories in our benchmark: World Knowledge, Human-Centric, Logic Reasoning, and Information-Based Reasoning. Each example consists of an input image and a generation prompt.

\begin{table*}[htp]
\centering
\begin{minipage}{\textwidth}
\begin{QAbox}{World Knowledge}

\begin{tabularx}{\textwidth}{@{}XXXX@{}}
\CasePromptFromJSON{1-a-4} &
\CasePromptFromJSON{1-a-19} &
\CasePromptFromJSON{1-a-31} &
\CasePromptFromJSON{1-a-64} \\[0.01em]
\CaseImage{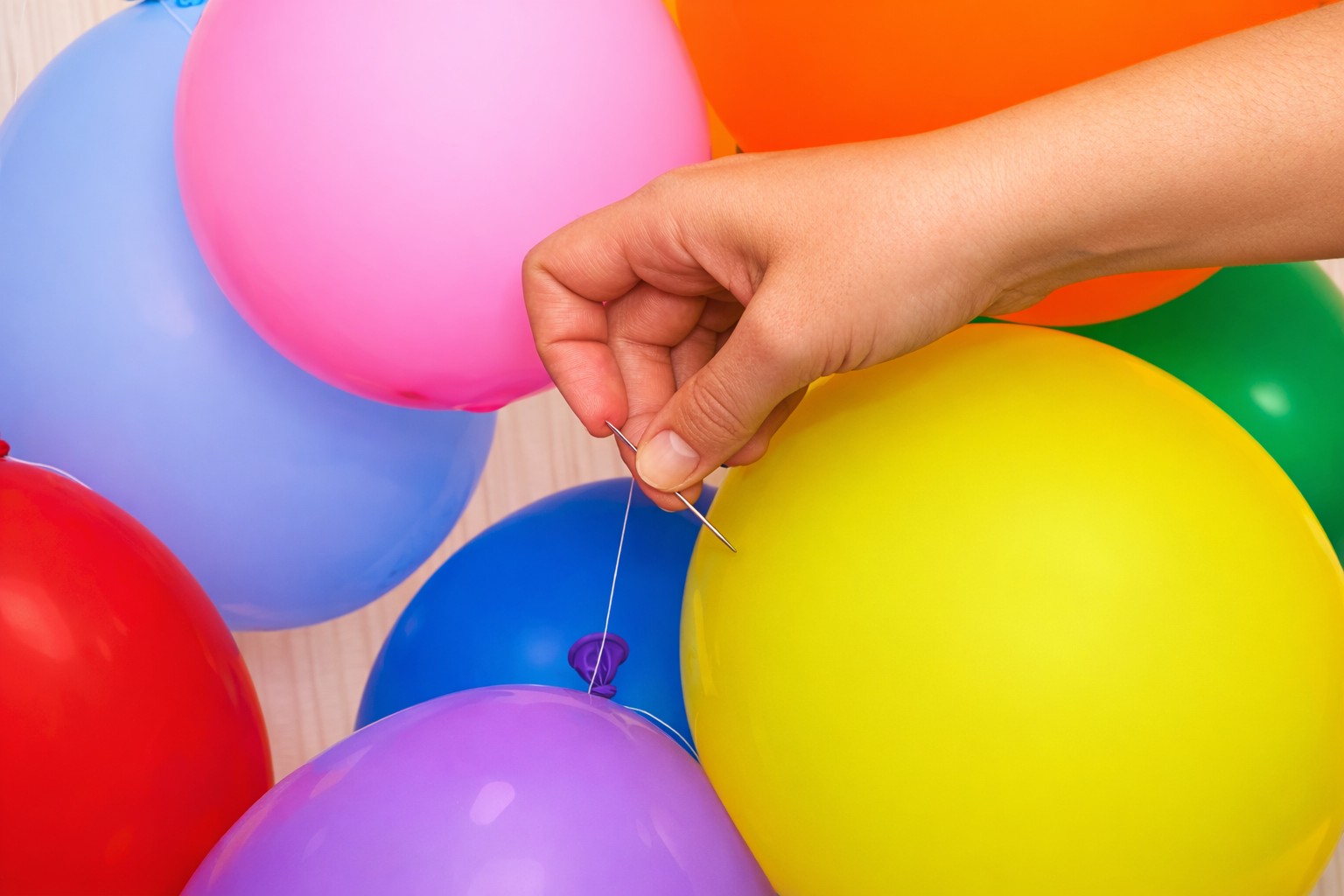} &
\CaseImage{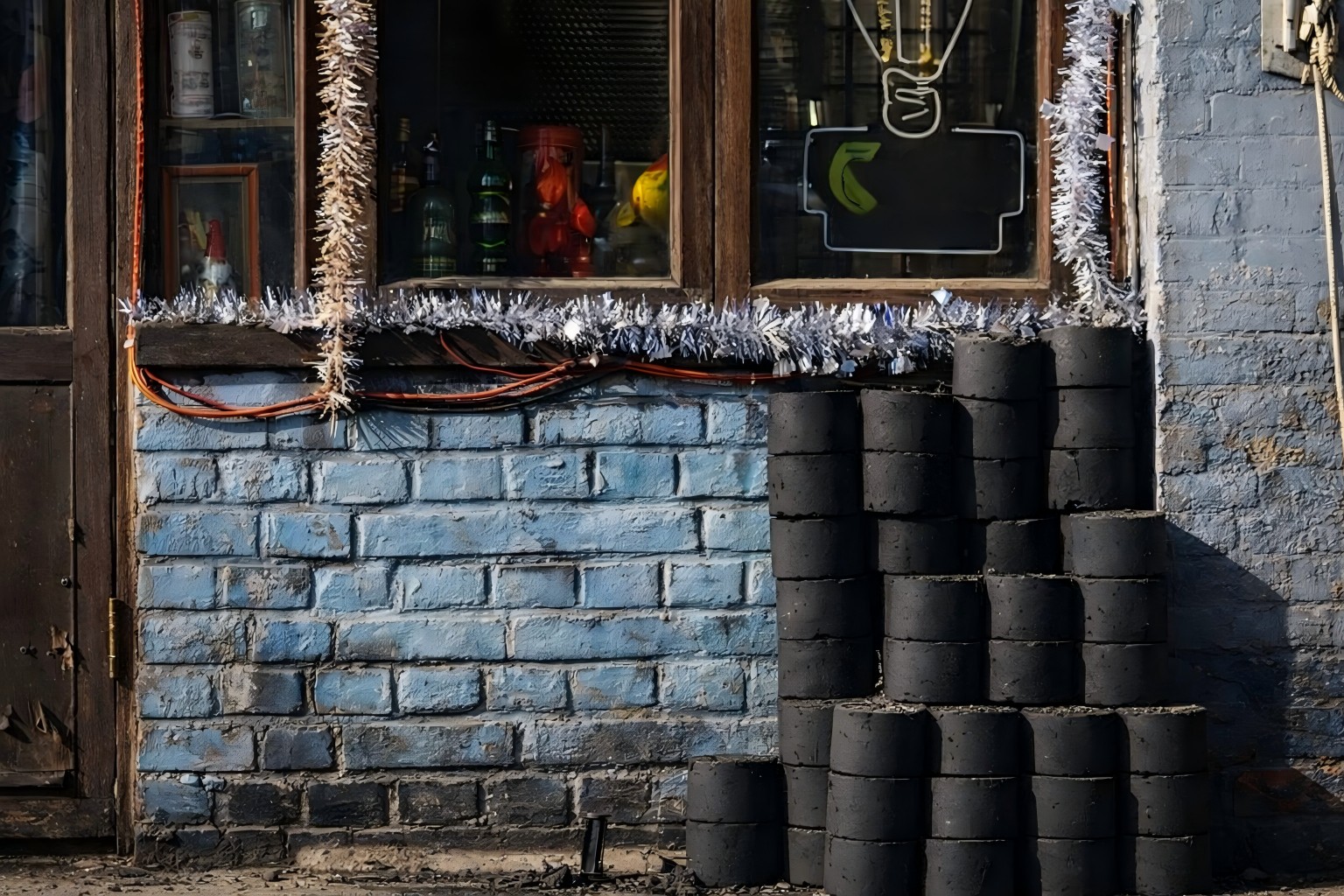} &
\CaseImage{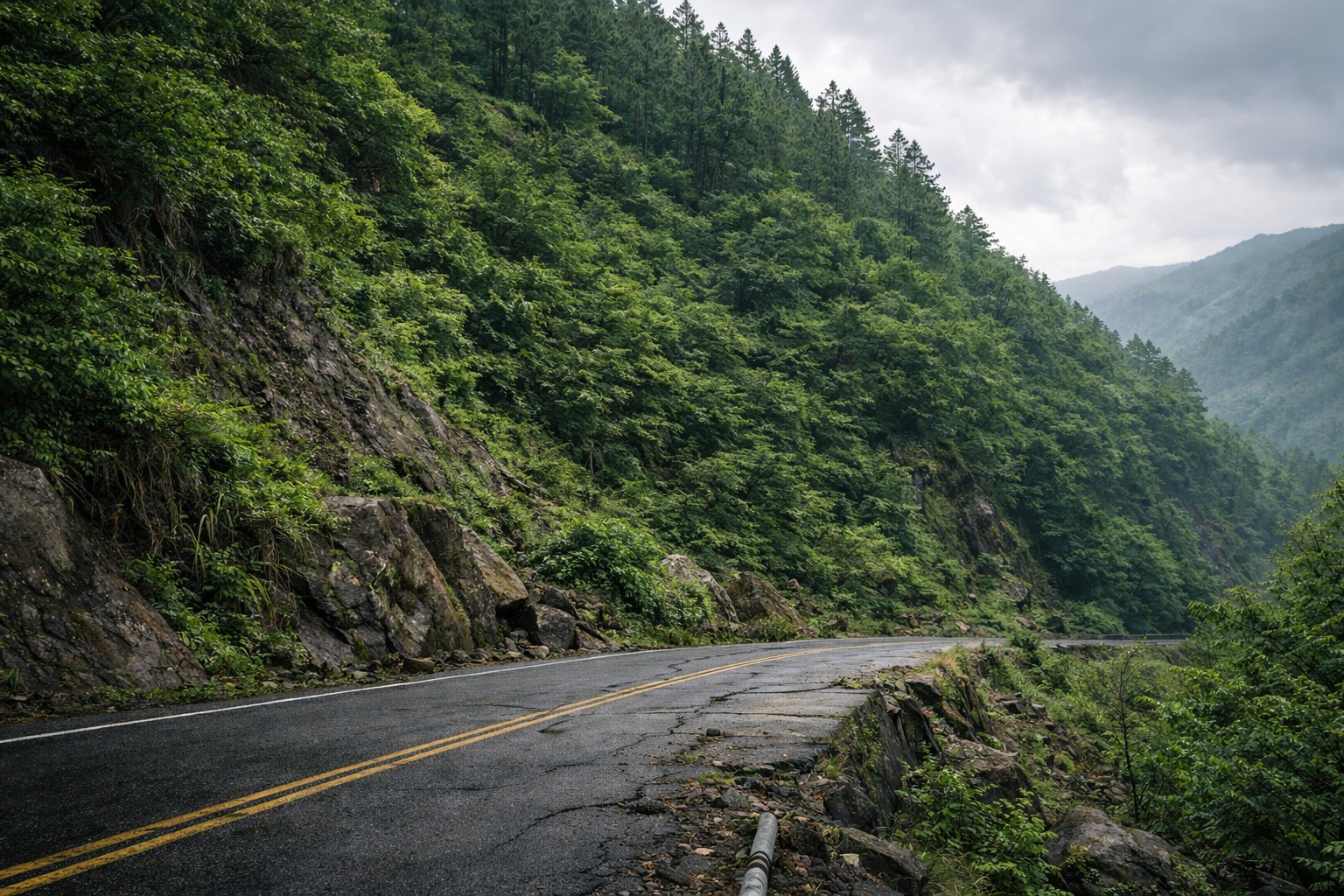} &
\CaseImage{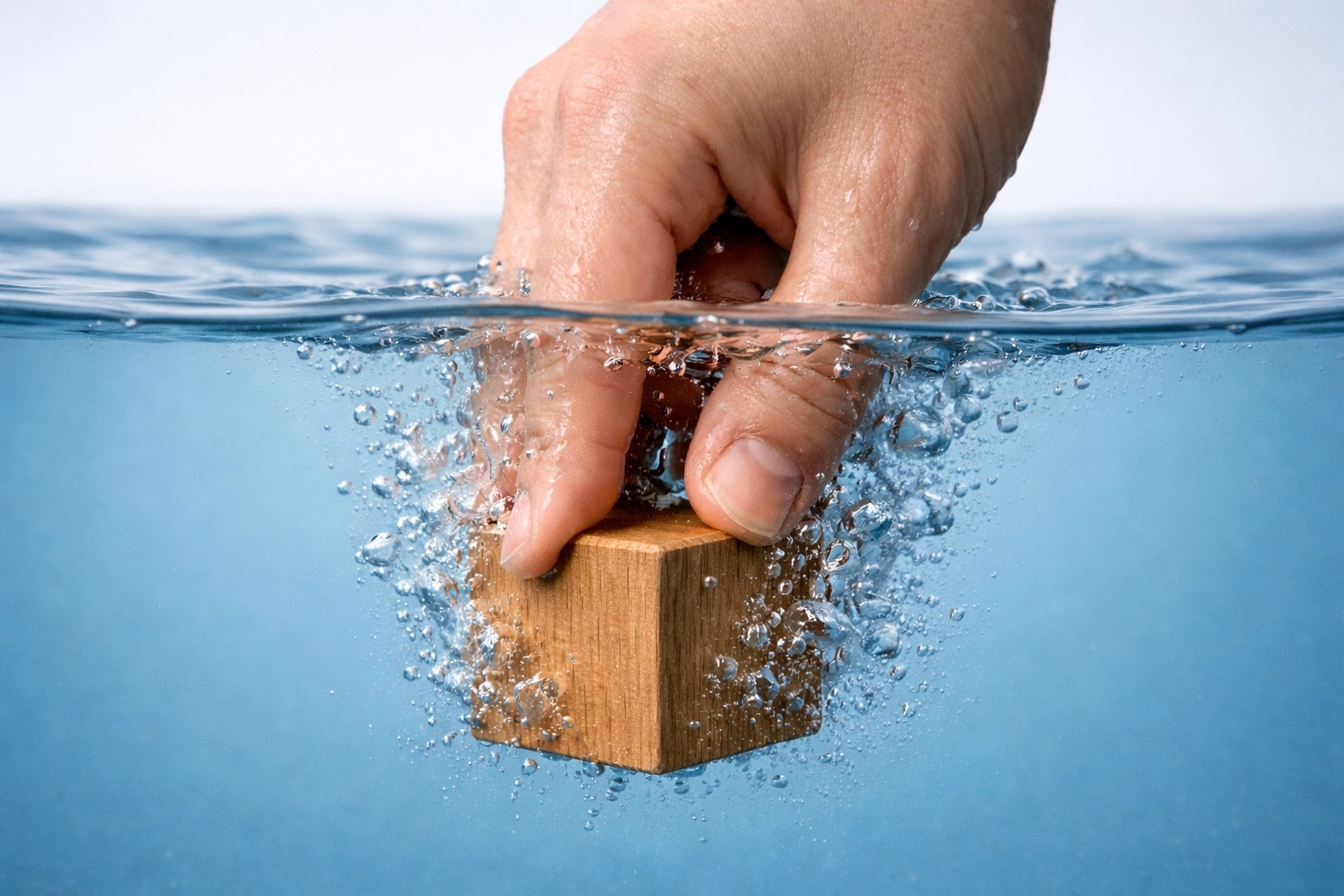} \\[0.0em]

\CasePromptFromJSON{1-a-34} &
\CasePromptFromJSON{1-a-47} &
\CasePromptFromJSON{1-a-45} &
\CasePromptFromJSON{1-a-24} \\[0.01em]
\CaseImage{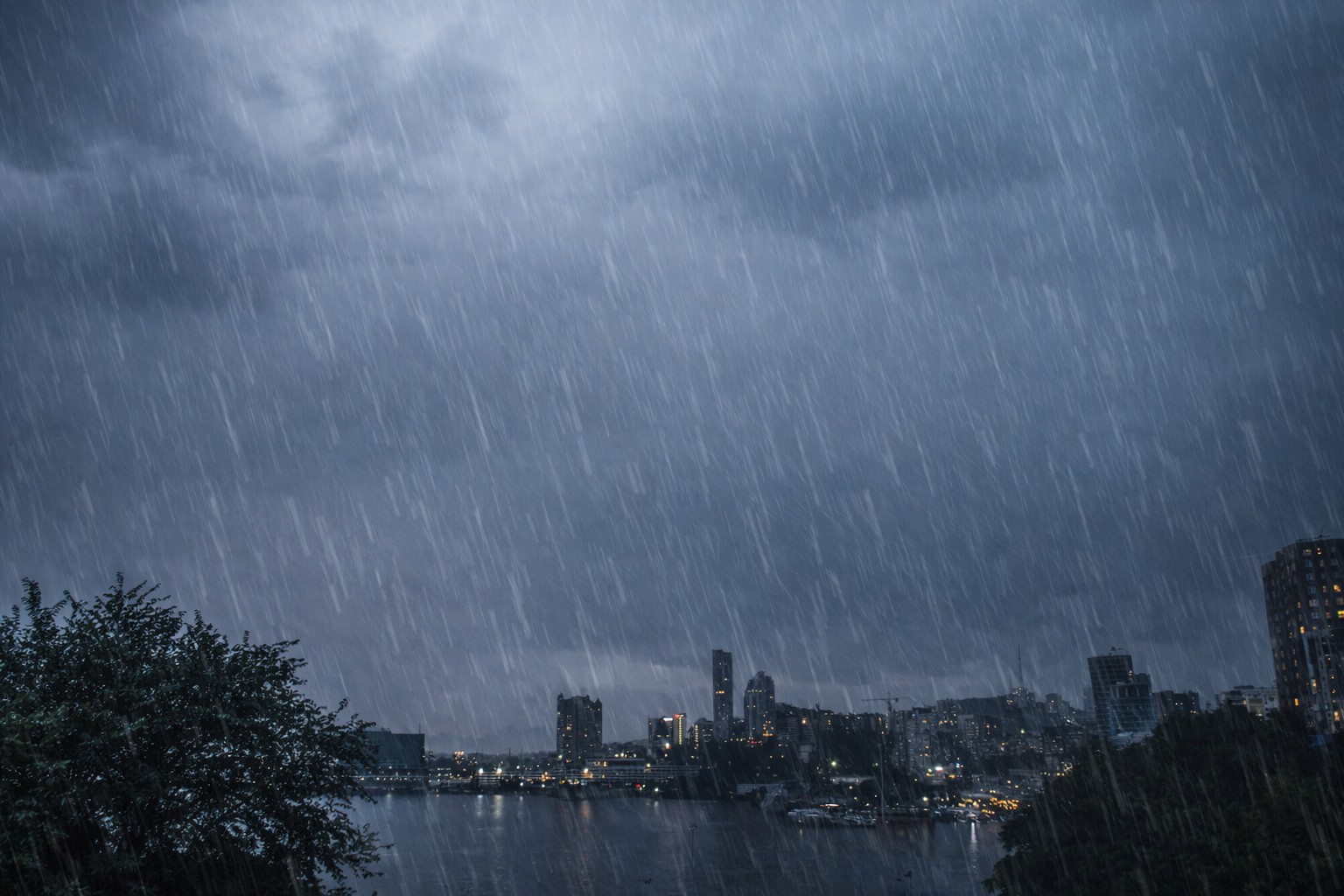} &
\CaseImage{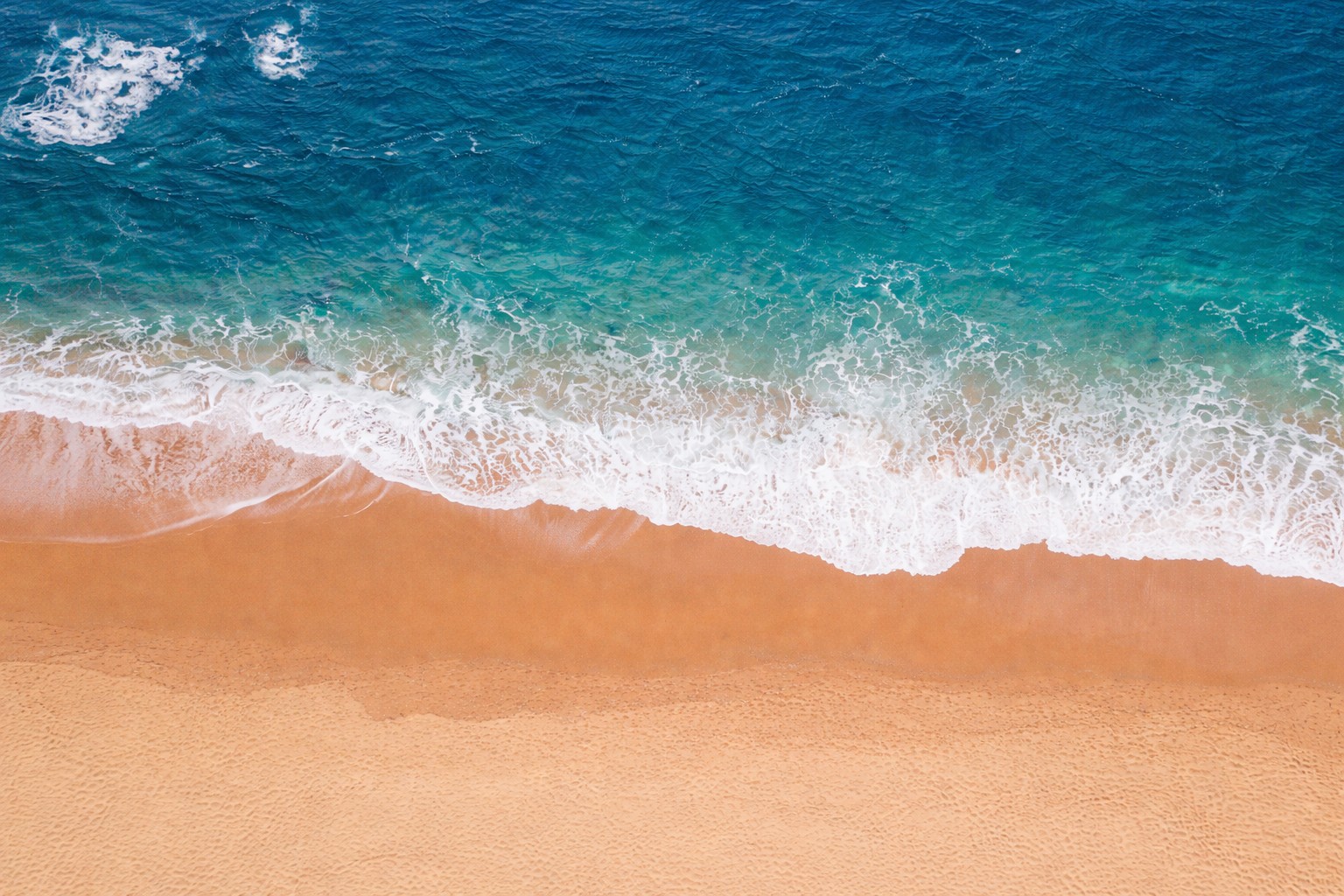} &
\CaseImage{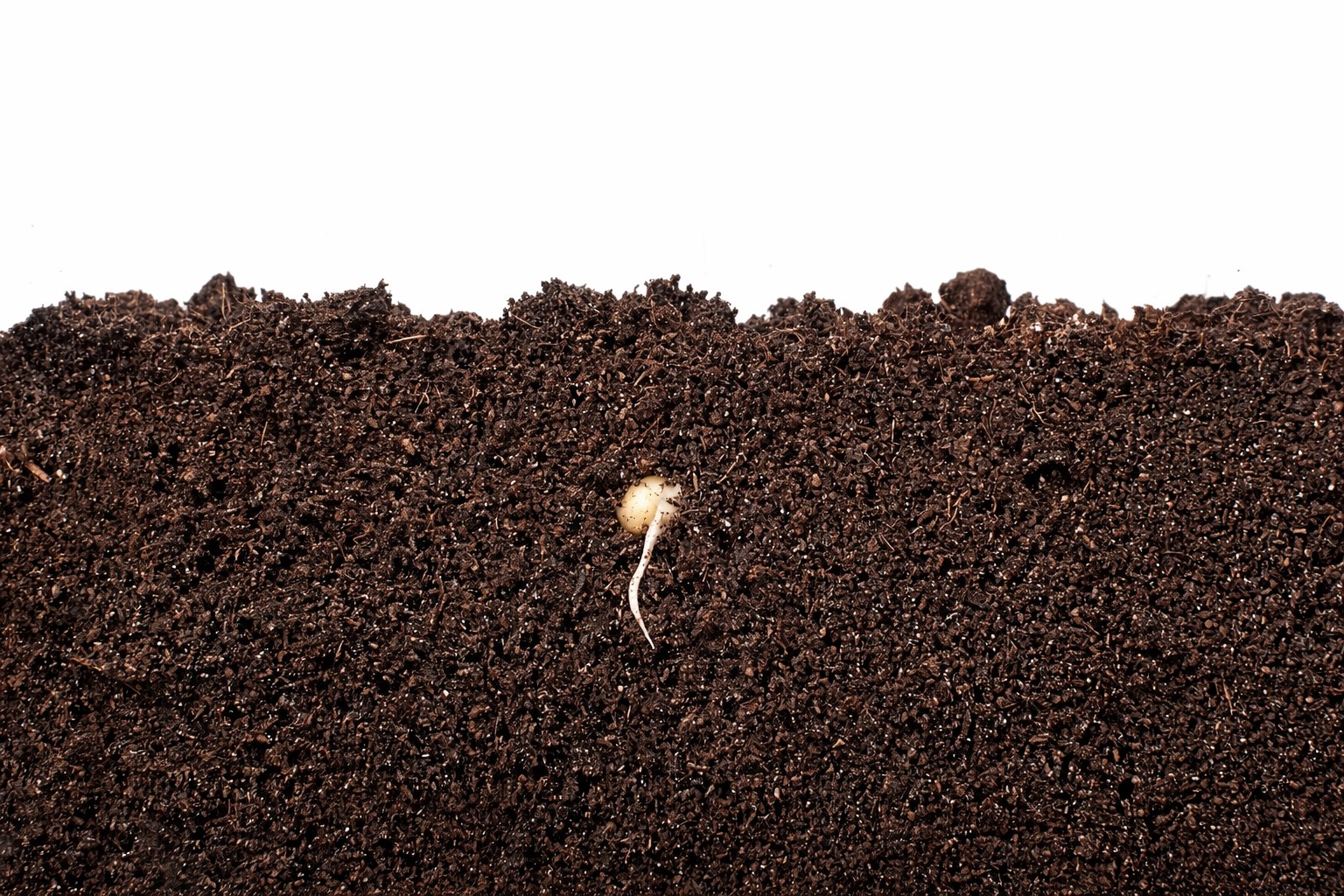} &
\CaseImage{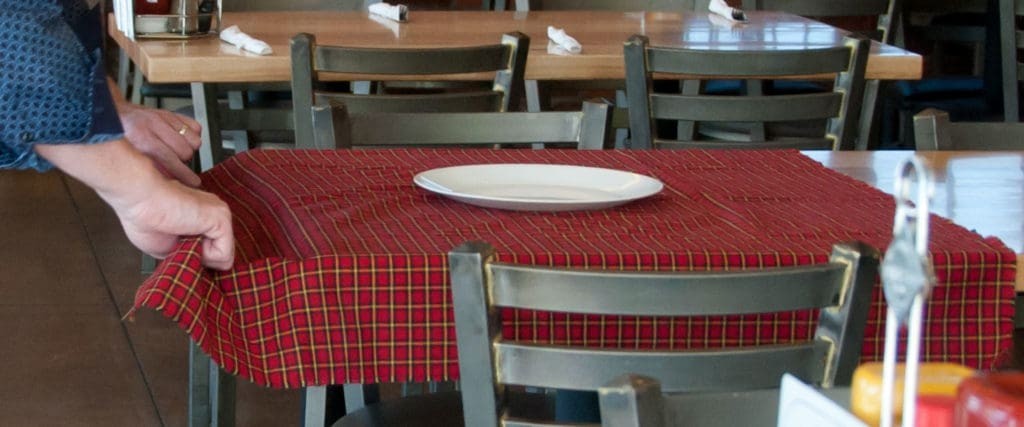} \\[0.0em]

\CasePromptFromJSON{1-b-02} &
\CasePromptFromJSON{1-b-05} &
\CasePromptFromJSON{1-a-32} &
\CasePromptFromJSON{1-c-10} \\[0.01em]
\CaseImage{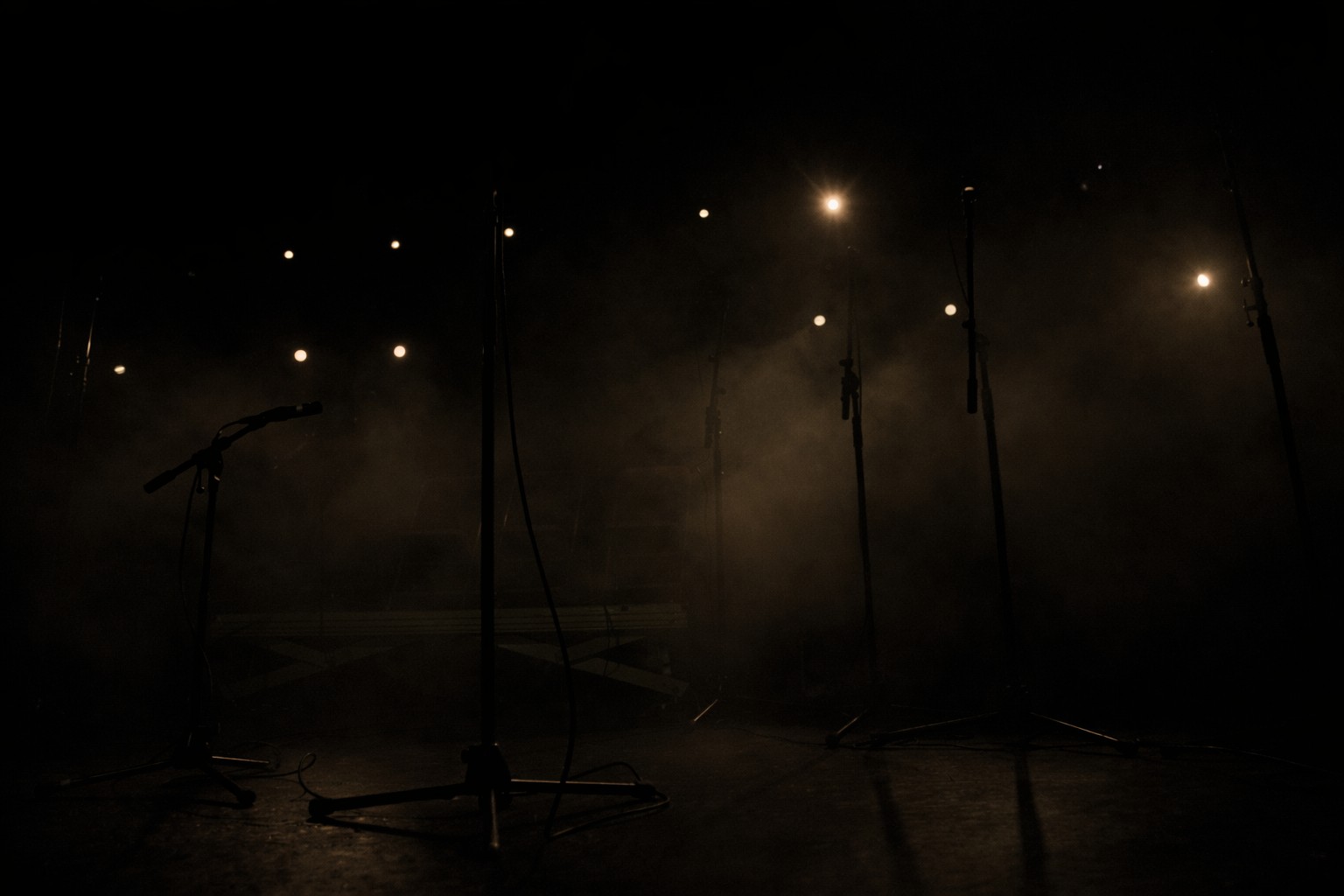} &
\CaseImage{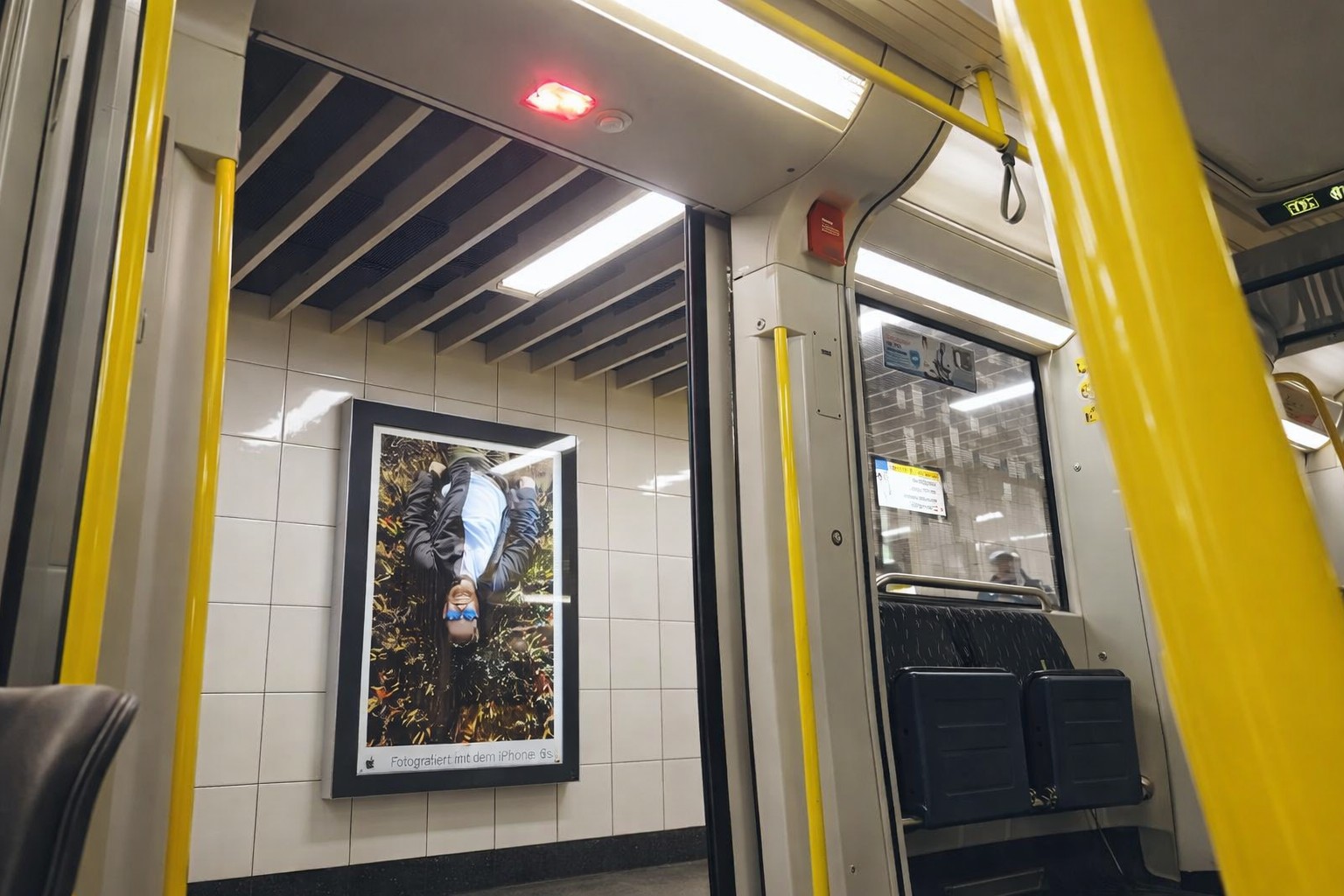} &
\CaseImage{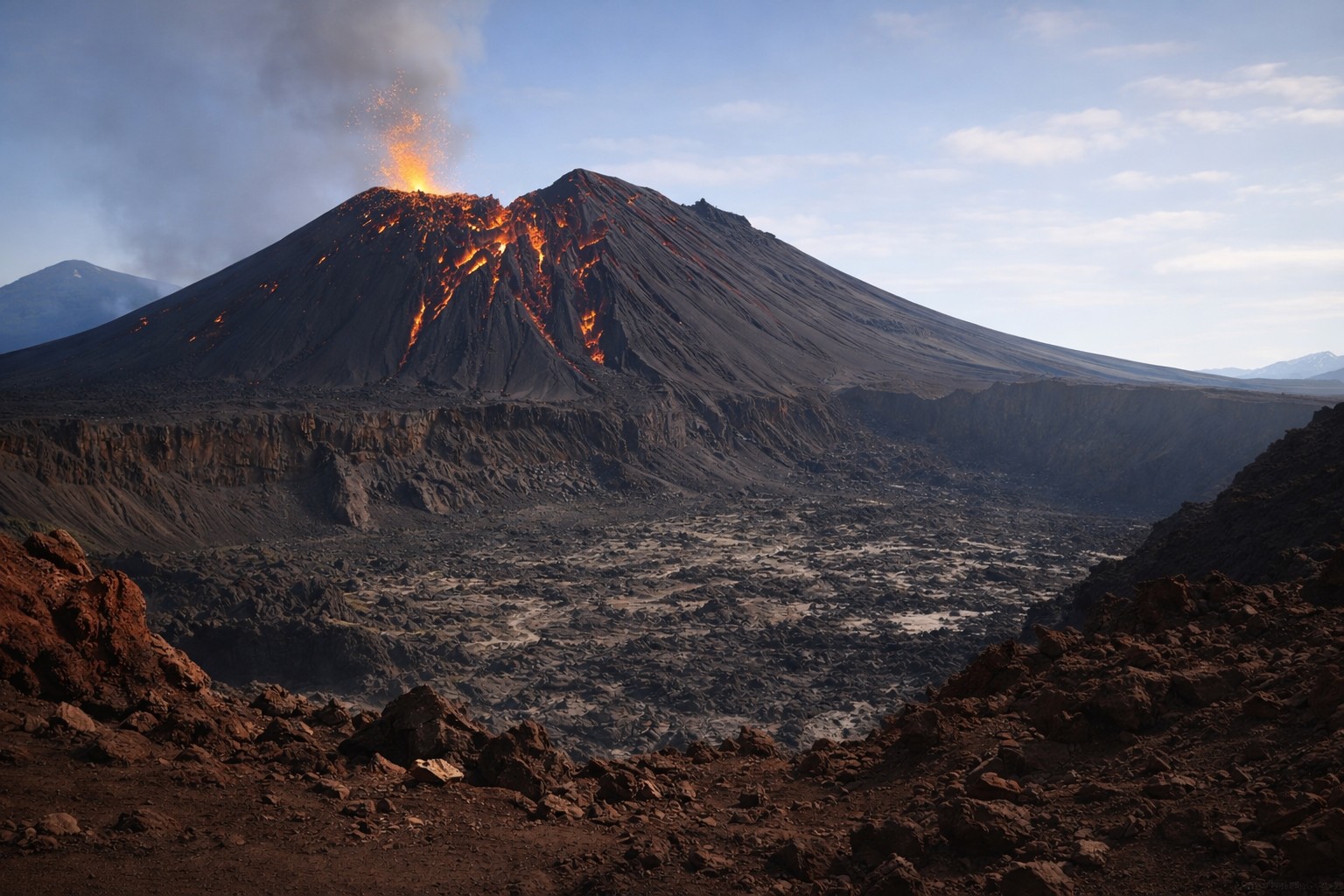} &
\CaseImage{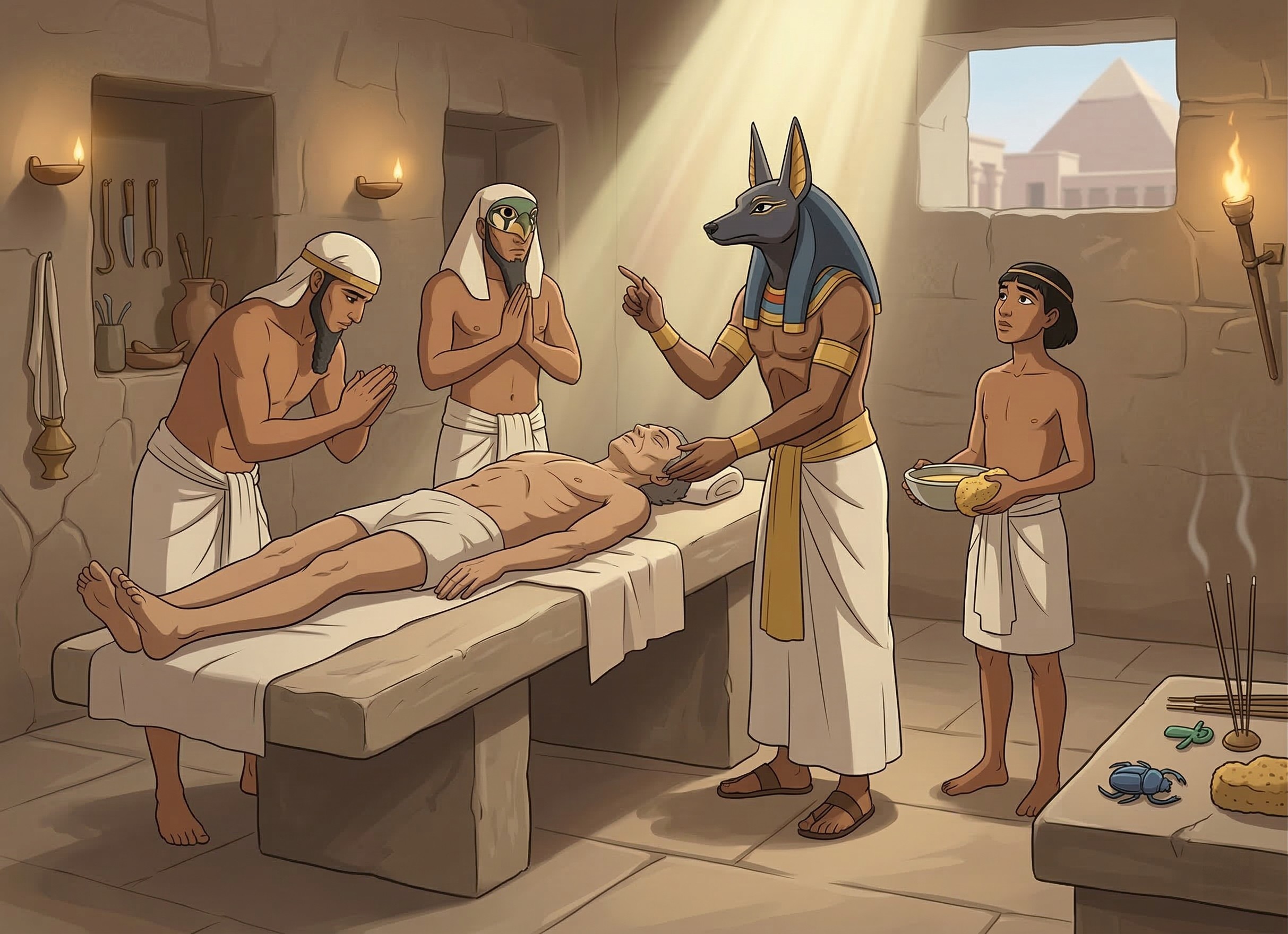} \\[0.0em]

\CasePromptFromJSON{1-b-49} &
\CasePromptFromJSON{1-a-5} &
\CasePromptFromJSON{1-b-29} &
\CasePromptFromJSON{1-c-03} \\[0.01em]
\CaseImage{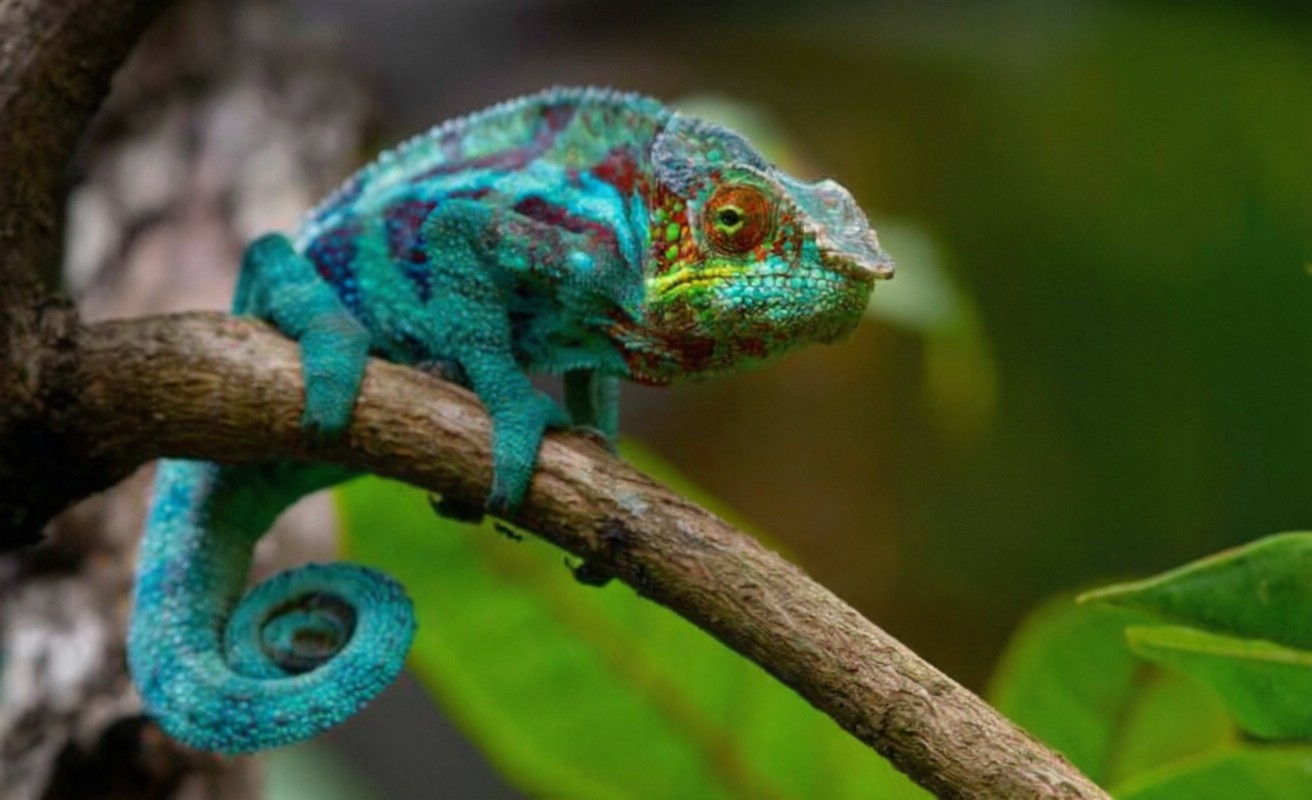} &
\CaseImage{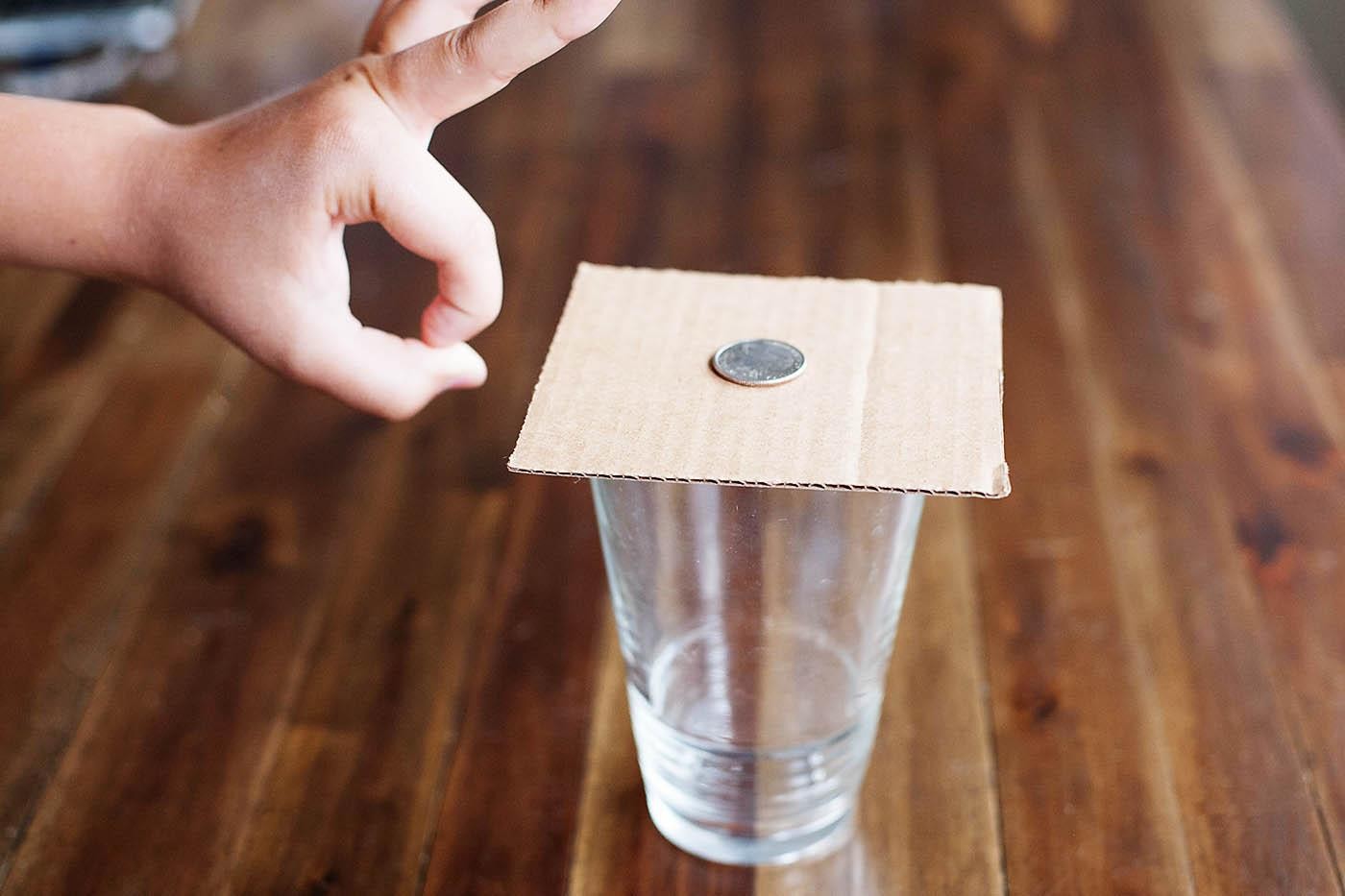} &
\CaseImage{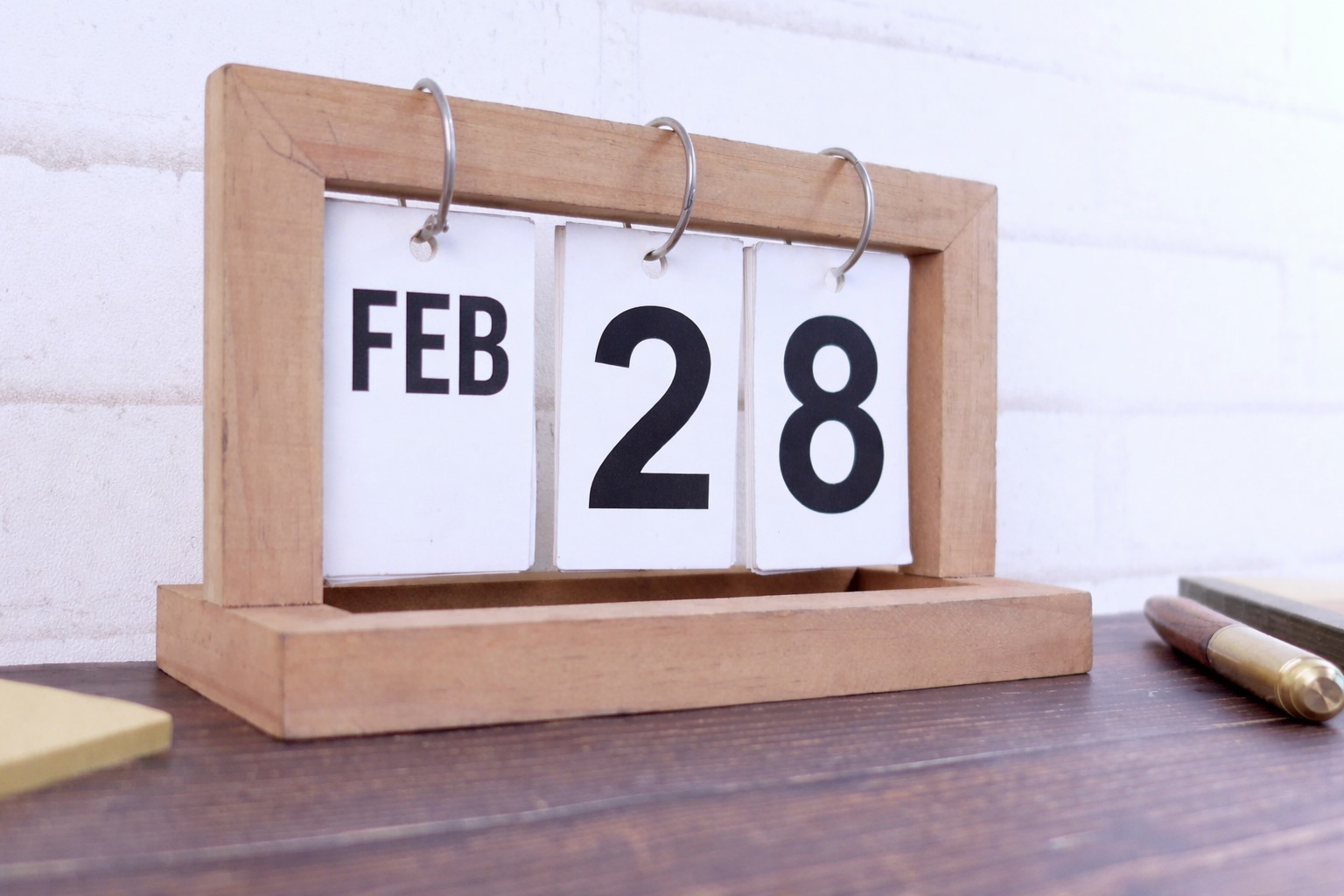} &
\CaseImage{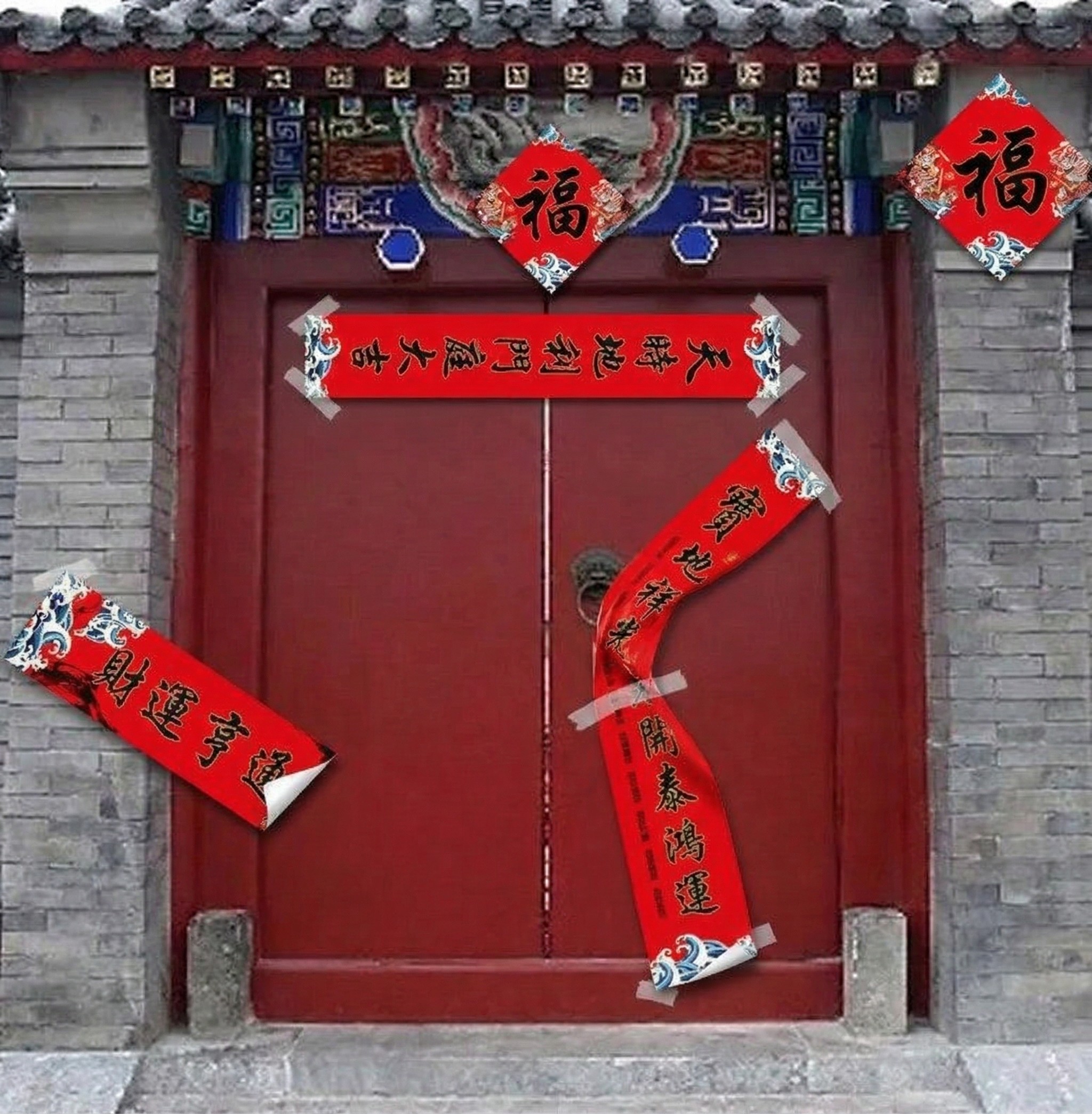} \\[0.0em]

\CasePromptFromJSON{1-a-9} &
\CasePromptFromJSON{1-a-25} &
\CasePromptFromJSON{1-a-49} &
\CasePromptFromJSON{1-c-12} \\[0.01em]
\CaseImage{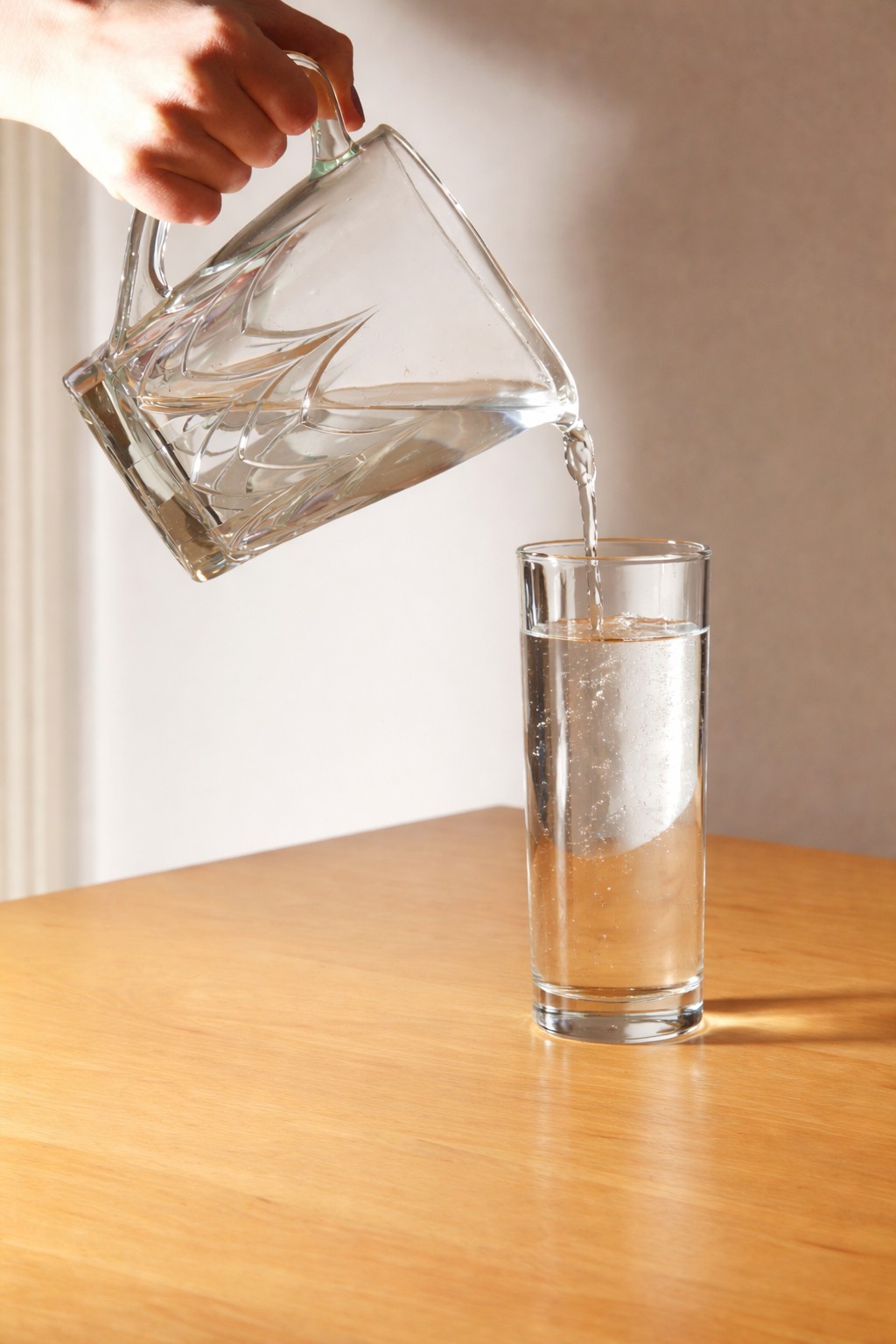} &
\CaseImage{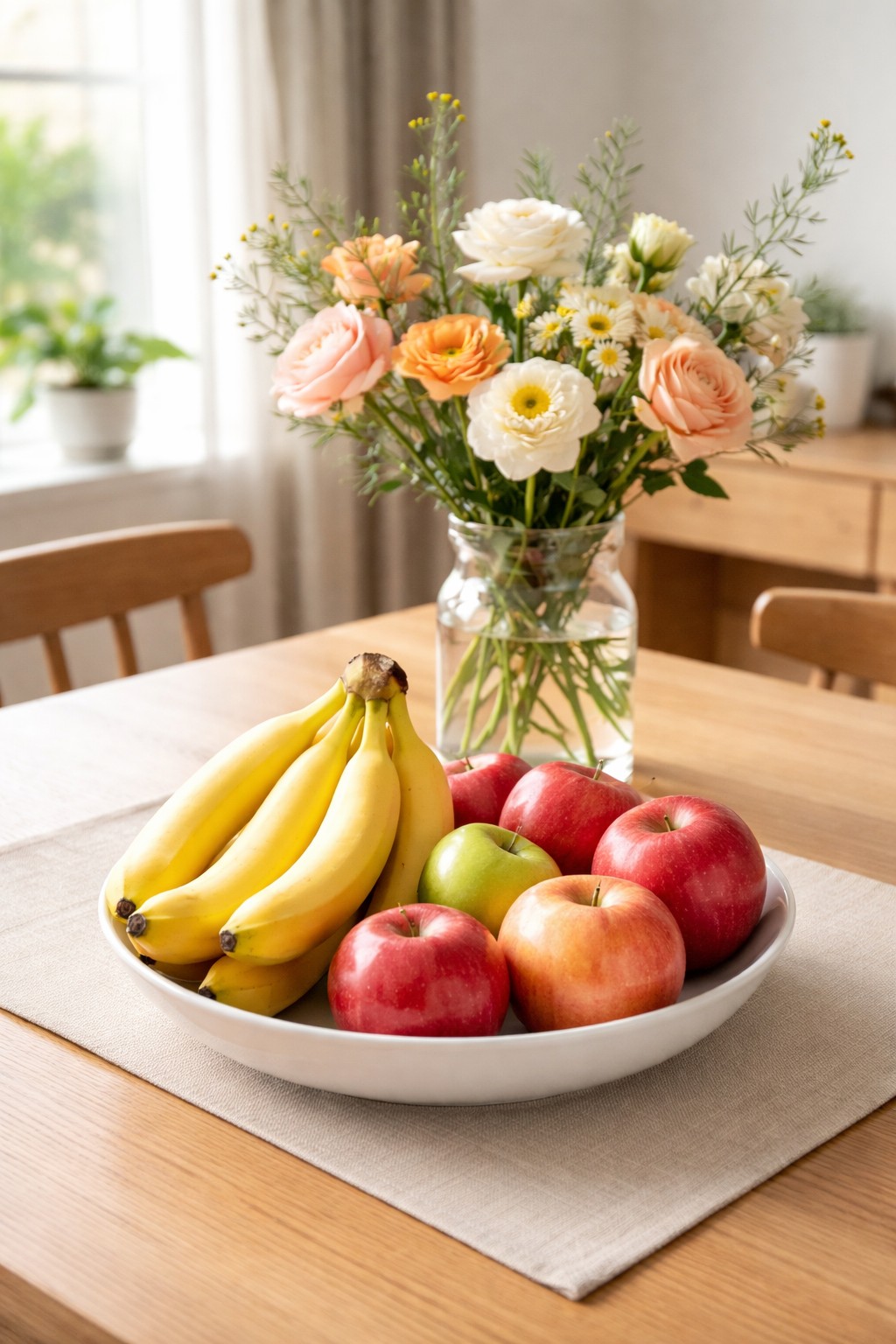} &
\CaseImage{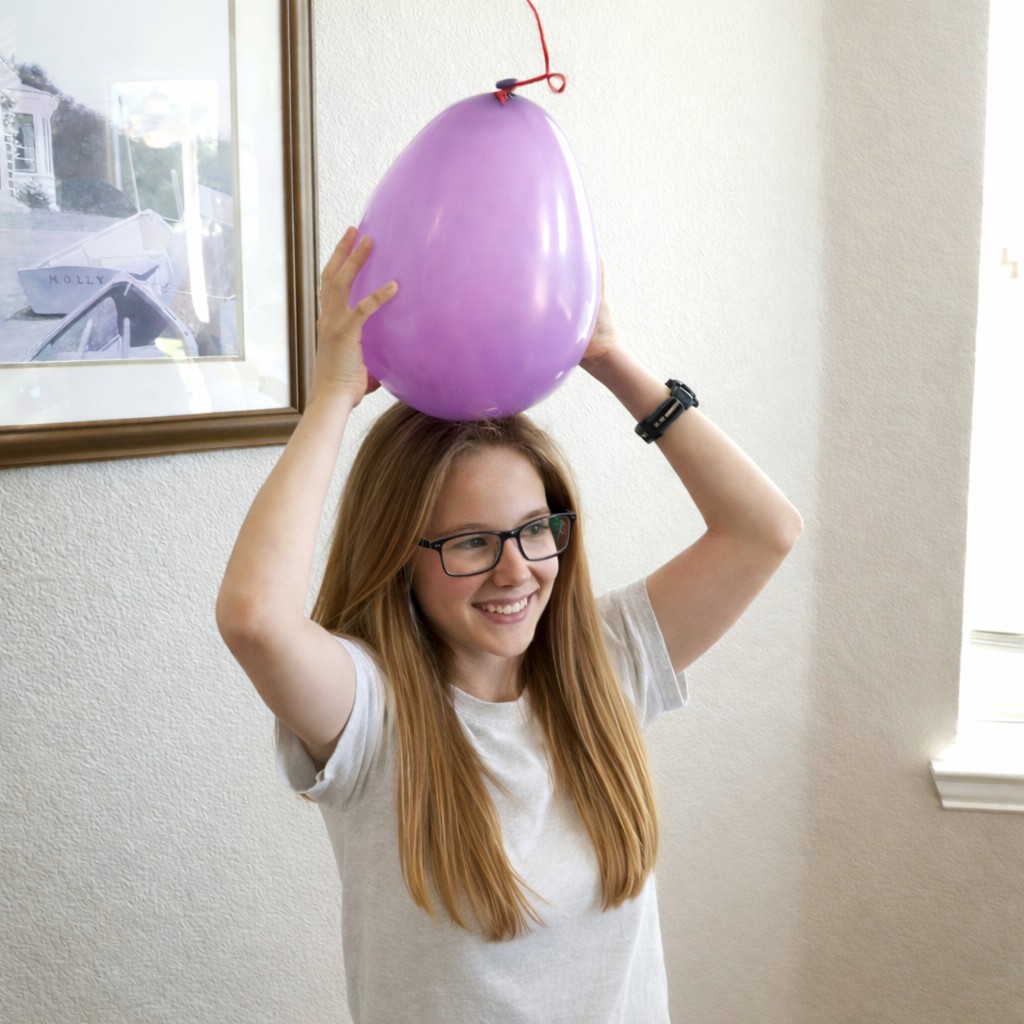} &
\CaseImage{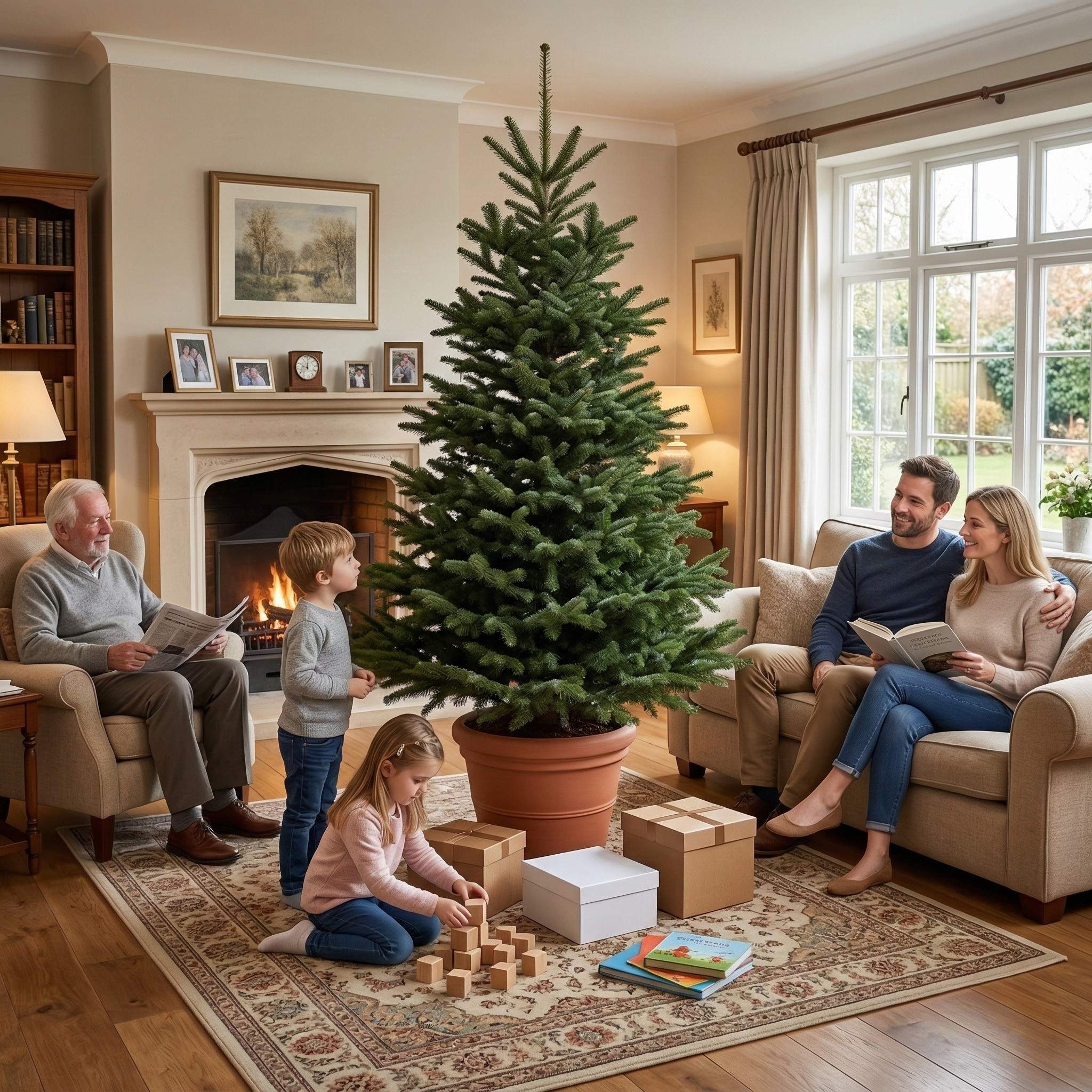} \\[-0.2em]

\end{tabularx}
\end{QAbox}
\end{minipage}

\caption{Representative examples from the World Knowledge category. These cases cover material change, public systems, world mechanics, cultural life, everyday living, earth cycles, and the living world, testing whether models can generate plausible state transitions grounded in physical, social, cultural, and natural-world knowledge.}
\label{tab:representative-world-knowledge-examples}
\end{table*}

\begin{table*}[htp]
\centering
\begin{minipage}{\textwidth}
\begin{QualBox}{Human-Centric}

\begin{tabularx}{\textwidth}{@{}XXXX@{}}
\CasePromptFromJSON{2-a-10} &
\CasePromptFromJSON{2-a-18} &
\CasePromptFromJSON{2-a-4} &
\CasePromptFromJSON{2-a-19} \\[0.01em]
\CaseImage{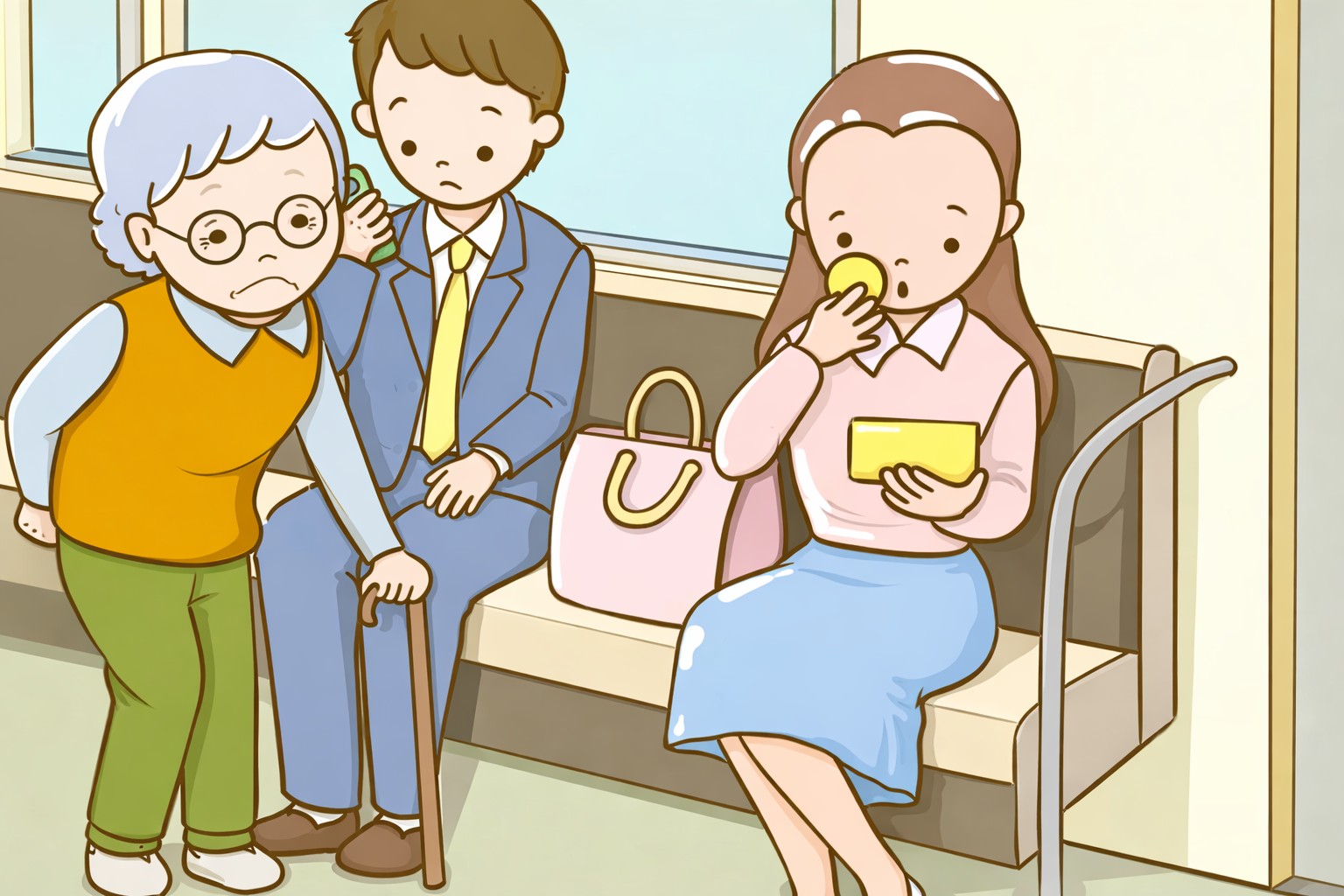} &
\CaseImage{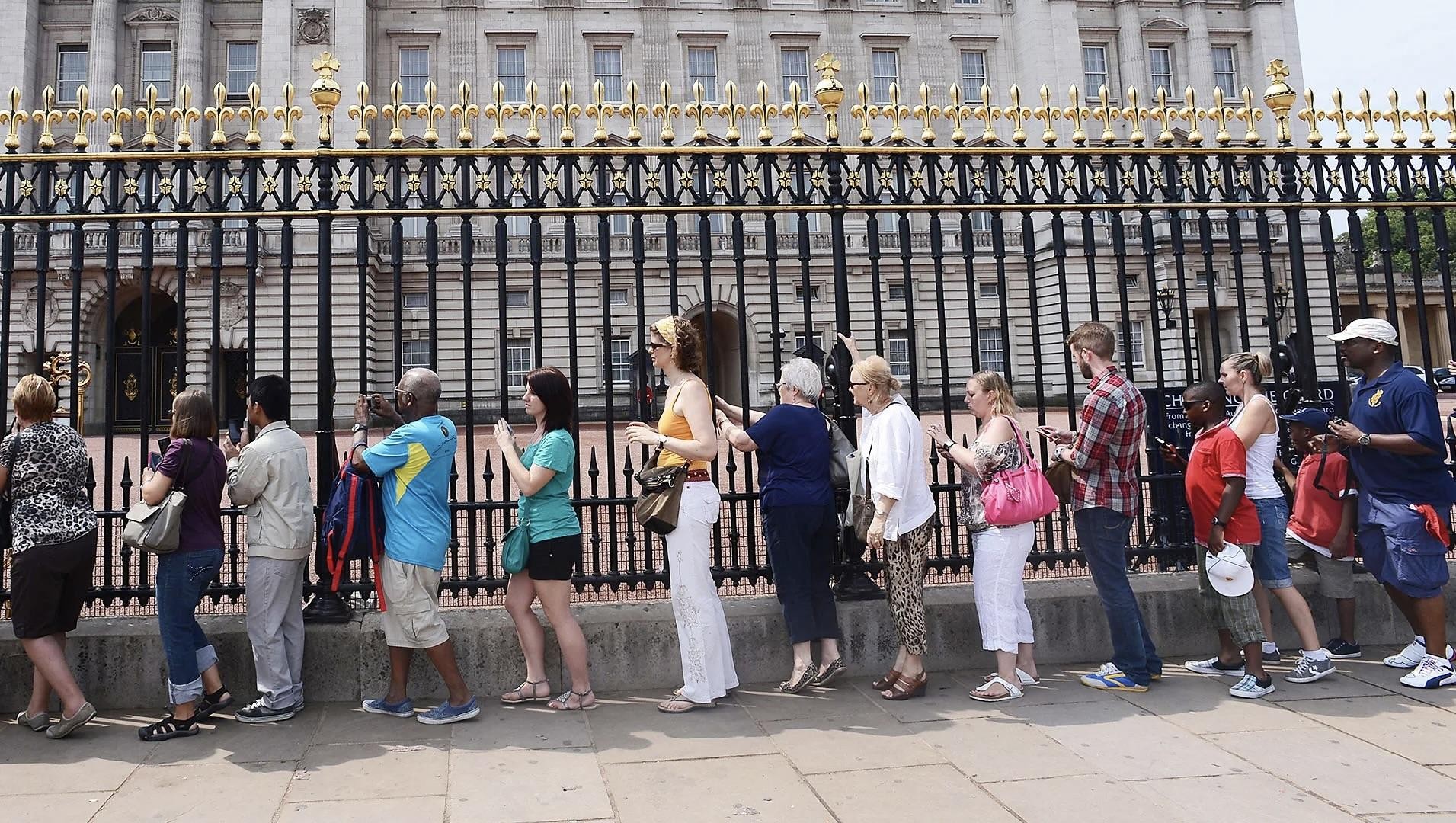} &
\CaseImage{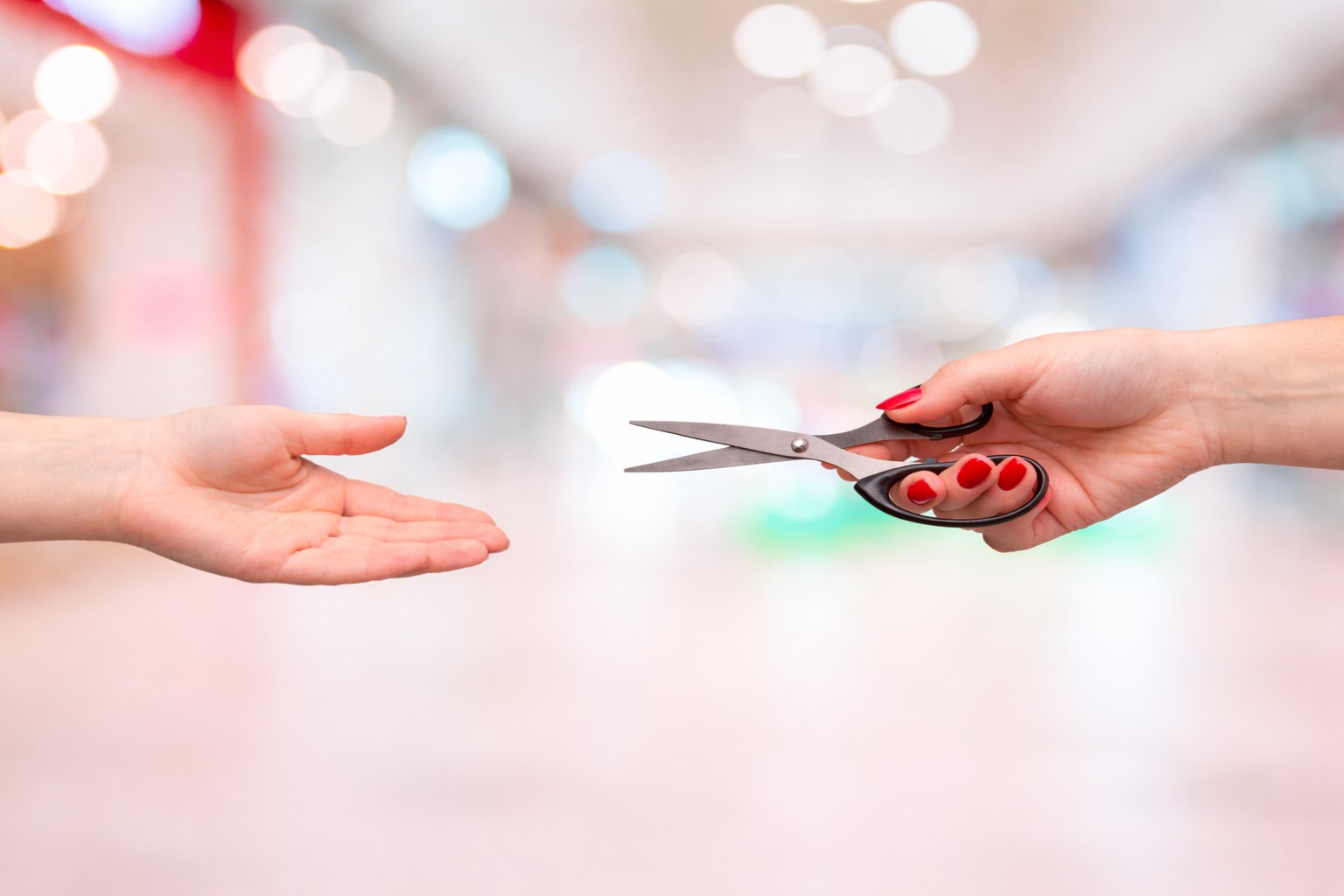} &
\CaseImage{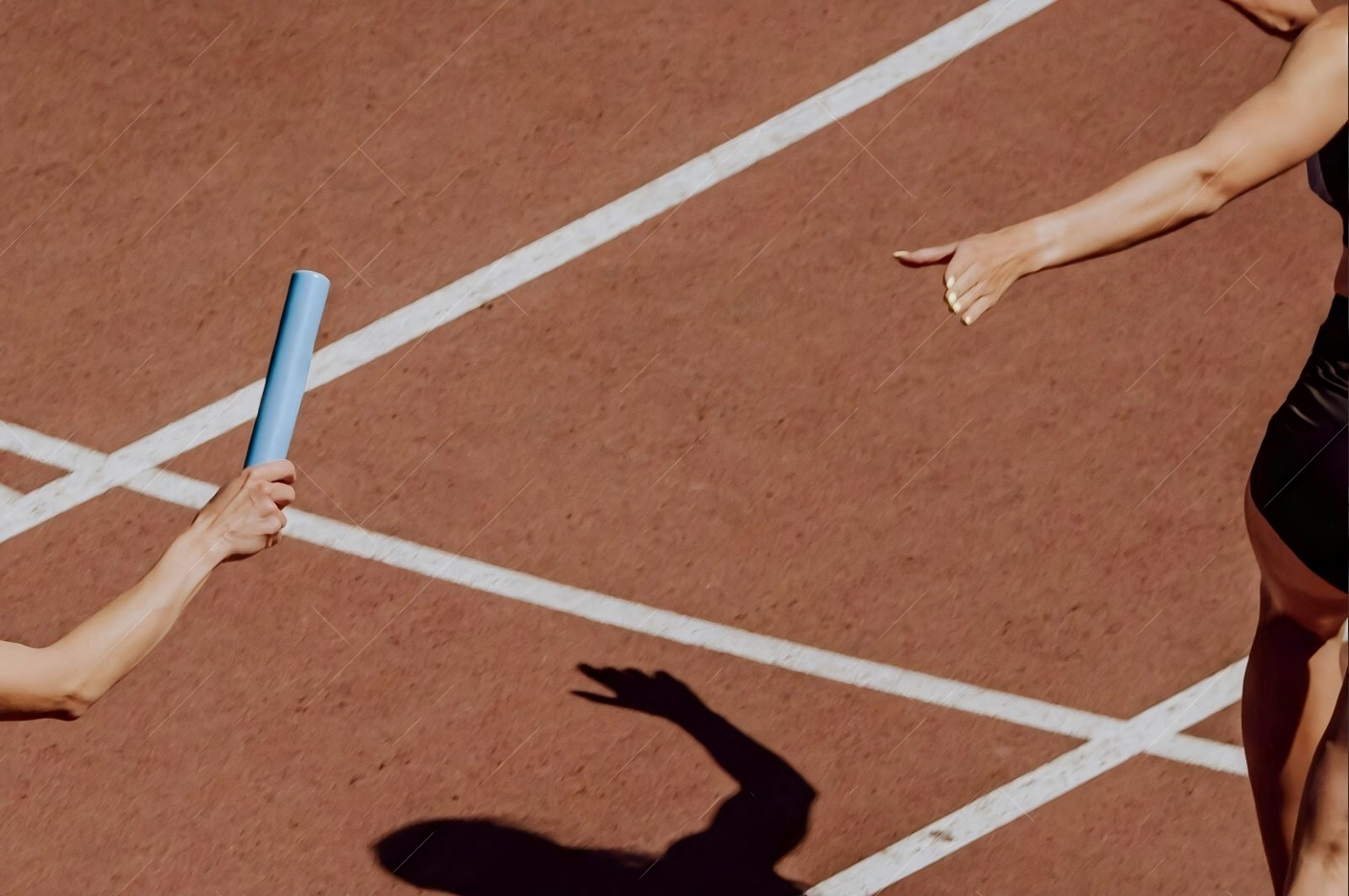} \\[0.0em]

\CasePromptFromJSON{2-a-21} &
\CasePromptFromJSON{2-a-23} &
\CasePromptFromJSON{2-a-3} &
\CasePromptFromJSON{2-a-6} \\[0.01em]
\CaseImage{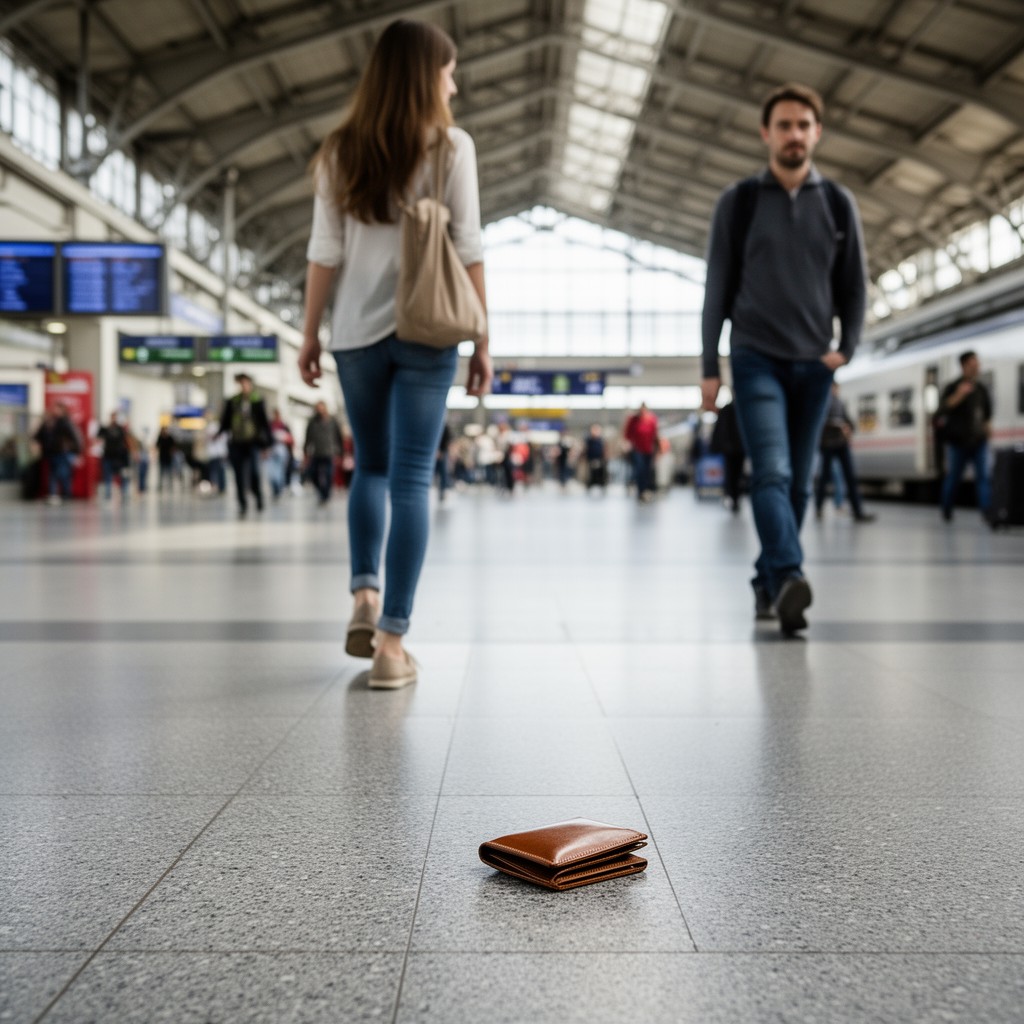} &
\CaseImage{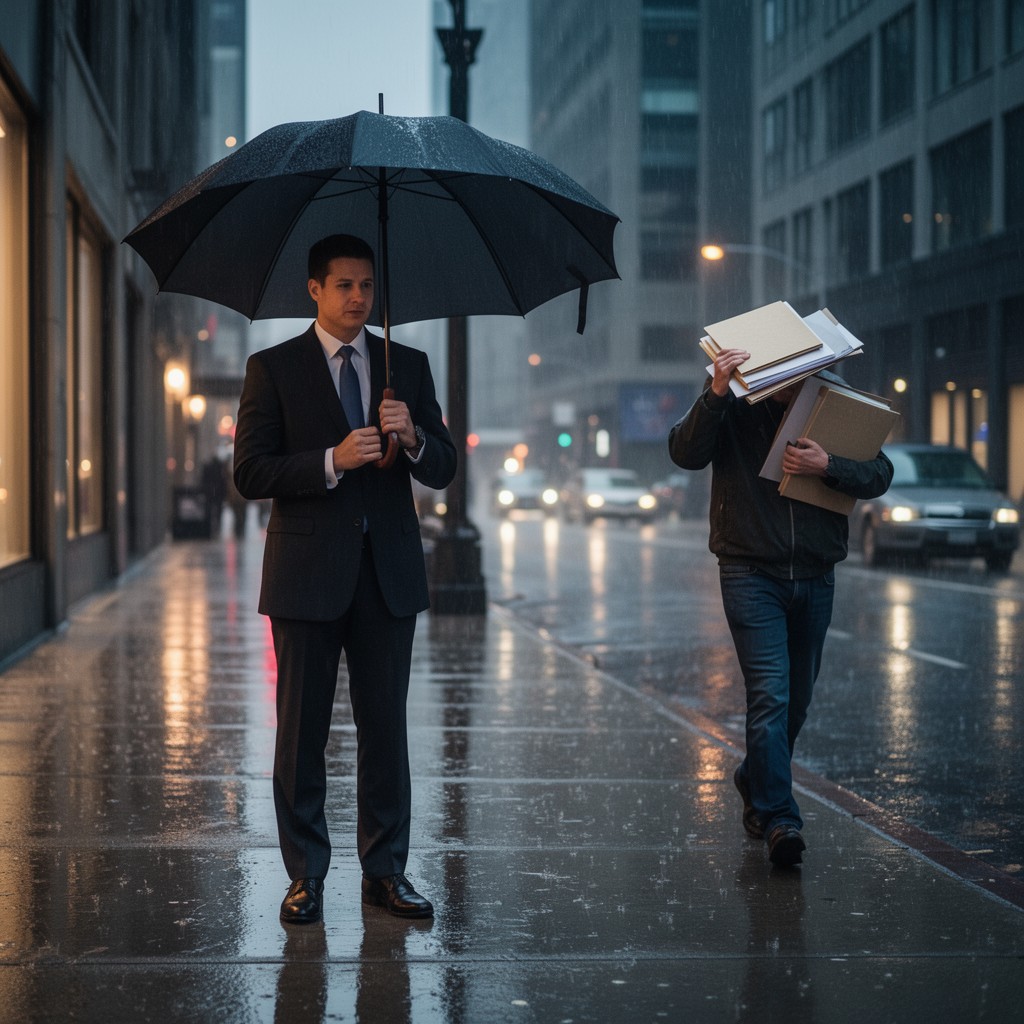} &
\CaseImage{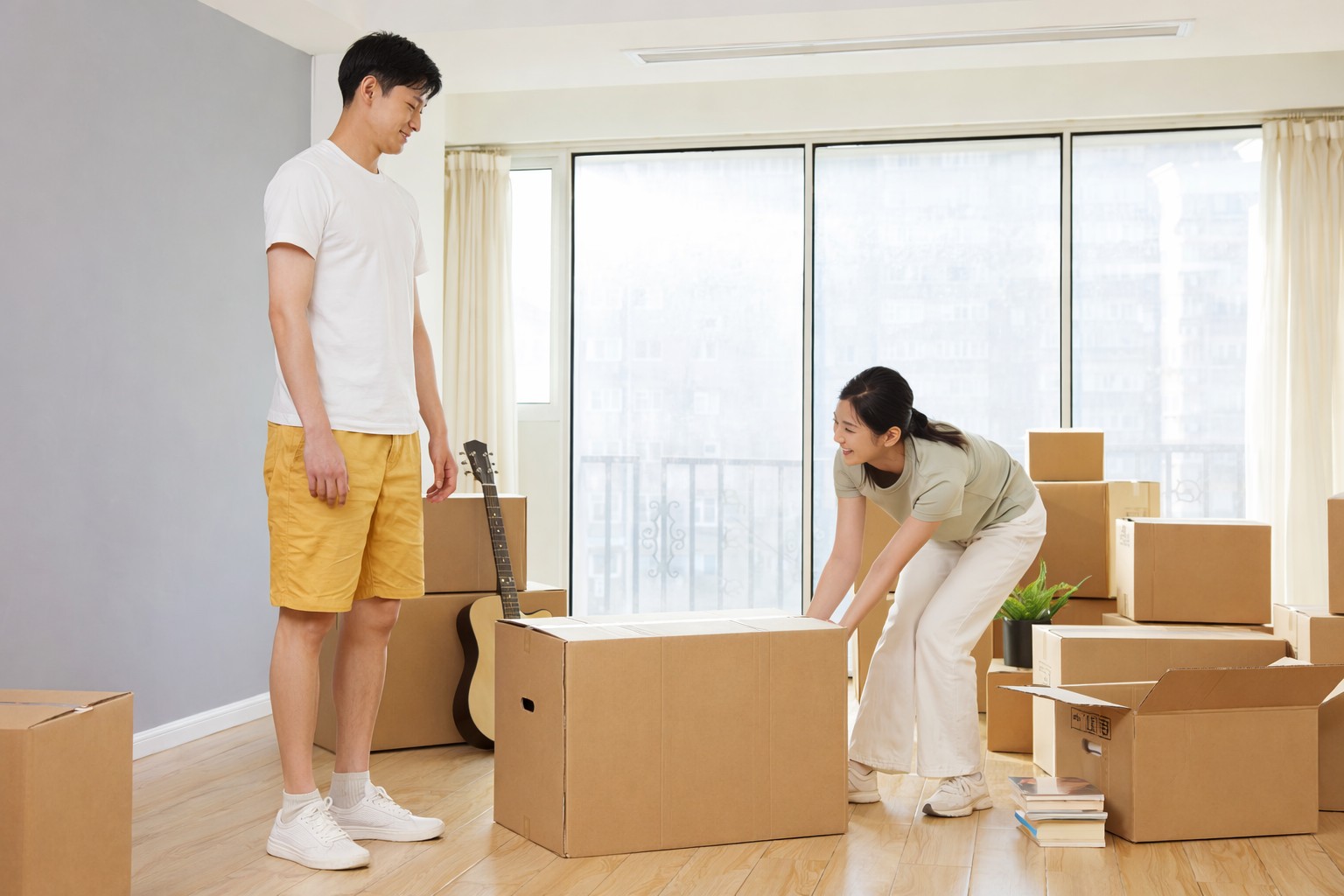} &
\CaseImage{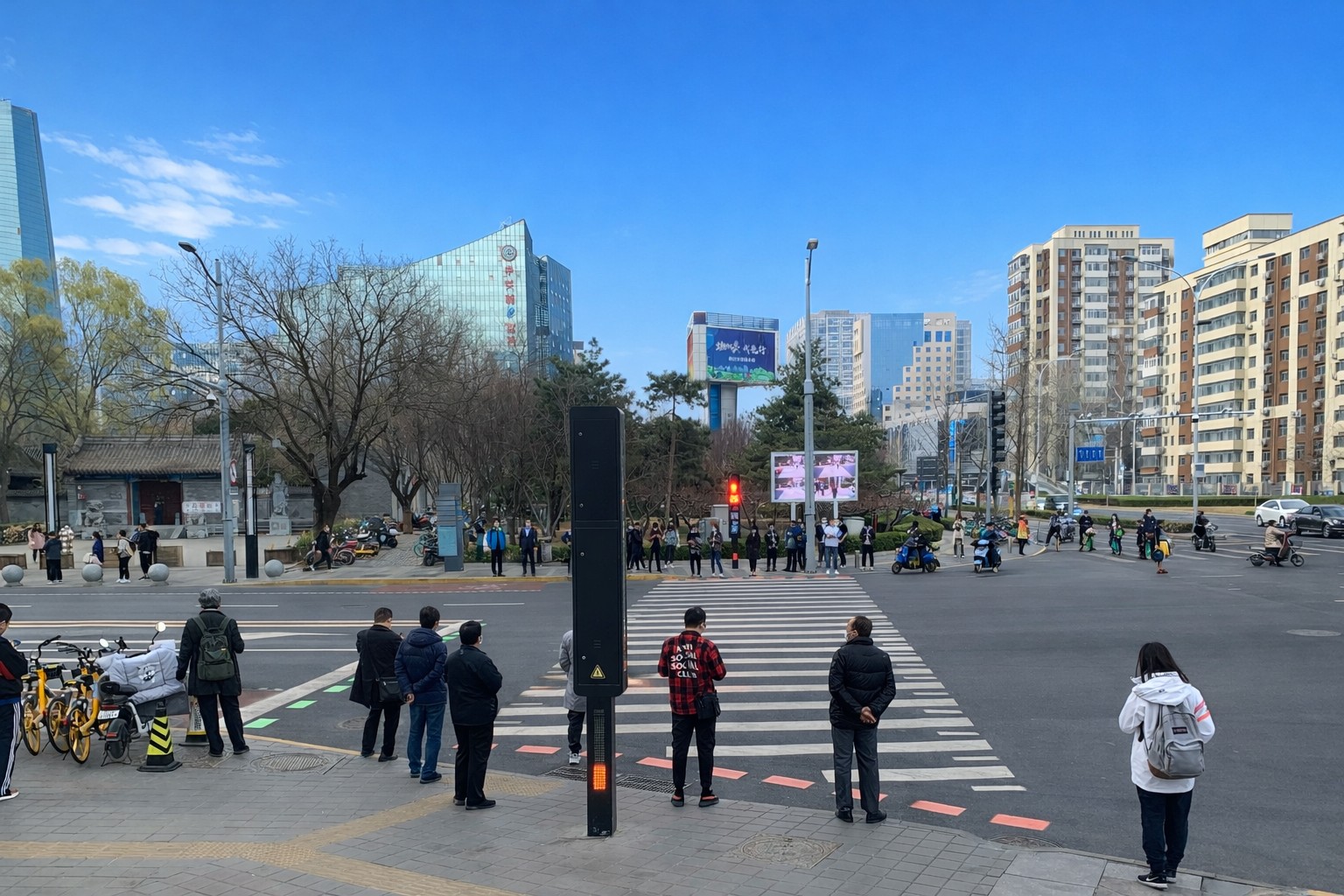} \\[0.0em]

\CasePromptFromJSON{2-b-2} &
\CasePromptFromJSON{2-b-6} &
\CasePromptFromJSON{2-b-8} &
\CasePromptFromJSON{2-b-11} \\[0.1em]
\CaseImage{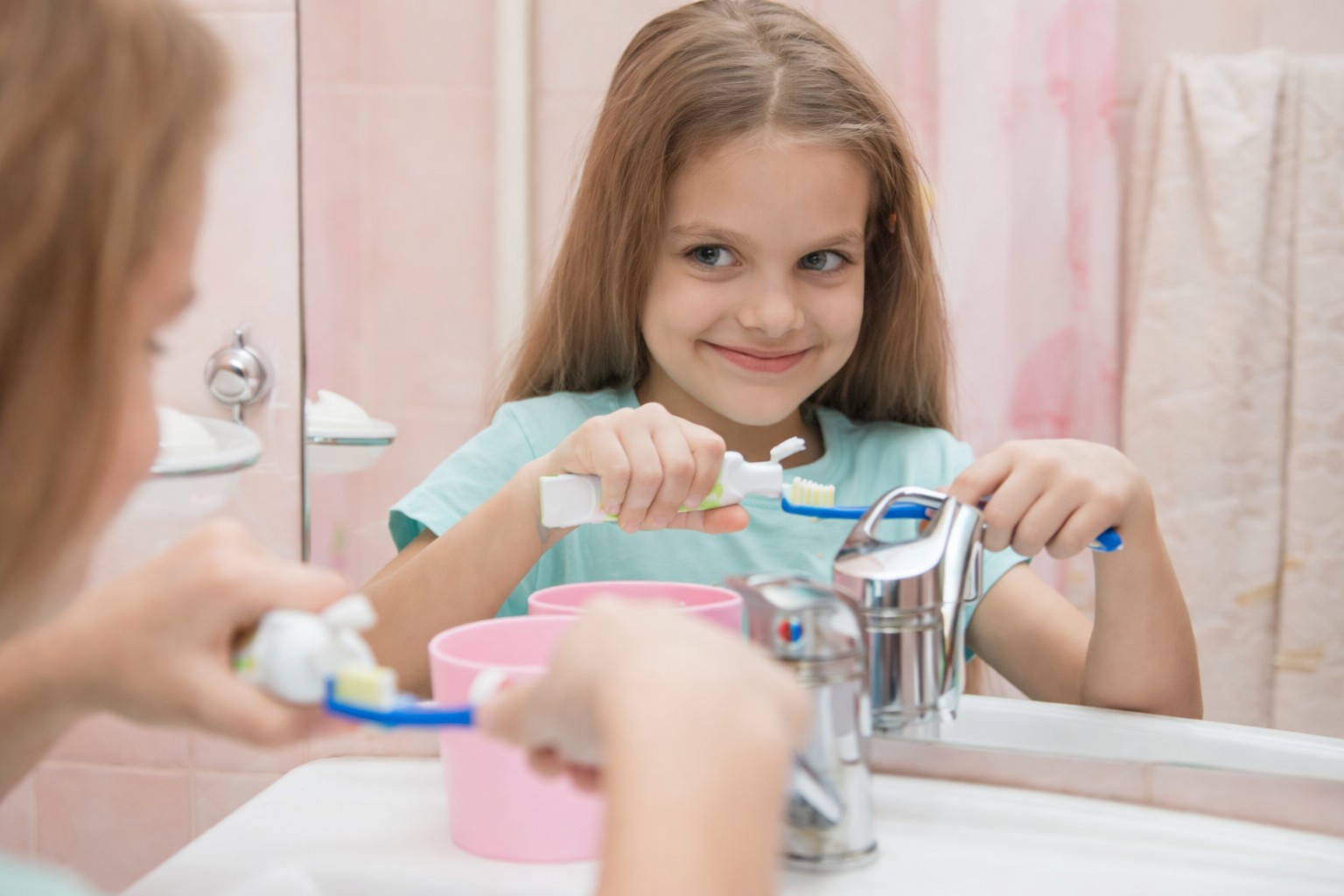} &
\CaseImage{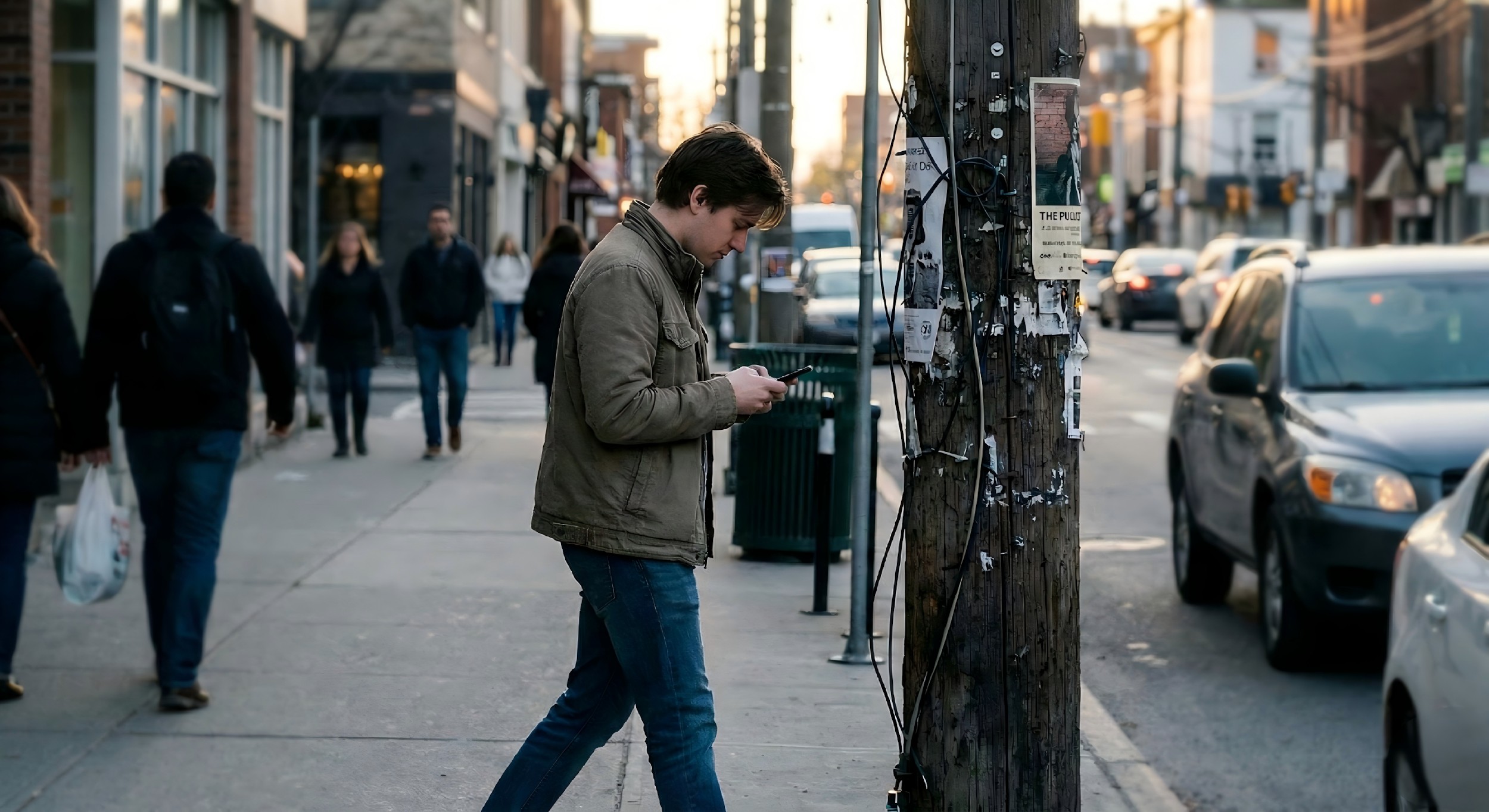} &
\CaseImage{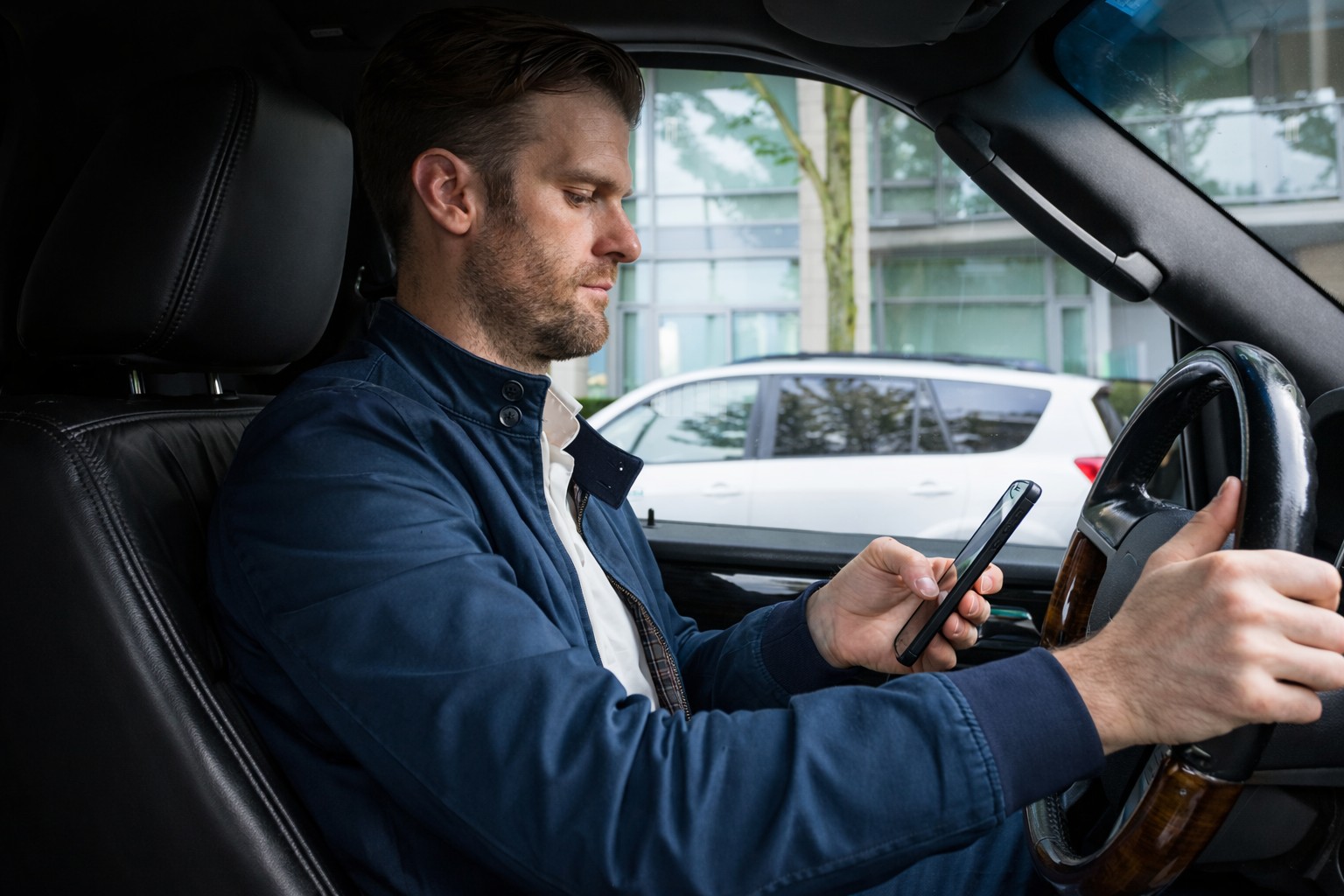} &
\CaseImage{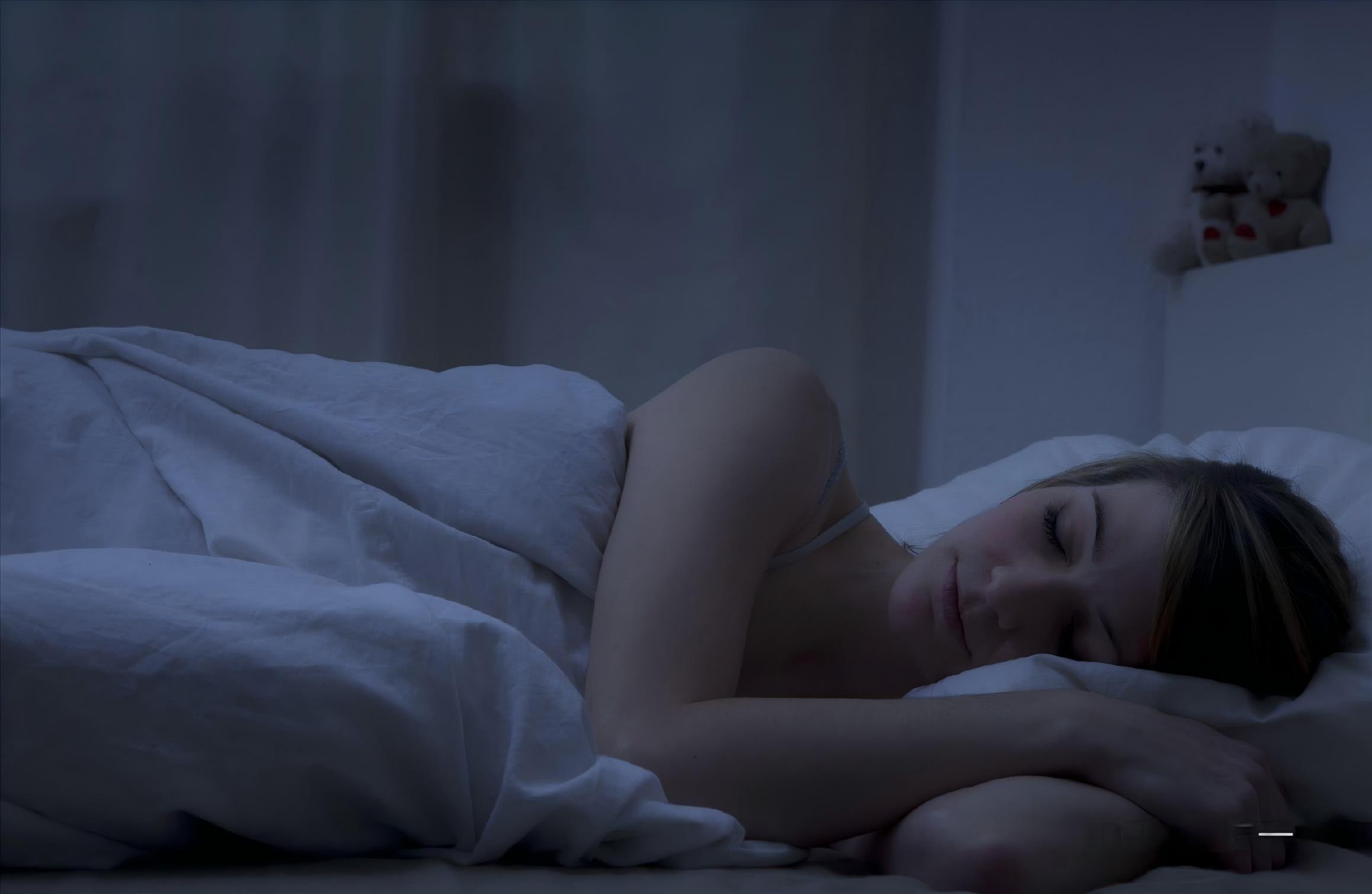} \\[0.0em]

\CasePromptFromJSON{2-b-12} &
\CasePromptFromJSON{2-b-15} &
\CasePromptFromJSON{2-b-21} &
\CasePromptFromJSON{2-b-19} \\[0.01em]
\CaseImage{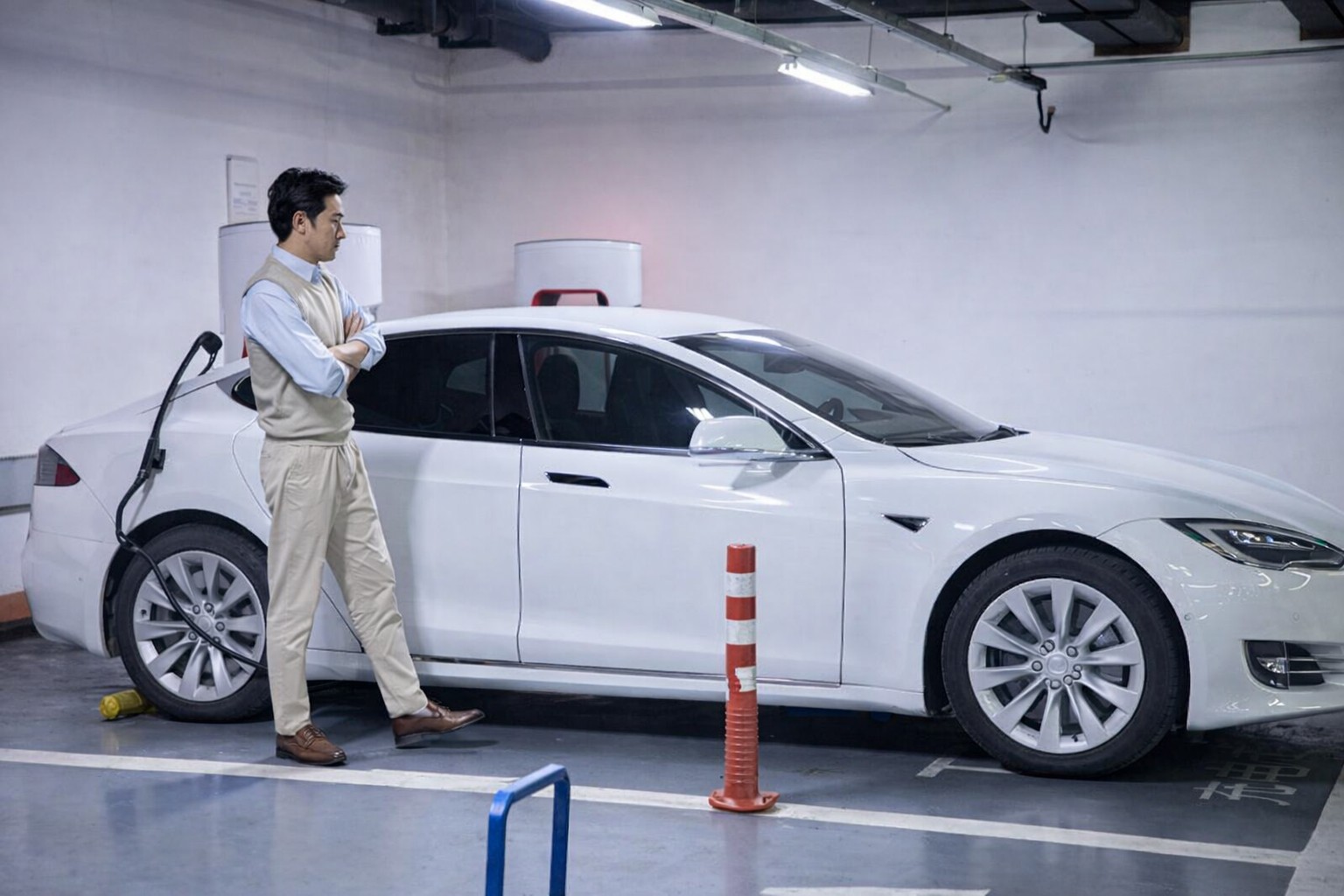} &
\CaseImage{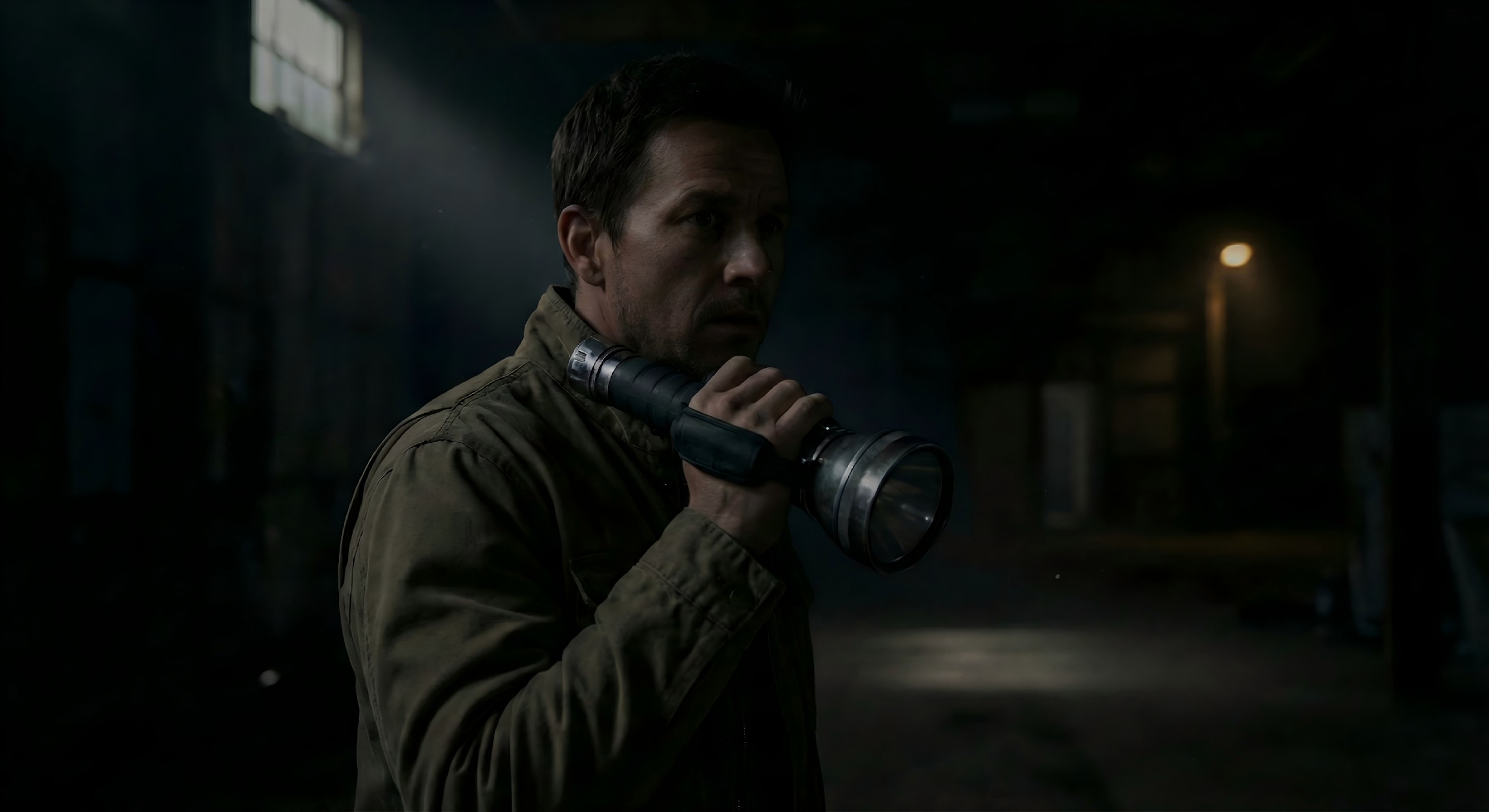} &
\CaseImage{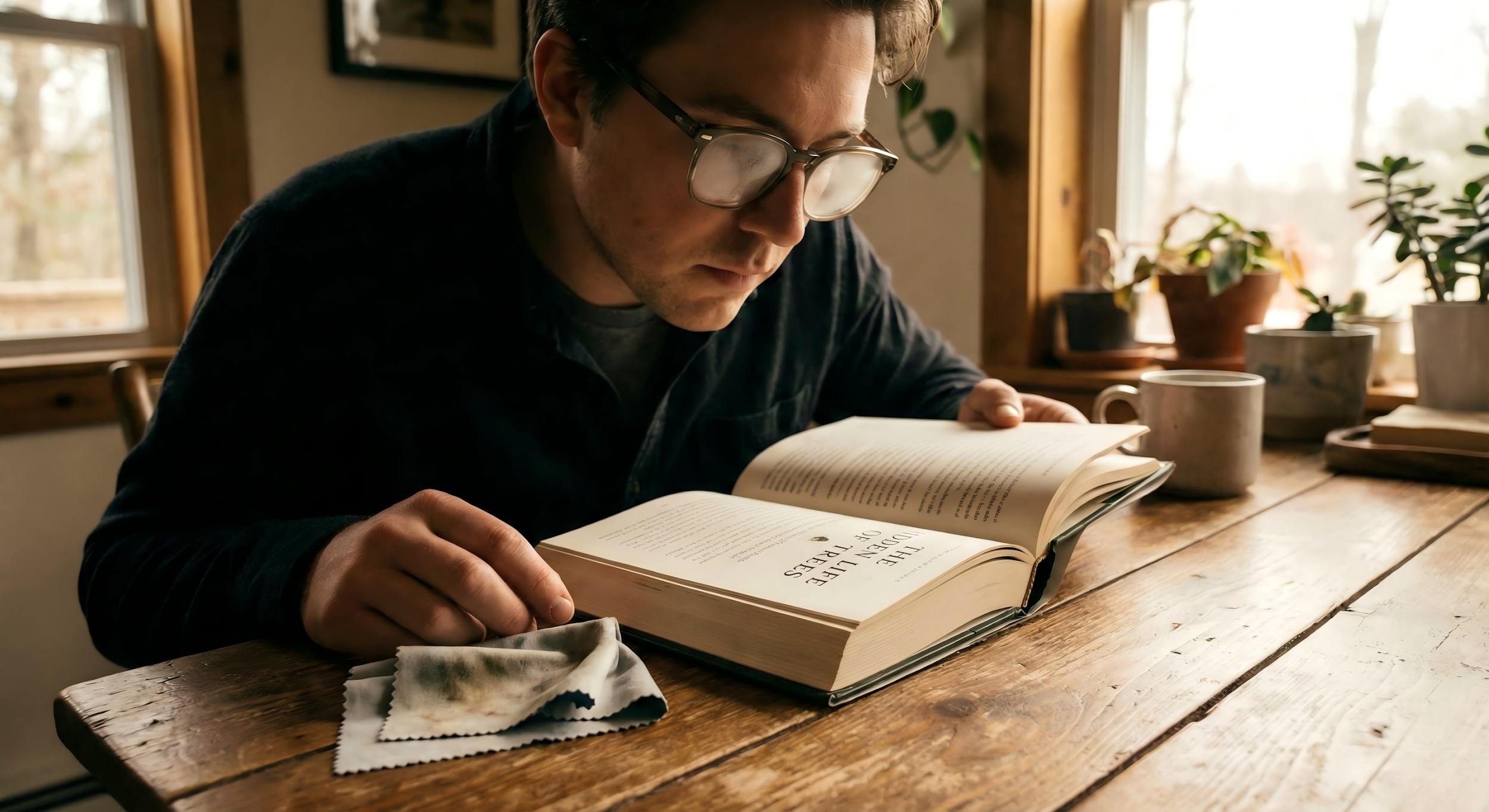} &
\CaseImage{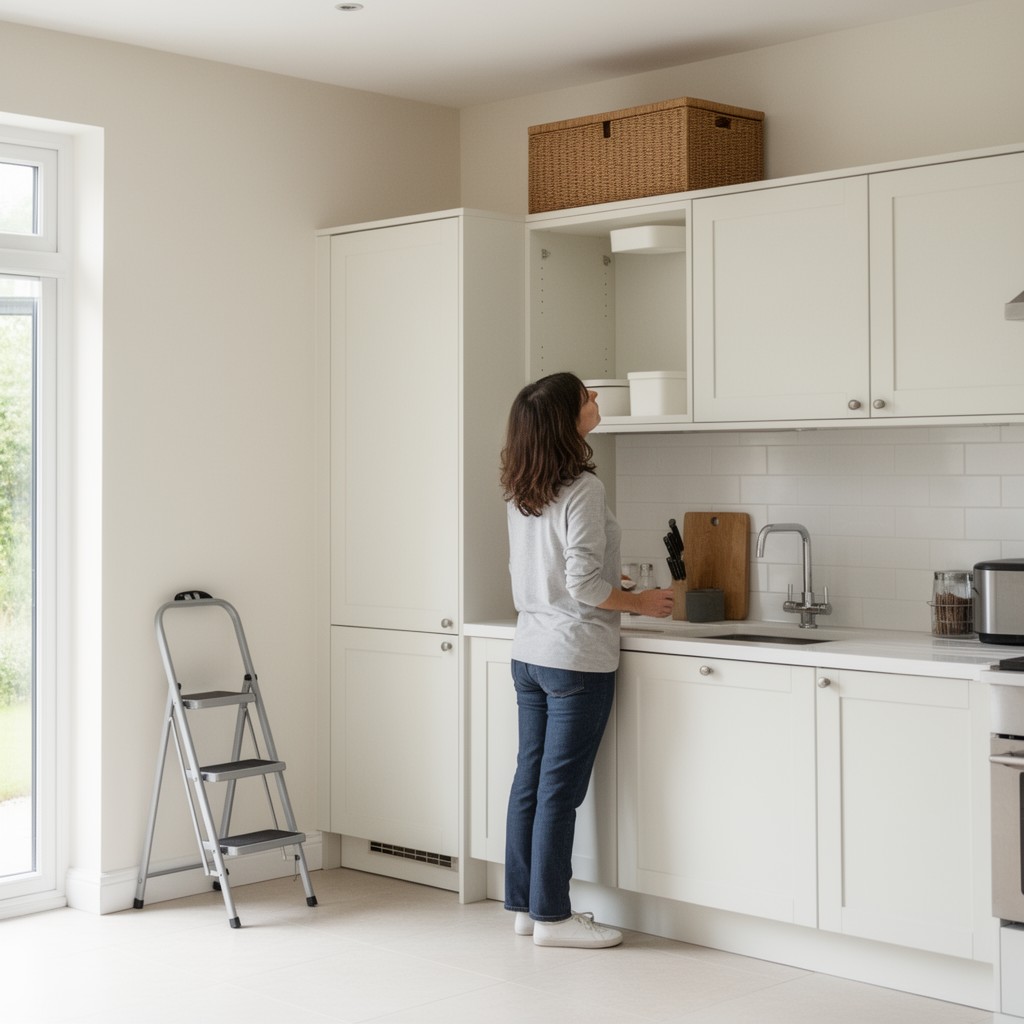} \\[0.0em]

\CasePromptFromJSON{2-c-12} &
\CasePromptFromJSON{2-c-15} &
\CasePromptFromJSON{2-c-24} &
\CasePromptFromJSON{2-c-26} \\[0.01em]
\CaseImage{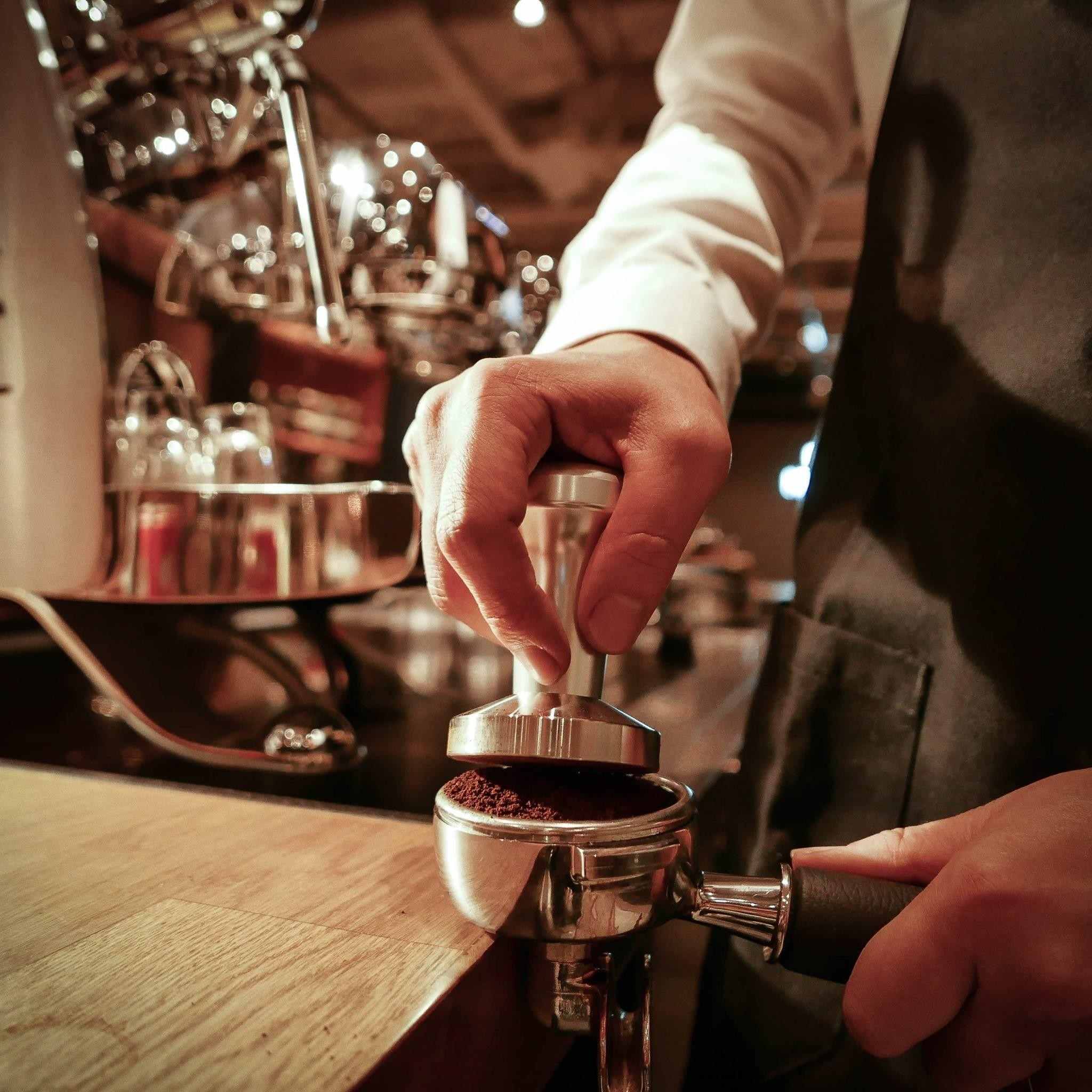} &
\CaseImage{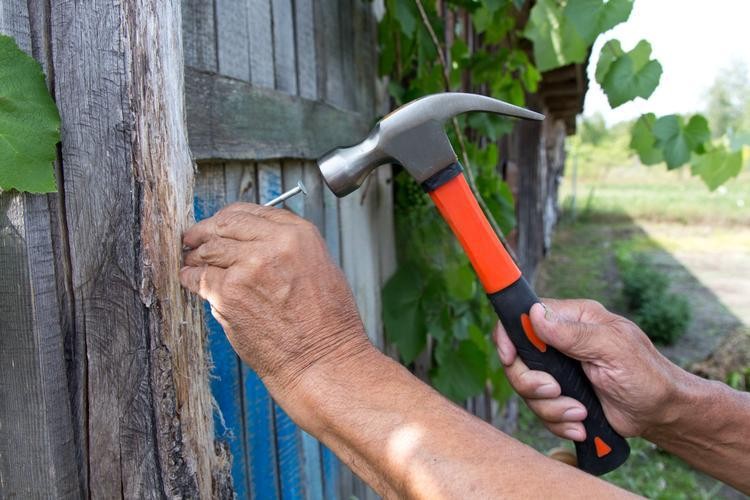} &
\CaseImage{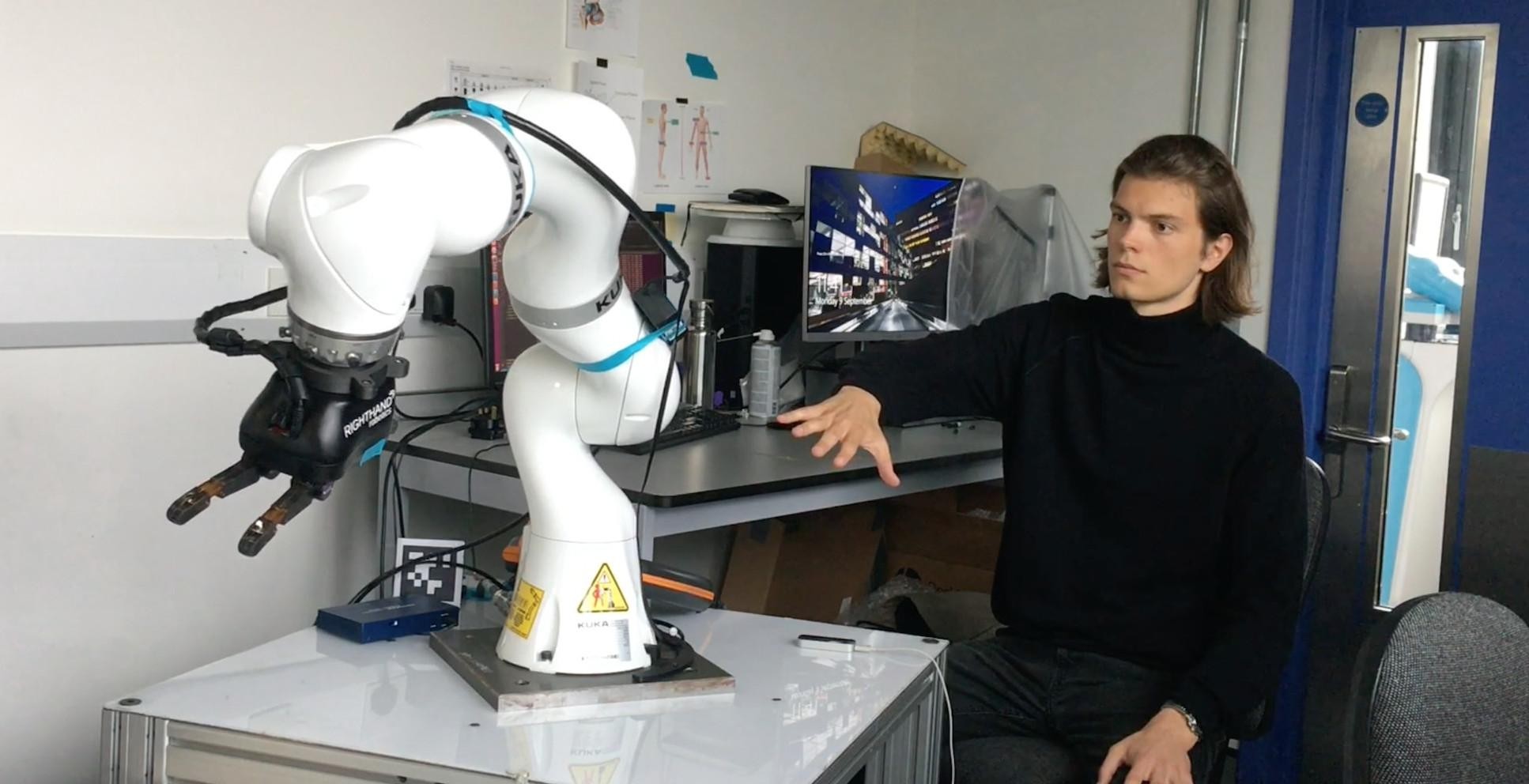} &
\CaseImage{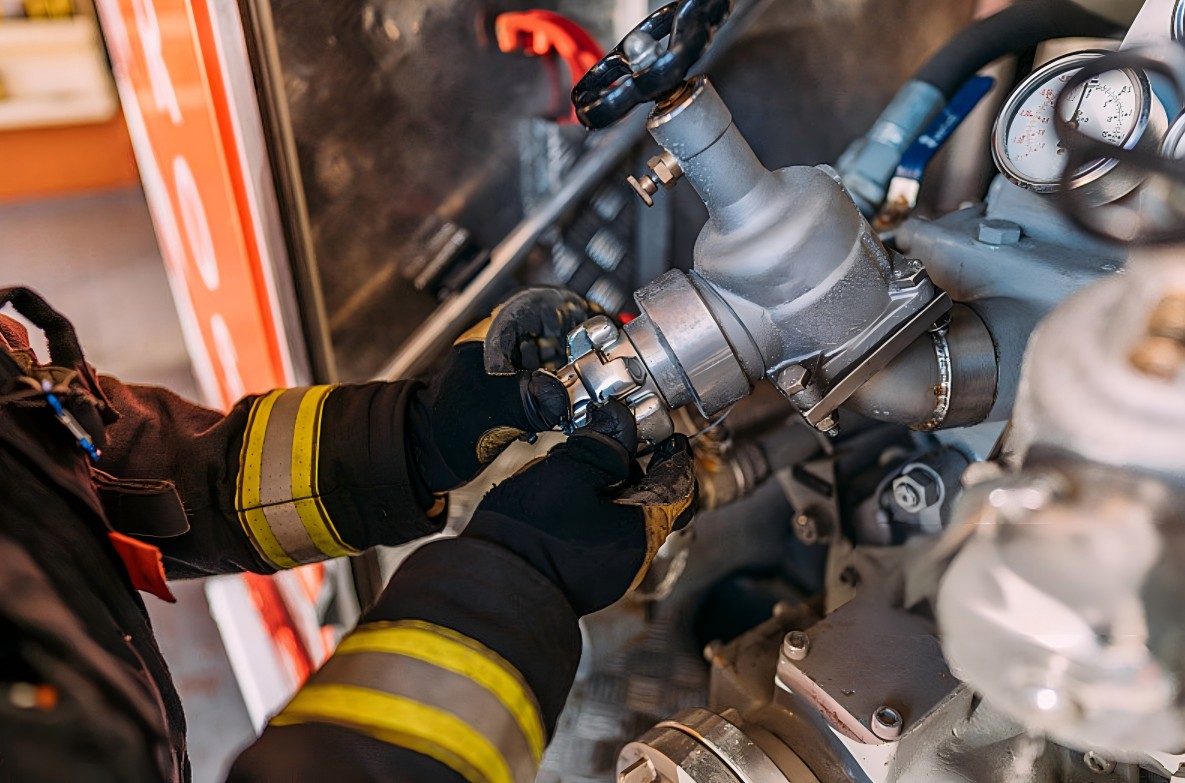} \\[0.0em]

\CasePromptFromJSON{2-c-19} &
\CasePromptFromJSON{2-c-30} &
\CasePromptFromJSON{2-c-5} &
\CasePromptFromJSON{2-c-6} \\[0.01em]
\CaseImage{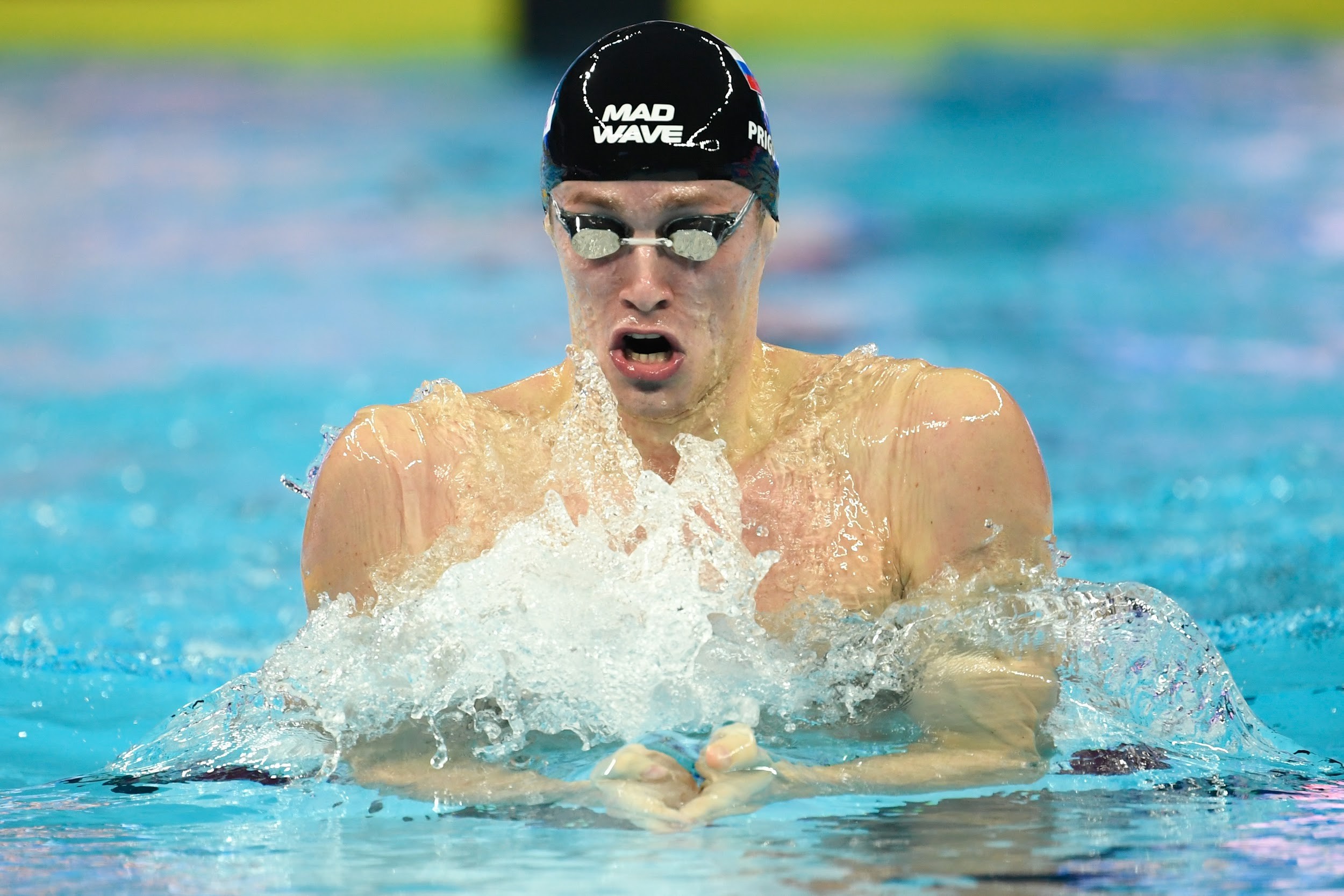} &
\CaseImage{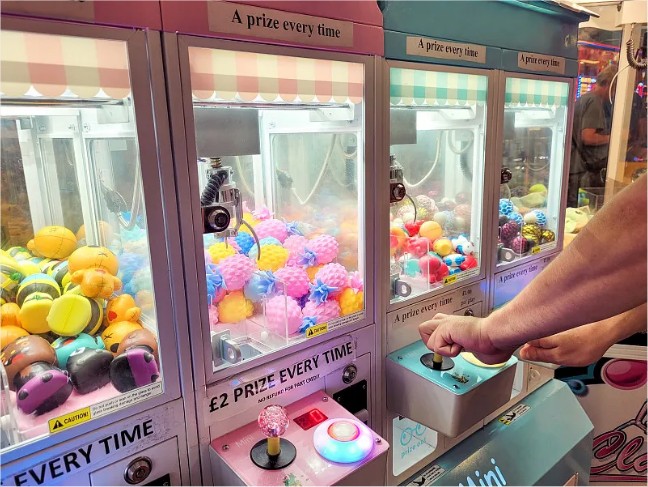} &
\CaseImage{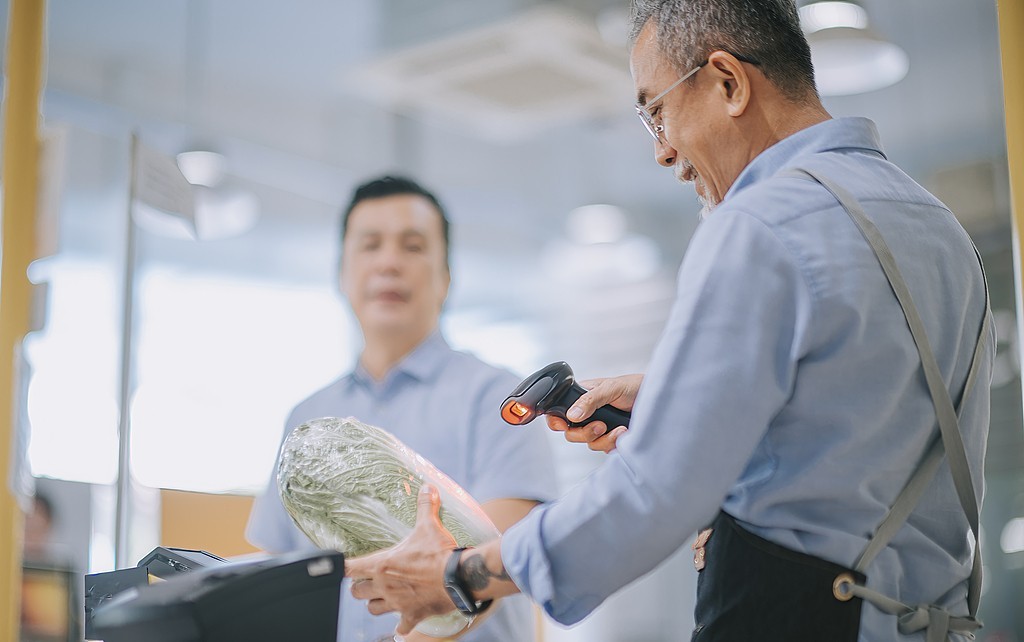} &
\CaseImage{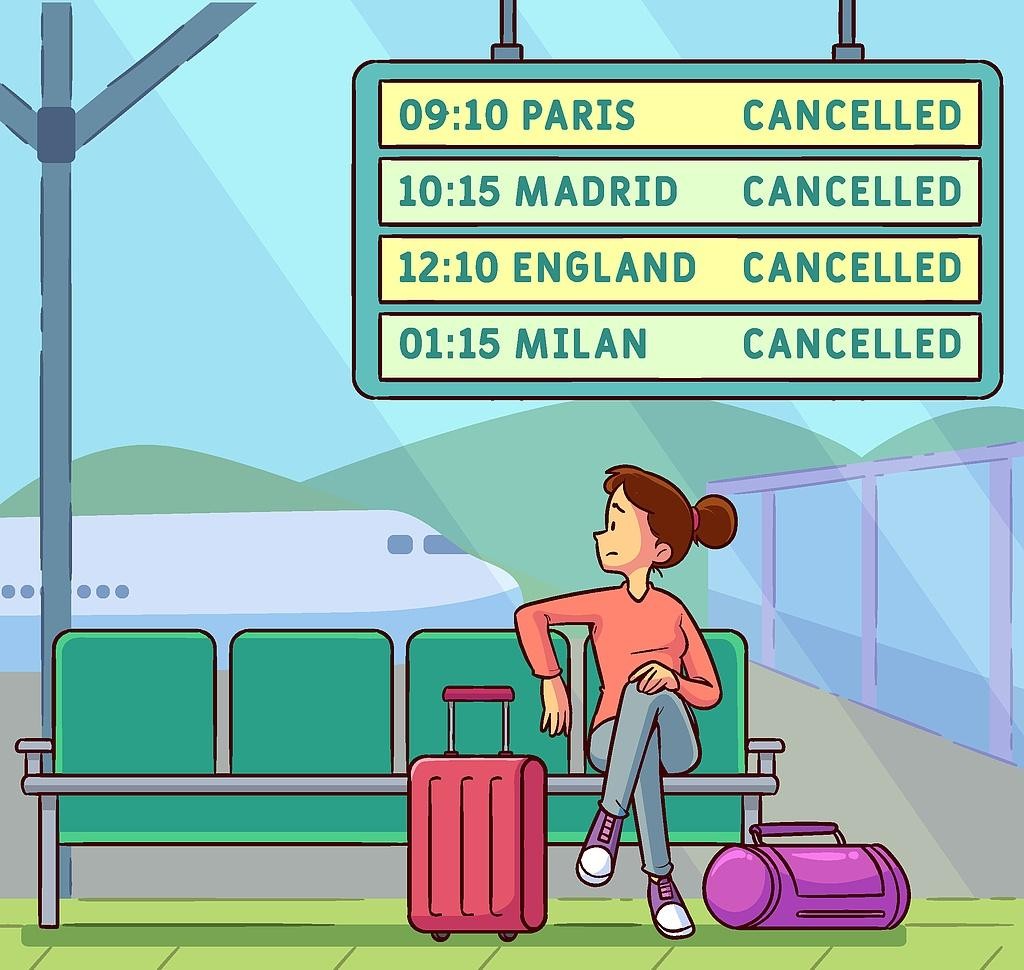} \\[-0.2em]

\end{tabularx}
\end{QualBox}
\end{minipage}
\caption{Representative examples from the Human-Centric category. These cases cover object handling, social scenes, skilled actions, personal routines, and public conduct, focusing on whether models can generate plausible human behaviors, interactions, and object-centered actions over time.}
\label{tab:representative-human-centric-examples}
\end{table*}

\begin{table*}[htp]
\centering
\begin{minipage}{\textwidth}
\begin{CASEBAX}{Logic Reasoning}

\begin{tabularx}{\textwidth}{@{}XXXX@{}}
\CasePromptFromJSON{3-a-1} &
\CasePromptFromJSON{3-a-10} &
\CasePromptFromJSON{3-a-8} &
\CasePromptFromJSON{3-a-23} \\[0.01em]
\CaseImage{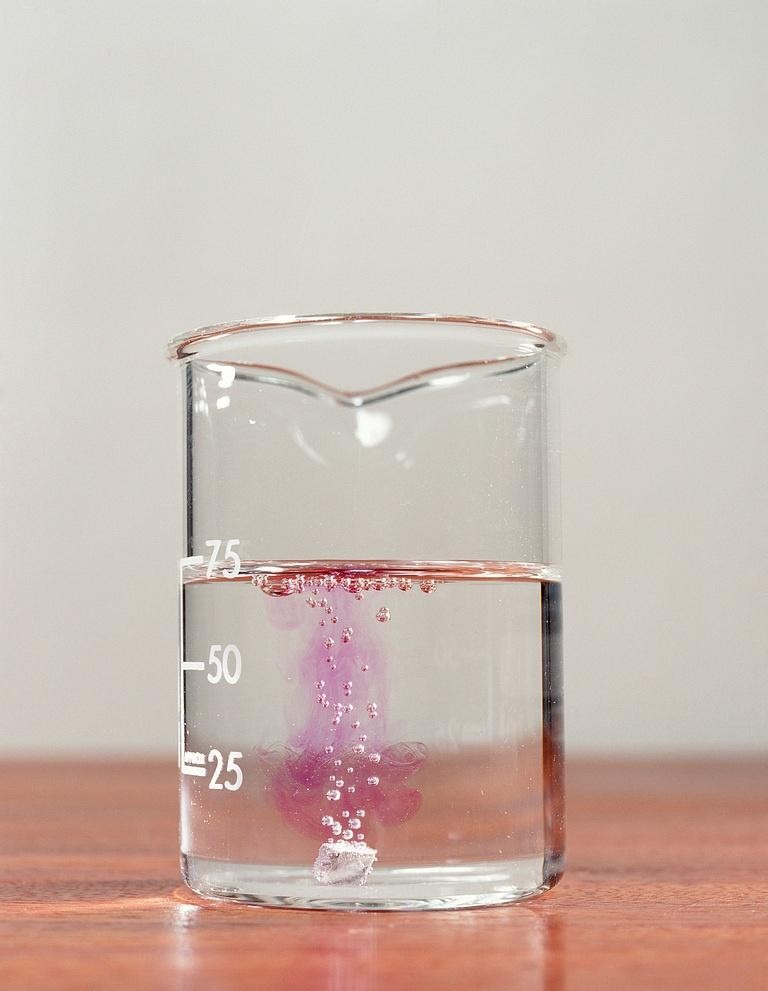} &
\CaseImage{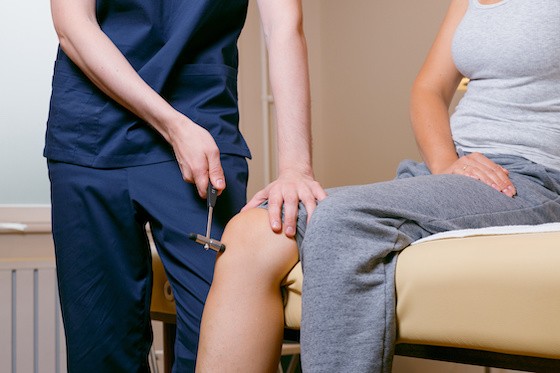} &
\CaseImage{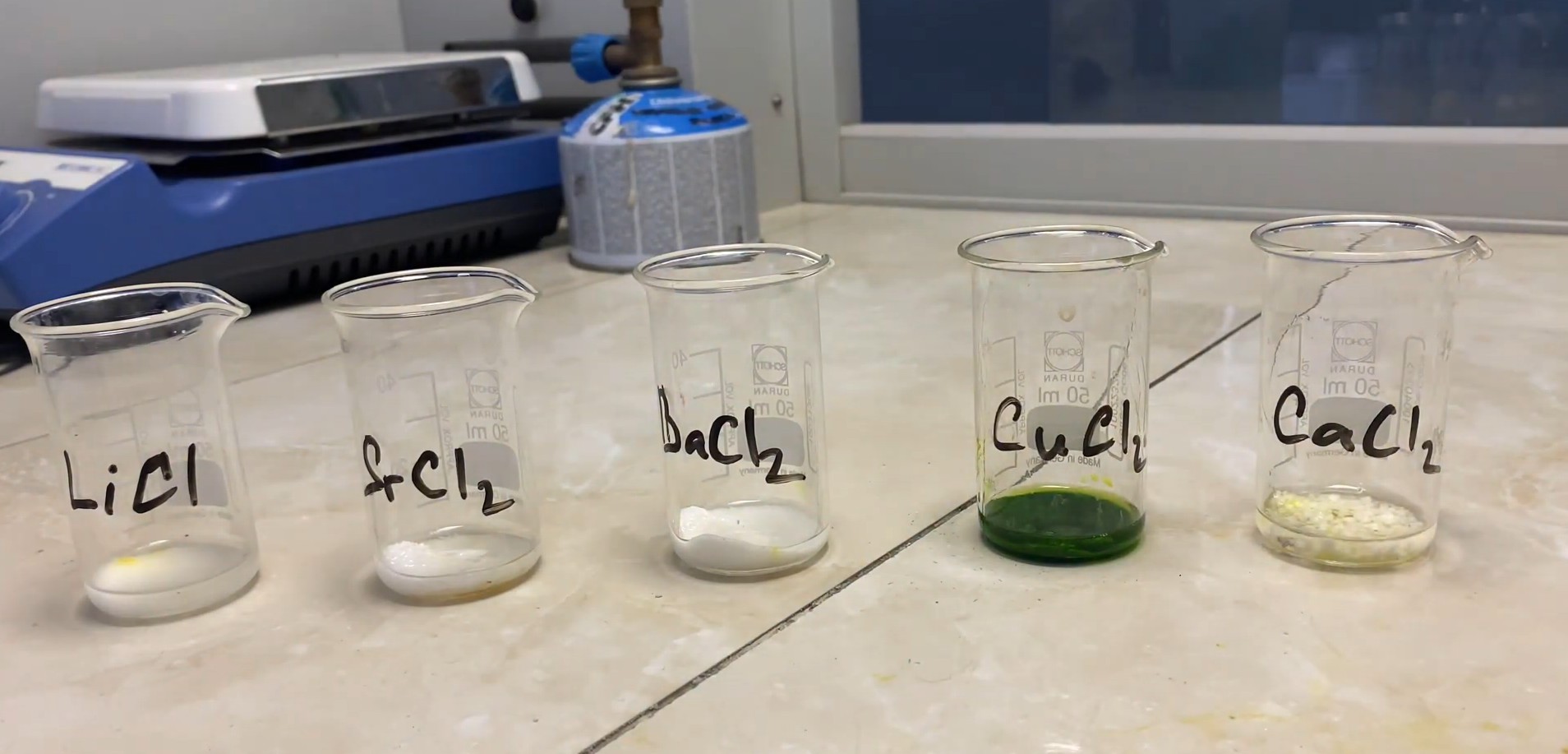} &
\CaseImage{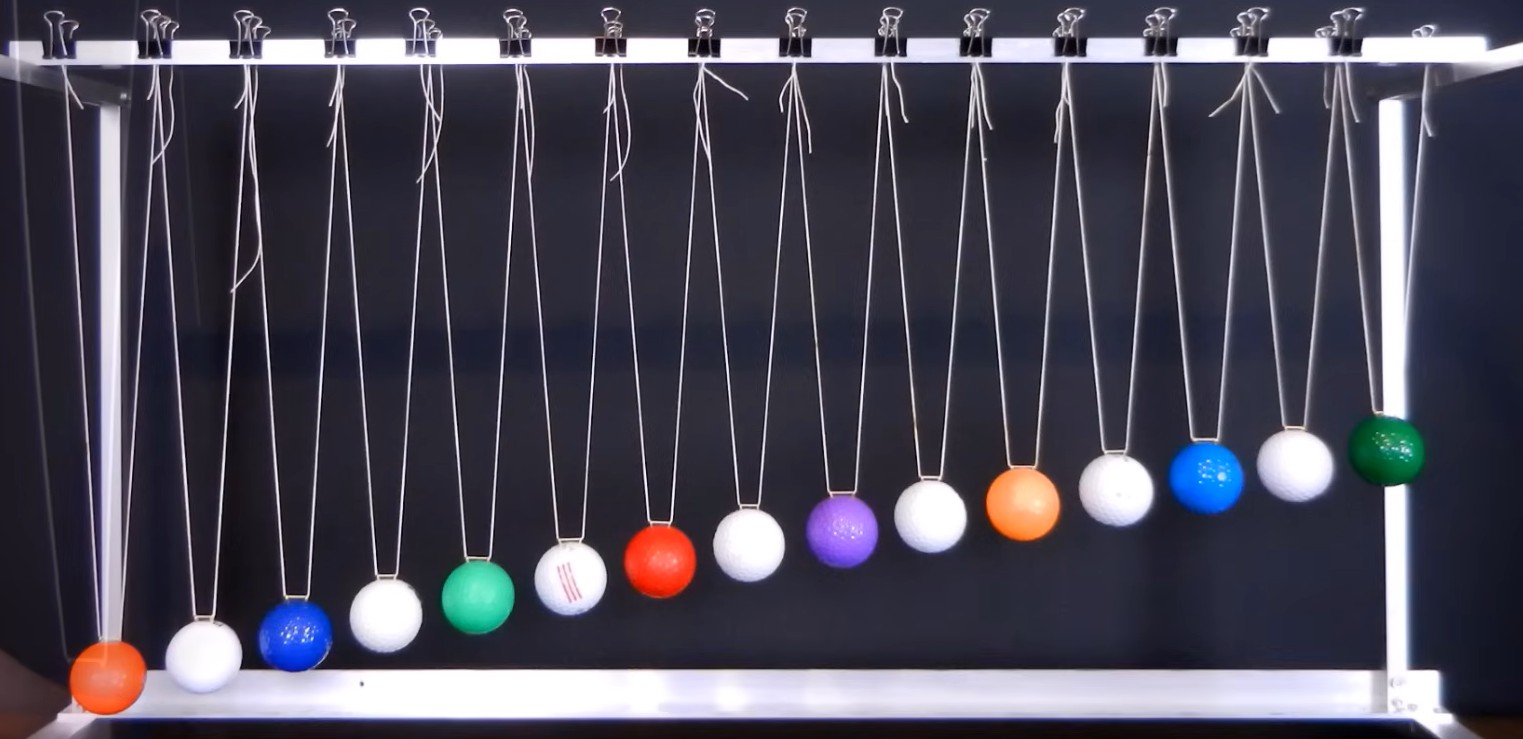} \\[0.0em]

\CasePromptFromJSON{3-a-5} &
\CasePromptFromJSON{3-b-22} &

\CasePromptFromJSON{3-b-2} &
\CasePromptFromJSON{3-b-3} \\[0.01em]
\CaseImage{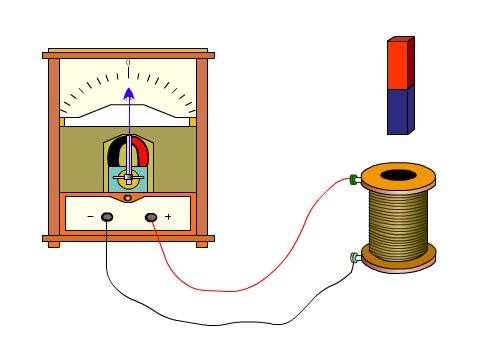} &
\CaseImage{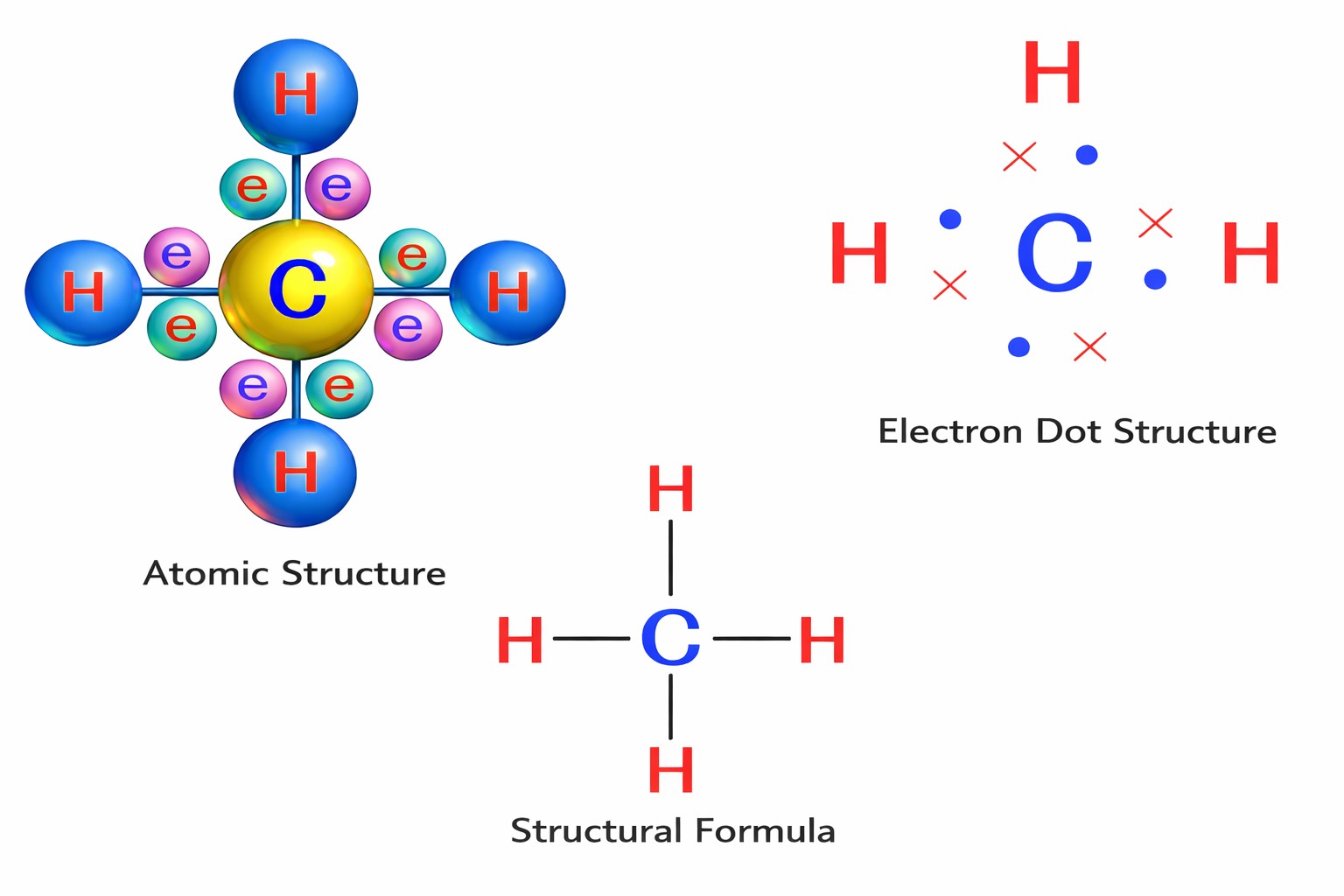} &
\CaseImage{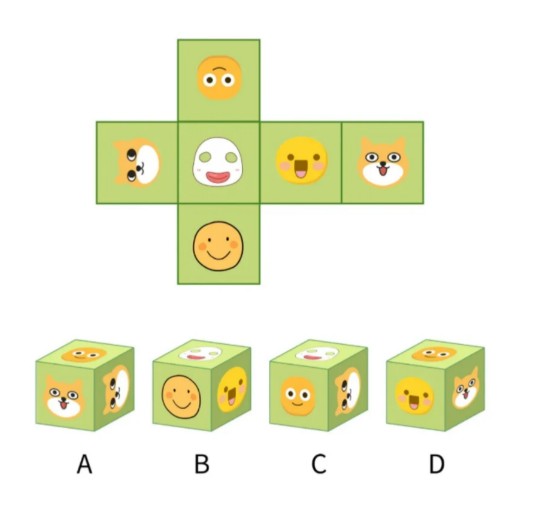} &
\CaseImage{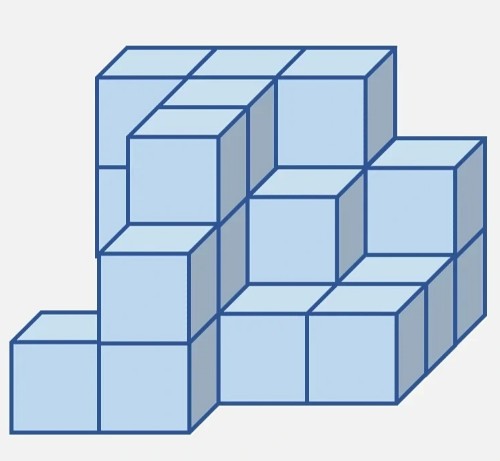} \\[0.0em]

\CasePromptFromJSON{3-b-6} &
\CasePromptFromJSON{3-b-19} &
\CasePromptFromJSON{3-a-28} &
\CasePromptFromJSON{3-b-32} \\[0.01em]
\CaseImage{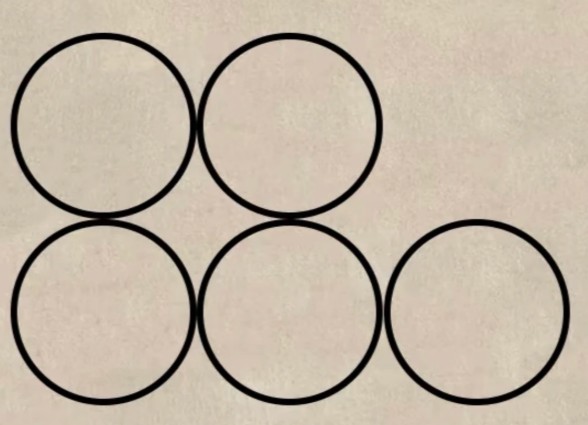} &
\CaseImage{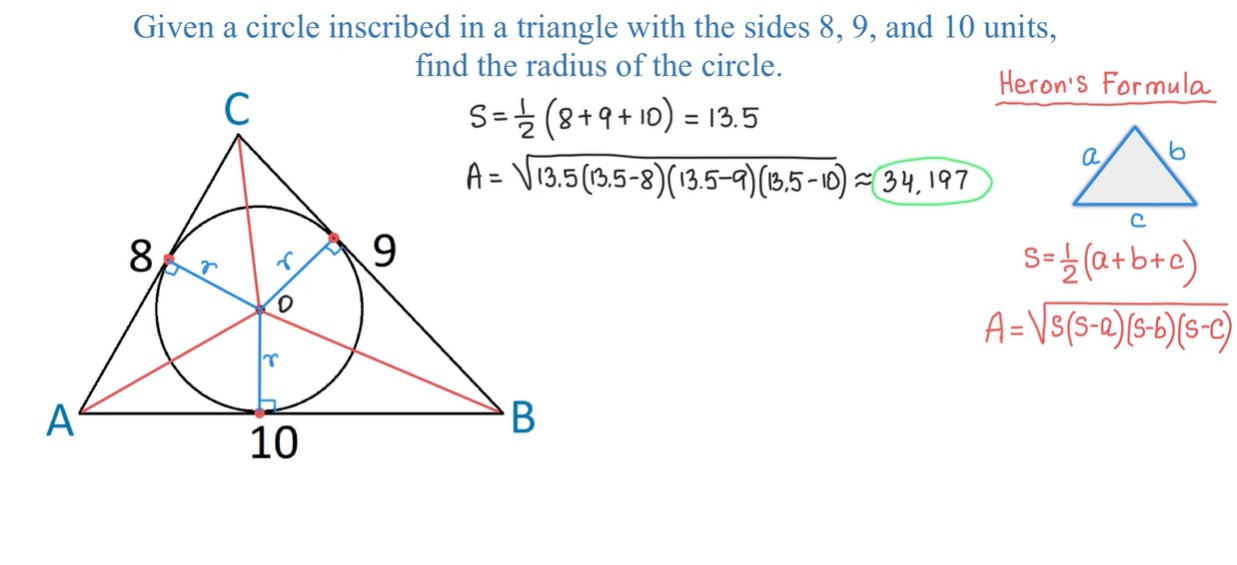} &
\CaseImage{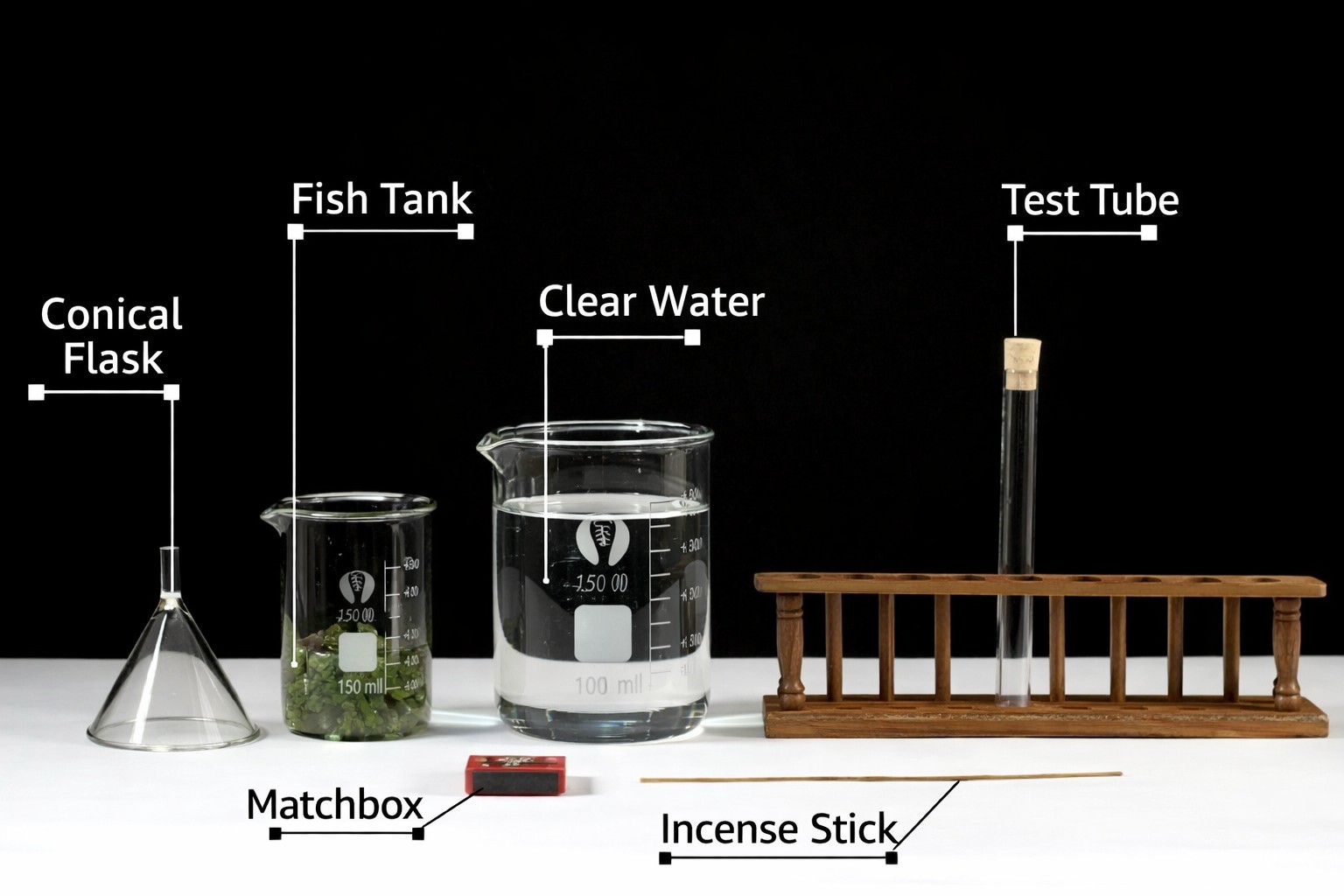} &

\CaseImage{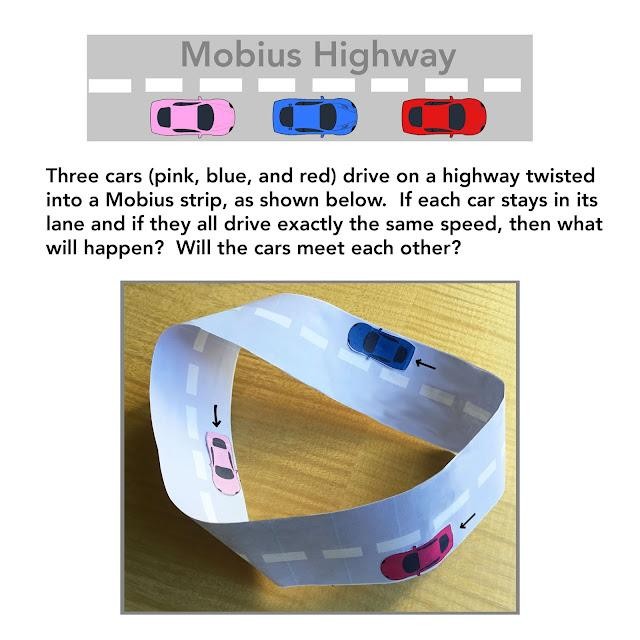} \\[0.0em]

\CasePromptFromJSON{3-c-1} &
\CasePromptFromJSON{3-c-11} &
\CasePromptFromJSON{3-c-13} &
\CasePromptFromJSON{3-c-23} \\[0.01em]
\CaseImage{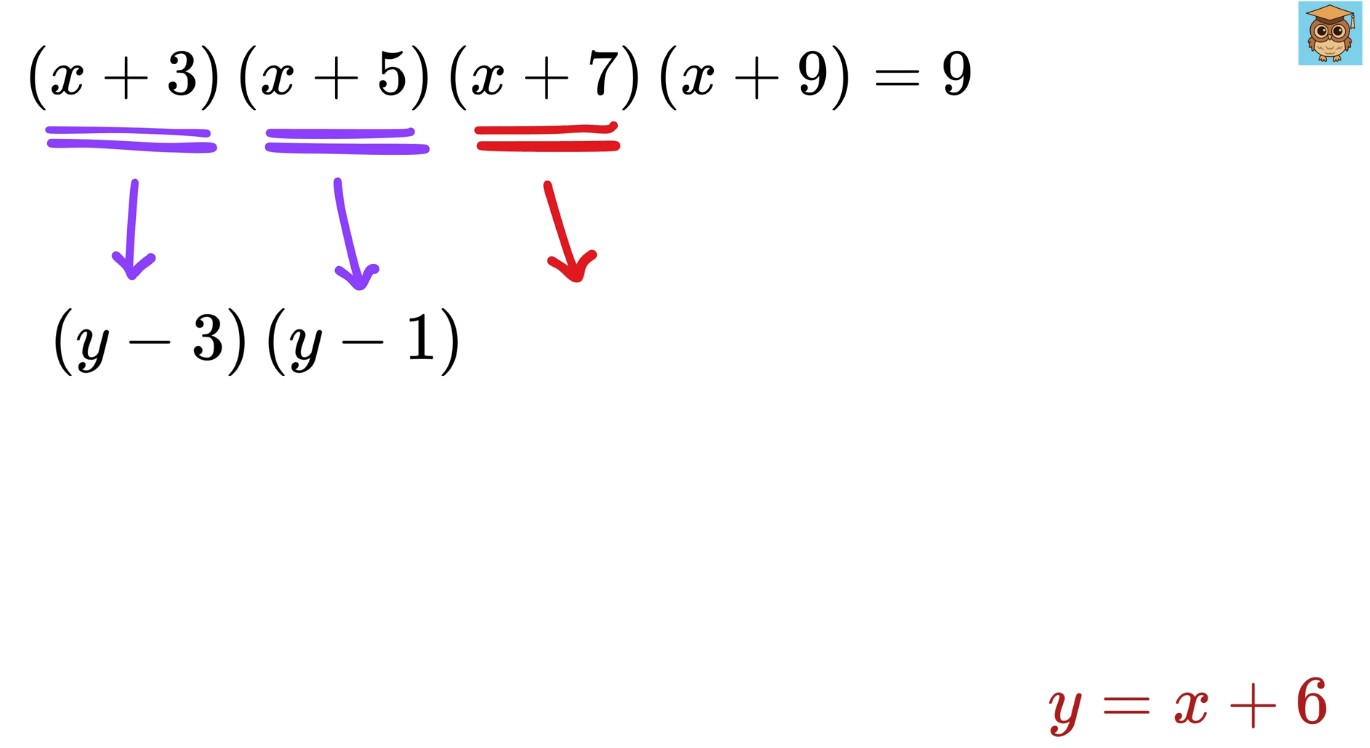} &
\CaseImage{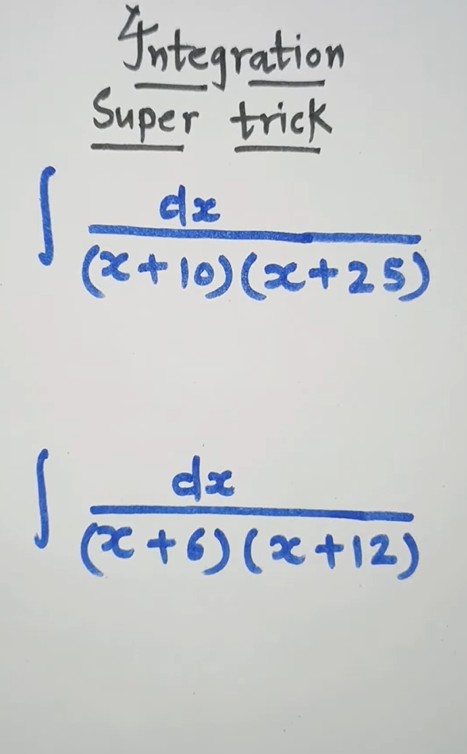} &
\CaseImage{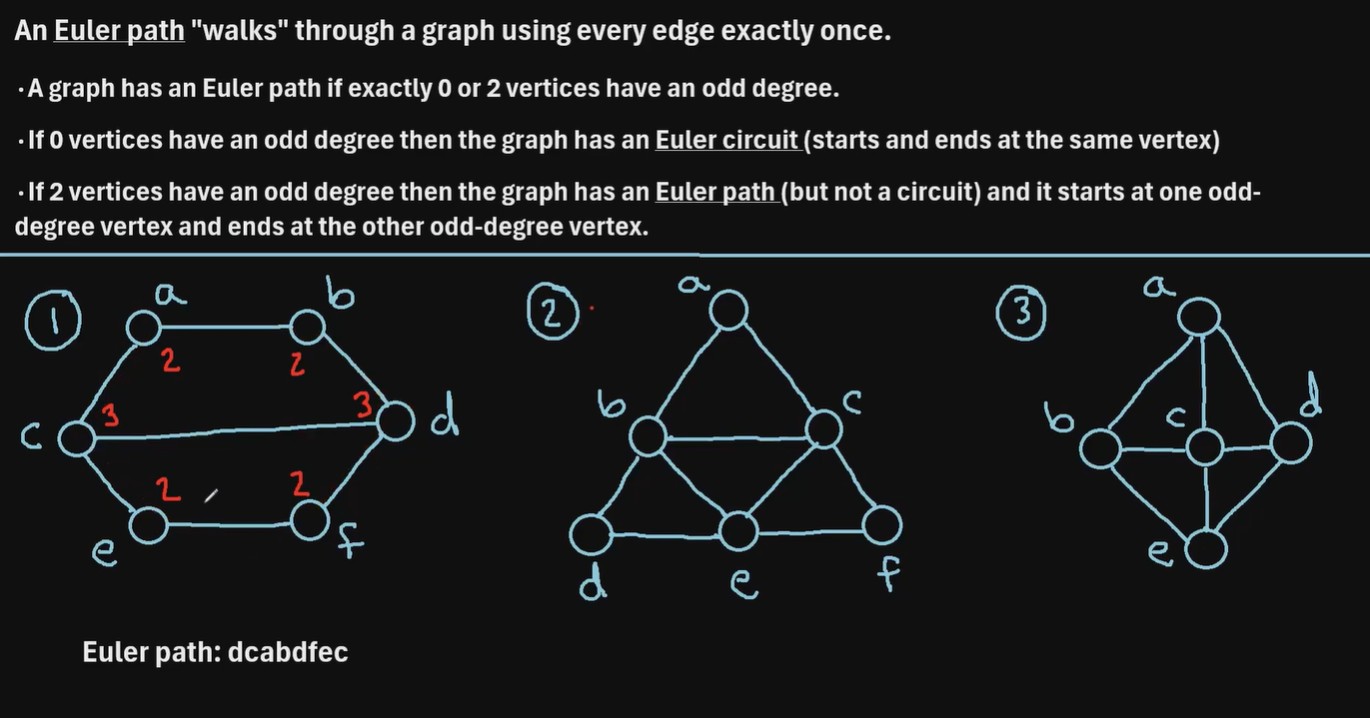} &
\CaseImage{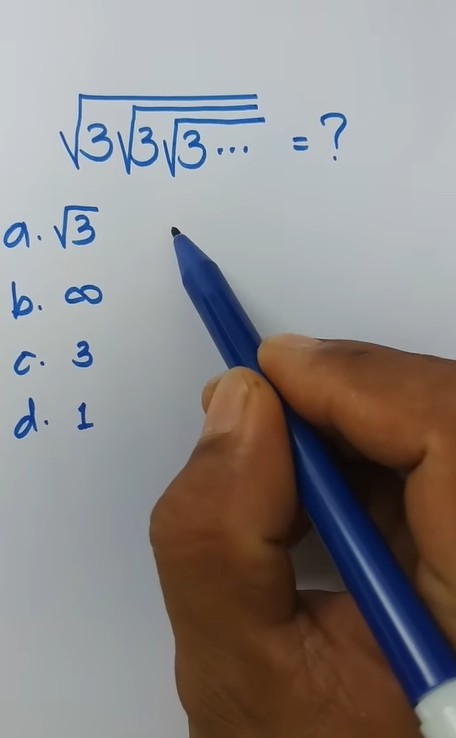} \\[0.0em]

\CasePromptFromJSON{3-c-25} &
\CasePromptFromJSON{3-c-31} &
\CasePromptFromJSON{3-c-36} &
\CasePromptFromJSON{3-d-2} \\[0.01em]
\CaseImage{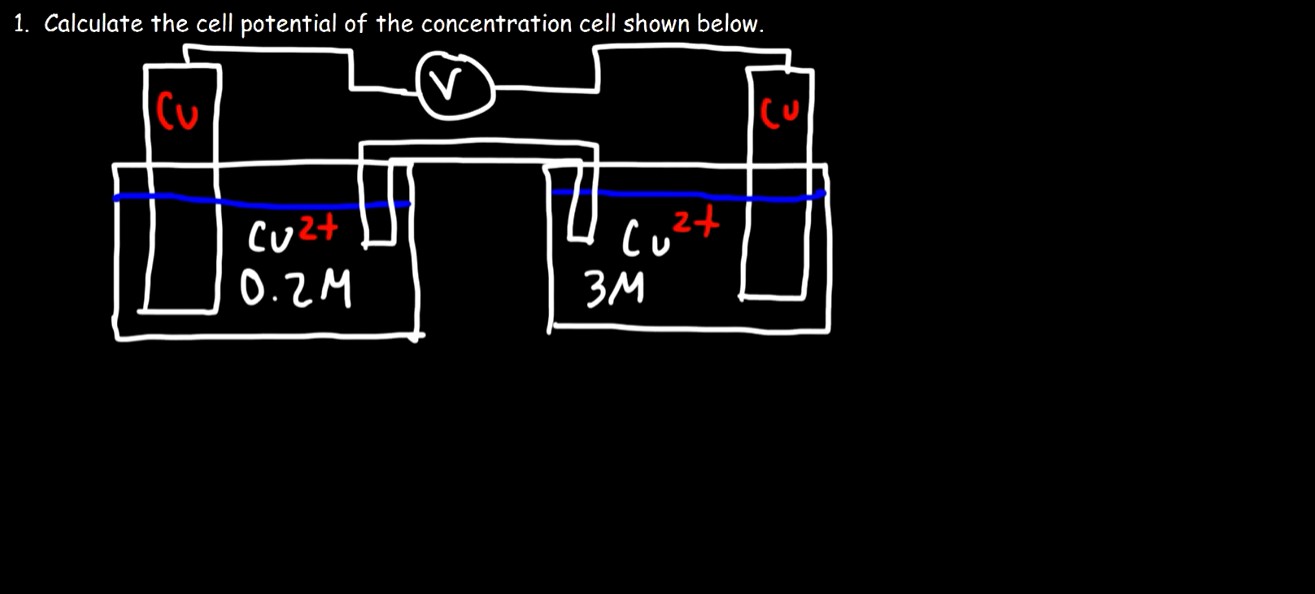} &
\CaseImage{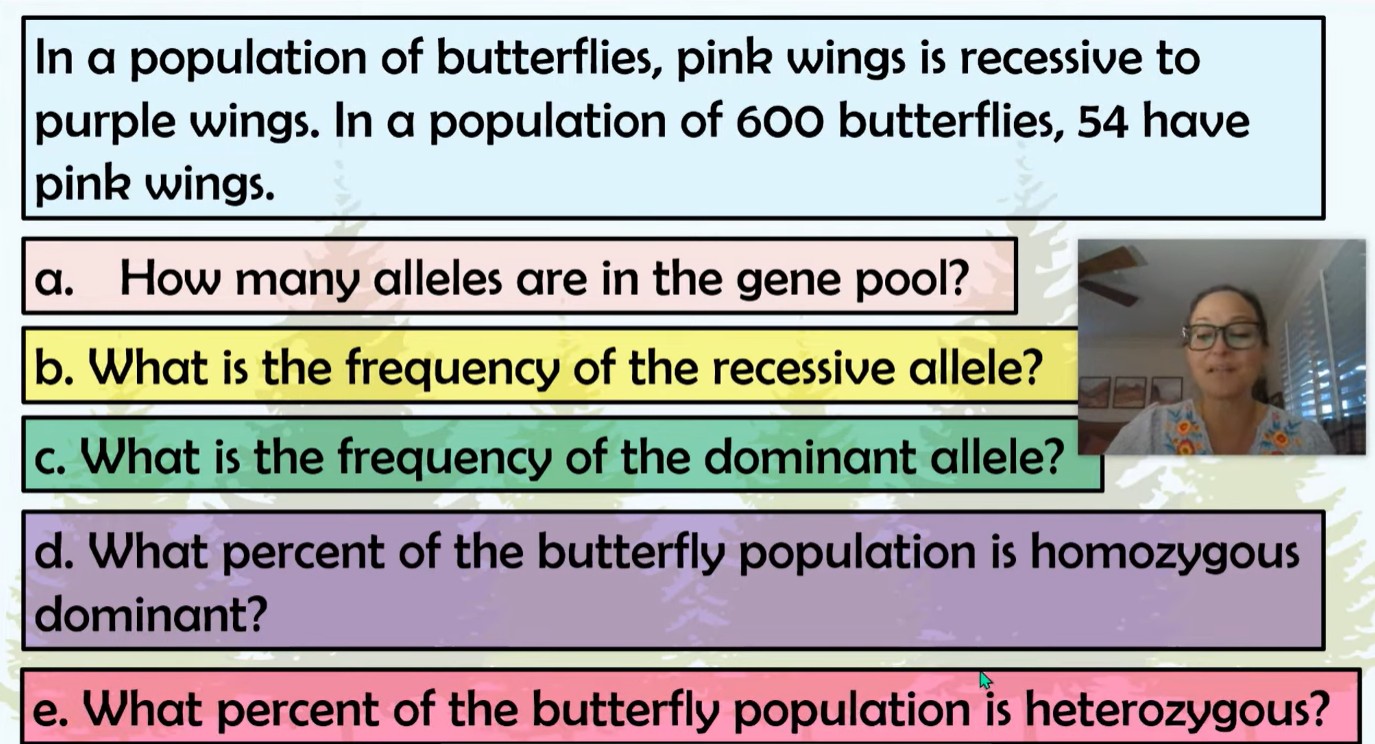} &
\CaseImage{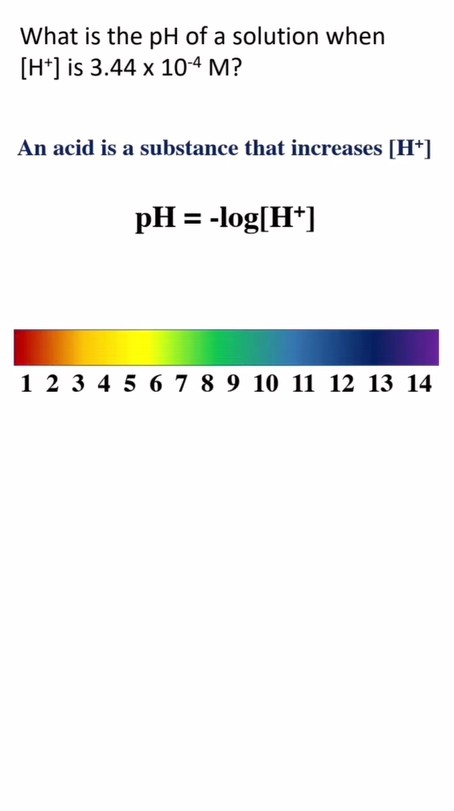} &
\CaseImage{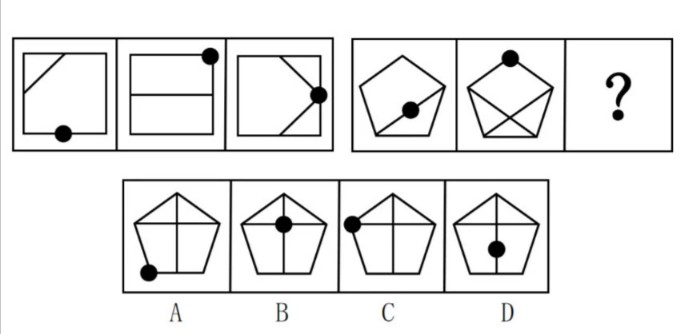} \\[0.0em]

\CasePromptFromJSON{3-d-6} &
\CasePromptFromJSON{3-d-11} &
\CasePromptFromJSON{3-d-12} &
\CasePromptFromJSON{3-d-22} \\[0.01em]
\CaseImage{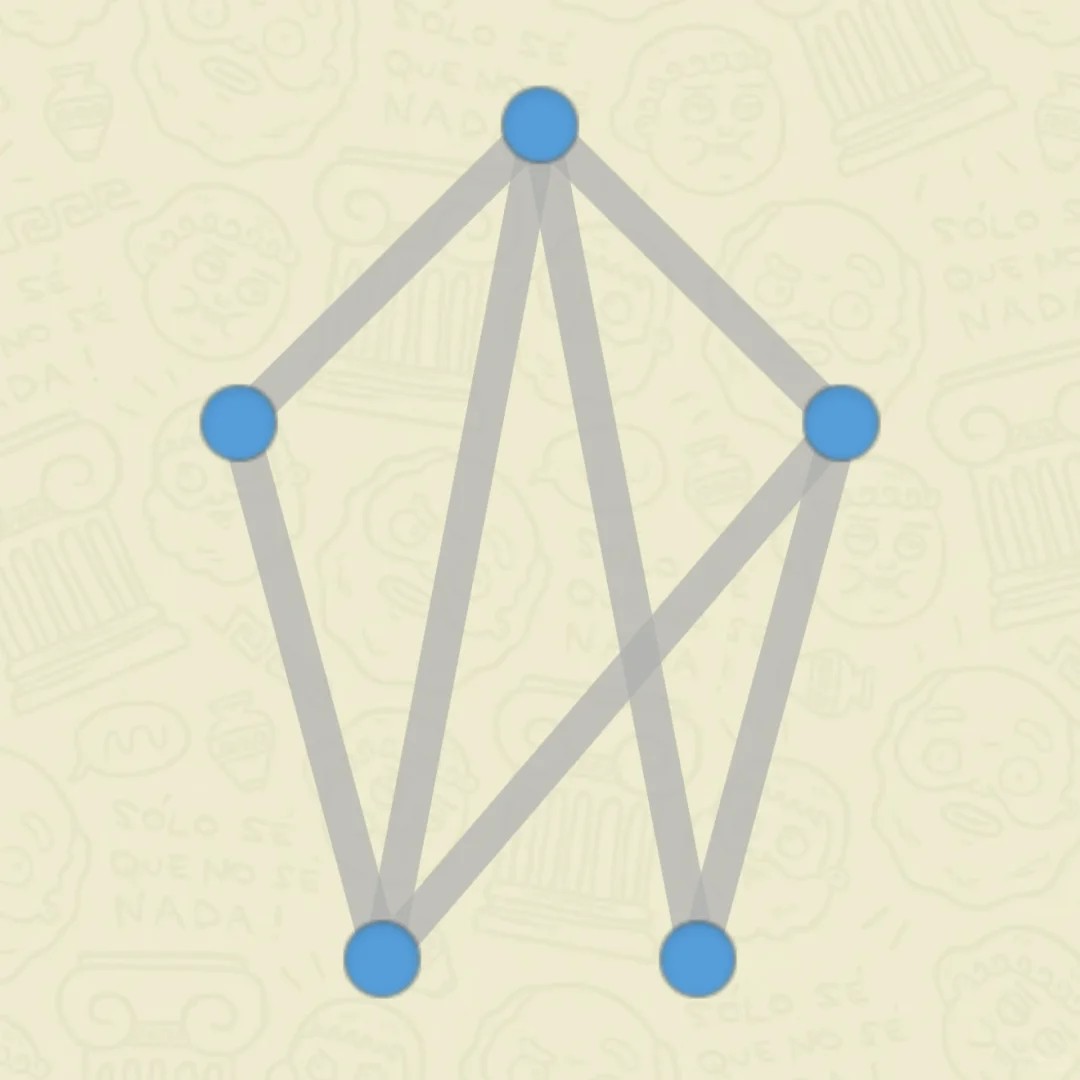} &
\CaseImage{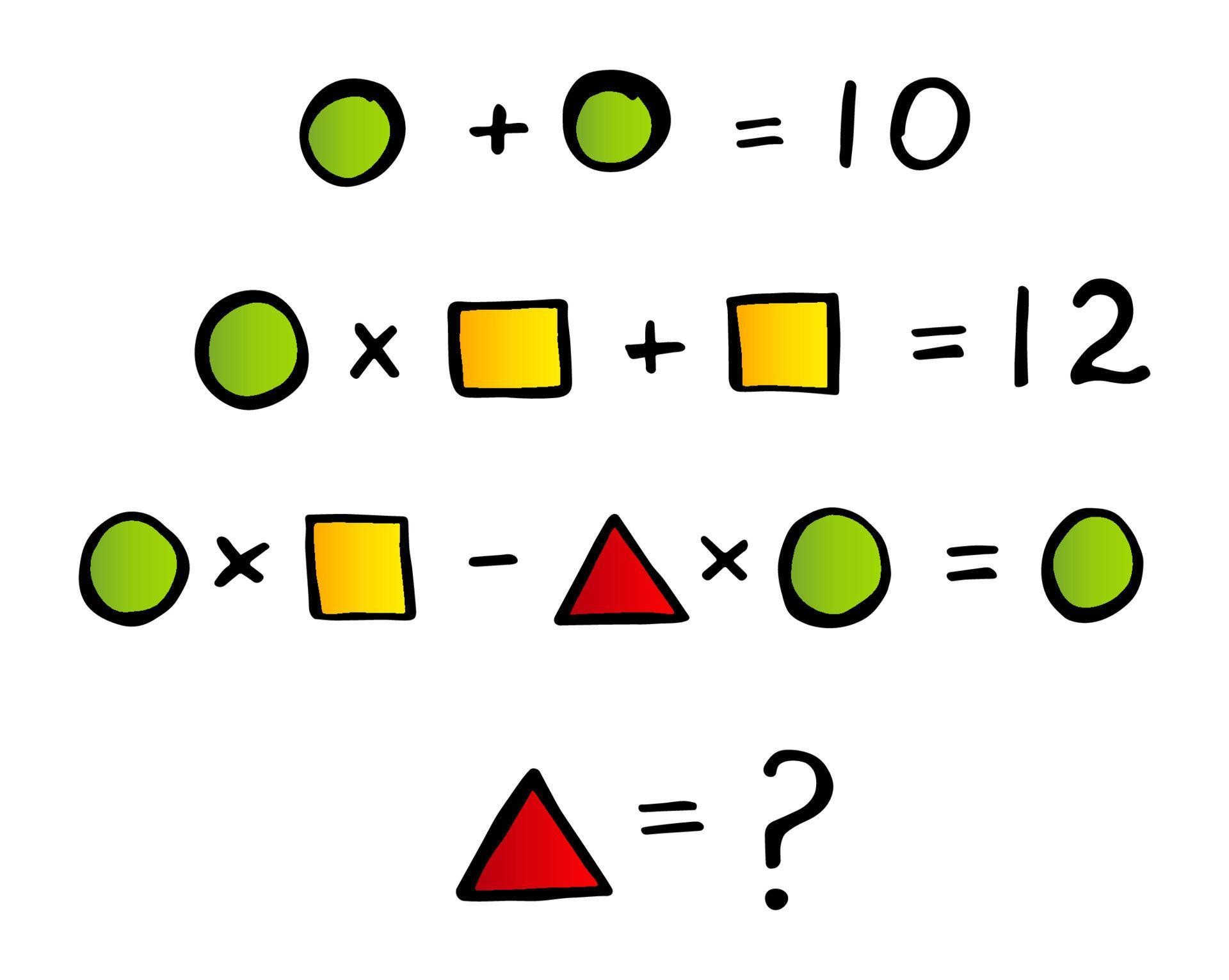} &
\CaseImage{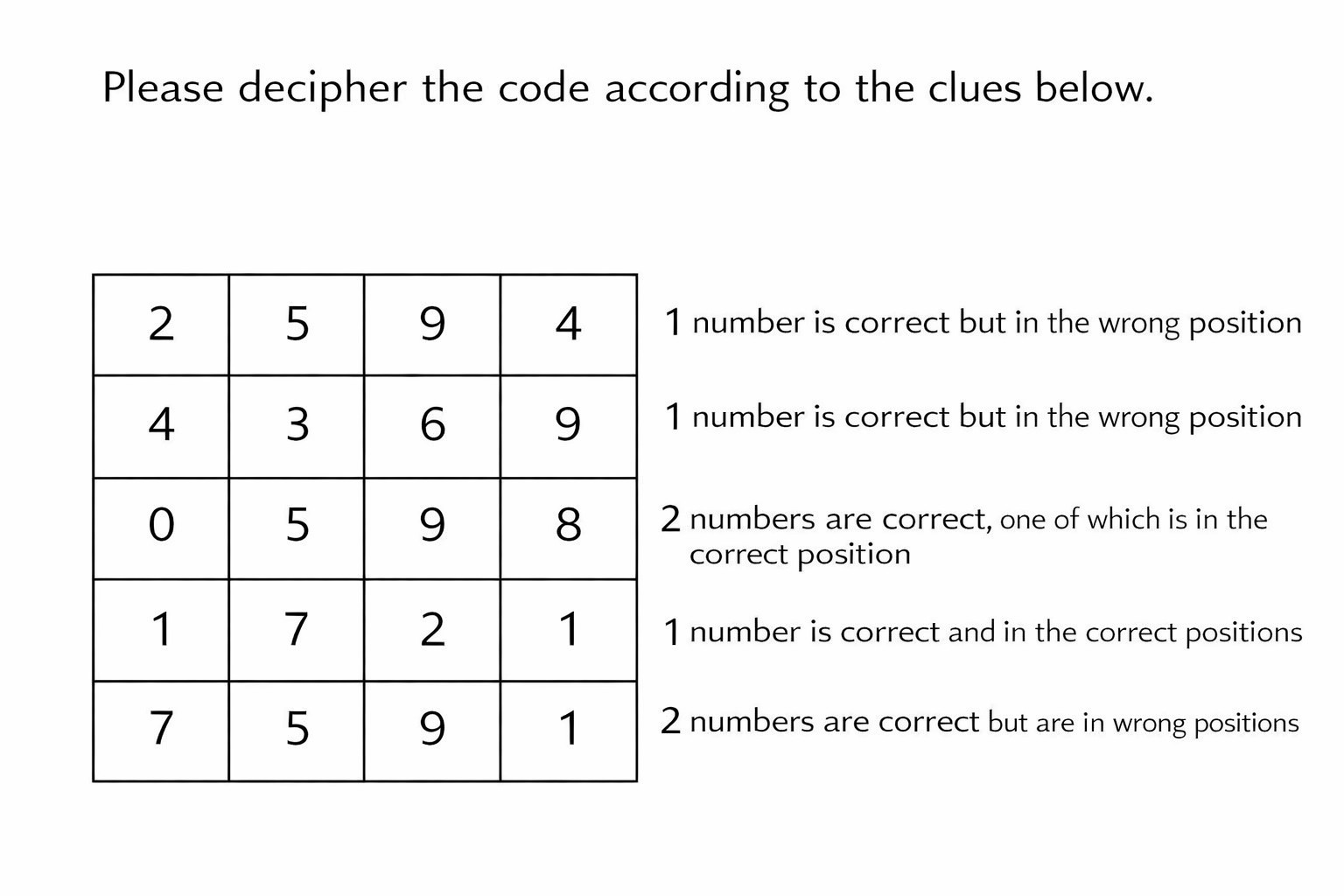} &
\CaseImage{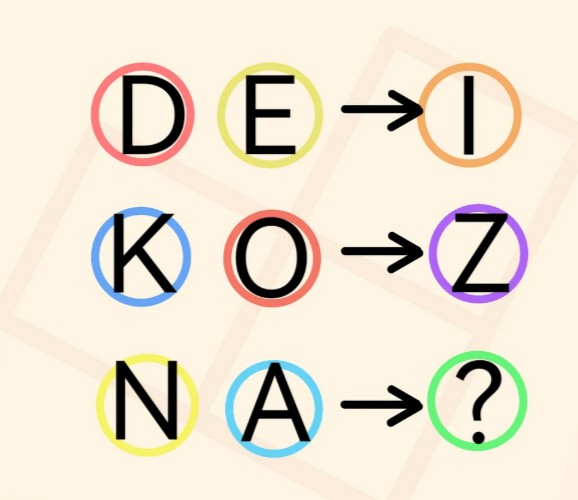} \\[-0.2em]

\end{tabularx}
\end{CASEBAX}
\end{minipage}
\caption{Representative examples from the Logic Reasoning category. These cases cover quantitative math, spatial geometry, experimental science, logic puzzles, and pattern discovery, evaluating whether models can maintain rule-based, spatial, symbolic, and causal relationships across generated frames.}
\label{tab:representative-logic-reasoning-examples}
\end{table*}

\begin{table*}[htp]
\centering
\begin{minipage}{\textwidth}
\begin{PairBox}{Information-Based Reasoning}

\begin{tabularx}{\textwidth}{@{}XXXX@{}}
\CasePromptFromJSON{4-a-1} &
\CasePromptFromJSON{4-a-5} &
\CasePromptFromJSON{4-a-7} &
\CasePromptFromJSON{4-a-3} \\[0.01em]
\CaseImage{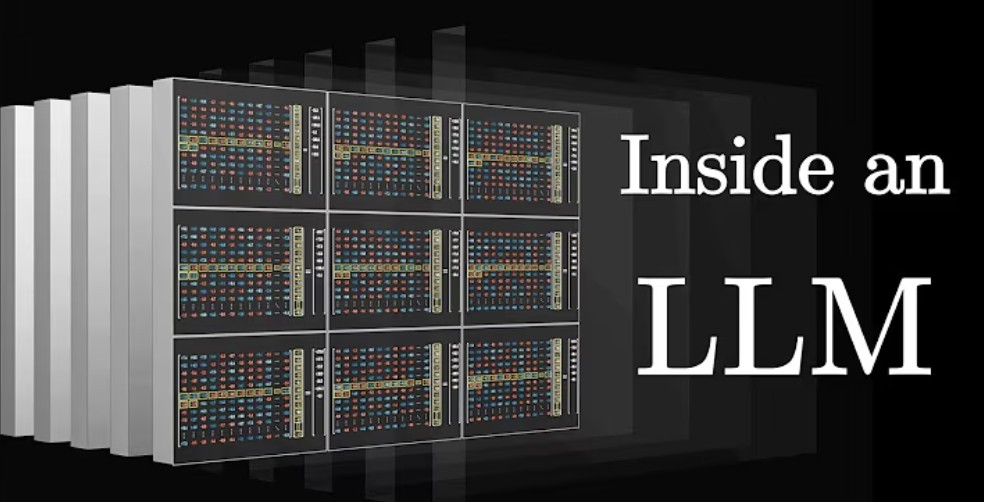} &
\CaseImage{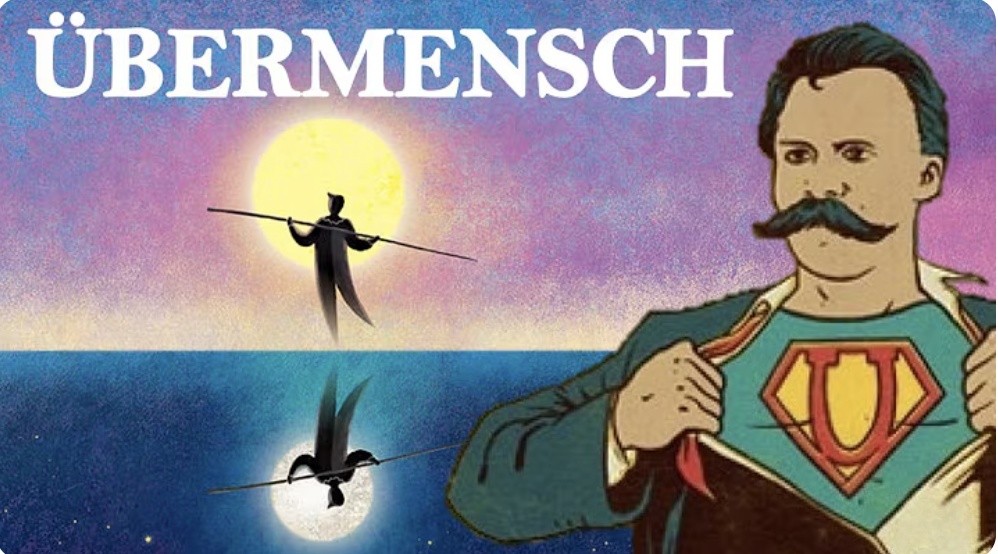} &
\CaseImage{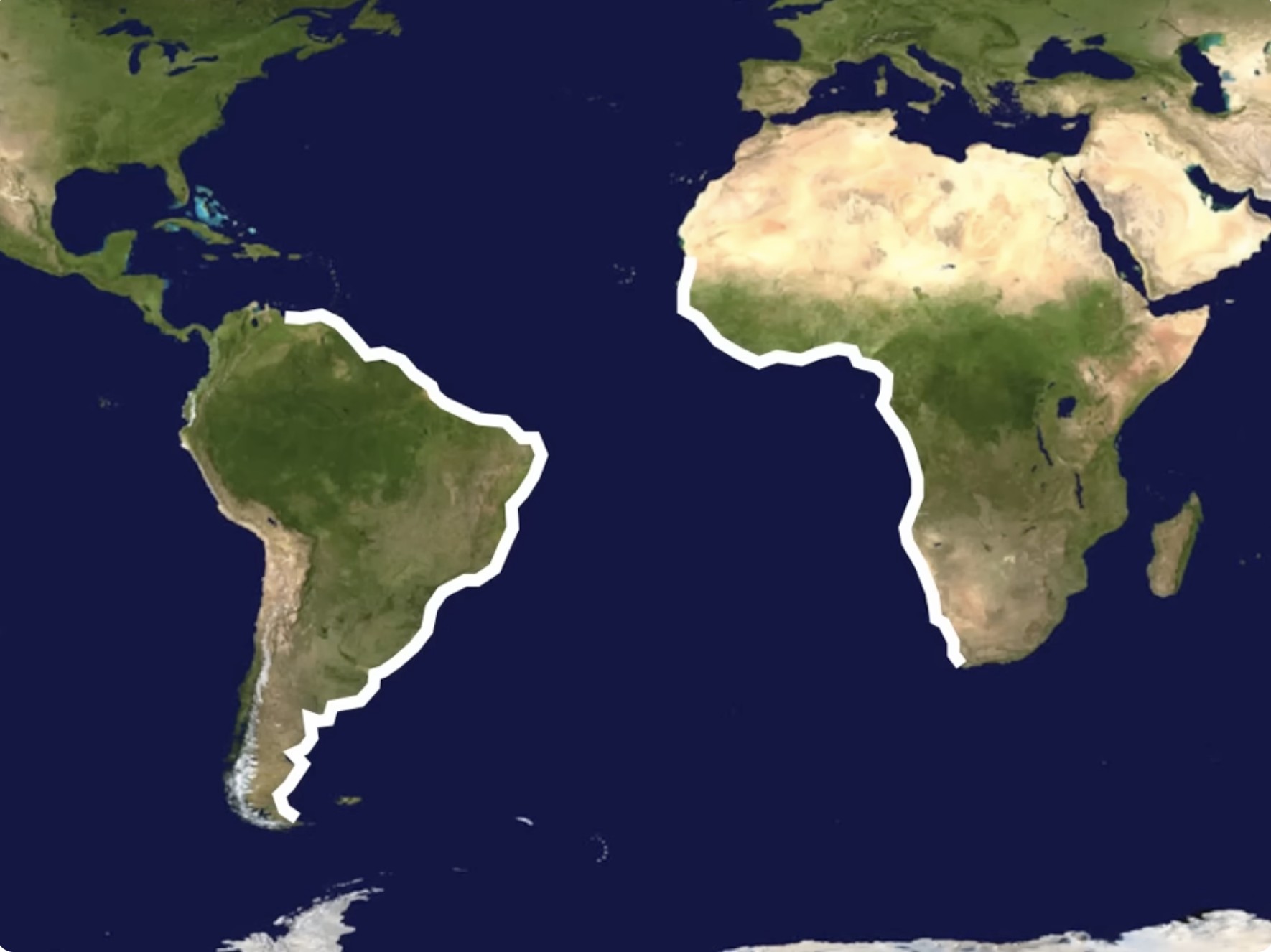} &
\CaseImage{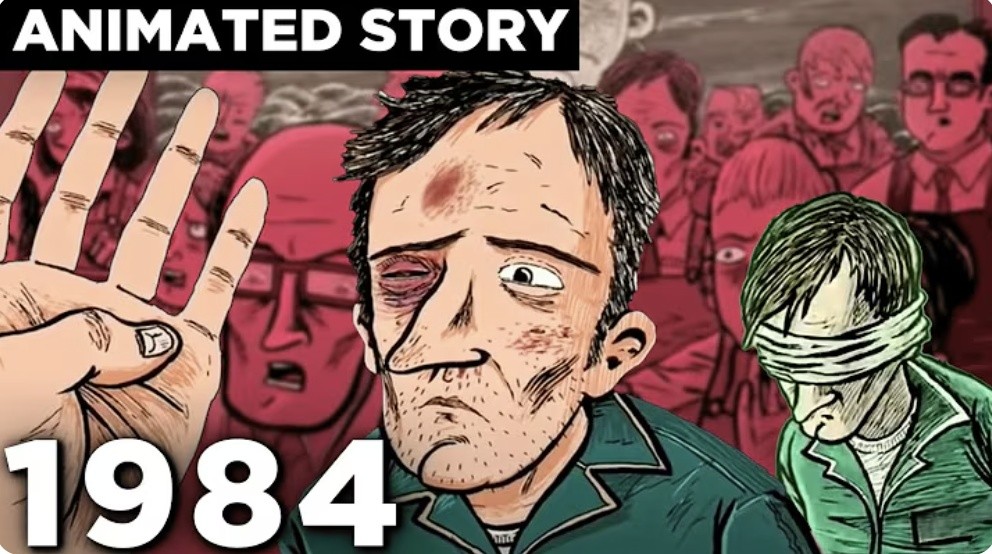} \\[0.0em]

\CasePromptFromJSON{4-b-26} &
\CasePromptFromJSON{4-b-29} &
\CasePromptFromJSON{4-b-31} &
\CasePromptFromJSON{4-b-39} \\[0.01em]
\CaseImage{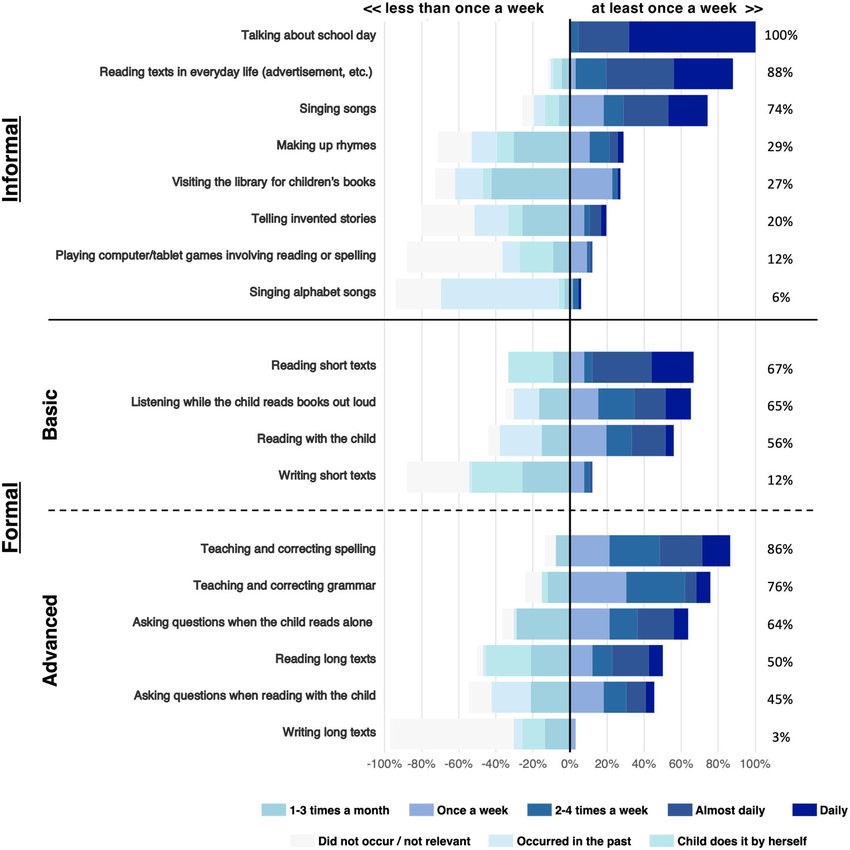} &
\CaseImage{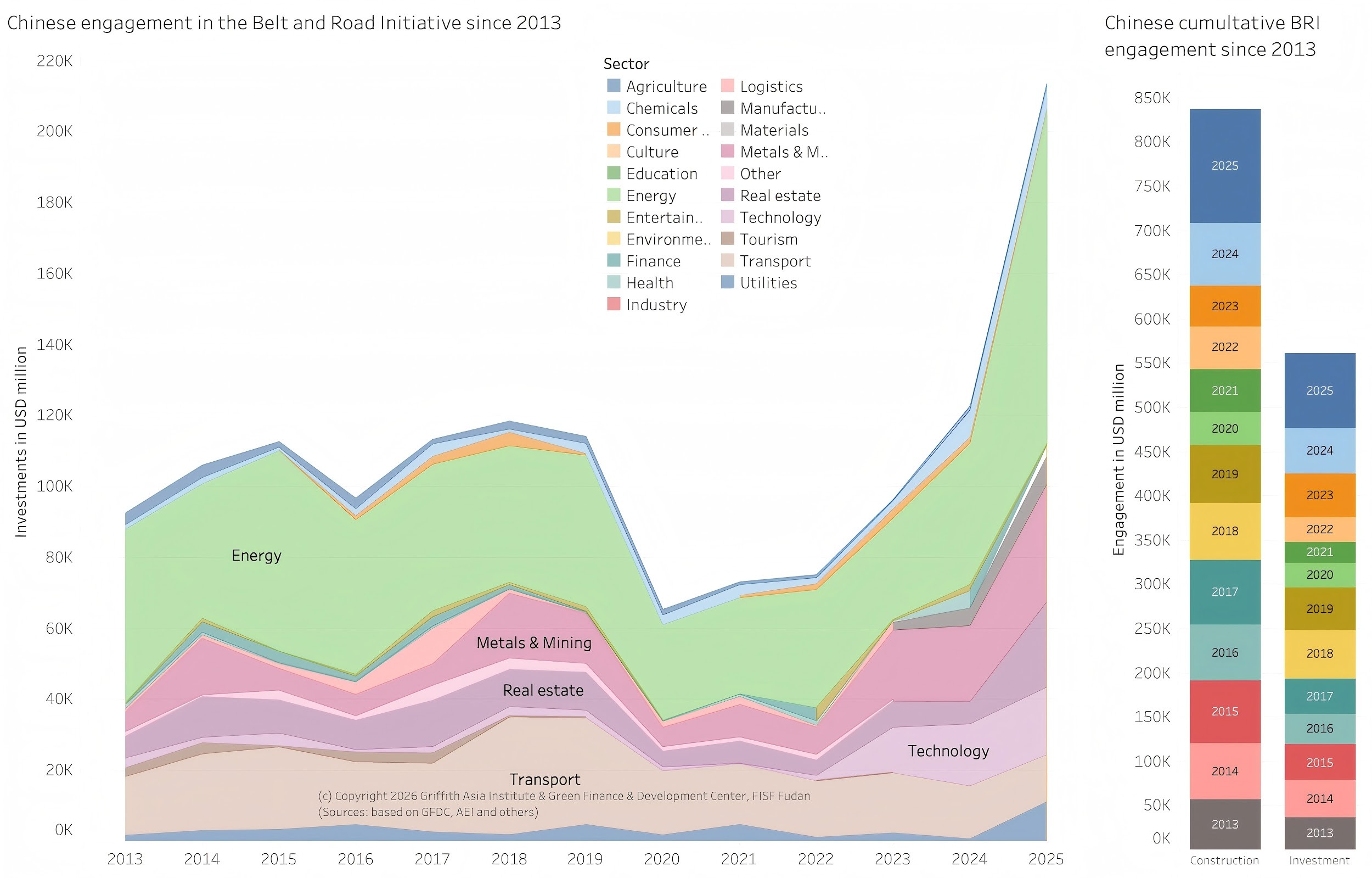} &
\CaseImage{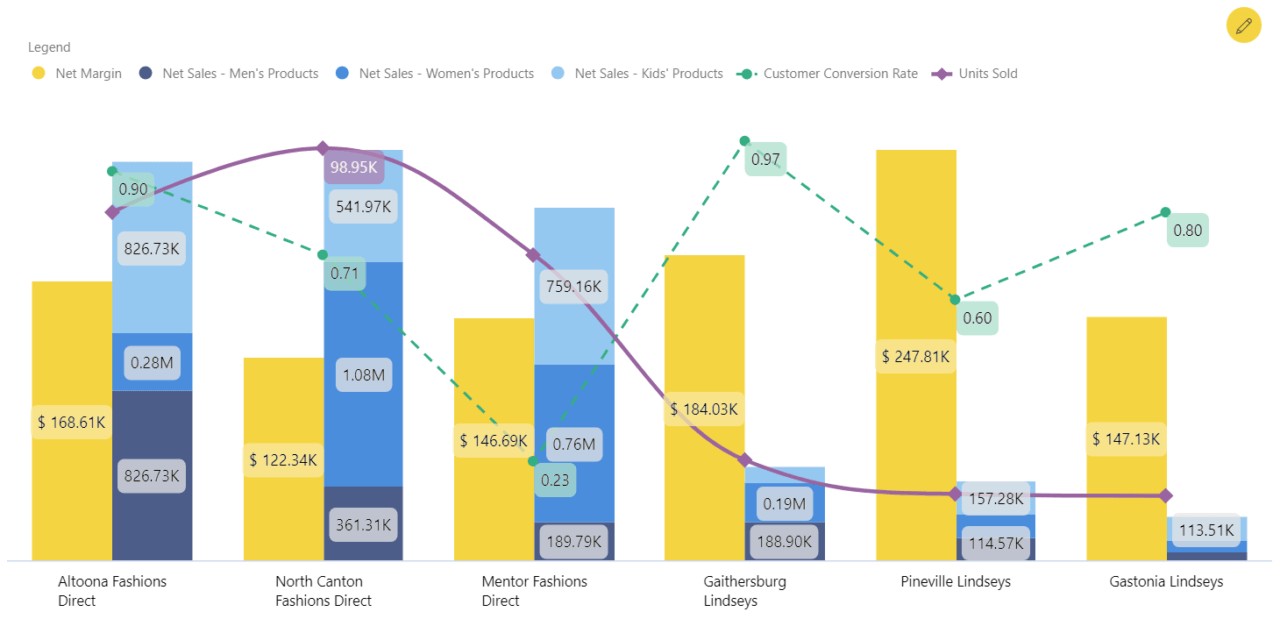} &
\CaseImage{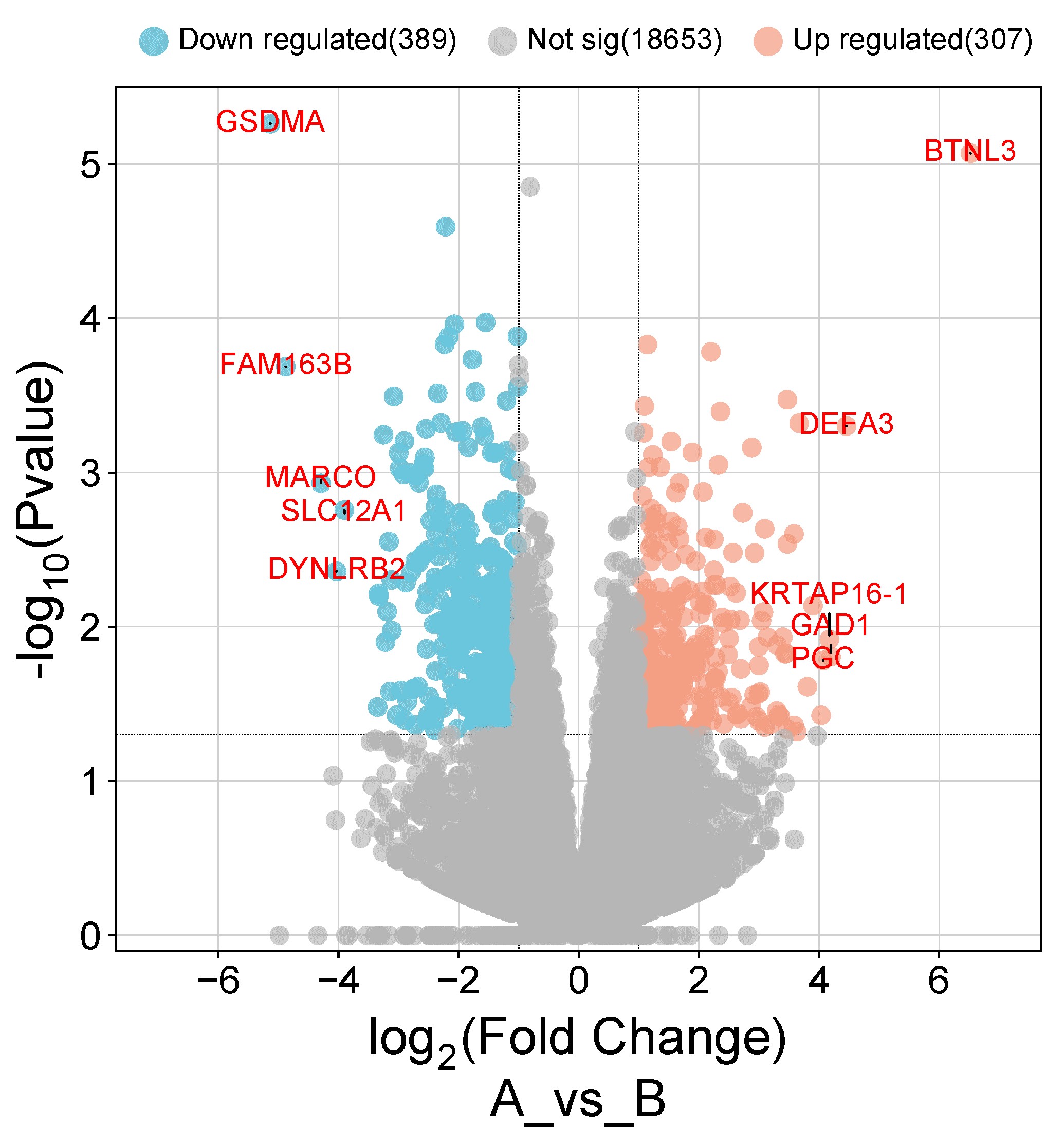} \\[0.0em]

\CasePromptFromJSON{4-b-16} &
\CasePromptFromJSON{4-b-01} &
\CasePromptFromJSON{4-b-07} &
\CasePromptFromJSON{4-b-08} \\[0.01em]
\CaseImage{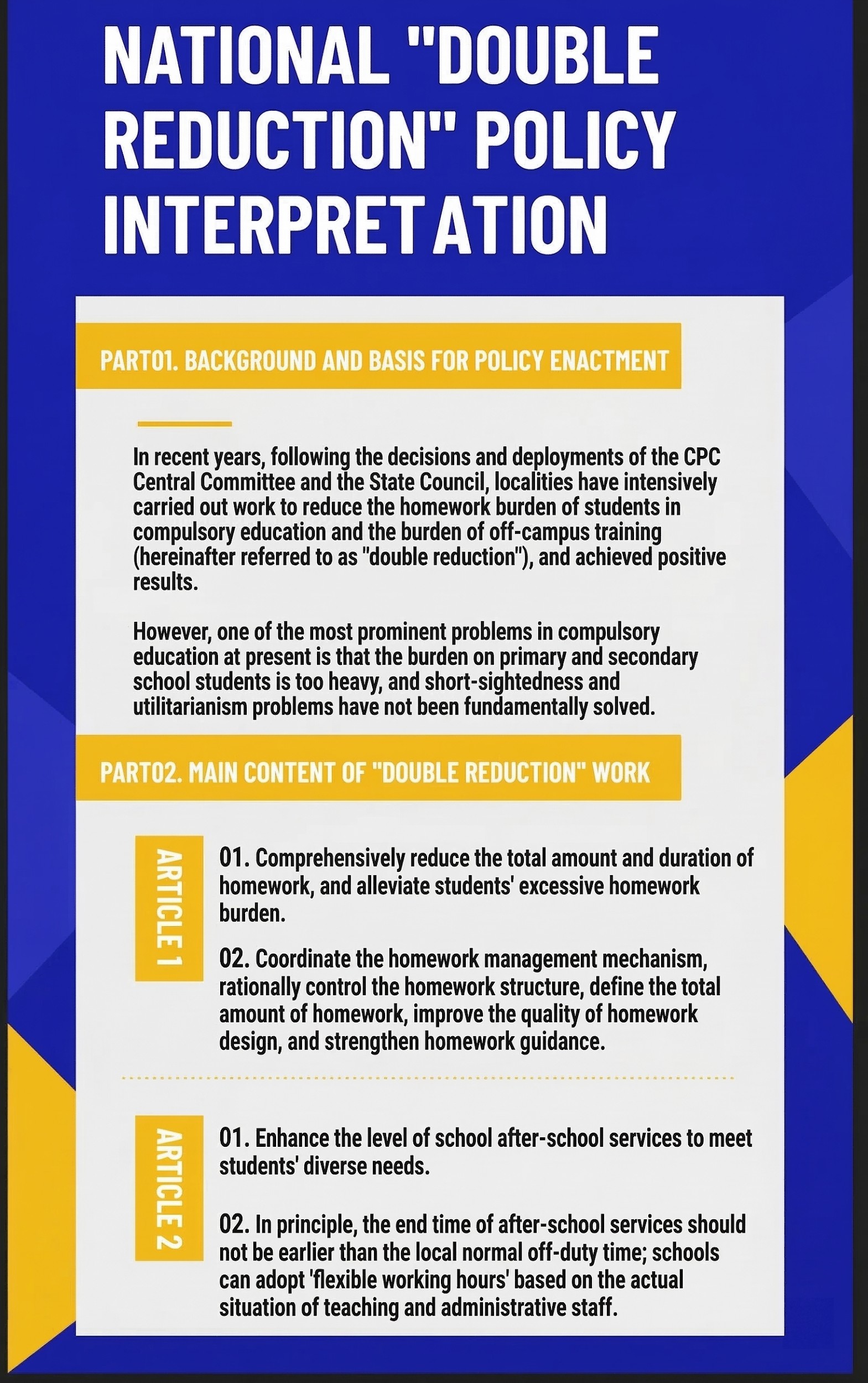} &
\CaseImage{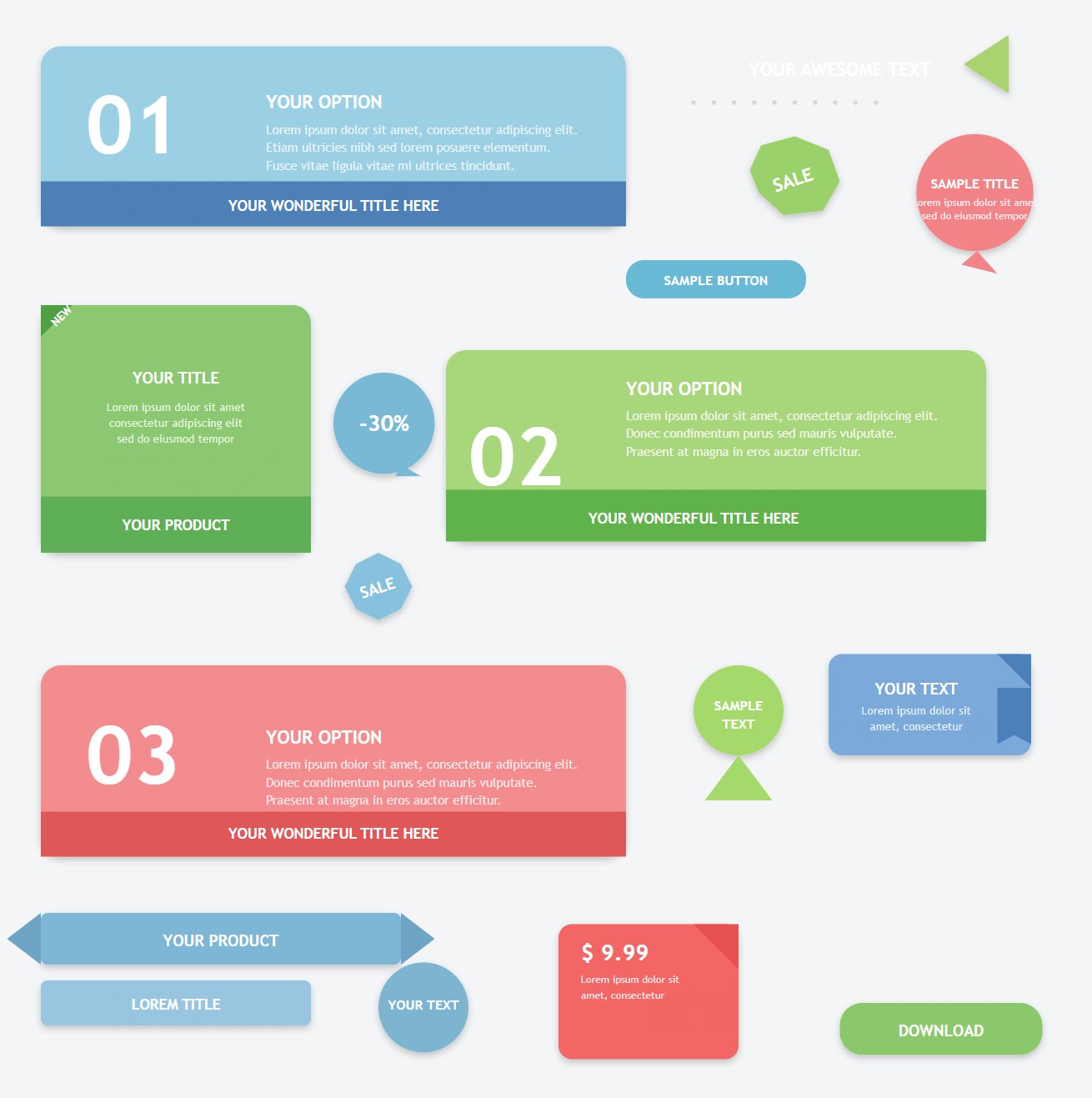} &
\CaseImage{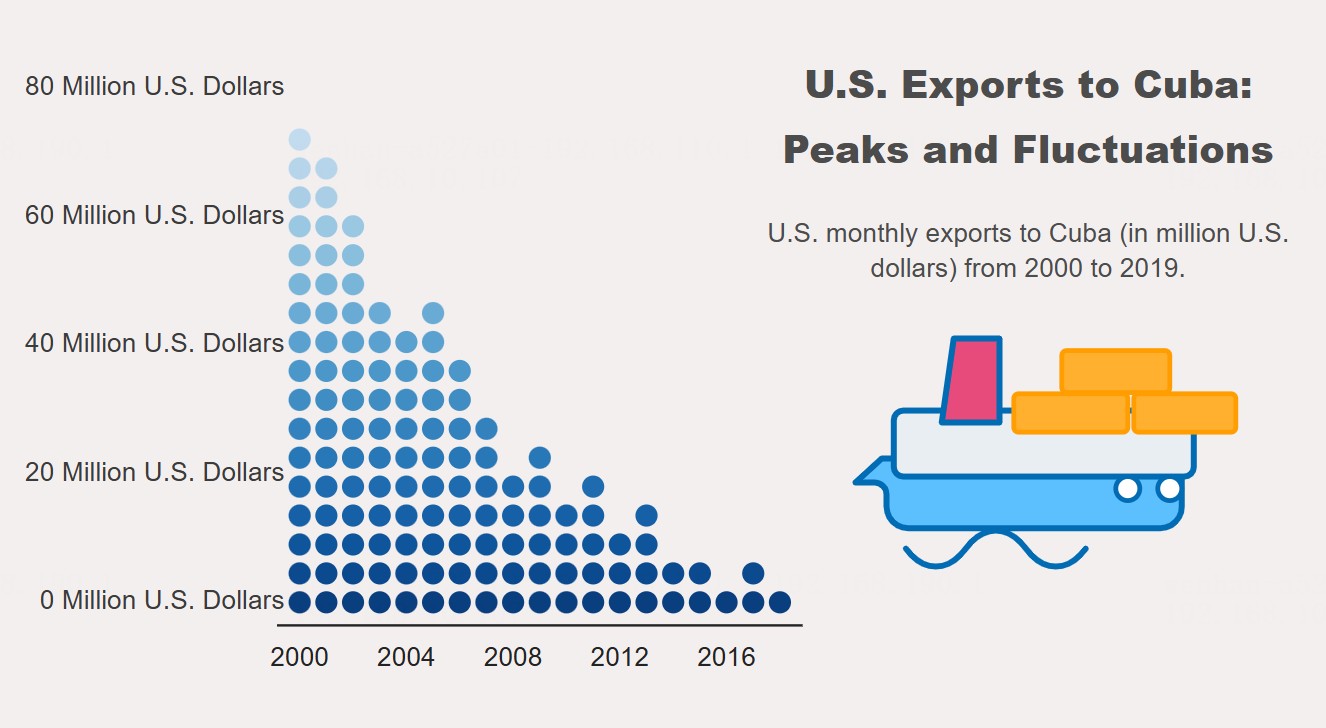} &
\CaseImage{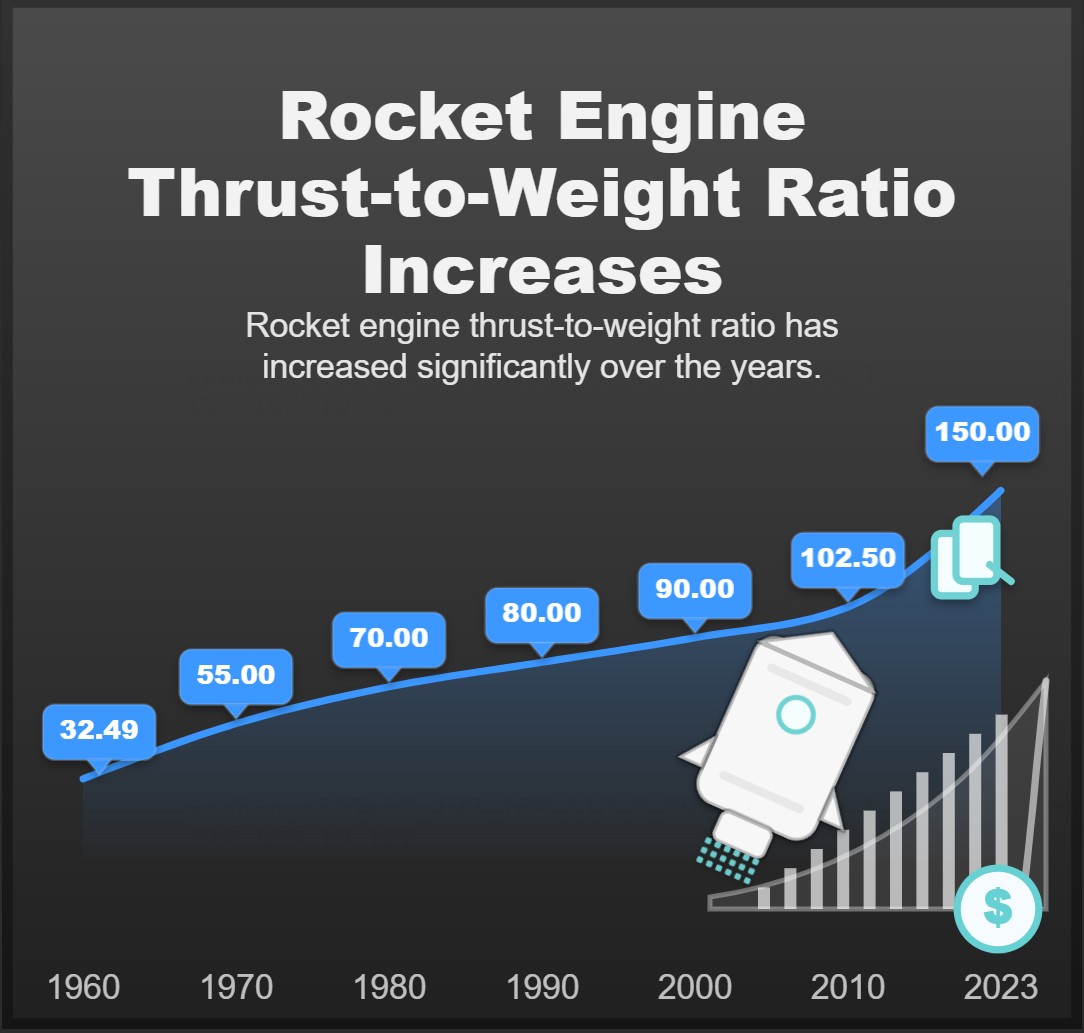} \\[0.0em]

\CasePromptFromJSON{4-c-1} &
\CasePromptFromJSON{4-c-2} &
\CasePromptFromJSON{4-c-19} &
\CasePromptFromJSON{4-c-21} \\[0.01em]
\CaseImage{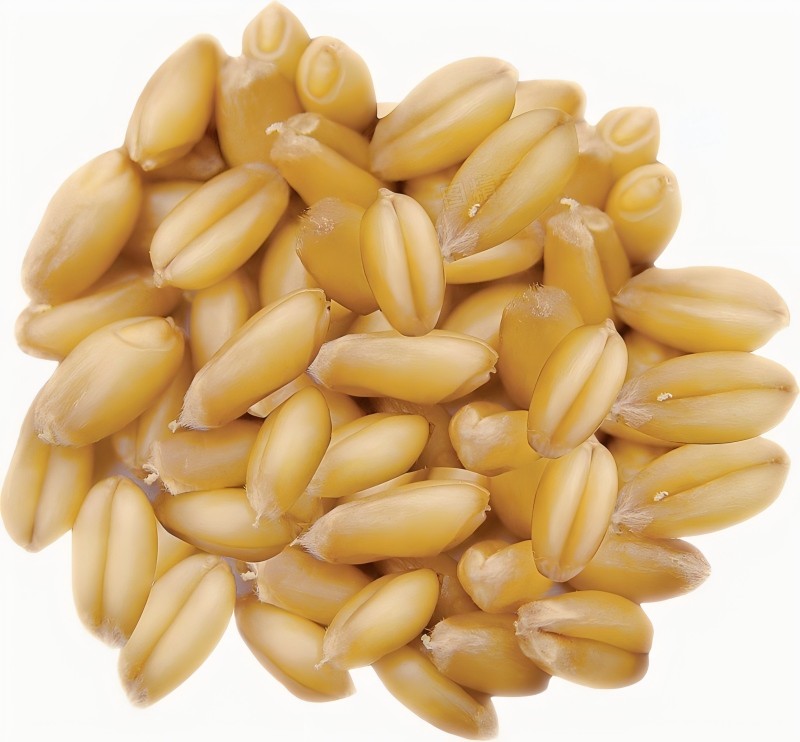} &
\CaseImage{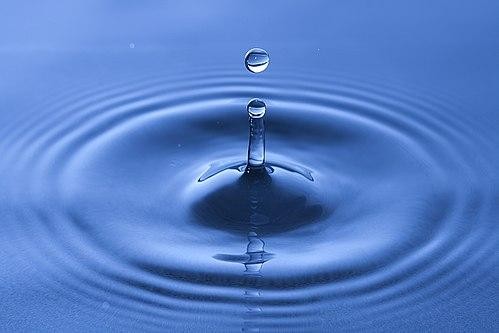} &
\CaseImage{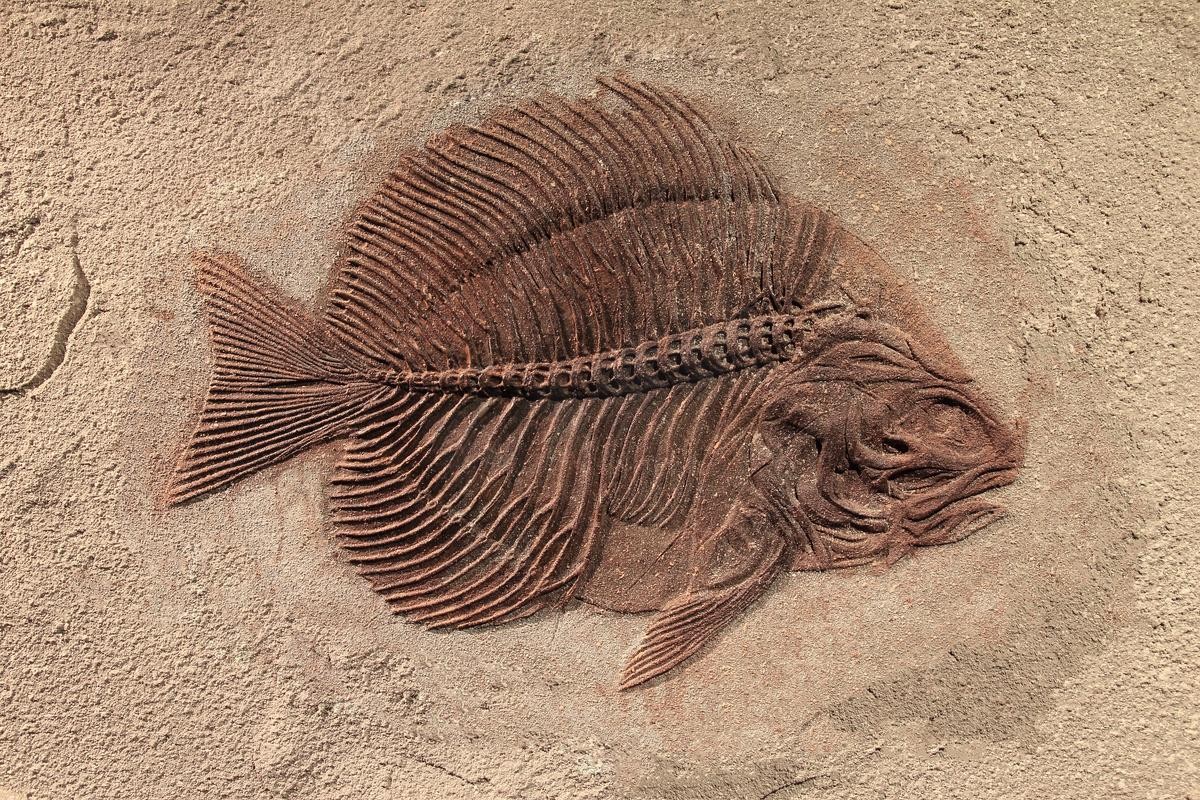} &
\CaseImage{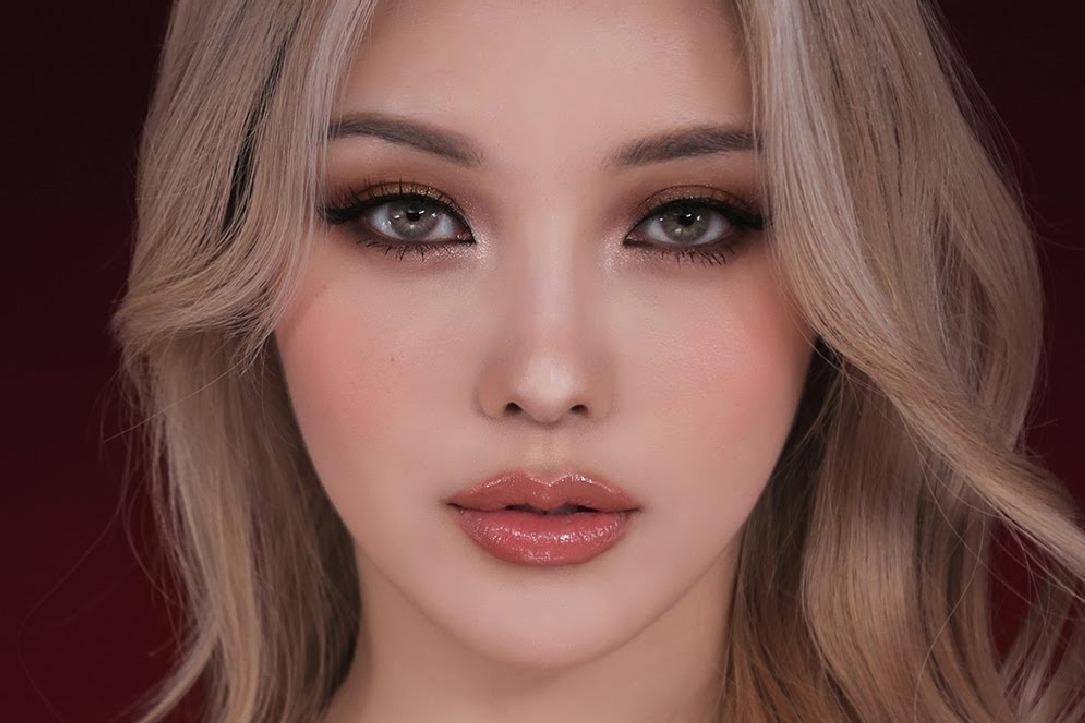} \\[0.0em]

\CasePromptFromJSON{4-c-27} &
\CasePromptFromJSON{4-d-2} &
\CasePromptFromJSON{4-d-6} &
\CasePromptFromJSON{4-d-7} \\[0.01em]
\CaseImage{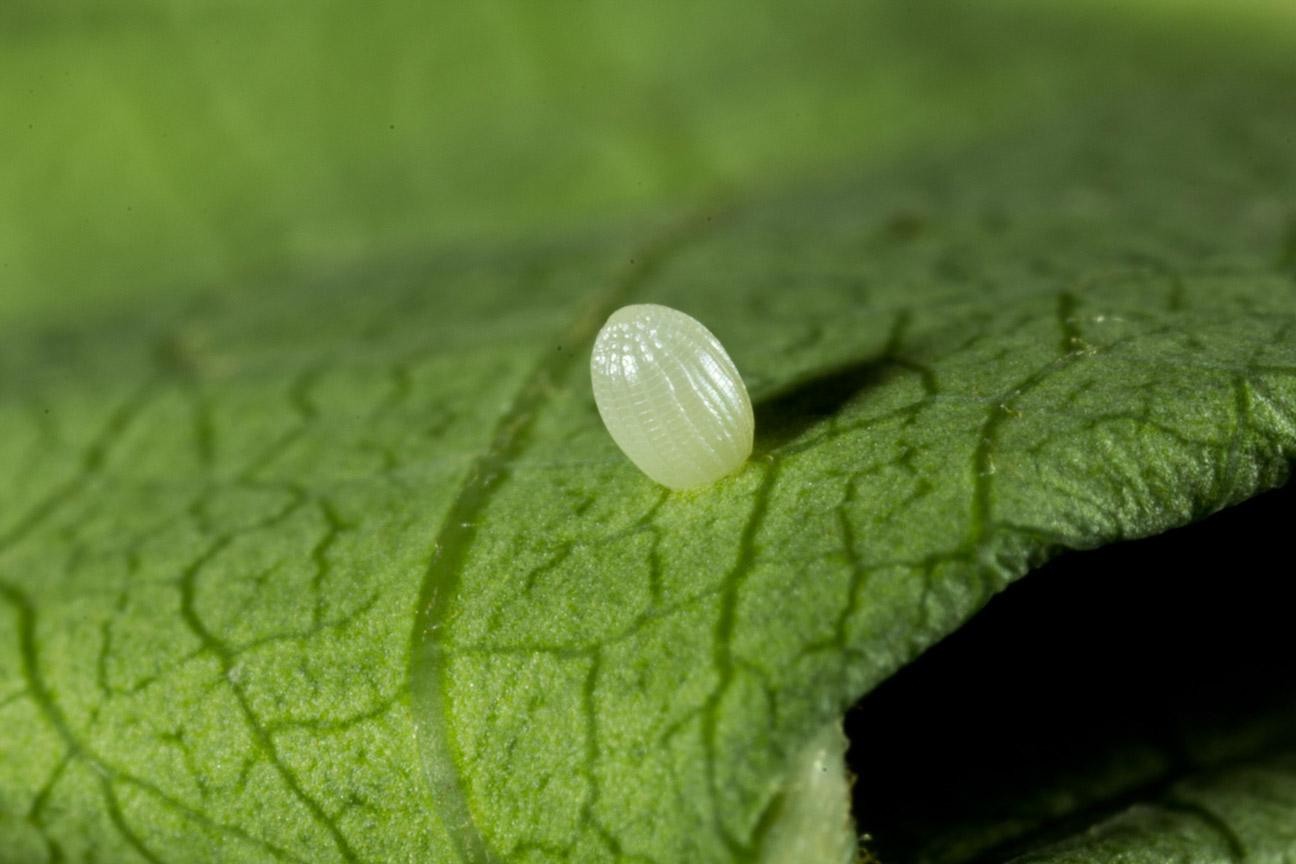} &
\CaseImage{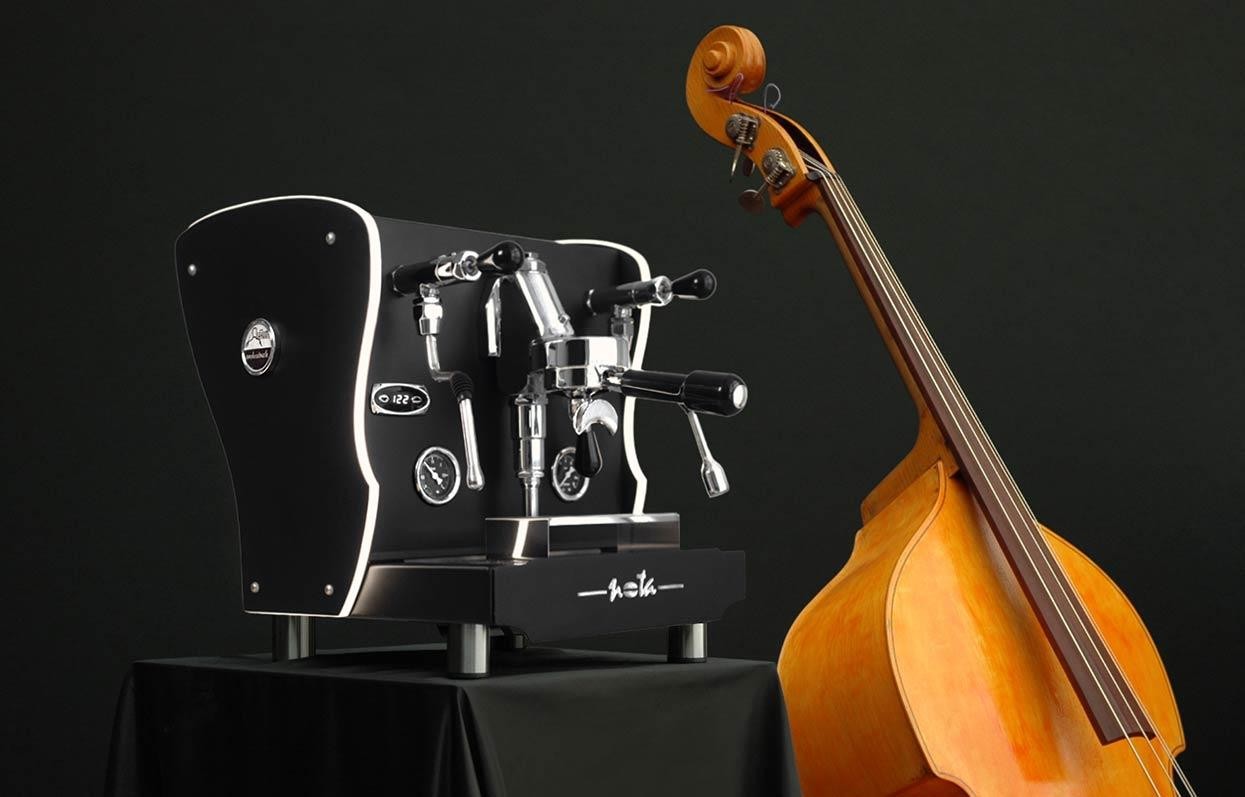} &
\CaseImage{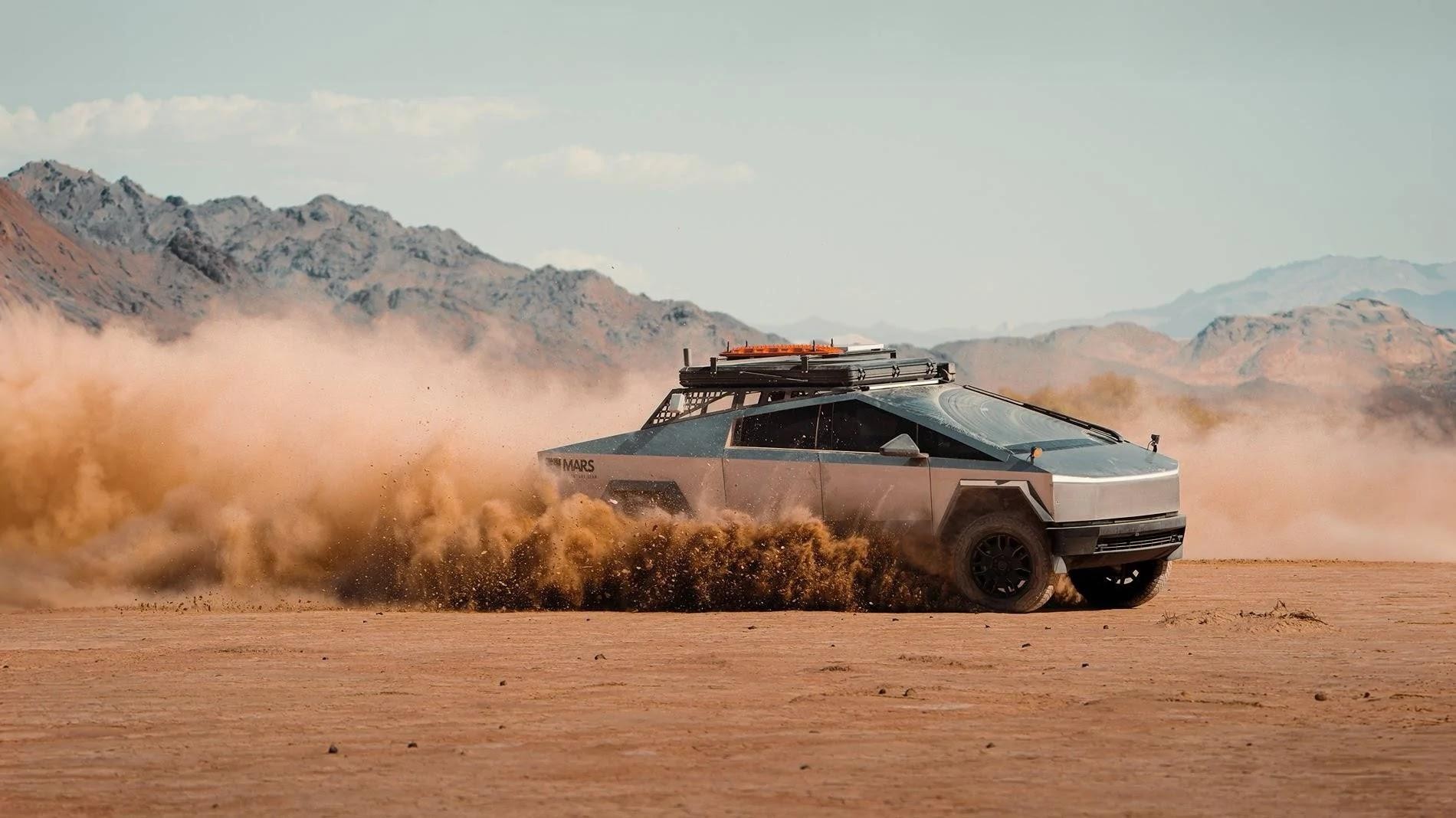} &
\CaseImage{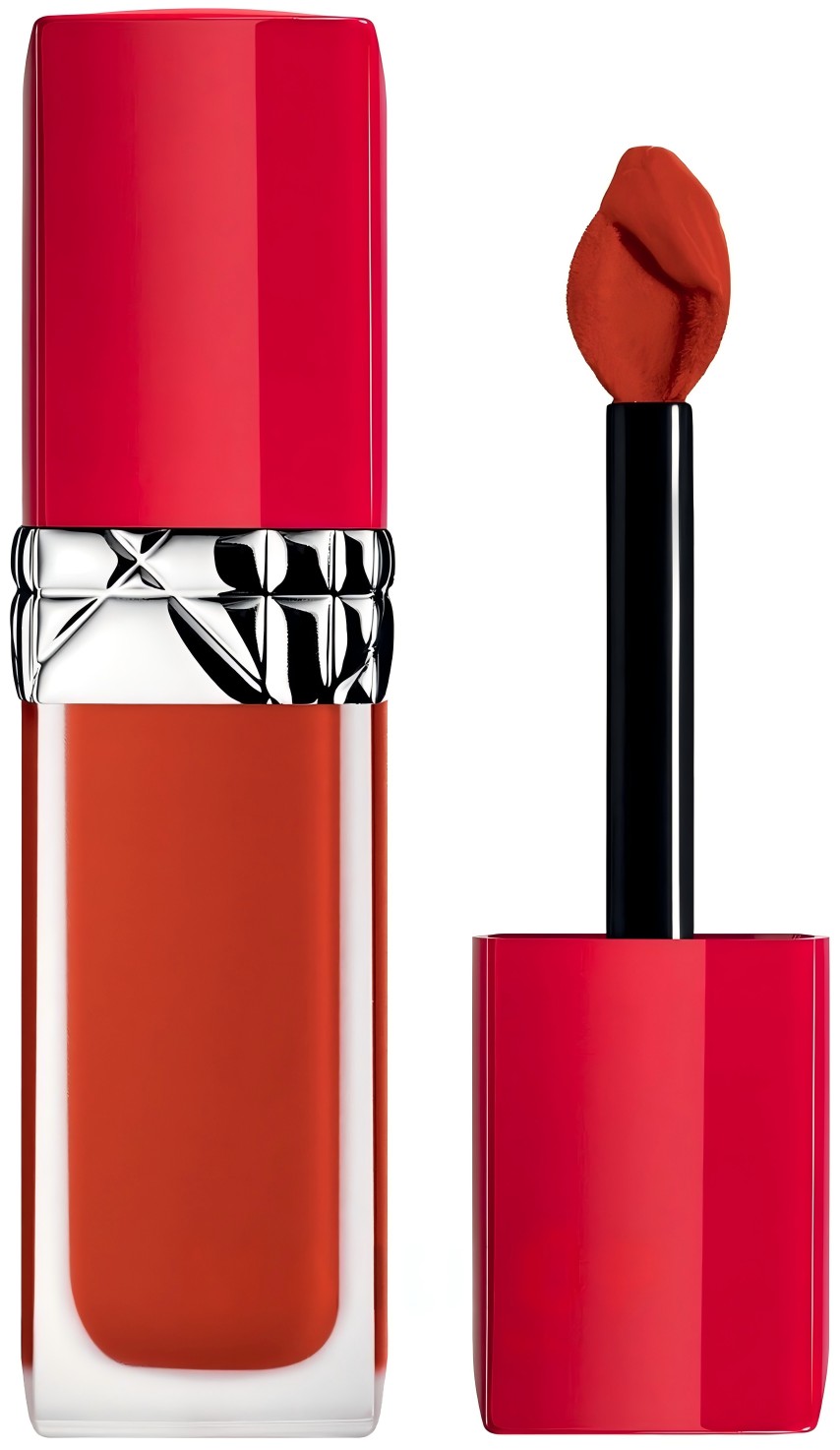} \\[0.0em]

\end{tabularx}
\end{PairBox}
\end{minipage}
\caption{Representative examples from the Information-Based Reasoning category. These cases involve data reading, process timelines, visual editing, knowledge media, and creative expression, testing whether models can preserve explicit information, temporal order, and semantic consistency during video generation.}
\label{tab:representative-information-based-examples}
\end{table*}

\section{Evaluation Prompts}
\label{app:prompts}

This section lists all prompts used in the two evaluation components. Each prompt is shown in a colored box corresponding to its component: \textcolor{qaBlueText}{\textbf{blue}} for Process-aware Reasoning Verification (QA pipeline), \textcolor{qualGreenText}{\textbf{green}} for Multi-dimensional Quality Assessment (point-wise scoring), and \textcolor{pairPurpleText}{\textbf{purple}} for pair-wise comparison.

\subsection{Process-aware Reasoning Verification Prompts}

The Process-aware Reasoning Verification prompts implement the two-stage QA pipeline used for the main reasoning metric. Stage~1 asks the VLM to answer a video-grounded question using only visible evidence, and Stage~2 converts the free-form answer into a binary correctness label against the ground truth.

\begin{table*}[htp]
\begin{minipage}{0.99\textwidth}
\begin{QAbox}{Stage 1: Video Question Answering}
\begin{lstlisting}[basicstyle=\scriptsize,breaklines=true]
You are a precise and conservative video QA assistant. Answer only from directly visible video content. Do not infer or guess. You may briefly analyze first, but the response must end with Final JSON: followed by one valid JSON object.

Sample metadata:
{
  "id": "<case_id>",
  "category": "<dimension>",
  "sub_category": "<subcategory>",
  "video_prompt": "<generation prompt>",
  "question": {
    "question": "<evaluation question>",
    "question_type": "<factual|temporal|detail|reasoning>",
    "difficulty": "<easy|medium|hard>"
  }
}

Answer ONLY using facts that are directly visible in the video frames.
Do NOT infer, speculate, guess, or complete missing information.

Output format:
Final JSON:
{"answer": "<short answer grounded in visible evidence>"}
\end{lstlisting}
\end{QAbox}
\end{minipage}
\caption{Stage~1 prompt for Process-aware Reasoning Verification. The VLM watches the video and answers one evaluation question grounded in visible evidence.}
\label{tab:prompt-qa-stage1}
\end{table*}

\begin{table*}[htp]
\begin{minipage}{0.99\textwidth}
\begin{QAbox}{Stage 2: Binary Answer Judging}
\begin{lstlisting}[basicstyle=\scriptsize,breaklines=true]
You are a strict evaluator for video QA.
You must score this single QA as binary: 1 or 0.
You may briefly analyze first, but keep it short.

Scoring rule:
- score = 1 only if the predicted answer is sufficiently correct according to BOTH the ground truth and the evaluation criteria.
- score = 0 if it is incorrect, incomplete on the key point, contradicts the criteria, or unsupported.
- Be strict. Do not give partial credit.
- If the predicted answer says 'unclear from the video', give 1 only if the criteria truly cannot be verified; otherwise give 0.

Question type: <factual|temporal|detail|reasoning>
Type-specific judging focus:
- factual: Judge whether the predicted answer correctly describes the relevant visible fact, state, object, or outcome.
- detail: Judge whether the predicted answer correctly captures the specific required detail (color, count, text, position, identity).
- temporal: Judge whether the predicted answer correctly captures event order, progression, or before/after relations.
- reasoning: Judge whether the predicted answer correctly captures causality, physical rule, or mechanism.

Question: <question>
Ground truth: <expected_answer>
Evaluation criteria: <criteria>
Difficulty: <easy|medium|hard>
Predicted answer: <model_answer>

Output format:
Final JSON:
{"score": 0_or_1, "reason": "<brief justification>", "verdict": "correct_or_incorrect"}
\end{lstlisting}
\end{QAbox}
\end{minipage}
\caption{Stage~2 prompt for Process-aware Reasoning Verification. A separate LLM judge compares the predicted answer against the ground truth and assigns a binary score.}
\label{tab:prompt-qa-stage2}
\end{table*}

\subsection{Multi-dimensional Quality Assessment Prompts}

The Multi-dimensional Quality Assessment prompts support both point-wise scoring and pair-wise preference comparison. Point-wise scoring produces calibrated per-video scores on reasoning quality, temporal consistency, and visual aesthetics, while pair-wise comparison directly estimates human preference between two candidate videos.

\begin{table*}[htp]
\begin{minipage}{0.99\textwidth}
\begin{QualBox}{Point-wise Video Scoring}
\begin{lstlisting}[basicstyle=\scriptsize,breaklines=true]
You are an impartial expert judge for AI-generated image-to-video outputs. The goal is to evaluate whether the generated video successfully models the world-state transition implied by the input image together with the prompt or instruction.

All input videos are AI-generated. Any humans appearing in the videos are also AI-generated.
You will be given an input image, a text prompt, and the generated video conditioned on them.

Evaluate the generated video on exactly the following three dimensions, using a score from 1 to 5 for each:

1. Reasoning Correctness
(
    1 indicates that the video fails to model the intended world-state transition from the input image, violates the required causal / physical / logical relation, or clearly misunderstands the instruction.
    3 indicates that the video captures part of the intended transition from the input image but still contains noticeable reasoning errors or inconsistencies.
    5 indicates that the video correctly and convincingly evolves the input image into the intended future state with strong causal, physical, logical, or informational consistency.
)

2. Content Fidelity & Continuity
(
    1 indicates that key entities, attributes, or temporal states from the input image are missing, unstable, or incoherent across frames.
    3 indicates that the core content from the input image is present but continuity is only partially preserved, with visible instability or abrupt state changes.
    5 indicates that the required content is faithfully preserved from the input image and the temporal presentation is coherent, stable, and continuous.
)

3. Visual Aesthetics
(
    1 indicates severe visual defects such as distortion, implausible or distracting motion, poor composition, or low rendering quality.
    3 indicates acceptable but imperfect visual quality with noticeable artifacts, limited appeal, or temporal presentation that is only partially convincing.
    5 indicates strong overall visual quality with clean rendering, coherent temporal presentation, and appealing composition. Motion should be rewarded only when it is necessary, plausible, and supportive of the prompt.
)

Reasoning Correctness is the most important dimension. If visual quality is strong but the implied world dynamics from the input image are wrong, the video should still receive a low overall assessment.

Important judging rules:
- Do not reward mere motion, camera movement, or visible change by itself.
- A static but faithful video can still receive a high score if it better matches the prompt and implied world-state transition.
- If a video is represented as sampled frames, interpret them as an ordered sequence from start to end rather than as unrelated still images.
- Reward motion only when it improves correctness, continuity, and prompt fidelity.

Input Image: <image>
Text Prompt: <prompt>
Generated Video: <video>

Give your output in the following JSON format:
{
    "reasoning": "Briefly discuss the video in terms of Reasoning Correctness, Content Fidelity & Continuity, and Visual Aesthetics, then give an overall judgement.",
    "score": [reasoning_correctness_score, content_fidelity_continuity_score, visual_aesthetics_score]
}

Output only valid JSON.
\end{lstlisting}
\end{QualBox}
\end{minipage}
\caption{Point-wise scoring prompt for Multi-dimensional Quality Assessment. The VLM scores each video on three dimensions (reasoning quality, temporal consistency, visual aesthetics) on a 1--5 scale.}
\label{tab:prompt-pointwise}
\end{table*}

\begin{table*}[htp]
\begin{minipage}{0.99\textwidth}
\begin{PairBox}{Pair-wise Video Comparison}
\begin{lstlisting}[basicstyle=\scriptsize,breaklines=true]
Please act as an impartial judge for two AI-generated image-to-video outputs produced from the same input image and prompt or instruction. You will be given the shared input image, Model A's generated video, and Model B's generated video. Your job is to determine which video is better.

Evaluate the two videos using exactly the following three dimensions:

1. Reasoning Correctness
   Whether the video correctly models the intended world-state transition implied by the input image together with the prompt or instruction. This includes causal validity, physical plausibility, logical consistency, and correct state evolution over time.

2. Content Fidelity & Continuity
   Whether the required visual content from the input image is faithfully preserved and whether entities, attributes, and temporal states remain coherent and continuous throughout the video.

3. Visual Aesthetics
   Whether the video is visually appealing, well-rendered, naturally animated, and free from obvious artifacts or severe distortions.

Use the following priority when making the final decision:
- Reasoning Correctness is the primary criterion.
- Content Fidelity & Continuity is the secondary criterion.
- Visual Aesthetics is the tertiary criterion.

In other words, if one video reasons substantially better about the implied world dynamics from the input image, it should usually be preferred even if the other video is slightly more attractive visually.

Important judging rules:
- Do not reward mere motion, camera movement, or any visible change by itself.
- A static but faithful video can be better than a dynamic but implausible or artifact-heavy one.
- If a video is represented as sampled frames, interpret them as an ordered sequence from start to end rather than as unrelated still images.

Input Image: <image>
Text Prompt: <prompt>
Model A Generated Video: <left_video>
Model B Generated Video: <right_video>

Provide a concise comparison covering all three dimensions. After your explanation, output only one final verdict label from the list below:

1. Model A is better: [[A>B]]
2. Model B is better: [[B>A]]
3. Tie, relatively the same acceptable quality: [[A=B=Good]]
4. Both are bad: [[A=B=Bad]]
\end{lstlisting}
\end{PairBox}
\end{minipage}
\caption{Pair-wise comparison prompt for Multi-dimensional Quality Assessment. The VLM directly compares two candidate videos and outputs a preference verdict.}
\label{tab:prompt-pairwise}
\end{table*}

\section{Reasoning Taxonomy Details}
\label{app:taxonomy-details}

This section provides the detailed interpretation of the 22 subcategories used in \WorldReasonBench{}. The concise taxonomy in Section~\ref{subsec:taxonomy} is used for main-text reporting, while the descriptions below specify the intended scope of each subcategory.

\paragraph{World Knowledge.}
This dimension tests whether generated videos respect established knowledge about how the physical, social, and cultural world evolves. 
World Knowledge contains 127 cases in the full set and 32 cases in the representative subset.
It includes \emph{Material Change} (23/5 cases; fluids, heat, optics, and observable physical transformations), 
\emph{Public Systems} (21/2 cases; traffic, logistics, civic infrastructure, and rule-governed public procedures), 
\emph{World Mechanics} (20/8 cases; motion, impact, force, balance, and statics), 
\emph{Cultural Life} (19/3 cases; rituals, performance, architecture, and artistic conventions), 
\emph{Everyday Living} (19/5 cases; household devices, food preparation, and routine daily activities), 
\emph{Earth Cycles} (17/6 cases; weather, astronomy, and large-scale periodic processes), 
and \emph{Living World} (8/3 cases; animal behavior and ecological change), where each pair denotes full-set / representative-subset counts.

\paragraph{Human-Centric Reasoning.}
This dimension targets scenes where humans are central actors and where correct generation requires plausible behavior, interaction, and motion. 
Human-Centric Reasoning contains 78 cases in the full set and 31 cases in the representative subset.
It includes \emph{Object Handling} (27/9 cases; tool use, affordances, interface control, and everyday manipulation), 
\emph{Social Scenes} (15/6 cases; conversation, helping, service interaction, and group events), 
\emph{Skilled Action} (13/3 cases; sports, motor coordination, and craft-like operation), 
\emph{Personal Routine} (13/4 cases; sleep, grooming, comfort management, eating, and self-directed daily behavior), 
and \emph{Public Conduct} (10/9 cases; norm-following behavior in public spaces, mobility, and risk-aware correction),

\paragraph{Logic Reasoning.}
This dimension evaluates whether generated videos can preserve structured logical relations over time. 
Logic Reasoning contains 131 cases in the full set and 35 cases in the representative subset.
It includes \emph{Quantitative Math} (38/11 cases; symbolic calculation, algebra, calculus, and quantitative science), 
\emph{Spatial Geometry} (38/10 cases; spatial transformations, geometric structure, and 3D reasoning), 
\emph{Experimental Science} (32/8 cases; scientific procedures and controlled cause-effect demonstrations), 
\emph{Logic Puzzles} (12/3 cases; constraint satisfaction and rule-governed problem solving), 
and \emph{Pattern Discovery} (11/3 cases; sequence completion, structural analogy, and pattern induction).

\paragraph{Information-Based Reasoning.}
This dimension covers cases with explicit information such as text, numbers, charts, diagrams, and data visualizations that must be preserved or transformed faithfully. 
Information-Based Reasoning contains 100 cases in the full set and 32 cases in the representative subset.
It includes \emph{Data Reading} (29/12 cases; chart, table, and exact value interpretation), 
\emph{Process Timeline} (28/8 cases; temporally ordered transformations and stage progression), 
\emph{Visual Editing} (18/4 cases; diagrammatic modification and layout-consistent changes), 
\emph{Knowledge Media} (16/5 cases; document-, history-, and explainer-style content grounded in external knowledge), 
and \emph{Creative Expression} (9/3 cases; metaphorical or stylistically transformed information presentation).

\section{Full Two-Component \WorldReasonBench{} Results}
\label{app:evaluation-full}

\newcommand{\wrappplaceholdercells}{ & -- & -- & -- & -- & -- & -- & -- & -- & -- & -- & -- }
\newcommand{\wrappqarow}[1]{#1 \wrappplaceholdercells \\}
\newcommand{\wrappqualityrows}[1]{%
  \multirow{3}{*}{#1} & \begin{tabular}[c]{@{}l@{}}Reasoning\\Quality\end{tabular} \wrappplaceholdercells \\
  & \begin{tabular}[c]{@{}l@{}}Temporal\\Consistency\end{tabular} \wrappplaceholdercells \\
  & \begin{tabular}[c]{@{}l@{}}Visual\\Aesthetics\end{tabular} \wrappplaceholdercells \\
}

We separate the detailed \WorldReasonBench{} results by evaluation component and top-level reasoning dimension. Process-aware Reasoning Verification tables report outcome QA accuracy (\%), while Multi-dimensional Quality Assessment tables expand each sub-category into three raw quality axes on the 1--5 scale.

\subsection{Auxiliary Process-Aware Metric Definitions}
\label{app:aux-process-metrics}

The main text uses $\mathrm{Score}_{\mathrm{PR}}$ as the headline Process-aware Reasoning Verification metric because it preserves ranking discriminability while emphasizing temporal and mechanistic correctness. We additionally define two auxiliary diagnostics for ablations and detailed error analysis.

\paragraph{Difficulty-weighted score.}
Each QA pair is annotated with a \texttt{difficulty} label. To penalize failures on easy questions more heavily while rewarding successes on hard ones, we use an asymmetric weighting scheme:
\begin{equation}
  \begin{aligned}
  w_i &=
  \begin{cases}
    w_d^{+} & \text{if correct},\\
    w_d^{-} & \text{if incorrect},
  \end{cases} \\
  (w_{\mathrm{easy}}^+, w_{\mathrm{easy}}^-) &= (0.8, 1.5),\\
  (w_{\mathrm{med}}^+, w_{\mathrm{med}}^-) &= (1.0, 1.0),\\
  (w_{\mathrm{hard}}^+, w_{\mathrm{hard}}^-) &= (1.5, 0.6).
  \end{aligned}
  \label{eq:diff_weights}
\end{equation}
The difficulty-weighted score is $\mathrm{Score}_{\mathrm{wt}} = {\sum_i w_i \cdot \mathbb{1}[\hat{y}_i = y_i^{\mathrm{gt}}]}\big/{\sum_i w_i}$.

\paragraph{Bottleneck composite score.}
The bottleneck composite score emphasizes the \emph{short-board effect} by penalizing failures in any reasoning phase:
\begin{equation}
  \mathrm{Score}_{\mathrm{bn}} = s_{\mathrm{state}}^{\,\alpha_1}\;\cdot\;s_{\mathrm{proc}}^{\,\alpha_2}\;\cdot\;s_{\mathrm{fidel}}^{\,\alpha_1}\;\cdot\;s_{\mathrm{mech}}^{\,\alpha_2},
  \quad \alpha_1 = 0.2,\; \alpha_2 = 0.3.
  \label{eq:bottleneck}
\end{equation}
Because the geometric mean collapses when any phase approaches zero, $\mathrm{Score}_{\mathrm{bn}}$ is useful for identifying severe reasoning failures but can be overly conservative as a headline ranking metric.

\subsection{Full Subcategory-Level Results}

This subsection expands the compact main-results table into all 22 subcategories. It reports both the process-aware reasoning score and the multi-dimensional quality score so that category-specific strengths and failure modes can be inspected directly. Here $\mathrm{Score}_{\mathrm{PR}} = \mathrm{Acc}_{\mathrm{QA}}^{0.8} \cdot s_{\mathrm{dyn}}^{0.2}$, and $S(v)$ is linearly mapped from [1,5] to [0,100].

\begin{table*}[htp]
  \caption{\textbf{Full subcategory-level evaluation results.} For each of the 22 subcategories, we report $\mathrm{Score}_{\mathrm{PR}}$ (\%) and $S(v)$ (0--100). Best and second-best within each model family are \textbf{bold} and \underline{underlined}. Some cells are ``--'' due to limited video coverage for that model--subcategory pair. This table expands the category-level summary in Table~\ref{tab:main_results}.}
  \label{tab:main_results_full}
  \centering
  \scriptsize
  \setlength{\tabcolsep}{3pt}
  \renewcommand{\arraystretch}{0.94}
  \resizebox{\textwidth}{!}{
  \begin{tabular}{@{}lllccccccccccc@{}}
    \toprule
    \multicolumn{3}{c}{} & \multicolumn{5}{c}{\textbf{Closed-Source Models}} & \multicolumn{6}{c}{\textbf{Open-Source Models}} \\
    \cmidrule(lr){4-8} \cmidrule(lr){9-14}
    \textbf{Dimension} & \textbf{Sub-category} & \textbf{Metric} & \textbf{Sora2} & \textbf{Kling} & \textbf{Wan2.6} & \textbf{Seedance2.0} & \textbf{Veo3.1-Fast} & \textbf{LTX2.3} & \textbf{Wan2.2-14B} & \textbf{UniVideo} & \textbf{HunyuanVideo-1.5} & \textbf{Cosmos-Predict2.5} & \textbf{LongCat-Video} \\
    \midrule
    \multirow{16}{*}{\rotatebox[origin=c]{90}{\textbf{World Knowledge}}}%
      & \multirow{2}{*}{\begin{tabular}[c]{@{}l@{}}Material\\Change\end{tabular}} & $\mathrm{Score}_{\mathrm{PR}}$ & 24.3 & \underline{52.4} & 54.7 & 36.4 & \textbf{57.9} & 21.2 & \textbf{32.9} & 26.7 & \underline{34.8} & 34.4 & 28.0 \\
      & & $S(v)$ & 40.8 & \underline{70.0} & 46.2 & 60.0 & \textbf{80.5} & 37.5 & \textbf{55.2} & 45.0 & \underline{51.5} & 59.1 & 48.0 \\
    \cmidrule(lr){2-14}
      & \multirow{2}{*}{\begin{tabular}[c]{@{}l@{}}Public\\Systems\end{tabular}} & $\mathrm{Score}_{\mathrm{PR}}$ & \textbf{83.7} & \underline{71.0} & 36.4 & 50.0 & 65.4 & 11.7 & \textbf{20.7} & \underline{20.3} & 15.2 & 13.1 & 17.9 \\
      & & $S(v)$ & 47.5 & 73.8 & \textbf{85.0} & 81.2 & \underline{85.0} & 28.5 & \textbf{59.0} & 39.5 & 38.5 & 45.0 & \underline{52.5} \\
    \cmidrule(lr){2-14}
      & \multirow{2}{*}{\begin{tabular}[c]{@{}l@{}}World\\Mechanics\end{tabular}} & $\mathrm{Score}_{\mathrm{PR}}$ & \textbf{47.7} & 32.8 & 42.1 & 40.1 & \underline{47.1} & 19.9 & \textbf{26.9} & 15.0 & \underline{26.4} & 18.3 & 14.3 \\
      & & $S(v)$ & \underline{79.0} & \textbf{81.7} & 73.0 & 66.8 & 70.2 & 40.2 & \textbf{57.5} & 25.0 & 37.5 & 43.1 & \underline{45.2} \\
    \cmidrule(lr){2-14}
      & \multirow{2}{*}{\begin{tabular}[c]{@{}l@{}}Cultural\\Life\end{tabular}} & $\mathrm{Score}_{\mathrm{PR}}$ & 43.5 & \underline{47.3} & 28.9 & 36.4 & \textbf{57.9} & 15.5 & 19.5 & \underline{20.3} & \textbf{21.3} & 18.8 & 13.4 \\
      & & $S(v)$ & \underline{63.7} & 50.7 & 61.3 & 48.8 & \textbf{78.8} & 17.8 & \textbf{51.2} & 44.0 & \underline{47.2} & 47.9 & 41.0 \\
    \cmidrule(lr){2-14}
      & \multirow{2}{*}{\begin{tabular}[c]{@{}l@{}}Everyday\\Living\end{tabular}} & $\mathrm{Score}_{\mathrm{PR}}$ & -- & 32.0 & \textbf{57.9} & 43.6 & \underline{48.4} & 20.3 & 19.5 & 18.4 & \textbf{30.3} & 19.2 & \underline{24.9} \\
      & & $S(v)$ & 72.5 & \textbf{83.0} & \underline{83.0} & 60.0 & 75.0 & 26.0 & \underline{41.0} & 38.2 & \textbf{42.8} & 40.3 & 37.7 \\
    \cmidrule(lr){2-14}
      & \multirow{2}{*}{\begin{tabular}[c]{@{}l@{}}Earth\\Cycles\end{tabular}} & $\mathrm{Score}_{\mathrm{PR}}$ & -- & \underline{43.6} & 35.8 & \textbf{48.5} & 41.1 & -- & 15.7 & \underline{17.1} & \textbf{19.3} & 19.3 & 11.0 \\
      & & $S(v)$ & 57.0 & \textbf{87.5} & 60.8 & 79.5 & \underline{83.2} & 28.7 & 43.0 & 42.8 & \textbf{54.8} & 54.4 & \underline{44.0} \\
    \cmidrule(lr){2-14}
      & \multirow{2}{*}{\begin{tabular}[c]{@{}l@{}}Living\\World\end{tabular}} & $\mathrm{Score}_{\mathrm{PR}}$ & 27.9 & 24.3 & \underline{29.0} & 38.0 & \textbf{61.3} & -- & 3.0 & \underline{3.0} & -- & 9.1 & \textbf{7.8} \\
      & & $S(v)$ & 83.2 & 69.2 & 89.2 & \underline{97.5} & \textbf{100.0} & 26.7 & 29.0 & 27.2 & \textbf{33.8} & 32.5 & \underline{33.5} \\
    \cmidrule(lr){2-14}
      & \multirow{2}{*}{\textbf{Average}} & $\mathrm{Score}_{\mathrm{PR}}$ & 45.4 & 43.3 & 40.7 & 41.9 & \textbf{54.1} & 17.7 & 19.8 & 17.3 & \underline{24.6} & 20.3 & 16.8 \\
      & & $S(v)$ & 65.8 & \underline{76.2} & 70.2 & 71.2 & \textbf{79.5} & 30.0 & \textbf{50.2} & 38.2 & \underline{44.5} & 47.4 & 44.2 \\
    \midrule
    \multirow{12}{*}{\rotatebox[origin=c]{90}{\textbf{Human-Centric}}}
      & \multirow{2}{*}{\begin{tabular}[c]{@{}l@{}}Object\\Handling\end{tabular}} & $\mathrm{Score}_{\mathrm{PR}}$ & 20.9 & \textbf{55.6} & 42.1 & 31.6 & \underline{43.6} & 20.2 & \underline{27.2} & \textbf{28.8} & 17.8 & 23.1 & 25.1 \\
      & & $S(v)$ & 48.8 & 73.5 & \underline{82.5} & 74.0 & \textbf{83.0} & 30.0 & \textbf{49.5} & 42.2 & 44.2 & 50.6 & \underline{47.2} \\
    \cmidrule(lr){2-14}
      & \multirow{2}{*}{\begin{tabular}[c]{@{}l@{}}Social\\Scenes\end{tabular}} & $\mathrm{Score}_{\mathrm{PR}}$ & \textbf{57.9} & 22.9 & 29.8 & \underline{50.4} & 29.1 & -- & \textbf{29.1} & 12.7 & \underline{16.4} & 13.2 & 10.9 \\
      & & $S(v)$ & \textbf{92.5} & 63.2 & 69.5 & \underline{92.5} & 70.5 & 31.0 & \underline{50.7} & 40.8 & 42.8 & 48.2 & \textbf{54.8} \\
    \cmidrule(lr){2-14}
      & \multirow{2}{*}{\begin{tabular}[c]{@{}l@{}}Skilled\\Action\end{tabular}} & $\mathrm{Score}_{\mathrm{PR}}$ & 39.4 & 36.1 & \underline{47.3} & \textbf{47.9} & 33.3 & 9.5 & \underline{23.9} & 17.3 & 11.7 & 20.3 & \textbf{25.4} \\
      & & $S(v)$ & \underline{97.5} & \textbf{100.0} & 83.2 & 97.5 & 75.8 & 43.5 & \textbf{72.0} & 56.8 & 59.5 & 47.3 & \underline{68.0} \\
    \cmidrule(lr){2-14}
      & \multirow{2}{*}{\begin{tabular}[c]{@{}l@{}}Personal\\Routine\end{tabular}} & $\mathrm{Score}_{\mathrm{PR}}$ & -- & \textbf{36.6} & \underline{20.9} & -- & -- & -- & \textbf{22.8} & 12.1 & \underline{15.2} & 14.5 & 10.9 \\
      & & $S(v)$ & -- & \underline{63.0} & \textbf{78.0} & -- & 47.5 & 28.0 & \textbf{52.2} & 38.0 & 43.5 & 36.9 & \underline{46.8} \\
    \cmidrule(lr){2-14}
      & \multirow{2}{*}{\begin{tabular}[c]{@{}l@{}}Public\\Conduct\end{tabular}} & $\mathrm{Score}_{\mathrm{PR}}$ & \textbf{62.7} & \underline{46.7} & 30.2 & 40.3 & 42.7 & 17.4 & \textbf{27.4} & 16.0 & \underline{25.8} & 19.7 & 20.0 \\
      & & $S(v)$ & 57.5 & 76.0 & 54.5 & \textbf{81.7} & \underline{81.7} & 19.0 & \textbf{56.0} & \underline{53.2} & 49.8 & 45.0 & 45.8 \\
    \cmidrule(lr){2-14}
      & \multirow{2}{*}{\textbf{Average}} & $\mathrm{Score}_{\mathrm{PR}}$ & \textbf{45.2} & 39.6 & 34.1 & \underline{42.5} & 37.2 & 15.7 & \underline{26.1} & 17.4 & 17.4 & 17.3 & 18.5 \\
      & & $S(v)$ & 76.8 & 73.5 & 72.0 & \textbf{83.5} & \underline{78.0} & 30.8 & \textbf{54.8} & 45.2 & 47.2 & 46.7 & \underline{52.0} \\
    \midrule
    \multirow{12}{*}{\rotatebox[origin=c]{90}{\textbf{Logic Reasoning}}}%
      & \multirow{2}{*}{\begin{tabular}[c]{@{}l@{}}Experimental\\Science\end{tabular}} & $\mathrm{Score}_{\mathrm{PR}}$ & 21.8 & \textbf{50.6} & 30.1 & \underline{49.1} & 42.8 & 19.5 & \textbf{25.1} & 17.9 & \underline{23.6} & 10.7 & 22.4 \\
      & & $S(v)$ & 53.2 & \underline{60.2} & 55.8 & \textbf{76.2} & 47.5 & 25.7 & 28.5 & 24.2 & \underline{34.5} & 25.3 & \textbf{36.3} \\
    \cmidrule(lr){2-14}
      & \multirow{2}{*}{\begin{tabular}[c]{@{}l@{}}Spatial\\Geometry\end{tabular}} & $\mathrm{Score}_{\mathrm{PR}}$ & 25.3 & 17.7 & 24.6 & \textbf{30.6} & \underline{25.5} & \textbf{13.0} & 12.2 & 9.0 & 9.7 & 7.6 & \underline{12.6} \\
      & & $S(v)$ & \underline{54.8} & 45.2 & 31.8 & \textbf{56.2} & 32.8 & 12.0 & \textbf{17.5} & 9.3 & 11.8 & 26.8 & \underline{14.8} \\
    \cmidrule(lr){2-14}
      & \multirow{2}{*}{\begin{tabular}[c]{@{}l@{}}Quantitative\\Math\end{tabular}} & $\mathrm{Score}_{\mathrm{PR}}$ & \underline{27.4} & 7.9 & \textbf{28.7} & 23.1 & 22.2 & 8.5 & \underline{9.0} & 8.2 & \textbf{9.9} & 14.2 & 7.1 \\
      & & $S(v)$ & \textbf{45.8} & 24.8 & \underline{42.0} & 33.0 & 25.0 & \underline{12.2} & \textbf{14.0} & 8.8 & 8.3 & 24.5 & 10.7 \\
    \cmidrule(lr){2-14}
      & \multirow{2}{*}{\begin{tabular}[c]{@{}l@{}}Logic\\Puzzles\end{tabular}} & $\mathrm{Score}_{\mathrm{PR}}$ & \textbf{29.0} & -- & -- & \underline{24.3} & 13.9 & \textbf{16.7} & 11.8 & 12.1 & \underline{15.3} & 9.6 & 13.2 \\
      & & $S(v)$ & 22.5 & \underline{27.5} & 25.0 & \textbf{35.0} & 15.0 & \underline{20.2} & \textbf{22.2} & 14.0 & 13.8 & 36.9 & 15.3 \\
    \cmidrule(lr){2-14}
      & \multirow{2}{*}{\begin{tabular}[c]{@{}l@{}}Pattern\\Discovery\end{tabular}} & $\mathrm{Score}_{\mathrm{PR}}$ & \textbf{28.9} & \underline{19.3} & 19.3 & -- & -- & 8.2 & -- & 3.9 & \underline{8.5} & 11.7 & \textbf{8.5} \\
      & & $S(v)$ & 27.5 & 37.5 & 30.8 & \textbf{70.0} & \underline{46.8} & 6.0 & 13.5 & 11.2 & \underline{14.5} & 29.2 & \textbf{16.5} \\
    \cmidrule(lr){2-14}
      & \multirow{2}{*}{\textbf{Average}} & $\mathrm{Score}_{\mathrm{PR}}$ & 26.5 & 23.9 & 25.7 & \textbf{31.8} & \underline{26.1} & 13.2 & \underline{14.5} & 10.2 & 13.4 & 11.7 & 12.8 \\
      & & $S(v)$ & \underline{46.8} & 40.0 & 40.0 & \textbf{55.8} & 34.0 & 15.7 & \textbf{19.5} & 13.5 & 17.0 & 27.1 & \underline{19.2} \\
    \midrule
    \multirow{12}{*}{\rotatebox[origin=c]{90}{\textbf{Information-Based}}}%
      & \multirow{2}{*}{\begin{tabular}[c]{@{}l@{}}Data\\Reading\end{tabular}} & $\mathrm{Score}_{\mathrm{PR}}$ & \textbf{32.6} & 27.2 & 22.0 & \underline{32.0} & 12.9 & \textbf{31.5} & \underline{29.0} & 18.1 & 31.1 & -- & 29.2 \\
      & & $S(v)$ & \textbf{43.8} & 25.2 & \underline{30.0} & 19.8 & 19.5 & \underline{25.5} & \textbf{30.8} & 4.7 & 9.7 & 22.5 & 17.2 \\
    \cmidrule(lr){2-14}
      & \multirow{2}{*}{\begin{tabular}[c]{@{}l@{}}Process\\Timeline\end{tabular}} & $\mathrm{Score}_{\mathrm{PR}}$ & \textbf{56.8} & 18.9 & 41.3 & \underline{52.6} & 41.3 & 10.8 & 11.7 & 12.7 & \underline{16.4} & -- & \textbf{17.9} \\
      & & $S(v)$ & \textbf{85.0} & 64.5 & 54.2 & 53.5 & \underline{75.8} & 30.2 & \textbf{35.0} & 30.0 & \underline{30.5} & 32.4 & 28.2 \\
    \cmidrule(lr){2-14}
      & \multirow{2}{*}{\begin{tabular}[c]{@{}l@{}}Visual\\Editing\end{tabular}} & $\mathrm{Score}_{\mathrm{PR}}$ & 18.2 & \underline{50.0} & 41.8 & \textbf{61.7} & -- & 10.8 & 12.1 & 7.9 & \underline{13.7} & -- & \textbf{16.7} \\
      & & $S(v)$ & 20.5 & \underline{46.2} & 36.3 & \textbf{55.0} & 43.2 & 18.0 & \textbf{21.5} & 5.5 & 15.5 & 18.7 & \underline{18.5} \\
    \cmidrule(lr){2-14}
      & \multirow{2}{*}{\begin{tabular}[c]{@{}l@{}}Knowledge\\Media\end{tabular}} & $\mathrm{Score}_{\mathrm{PR}}$ & 20.9 & \underline{29.1} & 26.2 & \textbf{51.6} & 25.1 & \underline{20.7} & 13.5 & 12.7 & 17.7 & 10.0 & \textbf{22.5} \\
      & & $S(v)$ & 32.5 & 49.0 & 42.5 & \textbf{58.5} & \underline{57.5} & 30.0 & 31.0 & 23.8 & \underline{32.5} & 39.8 & \textbf{36.8} \\
    \cmidrule(lr){2-14}
      & \multirow{2}{*}{\begin{tabular}[c]{@{}l@{}}Creative\\Expression\end{tabular}} & $\mathrm{Score}_{\mathrm{PR}}$ & 50.1 & \textbf{69.7} & \underline{61.3} & 61.3 & 64.1 & \textbf{40.3} & 25.9 & 29.0 & \underline{30.6} & 24.7 & 27.5 \\
      & & $S(v)$ & 97.5 & \textbf{100.0} & 74.2 & 85.0 & \underline{100.0} & 45.2 & \underline{65.0} & 61.5 & \textbf{66.0} & 55.3 & 52.0 \\
    \cmidrule(lr){2-14}
      & \multirow{2}{*}{\textbf{Average}} & $\mathrm{Score}_{\mathrm{PR}}$ & 35.7 & 39.0 & 38.5 & \textbf{51.8} & 35.9 & \underline{22.8} & 18.5 & 16.1 & 21.9 & 17.1 & 22.7 \\
      & & $S(v)$ & \textbf{54.2} & \underline{49.0} & 43.0 & 44.5 & 49.0 & \underline{28.0} & \textbf{33.5} & 20.2 & 25.2 & 30.4 & 26.7 \\
    \midrule
    \multicolumn{2}{l}{\textbf{Overall Average}} & $\mathrm{Score}_{\mathrm{PR}}$ & 37.8 & 37.7 & 35.7 & \textbf{42.5} & \underline{40.8} & 17.5 & \underline{20.0} & 15.4 & 19.6 & 17.1 & 17.6 \\
    \multicolumn{2}{l}{} & $S(v)$ & 57.5 & \underline{59.5} & 56.2 & \textbf{60.8} & 59.2 & 25.7 & \textbf{38.5} & 28.2 & 32.8 & 37.6 & \underline{34.5} \\
    \bottomrule
  \end{tabular}}
\end{table*}

\paragraph{Result analysis.}
The full subcategory table exposes two complementary trends that are compressed in the main text. First, closed-source systems lead clearly on both $\mathrm{Score}_{\mathrm{PR}}$ and $S(v)$, but their advantages are not uniform across reasoning dimensions: Veo3.1-Fast is strongest on World Knowledge, Seedance2.0 leads overall and on Information-Based reasoning, while Sora2 remains competitive on Human-Centric process scores. Second, the gap between $S(v)$ and $\mathrm{Score}_{\mathrm{PR}}$ is substantial in several subcategories, indicating that visually plausible or temporally smooth outputs can still miss the intended state transition or causal mechanism. This is especially visible in Logic Reasoning and Information-Based categories, where quality scores are often moderate while process-aware scores remain low.

\subsection{Process-aware Reasoning Verification}

The following tables isolate outcome QA accuracy for each top-level reasoning dimension. They complement Table~\ref{tab:main_results_full} by showing which subcategories contribute most to each model's process-aware reasoning performance.

\begin{table*}[htp]
  \caption{\textbf{Process-aware Reasoning Verification detailed results on World Knowledge. Outcome QA accuracy (\%) across World Knowledge subcategories.}}
  \label{tab:app-qa-world-knowledge}
  \centering
  \scriptsize
  \renewcommand{\arraystretch}{1.10}
  \resizebox{\textwidth}{!}{
  \begin{tabular}{l|ccccc|cccccc}
    \toprule
    & \multicolumn{5}{c|}{\textbf{Closed-Source Models}} & \multicolumn{6}{c}{\textbf{Open-Source Models}} \\
    \cmidrule(lr){2-6} \cmidrule(lr){7-12}
    \textbf{Sub-category} & \textbf{Sora2} & \textbf{Kling} & \textbf{Wan2.6} & \textbf{Seedance2.0} & \textbf{Veo3.1-Fast} & \textbf{LTX2.3} & \textbf{Wan2.2-14B} & \textbf{UniVideo} & \textbf{HunyuanVideo-1.5} & \textbf{Cosmos-Predict2.5} & \textbf{LongCat-Video} \\
    \midrule
    \begin{tabular}[c]{@{}l@{}}Material\\Change\end{tabular} & 26.7 & 56.0 & 56.0 & 40.0 & 60.0 & 23.3 & 36.2 & 29.7 & 37.1 & 37.1 & 30.2 \\
    \cmidrule(lr){1-12}
    \begin{tabular}[c]{@{}l@{}}Public\\Systems\end{tabular} & 80.0 & 70.0 & 40.0 & 50.0 & 70.0 & 12.4 & 21.9 & 20.0 & 16.2 & 14.3 & 21.0 \\
    \cmidrule(lr){1-12}
    \begin{tabular}[c]{@{}l@{}}World\\Mechanics\end{tabular} & 50.0 & 36.6 & 48.8 & 41.5 & 48.8 & 20.8 & 28.7 & 17.8 & 28.7 & 21.8 & 19.8 \\
    \cmidrule(lr){1-12}
    \begin{tabular}[c]{@{}l@{}}Cultural\\Life\end{tabular} & 50.0 & 46.7 & 30.0 & 40.0 & 60.0 & 15.8 & 22.1 & 22.1 & 22.1 & 20.0 & 14.7 \\
    \cmidrule(lr){1-12}
    \begin{tabular}[c]{@{}l@{}}Everyday\\Living\end{tabular} & 40.0 & 36.0 & 60.0 & 46.7 & 48.0 & 24.0 & 22.9 & 22.9 & 32.3 & 24.0 & 26.0 \\
    \cmidrule(lr){1-12}
    \begin{tabular}[c]{@{}l@{}}Earth\\Cycles\end{tabular} & 33.3 & 46.7 & 43.3 & 53.3 & 43.3 & 8.2 & 20.0 & 22.4 & 23.5 & 22.4 & 15.3 \\
    \cmidrule(lr){1-12}
    \begin{tabular}[c]{@{}l@{}}Living\\World\end{tabular} & 26.7 & 26.7 & 33.3 & 46.7 & 60.0 & 7.5 & 2.5 & 2.5 & 2.5 & 10.0 & 7.5 \\
    \cmidrule(lr){1-12}
    \textbf{Average} & 40.5 & 43.5 & 47.4 & 45.4 & 52.6 & 17.1 & 24.3 & 21.3 & 25.4 & 22.7 & 20.7 \\
    \bottomrule
  \end{tabular}}
\end{table*}

\begin{table*}[htp]
  \caption{\textbf{Process-aware Reasoning Verification detailed results on Human-Centric. Outcome QA accuracy (\%) across Human-Centric subcategories.}}
  \label{tab:app-qa-human-centric}
  \centering
  \scriptsize
  \renewcommand{\arraystretch}{1.10}
  \resizebox{\textwidth}{!}{
  \begin{tabular}{l|ccccc|cccccc}
    \toprule
    & \multicolumn{5}{c|}{\textbf{Closed-Source Models}} & \multicolumn{6}{c}{\textbf{Open-Source Models}} \\
    \cmidrule(lr){2-6} \cmidrule(lr){7-12}
    \textbf{Sub-category} & \textbf{Sora2} & \textbf{Kling} & \textbf{Wan2.6} & \textbf{Seedance2.0} & \textbf{Veo3.1-Fast} & \textbf{LTX2.3} & \textbf{Wan2.2-14B} & \textbf{UniVideo} & \textbf{HunyuanVideo-1.5} & \textbf{Cosmos-Predict2.5} & \textbf{LongCat-Video} \\
    \midrule
    \begin{tabular}[c]{@{}l@{}}Object\\Handling\end{tabular} & 20.0 & 55.6 & 42.2 & 32.0 & 46.7 & 22.8 & 30.1 & 30.1 & 19.9 & 25.0 & 27.2 \\
    \cmidrule(lr){1-12}
    \begin{tabular}[c]{@{}l@{}}Social\\Scenes\end{tabular} & 60.0 & 26.7 & 30.0 & 60.0 & 32.0 & 9.3 & 32.0 & 16.0 & 17.3 & 18.7 & 14.7 \\
    \cmidrule(lr){1-12}
    \begin{tabular}[c]{@{}l@{}}Skilled\\Action\end{tabular} & 37.5 & 31.2 & 43.8 & 43.8 & 31.2 & 10.6 & 24.2 & 18.2 & 12.1 & 21.2 & 24.2 \\
    \cmidrule(lr){1-12}
    \begin{tabular}[c]{@{}l@{}}Personal\\Routine\end{tabular} & -- & 35.0 & 20.0 & -- & 40.0 & 8.3 & 23.3 & 13.3 & 16.7 & 15.0 & 11.7 \\
    \cmidrule(lr){1-12}
    \begin{tabular}[c]{@{}l@{}}Public\\Conduct\end{tabular} & 60.0 & 48.9 & 30.0 & 40.0 & 42.5 & 18.0 & 28.0 & 18.0 & 26.0 & 22.0 & 20.0 \\
    \cmidrule(lr){1-12}
    \textbf{Average} & 43.5 & 42.9 & 33.8 & 40.7 & 40.5 & 15.2 & 28.2 & 21.2 & 18.3 & 21.2 & 20.9 \\
    \bottomrule
  \end{tabular}}
\end{table*}

\begin{table*}[htp]
  \caption{\textbf{Process-aware Reasoning Verification detailed results on Logic Reasoning. Outcome QA accuracy (\%) across Logic Reasoning subcategories.}}
  \label{tab:app-qa-logic-reasoning}
  \centering
  \scriptsize
  \renewcommand{\arraystretch}{1.10}
  \resizebox{\textwidth}{!}{
  \begin{tabular}{l|ccccc|cccccc}
    \toprule
    & \multicolumn{5}{c|}{\textbf{Closed-Source Models}} & \multicolumn{6}{c}{\textbf{Open-Source Models}} \\
    \cmidrule(lr){2-6} \cmidrule(lr){7-12}
    \textbf{Sub-category} & \textbf{Sora2} & \textbf{Kling} & \textbf{Wan2.6} & \textbf{Seedance2.0} & \textbf{Veo3.1-Fast} & \textbf{LTX2.3} & \textbf{Wan2.2-14B} & \textbf{UniVideo} & \textbf{HunyuanVideo-1.5} & \textbf{Cosmos-Predict2.5} & \textbf{LongCat-Video} \\
    \midrule
    \begin{tabular}[c]{@{}l@{}}Experimental\\Science\end{tabular} & 23.3 & 52.5 & 37.5 & 52.5 & 47.5 & 21.2 & 28.1 & 20.5 & 25.3 & 21.2 & 26.0 \\
    \cmidrule(lr){1-12}
    \begin{tabular}[c]{@{}l@{}}Spatial\\Geometry\end{tabular} & 26.0 & 20.0 & 27.5 & 34.0 & 28.0 & 15.8 & 14.8 & 11.5 & 12.0 & 14.2 & 15.8 \\
    \cmidrule(lr){1-12}
    \begin{tabular}[c]{@{}l@{}}Quantitative\\Math\end{tabular} & 27.5 & 8.8 & 28.1 & 23.5 & 22.8 & 10.2 & 10.2 & 9.6 & 11.8 & 10.2 & 9.6 \\
    \cmidrule(lr){1-12}
    \begin{tabular}[c]{@{}l@{}}Logic\\Puzzles\end{tabular} & 33.3 & 20.0 & 6.7 & 26.7 & 13.3 & 16.7 & 11.7 & 13.3 & 15.0 & 15.0 & 13.3 \\
    \cmidrule(lr){1-12}
    \begin{tabular}[c]{@{}l@{}}Pattern\\Discovery\end{tabular} & 30.0 & 20.0 & 20.0 & 10.0 & 0.0 & 9.8 & 5.9 & 3.9 & 9.8 & 11.8 & 9.8 \\
    \cmidrule(lr){1-12}
    \textbf{Average} & 26.9 & 24.1 & 27.5 & 33.1 & 27.1 & 15.0 & 15.5 & 12.6 & 15.2 & 14.5 & 15.6 \\
    \bottomrule
  \end{tabular}}
\end{table*}

\begin{table*}[htp]
  \caption{\textbf{Process-aware Reasoning Verification detailed results on Information-Based. Outcome QA accuracy (\%) across Information-Based subcategories.}}
  \label{tab:app-qa-information-based}
  \centering
  \scriptsize
  \renewcommand{\arraystretch}{1.10}
  \resizebox{\textwidth}{!}{
  \begin{tabular}{l|ccccc|cccccc}
    \toprule
    & \multicolumn{5}{c|}{\textbf{Closed-Source Models}} & \multicolumn{6}{c}{\textbf{Open-Source Models}} \\
    \cmidrule(lr){2-6} \cmidrule(lr){7-12}
    \textbf{Sub-category} & \textbf{Sora2} & \textbf{Kling} & \textbf{Wan2.6} & \textbf{Seedance2.0} & \textbf{Veo3.1-Fast} & \textbf{LTX2.3} & \textbf{Wan2.2-14B} & \textbf{UniVideo} & \textbf{HunyuanVideo-1.5} & \textbf{Cosmos-Predict2.5} & \textbf{LongCat-Video} \\
    \midrule
    \begin{tabular}[c]{@{}l@{}}Data\\Reading\end{tabular} & 35.5 & 28.1 & 24.2 & 32.3 & 14.5 & 35.3 & 32.0 & 20.9 & 34.0 & 32.7 & 34.6 \\
    \cmidrule(lr){1-12}
    \begin{tabular}[c]{@{}l@{}}Process\\Timeline\end{tabular} & 56.0 & 22.5 & 40.0 & 54.3 & 42.9 & 11.6 & 13.0 & 14.4 & 17.1 & 9.6 & 19.9 \\
    \cmidrule(lr){1-12}
    \begin{tabular}[c]{@{}l@{}}Visual\\Editing\end{tabular} & 20.0 & 50.0 & 40.0 & 65.0 & 30.0 & 11.0 & 12.1 & 7.7 & 14.3 & 15.4 & 16.5 \\
    \cmidrule(lr){1-12}
    \begin{tabular}[c]{@{}l@{}}Knowledge\\Media\end{tabular} & 20.0 & 32.0 & 28.0 & 52.0 & 24.0 & 21.3 & 17.3 & 16.0 & 21.3 & 18.7 & 22.7 \\
    \cmidrule(lr){1-12}
    \begin{tabular}[c]{@{}l@{}}Creative\\Expression\end{tabular} & 46.7 & 66.7 & 60.0 & 60.0 & 60.0 & 42.2 & 28.9 & 33.3 & 35.6 & 28.9 & 31.1 \\
    \cmidrule(lr){1-12}
    \textbf{Average} & 37.1 & 33.8 & 34.0 & 47.1 & 28.7 & 22.7 & 20.6 & 17.1 & 23.9 & 20.6 & 25.1 \\
    \bottomrule
  \end{tabular}}
\end{table*}

\paragraph{Result analysis.}
The QA-only breakdown confirms that reasoning difficulty is highly category dependent. World Knowledge and Human-Centric categories show the clearest closed-source advantage, whereas Logic Reasoning remains difficult for all generators, with low averages even for the best systems. Information-Based reasoning is more polarized: models can perform well on structured timeline or creative-expression cases, but exact data reading and visual editing remain brittle. Open-source models occasionally approach closed-source performance on narrow subcategories such as Data Reading, yet their averages remain much lower because temporal and mechanistic consistency fails across the broader set.

\subsection{Multi-dimensional Quality Assessment}

The following tables decompose the raw 1--5 quality scores into reasoning quality, temporal consistency, and visual aesthetics. This view clarifies whether a model's aggregate $S(v)$ comes from genuine reasoning quality or from stronger temporal and visual presentation.

\begin{table*}[htp]
  \caption{\textbf{Multi-dimensional Quality Assessment detailed results on World Knowledge.} For each World Knowledge sub-category, we report reasoning quality, temporal consistency, and visual aesthetics on a 1--5 scale.}
  \label{tab:app-quality-world-knowledge}
  \centering
  \scriptsize
  \renewcommand{\arraystretch}{1.05}
  \resizebox{\textwidth}{!}{
  \begin{tabular}{ll|ccccc|cccccc}
    \toprule
    \multicolumn{2}{c|}{} & \multicolumn{5}{c|}{\textbf{Closed-Source Models}} & \multicolumn{6}{c}{\textbf{Open-Source Models}} \\
    \cmidrule(lr){3-7} \cmidrule(lr){8-13}
    \textbf{Sub-category} & \textbf{Metric} & \textbf{Sora2} & \textbf{Kling} & \textbf{Wan2.6} & \textbf{Seedance2.0} & \textbf{Veo3.1-Fast} & \textbf{LTX2.3} & \textbf{Wan2.2-14B} & \textbf{UniVideo} & \textbf{HunyuanVideo-1.5} & \textbf{Cosmos-Predict2.5} & \textbf{LongCat-Video} \\
    \midrule
    \multirow{3}{*}{\begin{tabular}[c]{@{}l@{}}Material\\Change\end{tabular}} & Reasoning & 1.33 & 3.50 & 2.25 & 3.00 & 4.00 & 1.59 & 2.64 & 2.29 & 2.41 & 2.41 & 2.05 \\
    & Temporal & 3.67 & 4.00 & 3.00 & 3.33 & 4.25 & 3.14 & 3.45 & 2.95 & 3.36 & 4.05 & 3.32 \\
    & Aesthetics & 3.33 & 4.00 & 3.50 & 4.00 & 4.50 & 3.09 & 3.73 & 3.33 & 3.64 & 3.95 & 3.68 \\
    \cmidrule(lr){1-13}
    \multirow{3}{*}{\begin{tabular}[c]{@{}l@{}}Public\\Systems\end{tabular}} & Reasoning & 2.00 & 3.50 & 3.50 & 3.50 & 3.50 & 1.38 & 2.86 & 1.95 & 1.86 & 1.86 & 2.43 \\
    & Temporal & 4.00 & 4.00 & 5.00 & 5.00 & 5.00 & 2.67 & 3.62 & 2.90 & 2.71 & 3.43 & 3.43 \\
    & Aesthetics & 3.00 & 4.50 & 5.00 & 4.50 & 5.00 & 2.62 & 3.76 & 3.10 & 3.29 & 3.43 & 3.67 \\
    \cmidrule(lr){1-13}
    \multirow{3}{*}{\begin{tabular}[c]{@{}l@{}}World\\Mechanics\end{tabular}} & Reasoning & 4.29 & 4.14 & 3.62 & 3.38 & 3.62 & 2.15 & 2.90 & 1.45 & 2.05 & 1.90 & 2.05 \\
    & Temporal & 4.00 & 4.29 & 4.00 & 3.62 & 3.88 & 2.90 & 3.45 & 2.10 & 2.50 & 3.15 & 3.20 \\
    & Aesthetics & 4.14 & 4.43 & 4.25 & 4.12 & 4.00 & 2.95 & 3.70 & 2.65 & 3.10 & 3.40 & 3.45 \\
    \cmidrule(lr){1-13}
    \multirow{3}{*}{\begin{tabular}[c]{@{}l@{}}Cultural\\Life\end{tabular}} & Reasoning & 2.50 & 2.33 & 3.00 & 2.50 & 4.00 & 1.11 & 2.53 & 2.16 & 2.42 & 2.17 & 1.95 \\
    & Temporal & 5.00 & 3.33 & 3.50 & 3.00 & 4.00 & 1.94 & 3.21 & 3.11 & 3.00 & 3.33 & 2.89 \\
    & Aesthetics & 3.50 & 3.67 & 4.00 & 3.50 & 4.50 & 2.28 & 3.58 & 3.21 & 3.42 & 3.50 & 3.32 \\
    \cmidrule(lr){1-13}
    \multirow{3}{*}{\begin{tabular}[c]{@{}l@{}}Everyday\\Living\end{tabular}} & Reasoning & 3.00 & 4.20 & 4.20 & 3.00 & 4.00 & 1.26 & 1.95 & 1.79 & 2.26 & 1.63 & 1.72 \\
    & Temporal & 5.00 & 4.40 & 4.20 & 3.33 & 3.80 & 2.32 & 2.79 & 2.84 & 2.63 & 3.21 & 2.83 \\
    & Aesthetics & 4.00 & 4.40 & 4.60 & 4.00 & 4.20 & 2.79 & 3.42 & 3.21 & 3.37 & 3.32 & 3.22 \\
    \cmidrule(lr){1-13}
    \multirow{3}{*}{\begin{tabular}[c]{@{}l@{}}Earth\\Cycles\end{tabular}} & Reasoning & 2.83 & 4.20 & 2.83 & 3.83 & 4.33 & 1.41 & 1.94 & 2.00 & 2.59 & 2.47 & 1.82 \\
    & Temporal & 3.83 & 4.80 & 3.67 & 4.33 & 4.33 & 2.71 & 3.00 & 3.00 & 3.35 & 3.53 & 3.41 \\
    & Aesthetics & 3.33 & 4.60 & 4.00 & 4.50 & 4.33 & 2.59 & 3.47 & 3.35 & 3.82 & 3.76 & 3.35 \\
    \cmidrule(lr){1-13}
    \multirow{3}{*}{\begin{tabular}[c]{@{}l@{}}Living\\World\end{tabular}} & Reasoning & 4.33 & 3.67 & 4.67 & 5.00 & 5.00 & 1.25 & 1.00 & 1.00 & 1.38 & 1.25 & 1.25 \\
    & Temporal & 4.33 & 3.67 & 4.33 & 5.00 & 5.00 & 2.50 & 2.75 & 2.88 & 2.88 & 3.00 & 2.75 \\
    & Aesthetics & 4.33 & 4.00 & 4.67 & 4.67 & 5.00 & 2.75 & 3.12 & 2.75 & 3.12 & 3.00 & 3.38 \\
    \cmidrule(lr){1-13}
    \multirow{3}{*}{\textbf{Average}} & Reasoning & 3.22 & 3.79 & 3.43 & 3.52 & 4.03 & 1.48 & 2.40 & 1.88 & 2.20 & 2.02 & 1.97 \\
    & Temporal & 4.09 & 4.17 & 3.90 & 3.93 & 4.20 & 2.62 & 3.24 & 2.81 & 2.92 & 3.43 & 3.16 \\
    & Aesthetics & 3.74 & 4.28 & 4.23 & 4.22 & 4.37 & 2.74 & 3.59 & 3.11 & 3.41 & 3.53 & 3.46 \\
    \bottomrule
  \end{tabular}}
\end{table*}

\begin{table*}[htp]
  \caption{\textbf{Multi-dimensional Quality Assessment detailed results on Human-Centric.} For each Human-Centric sub-category, we report reasoning quality, temporal consistency, and visual aesthetics on a 1--5 scale.}
  \label{tab:app-quality-human-centric}
  \centering
  \scriptsize
  \renewcommand{\arraystretch}{1.05}
  \resizebox{\textwidth}{!}{
  \begin{tabular}{ll|ccccc|cccccc}
    \toprule
    \multicolumn{2}{c|}{} & \multicolumn{5}{c|}{\textbf{Closed-Source Models}} & \multicolumn{6}{c}{\textbf{Open-Source Models}} \\
    \cmidrule(lr){3-7} \cmidrule(lr){8-13}
    \textbf{Sub-category} & \textbf{Metric} & \textbf{Sora2} & \textbf{Kling} & \textbf{Wan2.6} & \textbf{Seedance2.0} & \textbf{Veo3.1-Fast} & \textbf{LTX2.3} & \textbf{Wan2.2-14B} & \textbf{UniVideo} & \textbf{HunyuanVideo-1.5} & \textbf{Cosmos-Predict2.5} & \textbf{LongCat-Video} \\
    \midrule
    \multirow{3}{*}{\begin{tabular}[c]{@{}l@{}}Object\\Handling\end{tabular}} & Reasoning & 2.50 & 3.44 & 4.00 & 3.60 & 4.22 & 1.48 & 2.33 & 2.00 & 2.19 & 2.44 & 2.37 \\
    & Temporal & 3.00 & 4.33 & 4.44 & 4.20 & 4.22 & 2.63 & 3.19 & 3.00 & 2.89 & 3.33 & 3.15 \\
    & Aesthetics & 3.50 & 4.22 & 4.56 & 4.20 & 4.56 & 2.74 & 3.63 & 3.30 & 3.44 & 3.48 & 3.33 \\
    \cmidrule(lr){1-13}
    \multirow{3}{*}{\begin{tabular}[c]{@{}l@{}}Social\\Scenes\end{tabular}} & Reasoning & 5.00 & 3.33 & 3.33 & 5.00 & 3.40 & 1.40 & 2.21 & 1.87 & 2.13 & 2.07 & 2.33 \\
    & Temporal & 4.50 & 3.50 & 3.83 & 4.50 & 4.00 & 2.73 & 3.36 & 3.13 & 2.87 & 3.47 & 3.73 \\
    & Aesthetics & 4.50 & 3.83 & 4.33 & 4.50 & 4.20 & 2.87 & 3.79 & 3.13 & 3.33 & 3.53 & 3.80 \\
    \cmidrule(lr){1-13}
    \multirow{3}{*}{\begin{tabular}[c]{@{}l@{}}Skilled\\Action\end{tabular}} & Reasoning & 5.00 & 5.00 & 4.33 & 5.00 & 4.33 & 2.00 & 3.69 & 2.92 & 2.92 & 2.17 & 3.46 \\
    & Temporal & 5.00 & 5.00 & 4.33 & 4.67 & 3.67 & 3.31 & 4.00 & 3.46 & 3.62 & 3.42 & 3.77 \\
    & Aesthetics & 4.67 & 5.00 & 4.33 & 5.00 & 4.00 & 3.15 & 4.00 & 3.54 & 3.77 & 3.33 & 4.00 \\
    \cmidrule(lr){1-13}
    \multirow{3}{*}{\begin{tabular}[c]{@{}l@{}}Personal\\Routine\end{tabular}} & Reasoning & -- & 3.00 & 3.75 & -- & 2.00 & 1.50 & 2.67 & 1.92 & 2.17 & 1.50 & 1.92 \\
    & Temporal & -- & 4.00 & 4.00 & -- & 3.00 & 2.50 & 3.33 & 2.83 & 3.00 & 3.08 & 3.17 \\
    & Aesthetics & -- & 3.75 & 4.75 & -- & 4.00 & 2.58 & 3.42 & 3.00 & 3.25 & 3.17 & 3.83 \\
    \cmidrule(lr){1-13}
    \multirow{3}{*}{\begin{tabular}[c]{@{}l@{}}Public\\Conduct\end{tabular}} & Reasoning & 3.00 & 3.78 & 2.50 & 4.17 & 4.14 & 1.10 & 3.00 & 2.50 & 2.60 & 1.90 & 2.20 \\
    & Temporal & 3.00 & 4.22 & 3.25 & 4.33 & 4.14 & 1.80 & 3.30 & 3.50 & 3.00 & 3.50 & 3.00 \\
    & Aesthetics & 4.00 & 4.22 & 4.00 & 4.33 & 4.57 & 2.60 & 3.50 & 3.60 & 3.50 & 3.30 & 3.50 \\
    \cmidrule(lr){1-13}
    \multirow{3}{*}{\textbf{Average}} & Reasoning & 4.00 & 3.61 & 3.47 & 4.25 & 3.96 & 1.51 & 2.68 & 2.18 & 2.35 & 2.11 & 2.45 \\
    & Temporal & 4.00 & 4.16 & 3.93 & 4.38 & 4.04 & 2.64 & 3.39 & 3.14 & 3.04 & 3.36 & 3.35 \\
    & Aesthetics & 4.22 & 4.16 & 4.37 & 4.44 & 4.40 & 2.79 & 3.67 & 3.30 & 3.45 & 3.39 & 3.64 \\
    \bottomrule
  \end{tabular}}
\end{table*}

\begin{table*}[htp]
  \caption{\textbf{Multi-dimensional Quality Assessment detailed results on Logic Reasoning.} For each Logic Reasoning sub-category, we report reasoning quality, temporal consistency, and visual aesthetics on a 1--5 scale.}
  \label{tab:app-quality-logic-reasoning}
  \centering
  \scriptsize
  \renewcommand{\arraystretch}{1.05}
  \resizebox{\textwidth}{!}{
  \begin{tabular}{ll|ccccc|cccccc}
    \toprule
    \multicolumn{2}{c|}{} & \multicolumn{5}{c|}{\textbf{Closed-Source Models}} & \multicolumn{6}{c}{\textbf{Open-Source Models}} \\
    \cmidrule(lr){3-7} \cmidrule(lr){8-13}
    \textbf{Sub-category} & \textbf{Metric} & \textbf{Sora2} & \textbf{Kling} & \textbf{Wan2.6} & \textbf{Seedance2.0} & \textbf{Veo3.1-Fast} & \textbf{LTX2.3} & \textbf{Wan2.2-14B} & \textbf{UniVideo} & \textbf{HunyuanVideo-1.5} & \textbf{Cosmos-Predict2.5} & \textbf{LongCat-Video} \\
    \midrule
    \multirow{3}{*}{\begin{tabular}[c]{@{}l@{}}Experimental\\Science\end{tabular}} & Reasoning & 2.83 & 3.00 & 3.14 & 3.75 & 2.75 & 1.28 & 1.43 & 1.38 & 1.66 & 1.17 & 1.72 \\
    & Temporal & 3.00 & 3.25 & 3.14 & 4.00 & 2.50 & 2.34 & 2.29 & 2.24 & 2.52 & 2.21 & 2.79 \\
    & Aesthetics & 3.67 & 4.12 & 3.43 & 4.50 & 3.50 & 2.72 & 2.93 & 2.48 & 3.21 & 2.93 & 3.07 \\
    \cmidrule(lr){1-13}
    \multirow{3}{*}{\begin{tabular}[c]{@{}l@{}}Spatial\\Geometry\end{tabular}} & Reasoning & 2.80 & 2.43 & 1.71 & 2.88 & 1.71 & 1.03 & 1.08 & 1.03 & 1.06 & 1.26 & 1.14 \\
    & Temporal & 3.10 & 2.86 & 2.57 & 3.50 & 2.29 & 1.61 & 1.92 & 1.25 & 1.36 & 2.29 & 1.78 \\
    & Aesthetics & 3.80 & 3.29 & 2.71 & 3.50 & 3.14 & 1.94 & 2.31 & 1.94 & 2.14 & 2.94 & 2.00 \\
    \cmidrule(lr){1-13}
    \multirow{3}{*}{\begin{tabular}[c]{@{}l@{}}Quantitative\\Math\end{tabular}} & Reasoning & 1.83 & 1.30 & 2.11 & 1.67 & 1.67 & 1.14 & 1.10 & 1.00 & 1.00 & 1.07 & 1.00 \\
    & Temporal & 3.50 & 2.30 & 2.78 & 2.50 & 2.00 & 1.59 & 1.72 & 1.26 & 1.10 & 2.24 & 1.61 \\
    & Aesthetics & 3.50 & 2.60 & 3.33 & 3.00 & 2.44 & 1.86 & 2.00 & 1.90 & 2.00 & 2.93 & 1.82 \\
    \cmidrule(lr){1-13}
    \multirow{3}{*}{\begin{tabular}[c]{@{}l@{}}Logic\\Puzzles\end{tabular}} & Reasoning & 1.00 & 1.00 & 1.00 & 1.00 & 1.00 & 1.33 & 1.17 & 1.08 & 1.00 & 1.50 & 1.08 \\
    & Temporal & 2.00 & 2.33 & 2.33 & 3.00 & 1.33 & 1.92 & 2.17 & 1.33 & 1.50 & 3.00 & 1.67 \\
    & Aesthetics & 3.00 & 3.33 & 3.00 & 3.67 & 2.67 & 2.33 & 2.58 & 2.42 & 2.33 & 3.25 & 2.25 \\
    \cmidrule(lr){1-13}
    \multirow{3}{*}{\begin{tabular}[c]{@{}l@{}}Pattern\\Discovery\end{tabular}} & Reasoning & 1.50 & 2.00 & 1.33 & 3.50 & 2.67 & 1.00 & 1.00 & 1.00 & 1.10 & 1.00 & 1.00 \\
    & Temporal & 2.50 & 2.67 & 2.67 & 4.00 & 2.67 & 1.00 & 1.60 & 1.20 & 1.50 & 2.60 & 2.00 \\
    & Aesthetics & 2.50 & 3.00 & 3.00 & 4.00 & 3.33 & 1.80 & 2.20 & 2.30 & 2.30 & 3.30 & 2.20 \\
    \cmidrule(lr){1-13}
    \multirow{3}{*}{\textbf{Average}} & Reasoning & 2.30 & 2.03 & 2.07 & 2.70 & 2.00 & 1.15 & 1.17 & 1.11 & 1.19 & 1.19 & 1.23 \\
    & Temporal & 3.00 & 2.71 & 2.76 & 3.41 & 2.20 & 1.77 & 1.96 & 1.50 & 1.61 & 2.36 & 2.00 \\
    & Aesthetics & 3.52 & 3.26 & 3.14 & 3.74 & 3.00 & 2.15 & 2.40 & 2.14 & 2.40 & 3.00 & 2.27 \\
    \bottomrule
  \end{tabular}}
\end{table*}

\begin{table*}[htp]
  \caption{\textbf{Multi-dimensional Quality Assessment detailed results on Information-Based.} For each Information-Based sub-category, we report reasoning quality, temporal consistency, and visual aesthetics on a 1--5 scale.}
  \label{tab:app-quality-information-based}
  \centering
  \scriptsize
  \renewcommand{\arraystretch}{1.05}
  \resizebox{\textwidth}{!}{
  \begin{tabular}{ll|ccccc|cccccc}
    \toprule
    \multicolumn{2}{c|}{} & \multicolumn{5}{c|}{\textbf{Closed-Source Models}} & \multicolumn{6}{c}{\textbf{Open-Source Models}} \\
    \cmidrule(lr){3-7} \cmidrule(lr){8-13}
    \textbf{Sub-category} & \textbf{Metric} & \textbf{Sora2} & \textbf{Kling} & \textbf{Wan2.6} & \textbf{Seedance2.0} & \textbf{Veo3.1-Fast} & \textbf{LTX2.3} & \textbf{Wan2.2-14B} & \textbf{UniVideo} & \textbf{HunyuanVideo-1.5} & \textbf{Cosmos-Predict2.5} & \textbf{LongCat-Video} \\
    \midrule
    \multirow{3}{*}{\begin{tabular}[c]{@{}l@{}}Data\\Reading\end{tabular}} & Reasoning & 2.00 & 1.55 & 1.75 & 1.17 & 1.25 & 1.03 & 1.38 & 1.00 & 1.03 & 1.10 & 1.24 \\
    & Temporal & 3.33 & 2.18 & 2.42 & 2.08 & 1.83 & 2.41 & 2.66 & 1.07 & 1.38 & 2.03 & 1.97 \\
    & Aesthetics & 3.17 & 2.45 & 2.58 & 2.33 & 2.42 & 2.93 & 2.93 & 1.55 & 1.86 & 2.83 & 2.00 \\
    \cmidrule(lr){1-13}
    \multirow{3}{*}{\begin{tabular}[c]{@{}l@{}}Process\\Timeline\end{tabular}} & Reasoning & 4.40 & 3.12 & 2.50 & 2.71 & 3.86 & 1.45 & 1.48 & 1.45 & 1.52 & 1.45 & 1.45 \\
    & Temporal & 4.40 & 3.62 & 3.25 & 3.14 & 3.86 & 2.59 & 2.72 & 2.41 & 2.38 & 2.62 & 2.31 \\
    & Aesthetics & 4.40 & 4.12 & 4.00 & 3.71 & 4.43 & 2.86 & 3.31 & 3.00 & 3.00 & 3.10 & 2.86 \\
    \cmidrule(lr){1-13}
    \multirow{3}{*}{\begin{tabular}[c]{@{}l@{}}Visual\\Editing\end{tabular}} & Reasoning & 1.00 & 2.25 & 2.00 & 2.75 & 2.50 & 1.06 & 1.22 & 1.00 & 1.17 & 1.06 & 1.39 \\
    & Temporal & 2.00 & 3.00 & 2.25 & 3.25 & 2.50 & 1.83 & 2.00 & 1.17 & 1.56 & 1.76 & 1.72 \\
    & Aesthetics & 2.75 & 3.50 & 3.25 & 3.75 & 3.25 & 2.50 & 2.56 & 1.56 & 2.28 & 2.65 & 2.22 \\
    \cmidrule(lr){1-13}
    \multirow{3}{*}{\begin{tabular}[c]{@{}l@{}}Knowledge\\Media\end{tabular}} & Reasoning & 2.00 & 2.60 & 1.80 & 3.40 & 3.00 & 1.60 & 1.86 & 1.53 & 2.00 & 1.73 & 1.93 \\
    & Temporal & 2.00 & 3.00 & 3.20 & 3.20 & 3.40 & 2.47 & 2.43 & 2.07 & 2.40 & 3.00 & 2.80 \\
    & Aesthetics & 3.00 & 3.40 & 3.40 & 3.40 & 3.60 & 2.73 & 2.57 & 2.40 & 2.60 & 3.33 & 2.87 \\
    \cmidrule(lr){1-13}
    \multirow{3}{*}{\begin{tabular}[c]{@{}l@{}}Creative\\Expression\end{tabular}} & Reasoning & 5.00 & 5.00 & 3.67 & 4.00 & 5.00 & 2.11 & 3.00 & 2.89 & 3.44 & 2.44 & 2.44 \\
    & Temporal & 5.00 & 5.00 & 4.00 & 4.33 & 5.00 & 3.33 & 4.00 & 3.78 & 3.67 & 3.67 & 3.44 \\
    & Aesthetics & 4.67 & 5.00 & 4.33 & 5.00 & 5.00 & 3.22 & 4.00 & 3.89 & 3.89 & 3.78 & 3.56 \\
    \cmidrule(lr){1-13}
    \multirow{3}{*}{\textbf{Average}} & Reasoning & 2.68 & 2.55 & 2.16 & 2.35 & 2.65 & 1.34 & 1.60 & 1.38 & 1.56 & 1.41 & 1.54 \\
    & Temporal & 3.48 & 3.06 & 2.88 & 2.87 & 2.94 & 2.45 & 2.65 & 1.87 & 2.06 & 2.45 & 2.28 \\
    & Aesthetics & 3.52 & 3.42 & 3.31 & 3.26 & 3.42 & 2.83 & 3.02 & 2.31 & 2.56 & 3.04 & 2.56 \\
    \bottomrule
  \end{tabular}}
\end{table*}

\paragraph{Result analysis.}
The quality-score breakdown shows that the three axes capture different failure modes. Temporal consistency and visual aesthetics are generally higher than reasoning quality, especially for closed-source models, which explains why a model can obtain a strong $S(v)$ while still underperforming on process-aware reasoning. The gap is largest in Logic Reasoning and Information-Based categories: many videos remain visually coherent, but the reasoning-quality row drops sharply when exact symbolic, causal, or text-grounded structure must be preserved. Among open-source models, Wan2.2-14B and Cosmos-Predict2.5 often retain better perceptual quality than reasoning quality, reinforcing the need to report the axes separately rather than relying only on an aggregate score.

\section{Full Frame-Rate Ablation Results}
\label{app:fps-ablation-full}

Tables~\ref{tab:fps_ablation} and~\ref{tab:fps_ablation_full} provide the frame-rate ablation results. The compact table summarizes overall accuracy and reasoning gap across FPS settings, while the full table includes per-category QA accuracy, reasoning-phase scores, selected 4~FPS open-source results, and the average rows used to select the default setting. Average rows are computed over the six closed-source models with all three FPS settings. Token costs per 5s video are approximately 4.5k, 9.0k, and 12.2k visual tokens for 2, 4, and 8~FPS respectively when using Qwen3.5-27B.

\begin{table*}[htp]
  \caption{\textbf{Compact frame-rate ablation for Reasoning Verification.} We report only overall QA accuracy ($\mathrm{Acc}_{\mathrm{QA}}$, \%) and the reasoning gap ($\Delta_\mathrm{RG}$) across three frame rates for six closed-source models. Bold marks the best overall accuracy per model; full per-category, reasoning-phase, and open-source 4~FPS results are provided in Table~\ref{tab:fps_ablation_full}.}
  \label{tab:fps_ablation}
  \centering
  \small
  \renewcommand{\arraystretch}{1.08}
  \setlength{\tabcolsep}{4pt}
  \resizebox{\textwidth}{!}{
  \begin{tabular}{llccccccc}
    \toprule
    \textbf{FPS} & \textbf{Metric} & \textbf{Kling} & \textbf{Seedance2.0} & \textbf{Sora2-8s} & \textbf{Sora2-12s} & \textbf{Veo3.1-Fast} & \textbf{Wan2.6} & \textbf{Average} \\
    \midrule
    \multirow{2}{*}{2} & Overall & 34.0 & 37.8 & 34.4 & 36.3 & 34.8 & 31.9 & 34.9 \\
      & $\Delta_\mathrm{RG}$ & $+$0.144 & $+$0.169 & $+$0.101 & $+$0.072 & $+$0.079 & $+$0.152 & $+$0.120 \\
    \midrule
    \multirow{2}{*}{4} & Overall & \textbf{36.3} & \textbf{41.8} & 37.1 & 35.2 & 37.3 & \textbf{35.8} & 37.2 \\
      & $\Delta_\mathrm{RG}$ & $+$0.122 & $+$0.135 & $+$0.097 & $+$0.090 & $+$0.108 & $+$0.127 & $+$0.113 \\
    \midrule
    \multirow{2}{*}{8} & Overall & 35.3 & 40.5 & \textbf{41.0} & \textbf{36.9} & \textbf{38.0} & 33.7 & \textbf{37.6} \\
      & $\Delta_\mathrm{RG}$ & $+$0.130 & $+$0.129 & $+$0.146 & $+$0.045 & $+$0.096 & $+$0.074 & $+$0.098 \\
    \bottomrule
  \end{tabular}}
\end{table*}

\begin{table*}[htp]
  \caption{\textbf{Full frame-rate ablation for Reasoning Verification.} We evaluate three frame rates (2, 4, 8~FPS) on six closed-source models across four reasoning categories, plus five open-source models at the selected 4~FPS default. WK, HC, LR, and IB denote World Knowledge, Human-Centric, Logic Reasoning, and Information-Based. We report per-category QA accuracy ($\mathrm{Acc}_{\mathrm{QA}}$, \%), overall average, and the four reasoning-phase scores ($s_\mathrm{state}$, $s_\mathrm{proc}$, $s_\mathrm{fidel}$, $s_\mathrm{mech}$, \%). Bold indicates the best FPS setting per model.}
  \label{tab:fps_ablation_full}
  \centering
  \small
  \renewcommand{\arraystretch}{1.05}
  \setlength{\tabcolsep}{4pt}
  \resizebox{\textwidth}{!}{
  \begin{tabular}{llcccccccccc}
    \toprule
    \textbf{Model} & \textbf{FPS} & \textbf{WK} & \textbf{HC} & \textbf{LR} & \textbf{IB} & \textbf{Overall} & $s_\mathrm{state}$ & $s_\mathrm{proc}$ & $s_\mathrm{fidel}$ & $s_\mathrm{mech}$ & $\Delta_\mathrm{RG}$ \\
    \midrule
    \multirow{3}{*}{Kling}
      & 2 & 39.2 & 38.7 & 26.9 & 31.3 & 34.0 & 45.2 & 23.9 & 35.7 & 28.2 & $+$0.144 \\
      & \textbf{4} & \textbf{43.4} & \textbf{43.2} & 24.4 & \textbf{34.0} & \textbf{36.3} & 46.4 & 25.9 & 36.5 & 32.6 & $+$0.122 \\
      & 8 & 41.6 & 40.0 & \textbf{25.6} & 33.9 & 35.3 & 42.0 & 24.7 & 40.5 & 31.8 & $+$0.130 \\
    \midrule
    \multirow{3}{*}{Seedance2.0}
      & 2 & 41.8 & 33.8 & 35.1 & 40.5 & 37.8 & 55.5 & 24.8 & 34.9 & 31.7 & $+$0.169 \\
      & \textbf{4} & \textbf{45.2} & \textbf{41.0} & 33.3 & \textbf{47.5} & \textbf{41.8} & 56.3 & 31.7 & 39.0 & 36.6 & $+$0.135 \\
      & 8 & 41.7 & 39.8 & \textbf{37.6} & 43.0 & 40.5 & 52.0 & 28.8 & 39.3 & 36.6 & $+$0.129 \\
    \midrule
    \multirow{3}{*}{Sora2-8s}
      & 2 & 40.0 & 37.8 & 23.9 & 36.0 & 34.4 & 43.7 & 26.1 & 34.9 & 32.3 & $+$0.101 \\
      & 4 & 40.1 & 44.1 & 27.1 & 37.2 & 37.1 & 51.5 & 27.1 & 32.4 & 37.3 & $+$0.097 \\
      & \textbf{8} & \textbf{43.8} & \textbf{53.3} & \textbf{29.7} & 37.2 & \textbf{41.0} & 49.4 & 24.8 & 45.3 & 40.7 & $+$0.146 \\
    \midrule
    \multirow{3}{*}{Sora2-12s}
      & 2 & \textbf{35.7} & 44.4 & 28.7 & \textbf{36.4} & \textbf{36.3} & 46.5 & 34.2 & 32.7 & 30.6 & $+$0.072 \\
      & 4 & 35.9 & 42.2 & 28.7 & 34.0 & 35.2 & 40.3 & 27.5 & 38.4 & 33.1 & $+$0.090 \\
      & 8 & 33.8 & \textbf{46.7} & \textbf{30.0} & 37.2 & 36.9 & 40.2 & 32.7 & 38.5 & 37.0 & $+$0.045 \\
    \midrule
    \multirow{3}{*}{Veo3.1-Fast}
      & 2 & 51.2 & 36.9 & 25.6 & 25.6 & 34.8 & 39.0 & 29.4 & 38.0 & 31.7 & $+$0.079 \\
      & 4 & \textbf{52.4} & 40.8 & 27.3 & 28.8 & 37.3 & 41.9 & 30.2 & 40.9 & 30.9 & $+$0.108 \\
      & \textbf{8} & 51.2 & \textbf{42.3} & \textbf{27.1} & \textbf{31.3} & \textbf{38.0} & 43.1 & 27.9 & 40.8 & 36.8 & $+$0.096 \\
    \midrule
    \multirow{3}{*}{Wan2.6}
      & 2 & 39.5 & 28.7 & 27.9 & 31.8 & 31.9 & 44.1 & 20.8 & 33.2 & 26.2 & $+$0.152 \\
      & \textbf{4} & \textbf{47.2} & \textbf{34.0} & 27.9 & \textbf{34.2} & \textbf{35.8} & 47.0 & 26.0 & 36.4 & 31.9 & $+$0.127 \\
      & 8 & 42.8 & 33.8 & 27.9 & 30.4 & 33.7 & 39.0 & 25.2 & 34.1 & 33.1 & $+$0.074 \\
    \midrule
    \multicolumn{12}{l}{\textit{Open-source models (fps=4 only)}} \\
    \midrule
    LTX2.3 & 4 & 17.0 & 15.3 & 15.0 & 22.7 & 17.5 & 24.0 & 6.0 & 20.7 & 14.7 & $+$0.121 \\
    Wan2.2-14B & 4 & 24.2 & 28.3 & 15.5 & 20.4 & 22.1 & 30.9 & 9.3 & 26.1 & 18.5 & $+$0.146 \\
    UniVideo & 4 & 21.3 & 21.3 & 12.6 & 17.1 & 18.1 & 24.8 & 6.2 & 22.9 & 14.2 & $+$0.136 \\
    HunyuanVideo-1.5 & 4 & 25.2 & 18.4 & 15.2 & 23.9 & 20.7 & 29.0 & 8.7 & 23.1 & 18.4 & $+$0.124 \\
    LongCat-Video & 4 & 20.6 & 21.0 & 15.7 & 25.0 & 20.6 & 26.8 & 9.0 & 26.8 & 15.6 & $+$0.145 \\
    \midrule
    \rowcolor{gray!10}
    \multirow{3}{*}{\textbf{Average}}
      & 2 & 41.2 & 36.7 & 28.0 & 33.6 & 34.9 & 45.7 & 26.5 & 34.9 & 30.1 & $+$0.120 \\
      & \textbf{4} & \textbf{44.0} & \textbf{40.9} & 28.1 & \textbf{35.9} & \textbf{37.2} & 47.2 & 28.1 & 37.3 & 33.7 & $+$0.113 \\
      \rowcolor{gray!10}
      & 8 & 42.5 & 42.6 & \textbf{29.7} & 35.5 & 37.6 & 44.3 & 27.3 & 39.7 & 36.0 & $+$0.098 \\
    \bottomrule
  \end{tabular}}
\end{table*}

\paragraph{Result analysis.}
The ablation supports 4~FPS as the default evaluation setting. Moving from 2~FPS to 4~FPS improves the six-model average accuracy from $34.9$\% to $37.2$\%, with gains on most closed-source models and a lower reasoning gap than 2~FPS. Although 8~FPS slightly increases the average accuracy to $37.6$\%, the gain is small relative to the additional visual-token cost, and it does not consistently improve every model. The full table also shows that higher frame rate mainly helps categories requiring dense temporal evidence, while reasoning-phase bottlenecks such as process and mechanism scores remain low; therefore, simply increasing frames cannot replace process-aware verification.

\section{Point-wise Scoring Ablation Details}
\label{app:pointwise-ablation}

We ablate point-wise scoring strategies to test whether reducing inter-dimension halo effects improves induced pairwise accuracy and rank correlation. Table~\ref{tab:pointwise-ablation} reports full-set results on the 5{,}969-pair \WorldRewardBench{} benchmark. Vanilla uses Qwen3.5-27B-Thinking with a single prompt for three scores, No-Thinking disables extended thinking, and SDE denotes Sequential Dimension Evaluation with three independent calls. Correlations r-t and r-a measure reasoning--temporal-consistency and reasoning--aesthetics coupling, respectively; parse rates are computed per unique video.

\begin{table}[htp]
  \caption{\textbf{Point-wise scoring ablation.} We report parse rate, induced pairwise accuracy (w/ and w/o ties), Spearman~$\rho$, and inter-dimension Pearson correlations as a halo-effect diagnostic on the full 5{,}969-pair set. Vanilla Thinking is the most efficient protocol and achieves the best rank correlation.}
  \label{tab:pointwise-ablation}
  \centering
  \small
  \renewcommand{\arraystretch}{1.10}
  \setlength{\tabcolsep}{3.5pt}
  \begin{tabular}{lccccccc}
    \toprule
    \textbf{Method} & \textbf{Calls} & \textbf{Parse} & \textbf{Acc w/t} & \textbf{Acc w/o} & $\boldsymbol{\rho}$ & \textbf{r-t} & \textbf{r-a} \\
    \midrule
    Vanilla         & 1 & 98.0\% & \textbf{58.45} & \textbf{67.63} & \textbf{0.626} & 0.770 & 0.741 \\
    No-Thinking     & 1 & \textbf{99.9\%} & 53.15 & 65.55 & 0.591 & 0.764 & 0.803 \\
    SDE             & 3 & 95.0\% & 58.00 & 64.64 & 0.562 & \textbf{0.384} & \textbf{0.495} \\
    \bottomrule
  \end{tabular}
\end{table}

The results reveal a tension between halo reduction and preference recovery. SDE reduces the reasoning$\leftrightarrow$temporal-consistency correlation from $0.770$ to $0.384$, but this decoupling does not translate into better rank correlation or induced accuracy. Vanilla Thinking achieves the highest $\rho$ ($0.626$) and the best w/o-ties accuracy ($67.63$\%) despite exhibiting stronger inter-dimension correlation. This suggests that human annotators themselves often assign correlated dimension scores: a video that fails reasoning also tends to receive lower temporal-consistency and aesthetics scores. SDE is therefore useful as a diagnostic tool, but vanilla scoring is the better default for efficient preference recovery.

\section{Elo Ranking Details}
\label{app:elo-details}

For the arena-style ranking in Table~\ref{tab:elo_ranking}, each \WorldRewardBench{} pair carries a human or judge verdict: model~A wins, model~B wins, or tie. We fit a shared ability vector $\boldsymbol{\theta} \in \mathbb{R}^{11}$ using the Bradley-Terry model with the Davidson tie extension~\citep{davidson1970extending}:
\begin{equation}
  P(A \succ B) = \frac{e^{\theta_A}}{e^{\theta_A} + e^{\theta_B} + 2\nu\, e^{(\theta_A+\theta_B)/2}},
  \label{eq:elo_bt}
\end{equation}
where $\nu$ is the tie propensity parameter estimated from data. The fitted tie propensities are $\nu_\mathrm{human}=0.148$ for expert preferences and $\nu_\mathrm{judge}=0.070$ for VLM-judge verdicts. We apply $\ell_2$ regularization ($\alpha=1.0$) and compute 95\% confidence intervals via 1{,}000 bootstrap resamples at the prompt-cluster level (resampling by \texttt{task\_id} to preserve intra-prompt correlation). Scores are mapped to the Elo scale as $\mathrm{Elo}_i = 1000 + \theta_i \cdot 400/\ln 10$.

\paragraph{Result analysis.}
The estimated tie propensities indicate that human annotators use ties more often than the VLM judge ($\nu_\mathrm{human}=0.148$ vs. $\nu_\mathrm{judge}=0.070$), which is consistent with the judge making sharper decisions on near-equal pairs. This difference helps explain why judge Elo can separate visually similar closed-source models differently from humans, even when the overall ranking remains strongly aligned. The bootstrap confidence intervals in the main table should therefore be interpreted together with the tie model: close model pairs are inherently less stable than the large closed-source/open-source separation.


\section{Expert Human Annotation Protocol}
\label{app:human_annotation}

This appendix describes the human annotation protocol for WorldRewardBench. The goal is to collect reliable human preference references for evaluating whether automatic judges align with human judgments of world-state reasoning in generated videos. Annotators judge whether each video correctly realizes the transition implied by the input image and prompt, rather than only its visual appeal.

\subsection{Annotators and Privacy}

WorldRewardBench was annotated by fifteen trained annotators with diverse backgrounds related to video generation. The annotator group included researchers working on video generation and multimodal evaluation, as well as users with practical experience using video generation systems in different application scenarios. Before formal annotation, all annotators received a one-hour training session covering the benchmark objective, annotation interface, scoring dimensions, representative calibration examples, and common failure cases in image-to-video generation.

The annotation task did not require annotators to provide personal or sensitive information beyond their scoring decisions. All annotators participated voluntarily in the annotation process. The annotations were used only to construct aggregate video-level scores and pairwise preference labels for WorldRewardBench. Released annotation files contain only anonymized annotator identifiers, and no personally identifying information about annotators is collected or released.

\subsection{Annotation Interface and Scoring Rubric}

Each annotation unit is an image-prompt-video tuple. For each selected benchmark case, annotators were shown the input image, the corresponding generation prompt, and eight anonymized generated videos sampled from the candidate pool. Model identities were hidden to reduce model-name bias. Annotators evaluated the full generated videos rather than sampled frames, and they could replay each video before submitting scores. The annotation assignment was randomized to avoid model-specific ordering patterns, while ensuring that each video was rated by at least two annotators.

\begin{figure}[t]
    \centering
    \includegraphics[width=\linewidth]{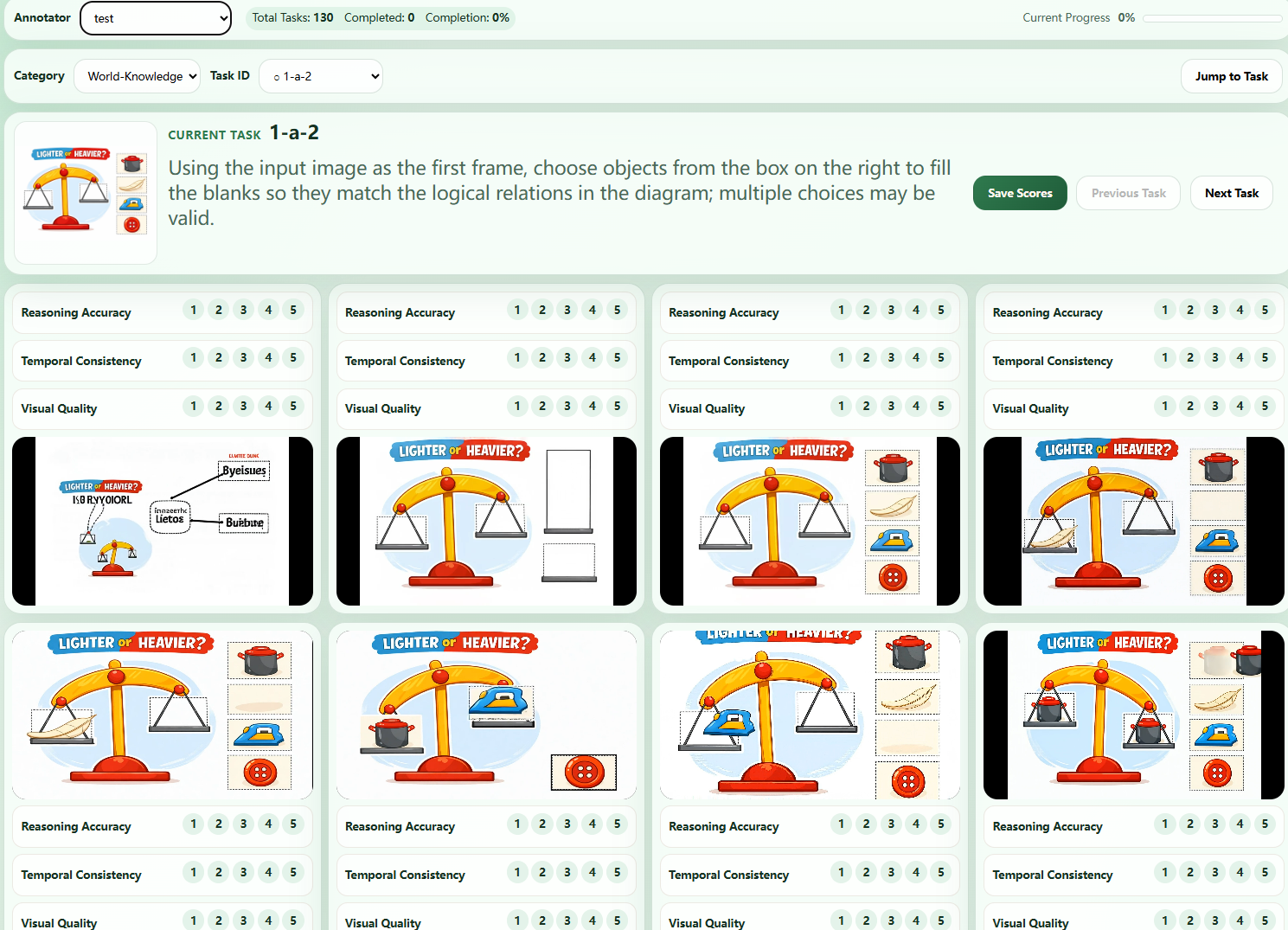}
    \caption{Human annotation interface for WorldRewardBench. Annotators see the input image, prompt, and eight anonymized generated videos. For each video, they provide three scores: Reasoning Quality, Temporal Consistency, and Visual Aesthetics. Model identities are hidden during annotation.}
    \label{fig:annotation_interface}
\end{figure}

Figure~\ref{fig:annotation_interface} shows the annotation interface. The upper panel displays the category, case ID, input image, and prompt, while the main panel presents the eight anonymized generated videos. Annotators scored each full video on three dimensions using a 1--5 scale: Reasoning Quality, Temporal Consistency, and Visual Aesthetics. The detailed scoring rubric is provided in Table~\ref{tab:human_scoring_rubric}. Annotators were instructed to treat Reasoning Quality as the primary criterion because WorldRewardBench evaluates world-state prediction rather than generic visual appeal, and not to reward visual quality alone when the video fails to satisfy the required causal, physical, logical, or informational relation.

\begin{table}[htbp]
\centering
\caption{Human scoring rubric for WorldRewardBench.}
\label{tab:human_scoring_rubric}
\scriptsize
\setlength{\tabcolsep}{2.5pt}
\renewcommand{\arraystretch}{1.08}
\begin{tabularx}{\linewidth}{cXXX}
\toprule
\multicolumn{1}{c}{\footnotesize\textbf{Score}} &
\multicolumn{1}{c}{\footnotesize\textbf{Reasoning Quality}} &
\multicolumn{1}{c}{\footnotesize\textbf{Temporal Consistency}} &
\multicolumn{1}{c}{\footnotesize\textbf{Visual Aesthetics}} \\
\midrule
1 
& Fails the prompt or required world-state transition. 
& Severe discontinuity, flickering, jumps, or deformation. 
& Very poor quality with blur, distortion, or major artifacts. \\

2 
& Weakly related to the prompt; main transition is incorrect. 
& Many visible temporal or motion problems. 
& Poor rendering with obvious artifacts. \\

3 
& Partially satisfies the prompt but has clear reasoning errors. 
& Mostly continuous, but with noticeable issues. 
& Acceptable but average visual quality. \\

4 
& Mostly satisfies the prompt with only minor reasoning issues. 
& Relatively smooth and natural, with minor glitches. 
& Good visual quality and generally natural appearance. \\

5 
& Fully satisfies the prompt with correct causal, physical, logical, or informational consistency. 
& Highly continuous with natural motion and coherent changes. 
& Refined, stable, and visually pleasing. \\
\bottomrule
\end{tabularx}
\end{table}

\subsection{Disagreement Detection}

To improve annotation reliability, we use a disagreement-based quality control procedure before final score aggregation. For each video \(v\) and each scoring dimension \(d\), we compute the score range among available annotators:
\[
R_d(v)=\max_{a\in A_v}s_{a,d}(v)-\min_{a\in A_v}s_{a,d}(v),
\]
where \(A_v\) denotes the set of annotators who rated video \(v\), and \(d\in\{r,c,a\}\) corresponds to Reasoning Quality, Temporal Consistency, and Visual Aesthetics.

A video is flagged as high-disagreement if the score range exceeds a predefined threshold. Specifically, if a video has two ratings and \(R_d(v)>1\), or if a video has three ratings and \(R_d(v)>2\), the video is assigned to additional annotators. Each high-disagreement video receives at least four valid ratings before final aggregation. We do not discard annotations solely because annotators disagree; instead, disagreement triggers additional annotation to obtain a more stable estimate.

After re-annotation, we compute inter-annotator reliability on the final annotation set. As shown in Table~\ref{tab:inter_annotator_reliability}, the final annotations show moderate-to-substantial agreement across all three dimensions. The aggregated score achieves the strongest reliability, with Krippendorff's \(\alpha=0.744\), ICC(2,k)\(=0.936\), and mean pairwise Spearman \(\rho=0.784\), indicating that the final video-level human scores are stable for constructing WorldRewardBench preferences.

For each video \(v\), we then average the final annotator scores within each dimension:
\[
\bar{s}_d(v)=\frac{1}{|A_v|}\sum_{a\in A_v}s_{a,d}(v),
\]
where \(d\in\{r,c,a\}\) denotes Reasoning Quality, Temporal Consistency, and Visual Aesthetics. The aggregated human score is computed as:
\[
S_{\mathrm{human}}(v)=0.4\bar{s}_r(v)+0.3\bar{s}_c(v)+0.3\bar{s}_a(v).
\]
The larger weight on Reasoning Quality reflects the goal of WorldRewardBench: evaluating whether generated videos correctly realize the intended world-state transition rather than merely rewarding visual appeal.

\begin{table}[htbp]
\centering
\caption{Inter-annotator reliability of human scores after re-annotation. Krippendorff's \(\alpha\) measures agreement on ordinal 1--5 ratings, ICC(2,k) measures the reliability of averaged scores, and mean pairwise Spearman \(\rho\) measures rank-level agreement between annotators.}
\label{tab:inter_annotator_reliability}
\small
\setlength{\tabcolsep}{5pt}
\renewcommand{\arraystretch}{1.12}
\begin{tabular}{lccc}
\toprule
Dimension & Krippendorff's $\alpha$ & ICC(2,k) & Mean Pairwise Spearman $\rho$ \\
\midrule
Reasoning Quality    & 0.723 & 0.939 & 0.766 \\
Temporal Consistency & 0.648 & 0.903 & 0.679 \\
Visual Aesthetics    & 0.649 & 0.903 & 0.676 \\
Aggregated Score     & 0.744 & 0.936 & 0.784 \\
\bottomrule
\end{tabular}
\end{table}

\subsection{QA Quality Audit}
\label{app:qa-audit}

To mitigate VLM bias in the automatically generated QA pairs of \WorldReasonBench{}, we conduct an independent audit on a stratified random sample of approximately $300$ QA pairs (balanced across the four reasoning dimensions and the four question types). Two trained auditors independently verify each pair against three criteria: (i) \emph{answerability}: the question is answerable from the prompt-induced visual evidence alone, without relying on hidden context; (ii) \emph{ground-truth correctness}: the labelled answer is consistent with physical, social, or factual common sense and with the prompt; (iii) \emph{ground-truth uniqueness}: no plausible alternative answer is supported by the prompt. Each pair is recorded with a binary \emph{accept}/\emph{reject} verdict and a free-form rejection reason. The two auditors reach Cohen $\kappa{=}0.78$ on the accept/reject verdict, indicating substantial agreement. The overall rejection rate is $7.8\%$, with rejections slightly more frequent on Information-Based questions ($11.4\%$) due to the stricter requirement on exact text and data preservation, and lowest on World Knowledge questions ($5.3\%$). The most common rejection reasons are \emph{ambiguous ground truth} ($43\%$ of rejections), \emph{multiple plausible answers} ($31\%$), and \emph{requires off-screen context} ($18\%$). Rejected QA pairs are either rewritten by the auditors or removed before release, so the released benchmark contains only audit-passed or human-corrected QA. The audit is treated as a quality-control pass rather than as a separate evaluation set; full audit logs and per-dimension breakdowns will be released alongside the benchmark.

\section{\WorldRewardBench{} Human Scoring Breakdown}
\label{app:reward-human-scoring}

Table~\ref{tab:rewardbench_category_breakdown} reports the expert human scores used to construct \WorldRewardBench{}, broken down by model and reasoning category. We keep this detailed annotation view in the appendix because the main text focuses on human-preference Elo and reward-model alignment.

\begin{table*}[t]
  \caption{\textbf{Human scoring breakdown on \WorldRewardBench{}.}
  We report expert annotation scores by model and reasoning category. The three raw dimensions are reasoning quality, temporal consistency, and visual aesthetics; category and overall weighted totals follow the human scoring protocol used to construct \WorldRewardBench{}.}
  \label{tab:rewardbench_category_breakdown}
  \centering
  \scriptsize
  \setlength{\tabcolsep}{3pt}
  \renewcommand{\arraystretch}{0.92}
  \resizebox{\textwidth}{!}{
  \begin{tabular}{@{}cllcccccc@{}}
    \toprule
    \textbf{Rank} 
    & \textbf{Model} 
    & \textbf{Category} 
    & \begin{tabular}[c]{@{}c@{}}\textbf{Reasoning}\\\textbf{Quality}\end{tabular}
    & \begin{tabular}[c]{@{}c@{}}\textbf{Temporal}\\\textbf{Consistency}\end{tabular}
    & \begin{tabular}[c]{@{}c@{}}\textbf{Visual}\\\textbf{Aesthetics}\end{tabular}
    & \begin{tabular}[c]{@{}c@{}}\textbf{Category}\\\textbf{Weighted Total}\end{tabular}
    & \begin{tabular}[c]{@{}c@{}}\textbf{Overall}\\\textbf{Weighted Total}\end{tabular}
    & \begin{tabular}[c]{@{}c@{}}\textbf{Sample Count}\\\textbf{per Dimension}\end{tabular} \\
    \midrule

    \multirow{4}{*}{1} & \multirow{4}{*}{seedance2.0} 
    & worldknowledge & 3.9151 & 4.2547 & 4.3396 & 4.1443 & \multirow{4}{*}{3.8225} & 106 \\
    & & humancentric & 4.3731 & 4.4179 & 4.4179 & 4.4000 & & 67 \\
    & & logicreasoning & 3.1298 & 3.5344 & 3.6260 & 3.4000 & & 131 \\
    & & informationbasedreasoning & 3.6140 & 3.6754 & 3.7368 & 3.6693 & & 114 \\
    \midrule

    \multirow{4}{*}{2} & \multirow{4}{*}{keling} 
    & worldknowledge & 3.4630 & 3.9352 & 3.9907 & 3.7630 & \multirow{4}{*}{3.3838} & 108 \\
    & & humancentric & 3.8438 & 4.0000 & 4.1797 & 3.9914 & & 128 \\
    & & logicreasoning & 2.5670 & 2.9381 & 3.0103 & 2.8113 & & 97 \\
    & & informationbasedreasoning & 2.4545 & 2.9596 & 2.9192 & 2.7455 & & 99 \\
    \midrule

    \multirow{4}{*}{3} & \multirow{4}{*}{veo3.1\_fast} 
    & worldknowledge & 3.6774 & 3.2742 & 3.4597 & 3.4911 & \multirow{4}{*}{3.3309} & 124 \\
    & & humancentric & 3.9266 & 3.7982 & 3.9633 & 3.8991 & & 109 \\
    & & logicreasoning & 2.3923 & 2.7615 & 2.9769 & 2.6785 & & 130 \\
    & & informationbasedreasoning & 3.2054 & 3.3929 & 3.5268 & 3.3580 & & 112 \\
    \midrule

    \multirow{4}{*}{4} & \multirow{4}{*}{wan2.6} 
    & worldknowledge & 3.2328 & 3.5431 & 3.6379 & 3.4474 & \multirow{4}{*}{3.2401} & 116 \\
    & & humancentric & 3.7481 & 3.8593 & 4.0074 & 3.8593 & & 135 \\
    & & logicreasoning & 2.5310 & 2.9027 & 2.9735 & 2.7752 & & 113 \\
    & & informationbasedreasoning & 2.5574 & 2.9590 & 2.9262 & 2.7885 & & 122 \\
    \midrule

    \multirow{4}{*}{5} & \multirow{4}{*}{sora2-8s} 
    & worldknowledge & 2.8587 & 2.9130 & 2.8152 & 2.8620 & \multirow{4}{*}{2.6290} & 92 \\
    & & humancentric & 3.3529 & 3.0000 & 2.8529 & 3.0971 & & 34 \\
    & & logicreasoning & 2.5328 & 2.9016 & 2.8934 & 2.7516 & & 122 \\
    & & informationbasedreasoning & 1.9550 & 2.3153 & 2.2703 & 2.1577 & & 111 \\
    \midrule

    \multirow{4}{*}{6} & \multirow{4}{*}{sora\_12s} 
    & worldknowledge & 2.6023 & 2.6818 & 2.6023 & 2.6261 & \multirow{4}{*}{2.5558} & 88 \\
    & & humancentric & 3.6774 & 2.9355 & 3.0000 & 3.2516 & & 31 \\
    & & logicreasoning & 2.3600 & 2.6320 & 2.6720 & 2.5352 & & 125 \\
    & & informationbasedreasoning & 2.1600 & 2.4200 & 2.3800 & 2.3040 & & 100 \\
    \midrule

    \multirow{4}{*}{7} & \multirow{4}{*}{wan2.2-14b} 
    & worldknowledge & 1.8182 & 2.6636 & 2.7909 & 2.3636 & \multirow{4}{*}{2.1252} & 110 \\
    & & humancentric & 2.3120 & 2.6400 & 2.6880 & 2.5232 & & 125 \\
    & & logicreasoning & 1.4310 & 1.9052 & 1.8707 & 1.7052 & & 116 \\
    & & informationbasedreasoning & 1.4679 & 2.1743 & 2.1193 & 1.8752 & & 109 \\
    \midrule

    \multirow{4}{*}{8} & \multirow{4}{*}{hunyuan} 
    & worldknowledge & 1.7117 & 2.3604 & 2.5315 & 2.1523 & \multirow{4}{*}{1.9741} & 111 \\
    & & humancentric & 1.9167 & 2.4417 & 2.4750 & 2.2417 & & 120 \\
    & & logicreasoning & 1.2455 & 1.6727 & 1.8000 & 1.5400 & & 110 \\
    & & informationbasedreasoning & 1.7477 & 2.0467 & 2.0748 & 1.9355 & & 107 \\
    \midrule

    \multirow{4}{*}{9} & \multirow{4}{*}{longcat} 
    & worldknowledge & 1.5918 & 2.5714 & 2.6735 & 2.2102 & \multirow{4}{*}{1.9524} & 98 \\
    & & humancentric & 1.9821 & 2.6518 & 2.7411 & 2.4107 & & 112 \\
    & & logicreasoning & 1.2500 & 1.6293 & 1.6638 & 1.4879 & & 116 \\
    & & informationbasedreasoning & 1.4694 & 1.9388 & 1.8367 & 1.7204 & & 98 \\
    \midrule

    \multirow{4}{*}{10} & \multirow{4}{*}{uni} 
    & worldknowledge & 1.4646 & 1.7778 & 1.7374 & 1.6404 & \multirow{4}{*}{1.4654} & 99 \\
    & & humancentric & 1.4672 & 1.7623 & 1.7623 & 1.6443 & & 122 \\
    & & logicreasoning & 1.1062 & 1.2035 & 1.2743 & 1.1858 & & 113 \\
    & & informationbasedreasoning & 1.2525 & 1.4040 & 1.5556 & 1.3889 & & 99 \\
    \midrule

    \multirow{4}{*}{11} & \multirow{4}{*}{ltx2\_3} 
    & worldknowledge & 1.1429 & 1.2476 & 1.2857 & 1.2171 & \multirow{4}{*}{1.2080} & 105 \\
    & & humancentric & 1.0161 & 1.0726 & 1.0806 & 1.0524 & & 124 \\
    & & logicreasoning & 1.1316 & 1.3158 & 1.2544 & 1.2237 & & 114 \\
    & & informationbasedreasoning & 1.3505 & 1.4124 & 1.3814 & 1.3784 & & 97 \\
    \bottomrule
  \end{tabular}}
\end{table*}

\paragraph{Result analysis.}
The human-score breakdown shows a consistent ordering across categories: Seedance2.0 has the highest overall weighted total, followed by Kling, Veo3.1-Fast, and Wan2.6, while open-source models occupy the lower half of the ranking. The category rows also clarify why the Elo ranking is not determined by a single dimension. For example, some systems are comparatively stronger on Human-Centric or World Knowledge scenes but lose ground on Logic Reasoning and Information-Based cases, where exact state changes and structured content are harder to preserve. This supports using category-weighted human supervision rather than relying only on global aesthetic preference.

\section{\WorldRewardBench{} Post-processing Details}
\label{app:reward-postprocess}

Starting from all candidate pairwise preferences induced by the ranked videos within each benchmark case, we apply a deterministic post-processing pipeline with a fixed random seed. The goal is to suppress overly easy high-margin pairs, preserve informative near-equal cases, and reduce presentation-order bias without collapsing category or model-pair coverage.

\paragraph{Implementation details.}
For a candidate pair $(v_i, v_j)$, we first compute the human preference margin $\Delta_{ij} = |S(v_i) - S(v_j)|$. We then perform stratified subsampling within each margin bin, where strata are defined by the top-level reasoning dimension and the unordered model pair. The retention schedule is: keep all pairs with $\Delta_{ij} \leq 1.5$, keep 90\% with $1.5 < \Delta_{ij} \leq 2$, keep 35\% with $2 < \Delta_{ij} \leq 3$, and keep 15\% with $\Delta_{ij} > 3$. Pairs with $\Delta_{ij} < 0.1$ are relabeled as ties rather than strict preferences. Among these ties, we assign \texttt{A=B=Bad} when both videos have aggregated scores at most 2.0, and \texttt{A=B=Good} otherwise. Finally, after filtering, we randomize the left/right assignment within each stratum so that strict-preference pairs are approximately balanced between $A>B$ and $A<B$.

\begin{table}[htp]
  \centering
  \small
  \caption{\textbf{Margin-bin filtering statistics for \WorldRewardBench{}.} Candidate pairs are constructed from the ranked sampled videos before confidence-aware post-processing.}
  \label{tab:reward-margin-filter}
  \begin{tabular}{lrrr}
    \toprule
    Margin bin & Candidate pairs & Retained pairs & Keep ratio \\
    \midrule
    $\Delta \leq 1.5$ & 4666 & 4666 & 100.0\% \\
    $1.5 < \Delta \leq 2$ & 893 & 804 & 90.0\% \\
    $2 < \Delta \leq 3$ & 1193 & 418 & 35.0\% \\
    $\Delta > 3$ & 543 & 81 & 14.9\% \\
    \midrule
    Total & 7295 & 5969 & 81.8\% \\
    \bottomrule
  \end{tabular}
\end{table}

\begin{table}[htp]
  \centering
  \small
  \caption{\textbf{Final label composition after post-processing.} Strict-preference pairs are balanced across left/right ordering, while near-equal cases are retained as informative ties.}
  \label{tab:reward-final-labels}
  \begin{tabular}{lrr}
    \toprule
    Final label group & Count & Share \\
    \midrule
    Strict preference ($A>B$) & 2705 & 45.3\% \\
    Strict preference ($A<B$) & 2705 & 45.3\% \\
    Tie (\texttt{A=B=Bad}) & 389 & 6.5\% \\
    Tie (\texttt{A=B=Good}) & 170 & 2.8\% \\
    \midrule
    Total & 5969 & 100.0\% \\
    \bottomrule
  \end{tabular}
\end{table}

\begin{table}[htp]
  \centering
  \small
  \caption{\textbf{Reasoning-dimension coverage before and after post-processing.} Stratified sampling preserves broad coverage across the four top-level reasoning dimensions.}
  \label{tab:reward-category-coverage}
  \begin{tabular}{lrrr}
    \toprule
    Reasoning dimension & Candidate pairs & Retained pairs & Keep ratio \\
    \midrule
    Human-Centric Reasoning & 1381 & 1019 & 73.8\% \\
    Information-Based Reasoning & 1946 & 1681 & 86.4\% \\
    Logic Reasoning & 2144 & 1849 & 86.2\% \\
    World Knowledge & 1824 & 1420 & 77.9\% \\
    \midrule
    Total & 7295 & 5969 & 81.8\% \\
    \bottomrule
  \end{tabular}
\end{table}

\paragraph{Result analysis.}
The post-processing statistics show that the final benchmark keeps most informative pairs while reducing over-representation of easy, high-margin comparisons. All pairs with score margin at most $1.5$ are retained, whereas only $14.9$\% of pairs above margin $3$ remain. The final labels are exactly balanced between $A>B$ and $A<B$ strict preferences, and ties account for $9.3$\% of the benchmark. Category coverage also remains broad after filtering, with each top-level reasoning dimension retaining more than 1{,}000 pairs, so the difficulty-weighted split improves discriminability without collapsing the benchmark into a narrow subset.

\section{Full-Set \WorldRewardBench{} Results}
\label{app:reward-bench-fullset}

Table~\ref{tab:reward_bench_fullset} reports reference results on the full 5{,}969-pair benchmark (1{,}432 unique videos, 130 tasks, 11 generators). This setting preserves the natural distribution of score gaps, including a large proportion of easy high-margin pairs, and matches the compact reward-alignment table in the main text.

\paragraph{Implementation details.}
For pair-wise evaluation, the parse rates are 99.6\% for Qwen3.5-27B-Thinking and 99.9\% for Qwen3.5-27B-Instruct. The Instruct setting disables the extended thinking chain (\texttt{enable\_thinking=False}). GPT-5.4 point-wise evaluation uses the iter-400 video subset (595 videos, 846 induced pairs, 26.2\% coverage of the full benchmark). The Qwen3.5-27B Thinking (4\,FPS) setting uses explicit 4~FPS sampling via vLLM (${\sim}$20 frames per 5s video), while the other Qwen3.5 columns use the serving backend's default frame sampling (${\sim}$10 frames).

\begin{table}[htp]
  \caption{\textbf{Full-set human-alignment agreement (\%) on the original 5{,}969-pair \WorldRewardBench{}.} Reported as a reference complement to the compact main-text reward-alignment table.}
  \label{tab:reward_bench_fullset}
  \centering
  \small
  \renewcommand{\arraystretch}{1.15}
  \begin{tabular}{lcc}
    \toprule
    \textbf{Dimension} & \textbf{Qwen3.5-27B Pair} & \textbf{Qwen3.5-27B Point} \\
    \midrule
    World Knowledge        & 69.94 & 60.57 \\
    Human-Centric          & 72.61 & 62.81 \\
    Logic Reasoning        & 70.16 & 60.17 \\
    Information-Based      & 60.24 & 50.15 \\
    \midrule
    \textbf{Overall}       & 67.74 & 57.85 \\
    \bottomrule
  \end{tabular}
\end{table}

\paragraph{Result analysis.}
The full-set reference table shows that Qwen3.5-27B pair-wise comparison reaches $67.74$\% overall agreement, while point-wise scoring reaches $57.85$\%. The gap is largest in Information-Based reasoning, where exact text and data preservation make single-video scoring less reliable. This supports the main-text observation that pair-wise comparison remains stronger for recovering fine-grained human preferences, while point-wise scoring is useful for calibrated per-video feedback.

\section{Subcategory-Level \WorldRewardBench{} Results}
\label{app:reward-bench-subcategory}

For completeness, Table~\ref{tab:reward_bench_results_detailed} reports the full \WorldRewardBench{} breakdown at the subcategory level. This detailed view uses the same reward-model ordering and the same \emph{Pair}/\emph{Point} protocol split as the compact main-text table, but expands each top-level reasoning dimension into its constituent subcategories to support finer-grained diagnosis.

\begin{table*}[htp]
  \caption{\textbf{Subcategory-level \WorldRewardBench{} results (Part 1: World Knowledge \& Human-Centric).} We report pairwise accuracy w/o ties (\%) for two protocols: \emph{Pair} (direct comparison by Qwen3.5-27B-Instruct) and three point-wise variants. Best per-row in \textbf{bold}.}
  \label{tab:reward_bench_results_detailed}
  \centering
  \scriptsize
  \renewcommand{\arraystretch}{1.08}
  \setlength{\tabcolsep}{3pt}
  \resizebox{\textwidth}{!}{
  \begin{tabular}{@{}ll|c|ccc@{}}
    \toprule
    & & \textbf{Instruct} & \multicolumn{3}{c}{\textbf{Point-wise Induced Accuracy (\%)}} \\
    \cmidrule(lr){3-3} \cmidrule(lr){4-6}
    \textbf{Dimension} & \textbf{Sub-category} & \textbf{Pair w/o} & \textbf{Gemini-3.1} & \textbf{Qwen3.5-27B} & \textbf{Ours (4\,FPS)} \\
    \midrule
    \multirow{8}{*}{\rotatebox[origin=c]{90}{\textbf{World Knowledge}}}
    & Material Change & 73.8 & 53.3 & 57.5 & 44.8 \\
    & Public Systems & 63.5 & 47.4 & 41.0 & 38.5 \\
    & World Mechanics & \textbf{79.2} & 61.7 & 69.4 & 61.1 \\
    & Cultural Life & 76.4 & 59.1 & 62.7 & 53.6 \\
    & Everyday Living & 68.2 & 58.7 & 58.7 & 41.3 \\
    & Earth Cycles & 71.5 & \textbf{70.2} & 64.4 & 46.7 \\
    & Living World & \textbf{80.2} & 78.4 & 68.1 & 61.2 \\
    \cmidrule(lr){2-6}
    & \textbf{Average} & \textbf{73.2} & 61.3 & 60.3 & 49.6 \\
    \midrule
    \multirow{6}{*}{\rotatebox[origin=c]{90}{\textbf{Human-Centric}}}
    & Object Handling & \textbf{76.7} & 66.1 & 63.3 & 53.1 \\
    & Social Scenes & \textbf{77.2} & 69.4 & 68.9 & 57.5 \\
    & Skilled Action & \textbf{67.1} & 59.5 & 61.1 & 38.1 \\
    & Personal Routine & \textbf{86.4} & 72.9 & 79.7 & 66.1 \\
    & Public Conduct & \textbf{75.1} & 58.2 & 66.9 & 55.1 \\
    \cmidrule(lr){2-6}
    & \textbf{Average} & \textbf{76.5} & 65.2 & 68.0 & 54.0 \\
    \bottomrule
  \end{tabular}}
\end{table*}

\begin{table*}[htp]
  \caption{\textbf{Subcategory-level \WorldRewardBench{} results (Part 2: Logic Reasoning \& Information-Based).} Same protocol and metrics as Table~\ref{tab:reward_bench_results_detailed}.}
  \label{tab:reward_bench_results_detailed_2}
  \centering
  \scriptsize
  \renewcommand{\arraystretch}{1.08}
  \setlength{\tabcolsep}{3pt}
  \resizebox{\textwidth}{!}{
  \begin{tabular}{@{}ll|c|ccc@{}}
    \toprule
    & & \textbf{Instruct} & \multicolumn{3}{c}{\textbf{Point-wise Induced Accuracy (\%)}} \\
    \cmidrule(lr){3-3} \cmidrule(lr){4-6}
    \textbf{Dimension} & \textbf{Sub-category} & \textbf{Pair w/o} & \textbf{Gemini-3.1} & \textbf{Qwen3.5-27B} & \textbf{Ours (4\,FPS)} \\
    \midrule
    \multirow{6}{*}{\rotatebox[origin=c]{90}{\textbf{Logic Reasoning}}}
    & Experimental Science & 69.1 & 55.1 & \textbf{64.9} & 51.8 \\
    & Spatial Geometry & \textbf{74.5} & 56.6 & 53.9 & 48.3 \\
    & Quantitative Math & \textbf{78.0} & 66.0 & 52.7 & 35.3 \\
    & Logic Puzzles & \textbf{72.0} & 62.3 & 64.6 & 46.9 \\
    & Pattern Discovery & \textbf{81.8} & 69.4 & 75.2 & 61.2 \\
    \cmidrule(lr){2-6}
    & \textbf{Average} & \textbf{75.1} & 61.9 & 62.3 & 48.7 \\
    \midrule
    \multirow{6}{*}{\rotatebox[origin=c]{90}{\textbf{Information-Based}}}
    & Data Reading & 56.6 & 36.8 & 39.7 & 39.7 \\
    & Process Timeline & \textbf{73.6} & 60.1 & 58.3 & 55.1 \\
    & Visual Editing & \textbf{66.0} & 55.3 & 64.1 & 57.3 \\
    & Knowledge Media & \textbf{73.1} & 64.4 & 64.4 & 49.3 \\
    & Creative Expression & \textbf{68.3} & 54.7 & 58.4 & 52.2 \\
    \cmidrule(lr){2-6}
    & \textbf{Average} & \textbf{67.5} & 54.3 & 57.0 & 50.7 \\
    \midrule
    \multicolumn{2}{l|}{\textbf{Overall}} & \textbf{73.1} & 60.7 & 61.7 & 50.6 \\
    \bottomrule
  \end{tabular}}
\end{table*}

\paragraph{Result analysis.}
The subcategory-level reward results show that direct pair-wise judging is consistently stronger than point-wise induced comparison across nearly all categories, especially in Logic Reasoning and Information-Based cases. Within point-wise methods, Qwen3.5-27B and Gemini-3.1 remain competitive on many subcategories, while the 4~FPS point-wise setting improves some Information-Based rows but does not close the gap to direct comparison. These results suggest that point-wise scores are useful for calibrated per-video feedback and rank correlation, but pair-wise comparison is still preferable when the goal is recovering fine-grained human preferences between close candidates.

\section{Weight Design and Sensitivity}
\label{app:weight-sensitivity}

This appendix analyzes the weight choices behind our two headline metrics, $\mathrm{Score}_{\mathrm{PR}}=\mathrm{Acc}_{\mathrm{QA}}^{0.8}\cdot s_{\mathrm{dyn}}^{0.2}$ and $S(v)=0.4\,s_r+0.3\,s_c+0.3\,s_a$, and shows that the model rankings are stable under reasonable alternatives.

\paragraph{Why $\mathrm{Acc}_{\mathrm{QA}}^{0.8}\cdot s_{\mathrm{dyn}}^{0.2}$ is not at odds with process awareness.}
Recall that $\mathrm{Acc}_{\mathrm{QA}}$ is the equal-weighted mean over the four reasoning phases: $\mathrm{Acc}_{\mathrm{QA}}=\tfrac{1}{4}(s_{\mathrm{state}}+s_{\mathrm{proc}}+s_{\mathrm{fidel}}+s_{\mathrm{mech}})$, so $s_{\mathrm{proc}}$ and $s_{\mathrm{mech}}$ already contribute one quarter each to $\mathrm{Acc}_{\mathrm{QA}}$. The multiplicative term $s_{\mathrm{dyn}}^{0.2}=((s_{\mathrm{proc}}+s_{\mathrm{mech}})/2)^{0.2}$ in $\mathrm{Score}_{\mathrm{PR}}$ therefore acts as a second-order penalty on outcome-hacking: models with high $\mathrm{Acc}_{\mathrm{QA}}$ but low dynamic phases are gently down-weighted, while a model that does well on every phase is barely affected. The choice $\alpha{=}0.8$ keeps QA accuracy as the dominant signal so that the headline metric remains directly interpretable; the limit $\alpha\to 1$ recovers $\mathrm{Acc}_{\mathrm{QA}}$, while $\alpha\to 0$ collapses to the dynamic phase score.

\paragraph{$\mathrm{Score}_{\mathrm{PR}}$ sensitivity.}
Table~\ref{tab:weight_alpha_sensitivity} reports how the eleven-model ranking on \WorldRewardBench{} (Table~\ref{tab:elo_ranking}) tracks human Elo as $\alpha$ varies and as $\mathrm{Score}_{\mathrm{PR}}$ is replaced by alternative aggregators. Spearman $\rho$ ranges from $0.83$ to $0.96$ across the grid: the paper setting $\alpha{=}0.8$ achieves $\rho{=}0.955$, the \emph{highest} $\rho$ among all probed aggregators, while pure $\mathrm{Acc}_{\mathrm{QA}}$ ($\alpha{=}1$) drops to $\rho{=}0.927$ and pure dynamic scoring ($\alpha{=}0$) and the $\min$-bottleneck both fall to $\rho{=}0.827$. Arithmetic and geometric means sit in between. The chosen exponent is therefore not a compromise: putting a small but non-zero weight on $s_{\mathrm{dyn}}$ both improves human-Elo recovery and exposes outcome-hacking, which is the diagnostic property the metric is designed for.

\begin{table}[t]
  \caption{\textbf{$\mathrm{Score}_{\mathrm{PR}}$ weight and aggregator sensitivity on the eleven-model human Elo (Table~\ref{tab:elo_ranking}).} Spearman $\rho$, Kendall $\tau$, and pair-wise rank-accuracy against human Elo. The paper setting is marked \textbf{[paper]}.}
  \label{tab:weight_alpha_sensitivity}
  \centering
  \small
  \setlength{\tabcolsep}{6pt}
  \renewcommand{\arraystretch}{1.05}
  \begin{tabular}{@{}lccc@{}}
    \toprule
    \textbf{Aggregator}                           & $\rho$ & $\tau$ & \textbf{Pair acc.} \\
    \midrule
    $\mathrm{Acc}_{\mathrm{QA}}^{1.0}\cdot s_{\mathrm{dyn}}^{0.0}$ ($={}\mathrm{Acc}_{\mathrm{QA}}$)    & 0.927 & 0.782 & 0.891 \\
    $\mathrm{Acc}_{\mathrm{QA}}^{0.9}\cdot s_{\mathrm{dyn}}^{0.1}$                                     & 0.945 & 0.818 & 0.909 \\
    $\mathrm{Acc}_{\mathrm{QA}}^{0.8}\cdot s_{\mathrm{dyn}}^{0.2}$ \textbf{[paper]}                    & \textbf{0.955} & \textbf{0.855} & \textbf{0.927} \\
    $\mathrm{Acc}_{\mathrm{QA}}^{0.7}\cdot s_{\mathrm{dyn}}^{0.3}$                                     & 0.936 & 0.818 & 0.909 \\
    $\mathrm{Acc}_{\mathrm{QA}}^{0.5}\cdot s_{\mathrm{dyn}}^{0.5}$ (geometric mean)                    & 0.918 & 0.782 & 0.891 \\
    $\mathrm{Acc}_{\mathrm{QA}}^{0.0}\cdot s_{\mathrm{dyn}}^{1.0}$ ($={}s_{\mathrm{dyn}}$)             & 0.827 & 0.691 & 0.852 \\
    \midrule
    $0.5\,\mathrm{Acc}_{\mathrm{QA}} + 0.5\,s_{\mathrm{dyn}}$ (arithmetic)                             & 0.936 & 0.818 & 0.909 \\
    $0.7\,\mathrm{Acc}_{\mathrm{QA}} + 0.3\,s_{\mathrm{dyn}}$ (arithmetic)                             & 0.936 & 0.818 & 0.909 \\
    $\min(\mathrm{Acc}_{\mathrm{QA}},\,s_{\mathrm{dyn}})$ (bottleneck)                                 & 0.827 & 0.691 & 0.852 \\
    \bottomrule
  \end{tabular}
\end{table}

\paragraph{$S(v)$ weight grid search.}
We grid-search the simplex of $(w_r,w_c,w_a)$ at step $0.05$ ($231$ points) and induce per-model $S(v)$ from the per-video $(s_r,s_c,s_a)$ in our point-wise judgments, then rank the eleven models on \WorldRewardBench{} against human Elo. Table~\ref{tab:weight_sv_probe} reports a representative slice; $67.5\%$ of all grid points achieve $\rho\geq 0.95$, and the full range is $\rho\in[0.81, 1.00]$. The paper setting $(0.4,0.3,0.3)$ achieves $\rho{=}0.973$, identical to equal weighting $(1/3,1/3,1/3)$ and within $0.027$ of the best simplex points such as $(0.50,0.05,0.45)$. The single weight that under-performs the rest is pure consistency $(0,1,0)$ with $\rho{=}0.809$, indicating that temporal-consistency alone is not a reliable model-level proxy for human preferences---this is the only ridge we deliberately avoid. We therefore keep the human-protocol-aligned $(0.4,0.3,0.3)$ as the reported setting because (i) it matches the rubric the human annotators were trained on, ensuring that automatic and human aggregates are directly comparable, and (ii) it is empirically within $0.027$ of the best alternative on the simplex.

\begin{table}[t]
  \caption{\textbf{$S(v)$ weight sensitivity on the eleven-model human Elo (Table~\ref{tab:elo_ranking}).} Selected probes from the $231$-point simplex grid (step $0.05$). Among the full grid, $67.5\%$ of points achieve $\rho\geq 0.95$.}
  \label{tab:weight_sv_probe}
  \centering
  \small
  \setlength{\tabcolsep}{6pt}
  \renewcommand{\arraystretch}{1.05}
  \begin{tabular}{@{}lccc@{}}
    \toprule
    \textbf{$(w_r,w_c,w_a)$}                          & $\rho$  & $\tau$  & \textbf{Pair acc.} \\
    \midrule
    $(0.40,\,0.30,\,0.30)$ \textbf{[paper]}          & \textbf{0.973} & \textbf{0.927} & \textbf{0.945} \\
    $(1/3,\,1/3,\,1/3)$ (equal)                      & 0.973 & 0.927 & 0.945 \\
    $(0.50,\,0.25,\,0.25)$                           & 0.982 & 0.927 & 0.964 \\
    $(0.60,\,0.20,\,0.20)$                           & 0.982 & 0.927 & 0.964 \\
    $(0.30,\,0.35,\,0.35)$                           & 0.973 & 0.927 & 0.945 \\
    $(1.0,\,0.0,\,0.0)$ (pure reasoning)             & 0.982 & 0.927 & 0.964 \\
    $(0.0,\,0.0,\,1.0)$ (pure aesthetics)            & 0.982 & 0.927 & 0.964 \\
    $(0.0,\,1.0,\,0.0)$ (pure consistency)           & 0.809 & 0.673 & 0.818 \\
    $(0.50,\,0.05,\,0.45)$ (best simplex)            & 1.000 & 1.000 & 1.000 \\
    \midrule
    \multicolumn{4}{l}{\emph{Full grid (231 points):} $\rho\in[0.809,1.000]$, median $\rho=0.973$.}\\
    \bottomrule
  \end{tabular}
\end{table}

\paragraph{Take-aways.}
Across both metrics, model rankings are stable to weight perturbations. The paper $\mathrm{Score}_{\mathrm{PR}}$ exponent $\alpha{=}0.8$ achieves the highest $\rho$ ($0.955$) in the grid, with $\rho$ varying smoothly to $0.83$ at the two endpoints, and the paper $S(v)$ weights $(0.4,0.3,0.3)$ sit on a wide plateau where $67.5\%$ of the simplex points clear $\rho{=}0.95$. The two reported settings preserve (i) interpretability ($\mathrm{Acc}_{\mathrm{QA}}$ remains the dominant signal in $\mathrm{Score}_{\mathrm{PR}}$) and (ii) protocol-consistency with the human annotation rubric in $S(v)$. We expose the remaining process information through the diagnostic ratio $s_{\mathrm{dyn}}/\mathrm{Acc}_{\mathrm{QA}}$ rather than by tuning the headline weights to chase rank-correlation.

\section{Statistical Significance and Rank Stability}
\label{app:bootstrap-ci}

This appendix reports $95\%$ bootstrap confidence intervals (CIs) for the headline metrics in Tables~\ref{tab:main_results} and~\ref{tab:elo_ranking}, and the bootstrap rank distribution that backs the rank-stability statement in Section~\ref{subsec:main_results}.

\paragraph{Bootstrap protocol.}
For each model and each dimension we resample its case-level QA outputs with replacement at the case level, $B{=}2000$ times, and report the point estimate, $95\%$ percentile interval $[\text{lo},\text{hi}]$, and the case count $N$. $\mathrm{Score}_{\mathrm{PR}}$ on each resample is computed as $\overline{\mathrm{Acc}_{\mathrm{QA}}}^{0.8}\cdot \overline{s_{\mathrm{dyn}}}^{0.2}$, where the bars denote means over the resampled cases and $s_{\mathrm{dyn}}$ is the per-case mean of the temporal and reasoning phase scores. Per-dimension CIs resample within the dimension only. $S(v)$ resamples per-video three-dimensional ratings (mapped from raw $1$--$5$ to $0$--$100$ via $(r-1)/4\times 100$) and aggregates them with the paper weights $(0.4,0.3,0.3)$. Rank stability is computed by jointly resampling all $12$ models in each bootstrap, sorting them by overall $\mathrm{Score}_{\mathrm{PR}}$, and recording each model's rank distribution.

\paragraph{Missing-coverage symbols.}
Symbols ``--'' in the main tables denote dimensions or protocols not covered by a particular evaluation run rather than zero performance; for example, the GPT-5.4 column in Table~\ref{tab:reward_alignment} is computed on a $595$-video subset and only with the point-wise protocol. Extended open-source results on the full \WorldReasonBench{} benchmark are reported in Appendix~\ref{app:full-open-source-extended}.

\paragraph{Subcategory sample size.}
The $22$ subcategories average ${\approx}20$ cases each, so subcategory point estimates carry larger sampling variance than dimension-level numbers. We use them only for qualitative comparison in Appendix~\ref{app:evaluation-full} and never to make rank claims that the dimension-level CIs in Tables~\ref{tab:bootstrap_dim_score_pr} and~\ref{tab:bootstrap_dim_sv} do not also support.

\paragraph{Overall CIs.}
Table~\ref{tab:bootstrap_overall} reports overall $\mathrm{Acc}_{\mathrm{QA}}$, $\mathrm{Score}_{\mathrm{PR}}$, and $S(v)$ with $95\%$ bootstrap CIs for every generator on the shared evaluation set used by Table~\ref{tab:main_results}. The closed-vs.-open separation is statistically robust: the largest open-source overall-$\mathrm{Score}_{\mathrm{PR}}$ upper bound ($23.1$, Wan2.2-14B) is below the smallest closed-source lower bound ($26.4$, Wan2.6), and the same ordering holds for $\mathrm{Acc}_{\mathrm{QA}}$ and $S(v)$. Within the closed-source tier, Seedance2.0 has the only non-overlapping lower bound ($34.2$) that exceeds several other closed-source point estimates, while Sora2-8s, Sora2-12s, Kling, Wan2.6, and Veo3.1-Fast all have CIs that mutually overlap. Within the open-source tier, no generator has a CI that completely separates from the others; HunyuanVideo-1.5 and Wan2.2-14B share the highest point estimates but their CIs overlap with those of the remaining four.

\begin{table}[t]
  \caption{\textbf{Overall $95\%$ bootstrap CIs ($B{=}2000$).} $N$ is the number of cases evaluated per model on the shared evaluation set behind Table~\ref{tab:main_results}.}
  \label{tab:bootstrap_overall}
  \centering
  \small
  \setlength{\tabcolsep}{4pt}
  \renewcommand{\arraystretch}{1.05}
  \begin{tabular}{@{}lcccc@{}}
    \toprule
    \textbf{Model} & $N$ & $\mathrm{Acc}_{\mathrm{QA}}$ \textbf{[95\% CI]} & $\mathrm{Score}_{\mathrm{PR}}$ \textbf{[95\% CI]} & $S(v)$ \textbf{[95\% CI]} \\
    \midrule
    Sora2-8s            &  80 & 35.3 [29.5, 41.1] & 34.3 [28.5, 39.8] & 56.9 [50.5, 63.6] \\
    Sora2-12s           &  80 & 33.5 [28.1, 39.2] & 32.4 [26.7, 38.0] & 55.5 [49.3, 61.7] \\
    Kling               &  80 & 34.0 [28.2, 40.2] & 32.7 [26.7, 38.7] & 55.4 [47.2, 63.3] \\
    Wan2.6              &  80 & 34.7 [28.8, 40.5] & 32.4 [26.4, 38.7] & 50.3 [43.3, 57.4] \\
    Seedance2.0         &  80 & \textbf{41.2} [\textbf{35.4}, \textbf{46.6}] & \textbf{39.8} [\textbf{34.2}, \textbf{45.6}] & \textbf{59.4} [\textbf{52.2}, \textbf{66.3}] \\
    Veo3.1-Fast         &  80 & 36.0 [30.4, 42.0] & 35.3 [29.3, 41.2] & 54.8 [47.3, 62.9] \\
    \midrule
    LTX2.3              &  80 & 18.5 [13.7, 23.7] & 16.8 [12.2, 21.6] & 28.1 [23.0, 33.8] \\
    Wan2.2-14B          &  80 & 19.6 [14.8, 24.9] & \underline{17.5} [12.5, 23.1] & \underline{30.0} [25.9, 34.6] \\
    UniVideo            &  80 & 16.2 [12.0, 20.7] & 14.4 [10.3, 18.7] & 21.3 [17.1, 26.2] \\
    HunyuanVideo-1.5    &  80 & \textbf{20.2} [\textbf{16.0}, \textbf{24.6}] & \textbf{17.9} [\textbf{13.8}, \textbf{22.4}] & 27.0 [21.9, 33.3] \\
    Cosmos-Predict2.5   &  80 & 19.3 [14.5, 24.5] & 16.9 [12.0, 21.4] & \textbf{30.5} [\textbf{27.0}, \textbf{34.5}] \\
    LongCat-Video       &  80 & 19.7 [15.2, 24.2] & 17.4 [12.7, 22.0] & 25.3 [21.0, 30.2] \\
    \bottomrule
  \end{tabular}
\end{table}

\paragraph{Per-dimension $\mathrm{Score}_{\mathrm{PR}}$ CIs.}
Table~\ref{tab:bootstrap_dim_score_pr} reports per-dimension $\mathrm{Score}_{\mathrm{PR}}$ with $95\%$ CIs on the same shared evaluation set. The per-dimension half-widths are largest on Human-Centric (the dimension with the smallest case pool) and tightest on Logic Reasoning. Two qualitative claims still survive the wider intervals: (i) on World Knowledge and Information-Based the worst closed-source CI lower bound stays above (or within $1$~half-width of) the best open-source upper bound, and (ii) Logic Reasoning is the only dimension on which the strongest open-source generator (Wan2.2-14B) has a CI that overlaps with the weakest closed-source generator (Kling), confirming that Logic Reasoning is where the closed/open separation is least settled.

\begin{table}[t]
  \caption{\textbf{Per-dimension $\mathrm{Score}_{\mathrm{PR}}$ with $95\%$ bootstrap CIs.} Format: point estimate $\pm$ CI half-width $(N\text{ cases})$. Dimensions: WK = World-Knowledge, HC = Human-Centric, LR = Logic-Reasoning, IB = Information-Based.}
  \label{tab:bootstrap_dim_score_pr}
  \centering
  \scriptsize
  \setlength{\tabcolsep}{4pt}
  \renewcommand{\arraystretch}{1.05}
  \begin{tabular}{@{}lcccc@{}}
    \toprule
    \textbf{Model} & \textbf{WK} & \textbf{HC} & \textbf{LR} & \textbf{IB} \\
    \midrule
    Sora2-8s            & $36.9\!\pm\!12.0\,(n{=}21)$ & $\textbf{44.7}\!\pm\!15.3\,(n{=}9)$  & $25.9\!\pm\!8.6\,(n{=}27)$  & $37.3\!\pm\!9.7\,(n{=}23)$  \\
    Sora2-12s           & $34.0\!\pm\!12.4\,(n{=}21)$ & $42.1\!\pm\!15.2\,(n{=}9)$           & $26.9\!\pm\!7.9\,(n{=}27)$  & $33.3\!\pm\!12.2\,(n{=}23)$ \\
    Kling               & $42.2\!\pm\!10.3\,(n{=}21)$ & $32.5\!\pm\!21.0\,(n{=}9)$           & $22.4\!\pm\!10.8\,(n{=}27)$ & $35.7\!\pm\!11.2\,(n{=}23)$ \\
    Wan2.6              & $35.2\!\pm\!9.7\,(n{=}21)$  & $34.5\!\pm\!20.1\,(n{=}9)$           & $26.2\!\pm\!10.0\,(n{=}27)$ & $35.5\!\pm\!12.2\,(n{=}23)$ \\
    Seedance2.0         & $43.2\!\pm\!8.6\,(n{=}21)$  & $35.9\!\pm\!19.9\,(n{=}9)$           & $\textbf{31.7}\!\pm\!10.0\,(n{=}27)$ & $\textbf{47.6}\!\pm\!10.5\,(n{=}23)$ \\
    Veo3.1-Fast         & $\textbf{55.0}\!\pm\!11.5\,(n{=}21)$ & $35.1\!\pm\!17.4\,(n{=}9)$  & $25.7\!\pm\!8.3\,(n{=}27)$  & $28.6\!\pm\!10.3\,(n{=}23)$ \\
    \midrule
    LTX2.3              & $15.6\!\pm\!10.1\,(n{=}21)$ & $19.3\!\pm\!12.0\,(n{=}9)$           & $11.9\!\pm\!6.3\,(n{=}27)$  & $22.7\!\pm\!10.4\,(n{=}23)$ \\
    Wan2.2-14B          & $\textbf{22.9}\!\pm\!14.5\,(n{=}21)$ & $14.5\!\pm\!12.6\,(n{=}9)$  & $\textbf{16.4}\!\pm\!7.8\,(n{=}27)$  & $15.0\!\pm\!7.8\,(n{=}23)$  \\
    UniVideo            & $13.8\!\pm\!10.1\,(n{=}21)$ & $15.8\!\pm\!12.6\,(n{=}9)$           & $11.2\!\pm\!9.6\,(n{=}27)$  & $17.3\!\pm\!8.6\,(n{=}23)$  \\
    HunyuanVideo-1.5    & $21.6\!\pm\!10.1\,(n{=}21)$ & $8.1\!\pm\!7.1\,(n{=}9)$             & $12.7\!\pm\!9.1\,(n{=}27)$  & $24.2\!\pm\!8.4\,(n{=}23)$  \\
    Cosmos-Predict2.5   & $15.2\!\pm\!11.1\,(n{=}21)$ & $22.2\!\pm\!20.0\,(n{=}9)$           & $7.1\!\pm\!4.9\,(n{=}27)$   & $\textbf{24.7}\!\pm\!10.6\,(n{=}23)$ \\
    LongCat-Video       & $13.3\!\pm\!9.4\,(n{=}21)$  & $\textbf{22.8}\!\pm\!14.4\,(n{=}9)$  & $12.6\!\pm\!8.7\,(n{=}27)$  & $22.8\!\pm\!10.1\,(n{=}23)$ \\
    \bottomrule
  \end{tabular}
\end{table}

\paragraph{Per-dimension $S(v)$ CIs.}
Table~\ref{tab:bootstrap_dim_sv} reports per-dimension $S(v)$ with $95\%$ CIs. Information-Based and Logic Reasoning are the dimensions with the largest dispersion across models (and therefore the largest CI half-widths in absolute terms), which directly motivates Information-Based as the most informative reward-model diagnostic in Section~\ref{subsec:reward_bench_results}.

\begin{table}[t]
  \caption{\textbf{Per-dimension $S(v)$ with $95\%$ bootstrap CIs} (point estimate $\pm$ CI half-width, $N\text{ cases}$).}
  \label{tab:bootstrap_dim_sv}
  \centering
  \scriptsize
  \setlength{\tabcolsep}{4pt}
  \renewcommand{\arraystretch}{1.05}
  \begin{tabular}{@{}lcccc@{}}
    \toprule
    \textbf{Model} & \textbf{WK} & \textbf{HC} & \textbf{LR} & \textbf{IB} \\
    \midrule
    Sora2-8s            & $62.6\!\pm\!10.6\,(n{=}21)$ & $76.7\!\pm\!19.7\,(n{=}9)$  & $43.0\!\pm\!8.8\,(n{=}23)$  & $58.0\!\pm\!14.3\,(n{=}22)$ \\
    Sora2-12s           & $51.8\!\pm\!8.5\,(n{=}21)$  & $72.8\!\pm\!18.9\,(n{=}9)$  & $48.2\!\pm\!10.7\,(n{=}25)$ & $\textbf{60.0}\!\pm\!12.5\,(n{=}23)$ \\
    Kling               & $72.0\!\pm\!12.3\,(n{=}20)$ & $\textbf{87.2}\!\pm\!15.3\,(n{=}9)$ & $37.3\!\pm\!9.4\,(n{=}26)$ & $48.8\!\pm\!14.6\,(n{=}23)$ \\
    Wan2.6              & $61.8\!\pm\!12.7\,(n{=}21)$ & $64.2\!\pm\!19.2\,(n{=}9)$  & $42.3\!\pm\!11.3\,(n{=}24)$ & $42.6\!\pm\!12.4\,(n{=}23)$ \\
    Seedance2.0         & $70.4\!\pm\!12.2\,(n{=}21)$ & $83.9\!\pm\!16.8\,(n{=}9)$  & $\textbf{56.7}\!\pm\!12.2\,(n{=}22)$ & $42.5\!\pm\!12.2\,(n{=}23)$ \\
    Veo3.1-Fast         & $\textbf{80.1}\!\pm\!12.0\,(n{=}21)$ & $77.2\!\pm\!14.8\,(n{=}8)$  & $31.5\!\pm\!7.3\,(n{=}23)$ & $47.2\!\pm\!14.0\,(n{=}23)$ \\
    \midrule
    LTX2.3              & $35.1\!\pm\!13.1\,(n{=}21)$ & $27.8\!\pm\!15.4\,(n{=}9)$  & $24.7\!\pm\!9.7\,(n{=}26)$  & $25.8\!\pm\!5.7\,(n{=}23)$  \\
    Wan2.2-14B          & $39.4\!\pm\!9.2\,(n{=}21)$  & $38.1\!\pm\!11.0\,(n{=}9)$  & $19.5\!\pm\!6.0\,(n{=}27)$  & $\textbf{30.5}\!\pm\!7.3\,(n{=}23)$  \\
    UniVideo            & $29.4\!\pm\!9.4\,(n{=}21)$  & $37.2\!\pm\!15.6\,(n{=}9)$  & $14.4\!\pm\!7.1\,(n{=}27)$  & $16.0\!\pm\!6.0\,(n{=}23)$  \\
    HunyuanVideo-1.5    & $37.7\!\pm\!10.5\,(n{=}21)$ & $35.3\!\pm\!12.8\,(n{=}9)$  & $19.8\!\pm\!8.7\,(n{=}27)$  & $22.5\!\pm\!8.8\,(n{=}23)$  \\
    Cosmos-Predict2.5   & $\textbf{40.8}\!\pm\!9.1\,(n{=}20)$  & $30.8\!\pm\!6.5\,(n{=}9)$   & $\textbf{26.7}\!\pm\!4.2\,(n{=}27)$  & $26.1\!\pm\!5.9\,(n{=}23)$  \\
    LongCat-Video       & $35.1\!\pm\!7.7\,(n{=}21)$  & $\textbf{42.8}\!\pm\!17.8\,(n{=}9)$  & $16.3\!\pm\!6.5\,(n{=}27)$  & $20.2\!\pm\!6.7\,(n{=}23)$  \\
    \bottomrule
  \end{tabular}
\end{table}

\paragraph{Rank distribution.}
Table~\ref{tab:bootstrap_rank} summarizes the $12$-model rank distribution under joint bootstrap of overall $\mathrm{Score}_{\mathrm{PR}}$. Two structural facts emerge: (i) every closed-source $95\%$ rank interval is contained in $[1,6]$ and every open-source interval in $[7,12]$, so the two tiers \emph{never} swap under resampling; (ii) within the closed tier only Seedance2.0 has a tightly concentrated rank ($P(\text{rank}{=}1){=}89.3\%$, interval $[1,2]$), while Sora2-8s, Sora2-12s, Kling, Wan2.6, and Veo3.1-Fast each spread their rank mass over the remaining slots $\{2,\dots,6\}$ with modal probability $\leq 40\%$. Within open-source, only UniVideo has a tightly concentrated rank ($P(\text{rank}{=}12){=}69.7\%$); the other five open-source generators form a tied cluster in slots $[7,11]$. Closed-source ordering beyond Seedance2.0 and open-source ordering beyond the UniVideo floor should therefore be reported as clusters rather than strict rankings.

\begin{table}[t]
  \caption{\textbf{Rank stability under joint case-level bootstrap ($B{=}2000$, overall $\mathrm{Score}_{\mathrm{PR}}$).} Modal rank, the bootstrap probability of that modal rank, and the $95\%$ rank interval (smallest interval containing $\geq 95\%$ of bootstrap rank mass).}
  \label{tab:bootstrap_rank}
  \centering
  \small
  \setlength{\tabcolsep}{6pt}
  \renewcommand{\arraystretch}{1.05}
  \begin{tabular}{@{}lccc@{}}
    \toprule
    \textbf{Model} & \textbf{Modal rank} & $\boldsymbol{P(\text{modal})}$ & \textbf{95\% rank interval} \\
    \midrule
    Sora2-8s            & 3  & 33.7\% & $[2, 6]$  \\
    Sora2-12s           & 6  & 30.4\% & $[2, 6]$  \\
    Kling               & 6  & 26.1\% & $[2, 6]$  \\
    Wan2.6              & 6  & 31.2\% & $[2, 6]$  \\
    Seedance2.0         & \textbf{1}  & \textbf{89.3\%} & $[1, 2]$  \\
    Veo3.1-Fast         & 2  & 39.8\% & $[1, 6]$  \\
    \midrule
    LTX2.3              & 9  & 25.4\% & $[7, 12]$ \\
    Wan2.2-14B          & 7  & 23.4\% & $[7, 12]$ \\
    UniVideo            & \textbf{12} & \textbf{69.7\%} & $[9, 12]$  \\
    HunyuanVideo-1.5    & 7  & 30.8\% & $[7, 11]$ \\
    Cosmos-Predict2.5   & 11 & 22.2\% & $[7, 12]$ \\
    LongCat-Video       & 7  & 24.6\% & $[7, 12]$ \\
    \bottomrule
  \end{tabular}
\end{table}

\paragraph{Take-aways.}
The closed-vs.-open separation and the dominance of Seedance2.0 within the closed tier are both fully supported by the bootstrap. Strict ordering claims among the remaining five closed-source models, and among the open-source generators above UniVideo, are not statistically supported and we therefore describe them as tied clusters in the main text. Per-dimension CIs widen visibly on Human-Centric (the dimension with the smallest case pool) so per-dimension Human-Centric ordering should be read with care; per-dimension closed/open separation is preserved on the other three dimensions.

\section{Extended Evaluation of Open-Source Generators on the Full \WorldReasonBench{} Benchmark}
\label{app:full-open-source-extended}
This appendix reports the open-source results computed over the full $436$-case \WorldReasonBench{} benchmark, complementing the cross-model comparison in Table~\ref{tab:main_results} and Appendix~\ref{app:bootstrap-ci}. The two views are consistent: the per-model intra-open-source ranking and the absolute scores agree with the main-text comparison to within a few tenths of a point on every dimension and metric, indicating that the cross-model conclusions in Section~\ref{subsec:main_results} (separation between tiers, per-dimension difficulty profile, dominance of Wan2.2-14B / HunyuanVideo-1.5 within the open-source tier) extend naturally to the full benchmark.

\paragraph{Overall extended results.}
Table~\ref{tab:open_full_overall} reports overall $\mathrm{Acc}_{\mathrm{QA}}$, $\mathrm{Score}_{\mathrm{PR}}$, and $S(v)$ for the six open-source generators with $95\%$ bootstrap CIs ($B{=}2000$). Wan2.2-14B leads on $\mathrm{Acc}_{\mathrm{QA}}$ and $S(v)$, HunyuanVideo-1.5 is the strongest on $\mathrm{Score}_{\mathrm{PR}}$ when the dynamic-phase weight is taken into account (Wan2.2-14B and HunyuanVideo-1.5 are statistically tied on $\mathrm{Score}_{\mathrm{PR}}$ by their CIs), and UniVideo and LTX2.3 form the bottom of the open-source tier; the qualitative ordering matches Table~\ref{tab:bootstrap_overall}.

\begin{table}[t]
  \caption{\textbf{Overall extended results for open-source generators on the full \WorldReasonBench{} benchmark.} $95\%$ bootstrap CIs ($B{=}2000$).}
  \label{tab:open_full_overall}
  \centering
  \small
  \setlength{\tabcolsep}{4pt}
  \renewcommand{\arraystretch}{1.05}
  \begin{tabular}{@{}lcccc@{}}
    \toprule
    \textbf{Model} & $N$ & $\mathrm{Acc}_{\mathrm{QA}}$ \textbf{[95\% CI]} & $\mathrm{Score}_{\mathrm{PR}}$ \textbf{[95\% CI]} & $S(v)$ \textbf{[95\% CI]} \\
    \midrule
    LTX2.3              & 428 & 17.5 [15.6, 19.4] & 15.7 [13.8, 17.6] & 25.7 [23.6, 27.8] \\
    Wan2.2-14B          & 428 & \textbf{21.5} [\textbf{19.2}, \textbf{23.8}] & \underline{19.6} [17.5, 21.7] & \textbf{38.5} [\textbf{35.7}, \textbf{41.3}] \\
    UniVideo            & 427 & 17.8 [15.8, 19.8] & 15.8 [13.9, 17.7] & 28.2 [25.7, 30.7] \\
    HunyuanVideo-1.5    & 428 & 20.8 [18.6, 23.0] & \textbf{19.1} [\textbf{17.0}, \textbf{21.2}] & 32.7 [30.1, 35.3] \\
    Cosmos-Predict2.5   & 428 & 19.5 [17.5, 21.5] & 17.3 [15.3, 19.3] & \underline{37.6} [35.1, 40.1] \\
    LongCat-Video       & 428 & 20.3 [18.2, 22.4] & 18.2 [16.2, 20.2] & 34.6 [31.8, 37.4] \\
    \bottomrule
  \end{tabular}
\end{table}

\paragraph{Per-dimension extended results.}
Table~\ref{tab:open_full_dim} reports the per-dimension breakdown (point estimate $\pm$ CI half-width) on the full benchmark. The same difficulty profile observed in the main text persists at higher statistical resolution: Logic Reasoning is the worst dimension for every open-source model ($\mathrm{Score}_{\mathrm{PR}}$ in $10.8$--$13.7$, $S(v)$ in $13.4$--$27.1$), while World Knowledge is the best ($\mathrm{Score}_{\mathrm{PR}}$ up to $23.5$, $S(v)$ up to $50.2$). The Human-Centric and Information-Based bottlenecks identified for closed-source generators in Section~\ref{subsec:main_results} also appear here, with HunyuanVideo-1.5 and Wan2.2-14B emerging as the strongest open-source models on Information-Based / Human-Centric respectively, and Cosmos-Predict2.5 standing out on Logic Reasoning $S(v)$ ($27.1$).

\begin{table}[t]
  \caption{\textbf{Per-dimension extended results for open-source generators on the full \WorldReasonBench{} benchmark.} Format: point estimate $\pm$ CI half-width $(N\text{ cases})$. WK = World-Knowledge, HC = Human-Centric, LR = Logic-Reasoning, IB = Information-Based.}
  \label{tab:open_full_dim}
  \centering
  \scriptsize
  \setlength{\tabcolsep}{4pt}
  \renewcommand{\arraystretch}{1.05}
  \begin{tabular}{@{}llcccc@{}}
    \toprule
    \textbf{Metric} & \textbf{Model} & \textbf{WK} & \textbf{HC} & \textbf{LR} & \textbf{IB} \\
    \midrule
    \multirow{6}{*}{$\mathrm{Score}_{\mathrm{PR}}$}
    & LTX2.3              & $15.5\!\pm\!3.8\,(n{=}127)$ & $13.2\!\pm\!4.1\,(n{=}77)$  & $13.3\!\pm\!2.9\,(n{=}124)$ & $21.0\!\pm\!4.4\,(n{=}100)$ \\
    & Wan2.2-14B          & $21.9\!\pm\!4.4\,(n{=}127)$ & $\textbf{26.4}\!\pm\!4.9\,(n{=}77)$ & $\textbf{13.7}\!\pm\!3.5\,(n{=}124)$ & $18.5\!\pm\!4.1\,(n{=}100)$ \\
    & UniVideo            & $19.1\!\pm\!3.7\,(n{=}126)$ & $19.6\!\pm\!4.5\,(n{=}77)$  & $10.8\!\pm\!2.9\,(n{=}124)$ & $15.0\!\pm\!3.7\,(n{=}100)$ \\
    & HunyuanVideo-1.5    & $\textbf{23.5}\!\pm\!4.3\,(n{=}127)$ & $17.2\!\pm\!4.8\,(n{=}77)$ & $13.6\!\pm\!3.1\,(n{=}124)$ & $21.7\!\pm\!4.6\,(n{=}100)$ \\
    & Cosmos-Predict2.5   & $19.8\!\pm\!3.9\,(n{=}127)$ & $19.7\!\pm\!4.8\,(n{=}77)$  & $11.8\!\pm\!2.8\,(n{=}124)$ & $18.9\!\pm\!4.3\,(n{=}100)$ \\
    & LongCat-Video       & $18.4\!\pm\!3.6\,(n{=}127)$ & $19.8\!\pm\!5.4\,(n{=}77)$  & $13.2\!\pm\!3.0\,(n{=}124)$ & $\textbf{22.7}\!\pm\!4.8\,(n{=}100)$ \\
    \midrule
    \multirow{6}{*}{$S(v)$}
    & LTX2.3              & $30.0\!\pm\!4.4\,(n{=}125)$ & $30.8\!\pm\!5.4\,(n{=}77)$  & $15.8\!\pm\!3.1\,(n{=}116)$ & $28.0\!\pm\!4.3\,(n{=}100)$ \\
    & Wan2.2-14B          & $\textbf{50.2}\!\pm\!5.4\,(n{=}126)$ & $\textbf{54.8}\!\pm\!7.5\,(n{=}76)$ & $19.4\!\pm\!2.6\,(n{=}115)$ & $\textbf{33.5}\!\pm\!4.9\,(n{=}99)$ \\
    & UniVideo            & $38.2\!\pm\!4.5\,(n{=}124)$ & $45.1\!\pm\!6.5\,(n{=}77)$  & $13.4\!\pm\!2.7\,(n{=}118)$ & $20.1\!\pm\!4.7\,(n{=}100)$ \\
    & HunyuanVideo-1.5    & $44.5\!\pm\!5.2\,(n{=}126)$ & $47.2\!\pm\!6.2\,(n{=}77)$  & $16.9\!\pm\!3.3\,(n{=}117)$ & $25.2\!\pm\!4.5\,(n{=}100)$ \\
    & Cosmos-Predict2.5   & $47.4\!\pm\!5.0\,(n{=}125)$ & $46.7\!\pm\!6.1\,(n{=}76)$  & $\textbf{27.1}\!\pm\!3.1\,(n{=}115)$ & $30.4\!\pm\!4.3\,(n{=}99)$ \\
    & LongCat-Video       & $44.3\!\pm\!4.8\,(n{=}125)$ & $51.9\!\pm\!6.4\,(n{=}77)$  & $19.4\!\pm\!3.6\,(n{=}115)$ & $26.7\!\pm\!5.1\,(n{=}100)$ \\
    \bottomrule
  \end{tabular}
\end{table}

\section{Compute Resources}
\label{app:compute}

This appendix reports the hardware, model sizes, and approximate compute budget used to produce the main-text experiments, so that the evaluation pipeline can be reproduced.

\paragraph{Hardware.}
All open-source video generation and all on-premise VLM-judge inference were carried out on NVIDIA H100~80GB GPUs hosted on an internal cluster (Linux, NVLink-equipped 8-GPU nodes, CUDA 12.x, PyTorch~2.x). Closed-source video generators (Sora2, Kling, Wan2.6, Seedance2.0, Veo3.1-Fast) were accessed through their respective commercial APIs and therefore did not consume our local compute. Closed-source VLM judges (GPT-5.4, Gemini-3.1-Flash) were similarly accessed through APIs.

\paragraph{Open-source video generation.}
The six open-source generators (LTX2.3, Wan2.2-14B, UniVideo, HunyuanVideo-1.5, Cosmos-Predict2.5, LongCat-Video) were deployed on H100 nodes following each model's official inference recipe. Each model produces a $5$-second 480p--720p clip from an image-plus-text input; we generate one video per benchmark case under each instruction regime. With $436$ cases and two instruction regimes (Difficult / Easy), each open-source generator contributes ${\sim}872$ generations. Wall-clock per video ranges from approximately $30$~seconds (LTX2.3) to ${\sim}5$~minutes (Wan2.2-14B, HunyuanVideo-1.5) on a single H100, depending on model size and sampler steps.

\paragraph{Open-source VLM-judge inference.}
The headline judge Qwen3.5-27B (and the comparison configurations Qwen3.5-9B and Qwen3.5-27B-Thinking) is deployed on H100 nodes via the vLLM serving stack with tensor parallelism across $4$~H100 GPUs per replica; we run one to two replicas in parallel. The QA pipeline issues one VQA call per case (5--7 questions answered jointly per video) and one binary-judging call per question against ground truth, processed at $4$~FPS with ${\sim}9$k visual tokens per $5$-second video. Per-video judge latency averages $20$--$60$~seconds depending on whether extended thinking is enabled. The full Process-aware Reasoning Verification pass over ${\sim}5$K videos consumes on the order of $50$--$80$~H100$\cdot$hours; the Multi-dimensional Quality Assessment point-wise pass over the same video pool consumes a comparable budget.

\paragraph{Closed-source API usage.}
Closed-source video generation and the GPT-5.4 / Gemini-3.1-Flash judges incur API cost rather than local GPU time. Aggregate API call volume is approximately $5$K image-conditioned video generations across the five closed-source generators and approximately $25$K VLM-judge calls across all reward-model evaluations.

\paragraph{Total project compute.}
Counting only the experiments reported in this paper, the on-premise VLM-judge passes (process-aware verification, point-wise and pair-wise quality assessment, frame-rate ablation, weight-sensitivity bootstrap) consume on the order of $400$~H100$\cdot$hours, and open-source video generation consumes on the order of $1{,}500$~H100$\cdot$hours. Including preliminary experiments, prompt-engineering iterations, failed runs, and earlier versions of the data-curation pipeline that did not appear in the final paper, the cumulative project compute is roughly $2{-}3\times$ the reported figure. Closed-source generation and closed-source judge calls are paid via APIs and are not included in the H100-hour budget. We will release inference scripts, vLLM serving configurations, and seeds together with the benchmark to support reproduction on comparable H100-class hardware.

\section{Broader Impacts}
\label{app:broader-impact}

\paragraph{Positive impacts.}
\WorldReasonBench{} and \WorldRewardBench{} are intended to make the evaluation of modern video generators more trustworthy. By exposing where current systems fail at world-state reasoning---especially on Logic Reasoning and Information-Based content---the benchmark gives researchers and downstream users a structured way to detect ``visually polished but semantically wrong'' generations rather than relying on aggregate aesthetic scores. The paired human-preference data provides a reusable calibration target for any future automatic judge or reward model, which we expect to be useful for safer deployment, model auditing by third parties, and more rigorous comparison across closed-source commercial systems and open-source releases.

\paragraph{Potential negative impacts and mitigations.}
\WorldReasonBench{} does not release any new generative model and does not improve the generative capability of any video model on its own; it is an evaluation suite. Nevertheless, it could indirectly inform stronger video generators over time, which carries the well-known risks of generative video research: misuse for deepfake creation, misleading or fabricated visual evidence, and impersonation. We mitigate this in three concrete ways: (i) the benchmark and reward bench are designed to expose failure modes (mechanism violations, fabricated text/numbers, missing dynamics), so the same diagnostics that are useful for improving models are also directly useful for fake-content detection; (ii) we release only evaluation prompts, automatically generated QA pairs, expert preference labels, and aggregation scripts---we do not release pretrained video generators or fine-tuned reward models with weights; (iii) we report the benchmark's limitations transparently (Appendix~\ref{app:limitations}), so that headline ``score gains'' cannot be used to overstate world-modelling competence in safety-critical contexts. We do not foresee any direct surveillance, privacy, or fairness harm from the benchmark itself.

\paragraph{Annotator wellbeing.}
The five annotators are research-team members who rated AI-generated content and did not interact with end-users or sensitive private data. The annotated material consists of model-generated videos derived from publicly sourced prompts and images and was screened to exclude graphic, explicit, or otherwise harmful content before annotation.

\section{Licenses for Existing Assets}
\label{app:licenses}

This appendix enumerates the major external assets used in this paper, with the license terms we relied on at the time of the experiments. We cite the original works in the references and we use each asset within the scope of its respective license.

\paragraph{Open-source video generation models.}
We use the following open-source generators for inference only, downloaded from their official model cards or repositories: LTX2.3 (Apache~2.0), Wan2.2-14B (Apache~2.0)~\citep{wan2025wan}, UniVideo (research-use license, official repository), HunyuanVideo-1.5 (Tencent Hunyuan Community License Agreement)~\citep{kong2024hunyuanvideo}, Cosmos-Predict2.5 (NVIDIA Open Model License), and LongCat-Video (Meituan LongCat License, research and non-commercial use). Use is consistent with each model card; weights are not redistributed.

\paragraph{Closed-source video generators (commercial APIs).}
We access Sora2 (OpenAI), Kling (Kuaishou), Wan2.6 (Alibaba), Seedance2.0 (ByteDance), and Veo3.1-Fast (Google) through their public commercial APIs and abide by each provider's Terms of Service for evaluation and research use. No model weights are obtained or redistributed; only the videos generated in response to our prompts are stored, and only for the purpose of running this benchmark.

\paragraph{VLM judges.}
Qwen3.5-9B/-27B and the Thinking variants are used under the Tongyi Qianwen License Agreement (research and limited commercial use)~\citep{team2026qwen3}. Gemini-3.1-Flash is accessed via Google Cloud API under the Google API Terms of Service. GPT-5.4 is accessed via the OpenAI API under the OpenAI Terms of Service.

\paragraph{Reference metrics, datasets, and benchmarks.}
We report or cite numbers from VBench~\citep{huang2024vbench} and VBench-2.0~\citep{zheng2025vbench} (Apache~2.0), EvalCrafter~\citep{liu2024evalcrafter} (MIT), FETV~\citep{liu2023fetv} (research license), T2V-CompBench~\citep{sun2025t2v} (Apache~2.0), V-ReasonBench~\citep{luo2025v}, Gen-ViRe~\citep{liu2025can}, VIPER~\citep{li2025viper}, WorldSimBench~\citep{qin2024worldsimbench}, VideoVerse~\citep{wang2025videoverse}, PhyGenBench~\citep{meng2024towards}, and Ruler-Bench~\citep{he2025ruler}. We do not redistribute their data; only summary numbers and qualitative descriptions are quoted.

\paragraph{Software tooling.}
The pipeline relies on standard scientific-computing libraries used within their original licenses: PyTorch (BSD), HuggingFace Transformers (Apache~2.0), vLLM (Apache~2.0), NumPy (BSD), pandas (BSD), and Matplotlib (PSF-style).

\paragraph{Image inputs.}
The initial-state images used as conditioning inputs are sampled from publicly available imagery permitted for research use; no scraped or copyrighted imagery is redistributed. The released benchmark includes only the rewritten textual prompts, generated QA pairs, and expert annotations; image references in our distribution take the form of open-source URLs or hashes rather than re-hosted pixel data, except for the small set of representative figures used in this paper (which are AI-generated or otherwise free-to-distribute).

\paragraph{Released assets.}
\WorldReasonBench{} and \WorldRewardBench{}---including the prompt list, taxonomy, automatic QA pairs, expert preference pairs, and evaluation scripts---will be released under a Creative Commons CC-BY~4.0 license, with attribution to this paper and notification of any derived datasets that re-package our assets.

\section{Limitations}
\label{app:limitations}

We list the known limitations of \WorldReasonBench{} and \WorldRewardBench{} together with the concrete mitigations already in place and the next steps we plan to take.

\paragraph{VLM dependency in QA construction and judging.}
Both QA generation and automatic answer verification rely on VLMs (Qwen3.5 / Gemini-3.1), and the headline metrics in the main text are produced by Qwen3.5-27B. To control for systematic VLM bias we apply three independent checks: (i) a human audit of a stratified random sample of ${\sim}300$ generated QA pairs by two trained annotators, with Cohen $\kappa{=}0.78$ and $7.8\%$ rejected items rewritten or removed before release (Appendix~\ref{app:qa-audit}); (ii) calibration of every automatic ranking against expert-derived Human Elo over ${\sim}6$K preference pairs, where $\mathrm{Score}_{\mathrm{PR}}$ achieves Spearman $\rho{=}0.955$ (Section~\ref{subsec:human_alignment}); and (iii) cross-family comparison of Qwen, Gemini, and (subset) GPT judges, which preserves the closed-vs.-open separation and the Information-Based bottleneck across families (Section~\ref{subsec:ablation}). Residual judge bias persists on close pairs ($47.5\%$ judge accuracy when the human score gap is $\leq 0.5$), and we explicitly recommend that any new judge be calibrated on \WorldRewardBench{} before being used as a headline metric.

\paragraph{Cross-model evaluation set and ranking uncertainty.}
The main-text comparison in Table~\ref{tab:main_results} is performed on a shared evaluation set so that closed- and open-source generators are scored on identical case-level inputs. Joint bootstrap rank stability (Appendix~\ref{app:bootstrap-ci}) shows that the closed-vs.-open separation is statistically robust while strict ordering inside the closed tier is supported only for Seedance2.0; we therefore report the other five closed-source models as a tied cluster rather than a strict ranking. Per-dimension Human-Centric CIs are visibly wider than those on the other three dimensions, so per-dimension Human-Centric ordering should be read as suggestive rather than definitive. Extended open-source results on the full \WorldReasonBench{} benchmark are provided in Appendix~\ref{app:full-open-source-extended}, and broadening cross-model coverage at the per-dimension level is the highest priority for the next benchmark revision.

\paragraph{Scope of the reasoning taxonomy and coverage.}
\WorldReasonBench{} covers four world-state dimensions and $22$ subcategories that focus on initial-state-conditioned future-state prediction. Several reasoning aspects are deliberately out of scope: counterfactual or interventional ``what-if'' queries, multi-agent social dynamics beyond two-actor interactions, exact physics simulation against numerical ground truth (e.g., trajectory MSE), and long-horizon multi-event chains beyond a single transition. Released QA prompts and ground-truth answers are in English only, and case images are sampled from publicly available sources without restriction by region or domain. We do not claim taxonomic exhaustiveness; extending the taxonomy along these axes is intended community-driven future work, which we explicitly invite by open-sourcing the construction pipeline.

\paragraph{Reward-model evaluation scope.}
We evaluate five judges (Qwen3.5-9B / 27B in two configurations, Gemini-3.1-Flash, GPT-5.4) under both pair-wise and point-wise protocols. Three uses are deliberately not validated in this paper and remain future work: (i) end-to-end \emph{training} of a reward model from \WorldRewardBench{} preference pairs; (ii) downstream finetuning of generators guided by $S(v)$ or $\mathrm{Score}_{\mathrm{PR}}$ as a reward signal; and (iii) a comprehensive full-set GPT-5.4 evaluation (currently $595$-video subset, point-wise only). The present results therefore validate \WorldRewardBench{} as a \emph{calibration} benchmark for automatic judges, not yet as a training corpus.

\paragraph{Hint-gain interpretation.}
We report the Easy/Difficult split in Section~\ref{subsec:main_results} as a descriptive signal of how much a model relies on prompt-side guidance, and explicitly do \emph{not} interpret it as direct evidence of latent world reasoning, because ceiling effects, prompt length, and instruction-following capacity could each enlarge the asymmetry. Substantive process-vs-outcome attribution is instead carried by $\mathrm{Score}_{\mathrm{PR}}$ and $s_{\mathrm{dyn}}/\mathrm{Acc}_{\mathrm{QA}}$, both calibrated against human Elo.


\end{document}